\documentclass[final]{jmlrforarxiv} % camera-ready template (de-anonymized)

\title[Mixture of Many Zero-Compute Experts: A High-Rate Quantization Theory Perspective]{Mixture of Many Zero-Compute Experts:\\A High-Rate Quantization Theory Perspective}
\usepackage{times}

\usepackage{amssymb}
\usepackage{amsmath}
\usepackage{mathtools}

\usepackage{enumitem}

\renewcommand{\vec}[1]{\boldsymbol{#1}}
\newcommand{\vecgreek}[1]{\boldsymbol{#1}}
\newcommand{\mtx}[1]{\mathbf{#1}}

\newcommand{\vecrand}[1]{\mathbf{#1}}
\newcommand{\scrand}[1]{\mathrm{#1}}
\newcommand{\greekrand}[1]{#1}

\newcommand{\pdfxvec}{p_{\vecrand{x}}}
\newcommand{\pdfxsc}{p_{\scrand{x}}}
\newcommand{\ith}{i^{\sf th}}

\newcommand{\dint}{\mathrm{d}}

\DeclareMathOperator*{\argmin}{arg\,min}

\newcommand{\expectation}[1]{\mathbb{E}\left[{{#1}}\right]} 
\newcommand{\expectationwrt}[2]{\mathbb{E}_{#2}\left[{{#1}}\right]}
\DeclareMathOperator{\Prob}{Prob}
\newcommand{\eventprob}[1]{\Prob\left[{#1}\right]}
\newcommand{\Ltwonormsquared}[1]{\left\Vert{{#1}}\right\Vert_2^2} 
\newcommand{\Ltwonorm}[1]{\left\Vert{{#1}}\right\Vert_2}

\newcommand{\exind}[1]{^{({#1})}}

\newtheorem{assumption}{Assumption}

\usepackage{color}
\newcommand {\toself}[1]{}
\newcommand {\toselfimportant}[1]{}

\altauthor{%
 \Name{Yehuda Dar} \Email{ydar@bgu.ac.il}\\
 \addr Faculty of Computer and Information Science, Ben-Gurion University
}

\begin{document}

\maketitle

\begin{abstract}%
This paper uses classical high-rate quantization theory to provide new insights into mixture-of-experts (MoE) models for regression tasks. 
Our MoE is defined by a segmentation of the input space to regions, each with a single-parameter expert that acts as a constant predictor with zero-compute at inference. 
Motivated by high-rate quantization theory assumptions, we assume that the number of experts is sufficiently large to make their input-space regions very small. 
This lets us study the approximation error of our MoE model class: (i) for one-dimensional inputs, we formulate the test error and its minimizing segmentation and experts; (ii) for multidimensional inputs, we formulate an upper bound for the test error and study its minimization. 
Moreover, we consider the learning of the expert parameters from a training dataset, given an input-space segmentation, and formulate their statistical learning properties. 
This leads us to theoretically and empirically show how the tradeoff between approximation and estimation errors in MoE learning depends on the number of experts. 
\end{abstract}

\begin{keywords}%
  Mixture of experts, high-rate quantization theory, regression, approximation error, bias-complexity tradeoff.
\end{keywords}

\section{Introduction}

The mixture of experts (MoE) design for machine learning has been increasingly popular recently as a way to efficiently train large models on large datasets. 
The MoE design is usually implemented by routing inputs to one or more predictors, called experts, from an ensemble. 
Specifically, Sparse MoE (SMoE) designs \citep{shazeer2017outrageously,riquelme2021scaling,fedus2022switch,fedus2022review,liu2024deepseekv2} adaptively choose (for a given input) a relatively small number of experts from the overall set of experts.
This effectively reduces the data complexity that each expert should be responsible for its predictions; thus, each expert can be implemented by a simpler sub-model (e.g., less parameters, simpler functionality) that requires less training data. This leads to a more efficient training for the entire MoE model. 
While the recent trend is to use the MoE design for layer blocks within deep neural networks, the MoE was originally invented as a complete prediction model by itself. 

The current popularity of MoE models calls for their foundational understanding. Indeed, recent MoE theories were provided by \cite{zhao2024sparse,jelassi2025mixture,nguyen2023demystifying,nguyen2024towards,nguyen2024statistical,nguyen2024sigmoid,nguyen2024least}; see Section \ref{subsec:related works - Theory for Sparse Mixture of Experts} for an overview. 
Naturally, a comprehensive mathematical theory requires a relatively simple MoE; for example, the theory by \cite{jelassi2025mixture} is for a single-layer transformer model with 1-sparse MoE where a single expert is adaptively chosen for each input. Importantly, 1-sparse MoE designs are practically useful in deep learning \citep{shazeer2017outrageously}.

In this paper, we provide a high-rate quantization theory perspective on 1-sparse MoE models for regression tasks. Our MoE includes many simple experts, specifically, the experts are learnable constant values that can provide predictions at inference without compute beyond routing the input to its expert. Accordingly, we call our design Zero-Compute 1-Sparse MoE (ZC-1SMoE). Despite the per-expert simplicity, the use of many experts prevents over-simplicitly from our ZC-1SMoE. 
Moreover, the idea of zero-compute experts was found by \citet{jin2024moezerocomputation} as practically useful for neural network MoEs; this further motivates our theory for ZC-1SMoE. 

\begingroup
\emergencystretch=2em
The use of many experts facilitates our use of classical high-rate quantization assumptions \citep{bennett1948spectra,panter1951quantization,gersho1979asymptotically,lloyd1982least,na1995bennetts,gray1998quantization}; see Section \ref{sec:Related Work - High-Rate Vector Quantization Theory} and Appendix \ref{appendix:sec:High Rate Quantization - Formulations from Related Work}.  Our central assumption is that there are so many experts such that the input space is segmented to sufficiently small regions (each belongs to another expert) that allow us to analytically approximate the discrete segmentation as a continuous density of segments. This plays a major role in our ZC-1SMoE analysis. Importantly, while we borrow ideas from high-rate quantization theory and its related use for piecewise-constant approximation of functions \citep{dar2019high}, here we use these ideas for a \textit{predictor learning} problem and its aspects that emerge from jointly handling both the input probability distribution and the input-to-output function (the latter function is due to the prediction problem). 
For a fixed routing segmentation, least-squares learning of the constant experts corresponds to histogram regression \citep{nobel1996histogram}. Our analysis characterizes the input-space segmentation and approximation error using high-rate quantization theory, and studies the resulting tradeoff with estimation error when the expert constants are learned from noisy training data.
\par
\endgroup

The main contributions of this paper stem from the novel use of high-rate quantization theory for the research of MoE learning, which provides new insights via the following: 
\begin{enumerate}
    \item Analysis of the approximation error of the ZC-1SMoE model class for one-dimensional inputs, including formulation of the test error and its minimizing segmentation and experts. 
    Specifically, the optimal density of expert regions is formulated and shown how it can be used for forming the optimal one-dimensional input-space segmentation. 
    
    \item Analysis of the approximation error of the ZC-1SMoE model class for multidimensional inputs, including formulation of an upper bound for the test error and study of its minimization. The geometry of the expert regions is discussed, and our formulated segment density that minimizes the error's upper bound is used to guide a recursive rectangular segmentation method for two-dimensional inputs. Optimizing the region shapes remains an open challenge.
    
    \item Analysis of the learning of the expert parameters from a training dataset given an input-space segmentation, formulating their statistical learning properties. This includes theoretical and empirical analysis of the tradeoff between approximation and estimation errors in ZC-1SMoE learning and how it depends on the number of experts. 
\end{enumerate}

\section{Related Work}

\subsection{Theory for Mixture of Experts}
\label{subsec:related works - Theory for Sparse Mixture of Experts}

\citet{zhao2024sparse} empirically and theoretically analyzed sparse MoE (SMoE) to elucidate how the sparsity level (the number of experts that are adaptively used from the overall set of experts) and task complexity (in terms of data distribution that combines more tasks/``skills'') affect the compositional generalization of the learned SMoE. Their theory analytically formulates the generalization error. Moreover, they heuristically define the big-O behavior of the approximation error and estimation error to explain a tradeoff as function of the sparsity level; by this, they explain that there is an optimal sparsity level that depends on the specific setting. Importantly, their approximation error does not depend on the overall number of experts, in contrast to the estimation error that we propose in this paper for the ZC-1SMoE. Moreover, our ZC-1SMoE setting lets us propose a more analytically-justified decomposition of the test error to its approximation and estimation errors.

\citet{jelassi2025mixture} theoretically analyzed a single-layer transformer model with top-1 sparse MoE form. They show, also empirically on deep models, that increasing the overall number of experts is beneficial for memory-based tasks but not for reasoning tasks. 

\citet{nguyen2023demystifying,nguyen2024towards,nguyen2024statistical,nguyen2024sigmoid,nguyen2024least} theoretically studied the density and parameter estimation of MoE model for an unknown mixture data distribution using a Voronoi-cell loss definition. As the number of components (``experts'') in the true data distribution is unknown, the Voronoi cells assign the learned experts to their closest (unknown) true component (``expert'') of the data distribution. While Voronoi cells are related to quantization, the quantization perspective does not seem to appear in these works. Specifically, \cite{nguyen2023demystifying,nguyen2024towards,nguyen2024statistical,nguyen2024sigmoid,nguyen2024least} significantly differ from this paper where we focus on the high-rate quantization theory (namely, segmentation of the input space to \textit{many} regions) and consider a more general data distribution that relies on a non-mixture nonlinear unknown function plus noise.

MoE theories examine different learning approaches, for example, the maximum likelihood estimation is studied by \citet{nguyen2023demystifying,nguyen2024towards,nguyen2024statistical}, and learning via least squares is studied by \citet{nguyen2024sigmoid,nguyen2024least}. 
Here we also use the least squares learning approach and contribute to it the unique perspective of high-rate quantization theory.

\subsection{High-Rate Quantization Theory}
\label{sec:Related Work - High-Rate Vector Quantization Theory}

\subsubsection{High-Rate Scalar Quantization Theory}
The high-rate quantization theory literature traces back to \citet{bennett1948spectra} who provided an approximated error formula for nonuniform scalar quantization; remarkably, his error formula used integration over the continuous input domain, relaxing the discrete nature of the quantization cells. Namely, \citet{bennett1948spectra} effectively assumed that the quantization cells are many and of sufficiently small sizes, such that the quantizer can be described by a smooth function (which also acts as an input compressor function that as a preprocessing can turn a uniform quantizer into a nonuniform quantizer; this is the compander design, which also requires a postprocessing function). This seminal work motivated a long line of research works on high-rate scalar quantization; see the comprehensive overview by \citet{gray1998quantization}. Notably, \citet{panter1951quantization} used the integral error formula to characterize the optimal scalar quantizer based on the input probability distribution function. Moreover, \citet{lloyd1982least} explicitly defined a differentiable density function of the quantizer cells over the continuous input domain; this density function generalizes the role of the derivative of the compressor function defined earlier by \citet{bennett1948spectra}. 
See Appendix \ref{appendix:sec:High Rate Quantization - Formulations from Related Work} for formulations and more details. 

\subsubsection{High-Rate Vector Quantization}
The work by \citet{gersho1979asymptotically} importantly extended the scope of high-rate quantization theory from scalar quantizers (that operate in one-dimensional space) to vector quantizers (that operate in multidimensional space). The principal challenge is that vector quantizers are based on input-space segmentation to multidimensional regions that have shapes and not just sizes (i.e., volumes); this contrasts the simpler case of scalar quantizers whose regions are ``shapeless'' intervals in the one-dimensional input domain. 
To analytically formulate the quantization error, \citet{gersho1979asymptotically} introduced a conjecture on the shapes of the regions in asymptotically optimal vector quantization (optimality in the sense of minimum mean squared error) -- later on, this conjecture became famous and widely-believed \citep{gray1998quantization}. 
According to the conjecture, for high-rate (i.e., quantizer with many regions), most of the regions of an optimal $d$-dimensional vector quantizer are approximately translated, rotated, and rescaled versions of a $d$-dimensional convex polytope that tessellates the $d$-dimensional space and is optimal in the sense of minimal normalized moment of inertia. 
Additional discussion on this conjecture is provided in Appendix \ref{appendix:sec:Related Work: Gersho Conjecture for High-Rate Vector Quantization}.

Another useful assumption given by \citet{na1995bennetts} for high-rate vector quantization analysis is the existence of a smooth function (over the input space) that approximates the normalized moment of inertia of the segmentation regions. We use this assumption for upper bounding the approximation error of the MoE model class for the multidimensional inputs. 

\subsubsection{High-Rate Functional Quantization}
\label{sec:Related Work - Functional Quantization}
\citet{misra2011distributed,sun2013distributed} study distributed functional scalar quantization, where separately quantized scalar sources are used to reconstruct a known computing function of the original source values. Their high-rate analysis optimizes the scalar quantizers for this reconstruction objective.
The joint consideration of source probability density functions and the known computing function in their one-dimensional fixed-rate setting leads to distortion and optimizing point density formulas that mathematically coincide with our intermediate formulations in Section \ref{subsec:Optimizing the Routing Segmentation using Expert Segment Density} for asymptotic approximation error and optimizing segment density for one-dimensional inputs and a given $\beta$ function that defines the noiseless part of the input-to-output mapping in our regression problem.
However, we consider a learning problem that goes beyond the approximation error analysis; our problem requires the learning of an \textit{unknown} $\beta$ function, reflecting an unknown input-to-output relation. Hence, our learning induces an estimation error term (Section \ref{sec:Learning the Constant Experts}) that does not appear in the functional quantization problem of \citet{misra2011distributed,sun2013distributed} due to the known computing function for the quantizer designer. 
Moreover, the distributed scalar quantization via regular scalar companding of \citet{misra2011distributed,sun2013distributed} implies that the effective multidimensional quantization of the joint sources is a separable Cartesian-product segmentation whose quantization cells are multidimensional rectangles parallel to the axes; in contrast, in Section \ref{sec:Multi-Dimensional Inputs}, our theory considers multidimensional segmentations of the input space of more general forms, as in the vector quantization framework by \citet{gersho1979asymptotically,na1995bennetts}, and our empirical two-dimensional segmentation uses a density-guided recursive rectangular segmentation that allows locally varying rectangular splitting that is not restricted to an overall Cartesian-product segmentation.

\section{Problem and Model Definition}
\label{sec:Problem and Model Definition}

Consider a data model where the random input $\vecrand{x}\in[0,1]^d$ is drawn from a probability distribution $P_{\vecrand{x}}$ over the $d$-dimensional unit cube, and the output $\scrand{y}$ is a real value such that  
\begin{equation}
    \label{eq:data model - multidimensional}
    \scrand{y} = \beta(\vecrand{x})+{\greekrand{\epsilon}}
\end{equation}
where $\beta:[0,1]^d\rightarrow\mathbb{R}$ is a non-random function unknown to the learner; $\greekrand{\epsilon}\sim P_{\greekrand{\epsilon}}$ is a real-valued random noise component with zero mean and variance $\sigma_{\epsilon}^2$. $\vecrand{x}$ and $\greekrand{\epsilon}$ are statistically independent. 
The distribution $P_{\vecrand{x}}$ has a positive-valued probability density function $\pdfxvec(\vec{x})$ over $[0,1]^d$. Both $\pdfxvec(\vec{x})$ and $\beta$ are continuously differentiable in $[0,1]^d$.

The goal is to learn a mixture-of-experts predictor of the form 
\begin{equation}
    \label{eq:predictor - general form - multidimensional}
    \widehat{\beta}(\vec{x}) = c_i  ~~\text{for}~~\vec{x}\in A_i
\end{equation}
where the prediction for an input $\vec{x}$ is determined based on a segmentation of the input space $[0,1]^d$ into $m$ regions $\{A_i\}_{i=1}^m$ that satisfy
\begin{align}
\label{eq:segmentation regions conditions}
&A_i\subseteq[0,1]^d ~\text{is measurable, connected, and of nonzero volume},~\forall i\nonumber\\
&\cup_{i=1}^m A_i = [0,1]^d,~~ A_i\cap A_j=\emptyset~~\forall i\ne j.    
\end{align}
For any $i\in\{1,\dots,m\}$, the $i^{\rm th}$ region $A_i$ is associated with a constant $c_i\in\mathbb{R}$. 
This MoE prediction model implies that a given input is routed to an expert based on the region that the input belongs to, and then the prediction is the region's constant without additional compute. Namely, each of the constants $\{c_i\}_{i=1}^m$ represents a zero-compute expert. 
We call this prediction model a zero-compute 1-sparse mixture of experts, which is abbreviated as ZC-1SMoE; in the rest of this paper, the term MoE refers to ZC-1SMoE, unless otherwise specified. 

Let us denote the hypothesis class of the MoE for $d$-dimensional input as 
\begin{equation}
    \label{eq:hypothesis class of d-dimensional MoE - definition}
    \mathcal{H}_{m,d} = \left\{ \widehat{\beta}(\vec{x}) = c_i  ~~\text{for}~~\vec{x}\in A_i   ~\Big\vert~ c_1,\dots,c_m\in\mathbb{R},~\{A_i\}_{i=1}^m~\text{that satisfy (\ref{eq:segmentation regions conditions})}\right\}.
\end{equation}
Moreover, we define a simpler hypothesis class of MoE models for a \textit{given} routing segmentation $\{A_i\}_{i=1}^m$ that satisfies (\ref{eq:segmentation regions conditions}): 
\begin{equation}
    \label{eq:hypothesis class of d-dimensional MoE given segmentation- definition}
    \mathcal{H}_{m,d}^{c}\left(\{A_i\}_{i=1}^m\right) = \left\{ \widehat{\beta}(\vec{x}) = c_i  ~~\text{for}~~\vec{x}\in A_i   ~\Big\vert~ c_1,\dots,c_m\in\mathbb{R}\right\}, 
\end{equation}
namely, only the expert constants should be determined when learning/selecting a model from $\mathcal{H}_{m,d}^{c}\left(\{A_i\}_{i=1}^m\right)$.

The predictor form (\ref{eq:predictor - general form - multidimensional}) motivates the following question: 
\begin{itemize}
    \item What is the best MoE predictor from the hypothesis class $\mathcal{H}_{m,d}$ (\ref{eq:hypothesis class of d-dimensional MoE - definition}) for a given data distribution $P_{\vecrand{x},\scrand{y}}$ that follows the data model (\ref{eq:data model - multidimensional})?
\end{itemize} 
This question requires to answer two subsequent questions: 
\begin{itemize}
    \item What is the optimal input-space segmentation $\{A_i\}_{i=1}^m$ that defines the routing to experts?
    \item What are the optimal constants $\{c_i\}_{i=1}^{m}$ that define the experts' predictions?
\end{itemize}
To answer these questions, we need to define a performance criterion -- in our case this will be minimal prediction error on test data, which corresponds to the best generalization performance. 

We consider a regression problem, the mean squared error (MSE) is used here to quantify prediction performance on a random input-output test pair $(\vecrand{x},\scrand{y})$ drawn from the data model (\ref{eq:data model - multidimensional}) independently of other draws (such as the training data, which will be studied later in Section \ref{sec:Learning the Constant Experts}). The test error of a predictor $\widehat{\beta}$ is 
\begin{equation}
\label{eq:test error - general definition - multidimensional}    
\mathcal{E}_{\sf test} \left(\widehat{\beta}\right) = \expectationwrt{\left( \widehat{\beta}(\vecrand{x}) - \scrand{y} \right)^2}{\vecrand{x},\scrand{y}}.
\end{equation}
For the MoE model (\ref{eq:predictor - general form - multidimensional}) with segmentation $\{A_i\}_{i=1}^m$ and expert constants $\{c_i\}_{i=1}^m$, the test error can be formulated (see Appendix \ref{appendix:subsec:Proof of the Test Error Formula for Multidimensional MoE}) as 
\begin{equation}
\label{eq:test error - MoE - multidimensional}
\mathcal{E}_{\sf test} \left({\{A_i\}_{i=1}^m, \{c_i\}_{i=1}^m}\right) = \sigma_{\epsilon}^2 + \sum_{i=1}^{m} {\int\displaylimits_{\vec{x}\in A_i}{\left(c_i - \beta(\vec{x})\right)^2 \pdfxvec(\vec{x})\dint\vec{x}} }.
\end{equation}

For a start, let us formulate the best predictor in the simpler hypothesis class $\mathcal{H}_{m,d}^{c}\left(\{A_i\}_{i=1}^m\right)$, which minimizes the test error 
\begin{equation}
\label{eq:best in-class predictor that minimizes test error - simpler class given segmentation}    
\widehat{\beta}^{\sf opt} = \underset{\widehat{\beta}\in \mathcal{H}_{m,d}^{c}\left(\{A_i\}_{i=1}^m\right)}{\argmin} \mathcal{E}_{\sf test} \left(\widehat{\beta}\right).
\end{equation}
Solving this minimization problem (see Appendix \ref{appendix:subsec:Proof of the Optimal Constants}) yields that, for a given segmentation $\{A_i\}_{i=1}^m$, the optimal constants $\{c_i\}_{i=1}^m$ that minimize the test error are 
\begin{equation}
    \label{eq:optimal c_i - multidimensional}
    c_i^{\sf opt} = \frac{\int\displaylimits_{\vec{x}\in A_i}{\beta(\vec{x}) \pdfxvec(\vec{x})\dint\vec{x}}}{\int\displaylimits_{\vec{x}\in A_i}{\pdfxvec(\vec{x})\dint\vec{x}}},~~\forall i\in\{1,\dots,m\}.
\end{equation}
The denominator in (\ref{eq:optimal c_i - multidimensional}) is nonzero for any $i\in\{1,\dots,m\}$ due to the assumptions of positive $\pdfxvec$ and regions $\{A_i\}_{i=1}^m$ with nonzero volumes.
The test error of the best predictor in a hypothesis class $\mathcal{H}$ is known as the \textit{approximation error} of $\mathcal{H}$ \citep{shalev2014understanding}, a quantity that reflects the \textit{inductive bias} of using $\mathcal{H}$ for learning a predictor.  Here, the approximation error of the MoE hypothesis class $\mathcal{H}_{m,d}^{c}\left(\{A_i\}_{i=1}^m\right)$ is 
\begin{align}
    \label{eq:approximation error of the simpler hypothesis class given segmentation}
    \mathcal{E}_{\sf app} \left(\mathcal{H}_{m,d}^{c}\left(\{A_i\}_{i=1}^m\right)\right) &= \underset{\widehat{\beta}\in \mathcal{H}_{m,d}^{c}\left(\{A_i\}_{i=1}^m\right)}{\min} \mathcal{E}_{\sf test} \left(\widehat{\beta}\right).
\end{align}

The greater challenge is to formulate the best predictor in the hypothesis class $\mathcal{H}_{m,d}$, which minimizes the test error 
\begin{equation}
\label{eq:best in-class predictor that minimizes test error}    
\widehat{\beta}^{\sf opt} = \underset{\widehat{\beta}\in \mathcal{H}_{m,d}}{\argmin}~ \mathcal{E}_{\sf test} \left(\widehat{\beta}\right), 
\end{equation}
namely, to characterize the \textit{jointly optimal} segmentation $\{A_i\}_{i=1}^m$ and expert constants $\{c_i\}_{i=1}^m$. 
Here, the approximation error of the MoE hypothesis class $\mathcal{H}_{m,d}$ is 
\begin{align}
    \label{eq:approximation error of the hypothesis class}
    \mathcal{E}_{\sf app} \left(\mathcal{H}_{m,d}\right) &= \underset{\widehat{\beta}\in \mathcal{H}_{m,d}}{\min} \mathcal{E}_{\sf test} \left(\widehat{\beta}\right).
\end{align}

The formulation of the best in-class predictor and approximation error of $\mathcal{H}_{m,d}$ is the focus of Section \ref{sec:One-Dimensional Inputs} for one-dimensional inputs. The more intricate case of multidimensional inputs is addressed in Section \ref{sec:Multi-Dimensional Inputs} by formulating an upper bound to the approximation error and explaining the challenge in finding the optimal multidimensional segmentation. 
Section \ref{sec:Learning the Constant Experts} analyzes the tradeoff between the approximation error and estimation error in learning MoE from $\mathcal{H}_{m,d}^{c}\left(\{A_i\}_{i=1}^m\right)$.

\section{Approximation Error of MoE for One-Dimensional Inputs}
\label{sec:One-Dimensional Inputs}
In this section, we focus on the case of one-dimensional inputs that are drawn from a probability distribution $P_{\scrand{x}}$ over the unit interval $[0,1]$. Then, the data model (\ref{eq:data model - multidimensional}) for $d=1$ reduces to $\scrand{y}~=~\beta(\scrand{x})+{\greekrand{\epsilon}}$ where $\beta:[0,1]\rightarrow\mathbb{R}$ is a non-random function unknown to the learner; ${\greekrand{\epsilon}}\sim P_{\greekrand{\epsilon}}$ is a real-valued random noise component with zero mean and variance $\sigma_{\greekrand{\epsilon}}^2$.
The formulations in Section \ref{sec:Problem and Model Definition} should be considered with $d=1$ in this section. The notation of one-dimensional inputs and other scalar variables is not in bold font to emphasize differences from the multidimensional case.

Here, the regions $\{A_i\}_{i=1}^m$ that segment the input space are subintervals of the unit interval. Namely, $0=a_0<a_1<\dots<a_{m-1}<a_m=1$ are parameters that define $m$ subintervals $A_i=[a_{i-1},a_i)$, $i\in\{1,\dots,m\}$ that segment $[0,1]$; formally, the $m^{\sf th}$ subinterval is $[a_{m-1},a_m]$ but for writing simplicity it will be referred to as $[a_{m-1},a_m)$ as the other subintervals. Accordingly, the multidimensional MoE from (\ref{eq:predictor - general form - multidimensional}) reduces to the 1D-MoE model: 
\begin{equation}
    \label{eq:predictor - general form - one-dimensional}
    \widehat{\beta}(x) = c_i  ~~\text{for}~~x\in[a_{i-1},a_i)
\end{equation}
where $c_i\in\mathbb{R}$ is the constant expert of the $i^{\rm th}$ subinterval $[a_{i-1},a_i)$. 
%Similarly to the multidimensional case, this prediction model implies that a given input should be routed to its subinterval and then the prediction is provided as the subinterval's constant without additional compute. 

For formulating the best predictor in the hypothesis class and its approximation error (i.e., solving (\ref{eq:best in-class predictor that minimizes test error}), (\ref{eq:approximation error of the hypothesis class})), 
\textit{the important aspect of the 1D-MoE is that optimizing the input space segmentation is tractable}, in contrast to the multidimensional case. This is because multidimensional regions $\{A_i\}_{i=1}^m$ are defined by their volumes and \textit{shapes}. In contrast, segmenting the unit interval requires to optimize $m-1$ scalar variables $\{a_i\}_{i=1}^{m-1}$ that create subintervals by determining their lengths (which take the role of region volumes) without any other optimizable aspect of shape.

For the 1D-MoE from (\ref{eq:predictor - general form - one-dimensional}) with segmentation parameters $\{a_i\}_{i=0}^m$ and expert constants $\{c_i\}_{i=1}^m$, the test error (\ref{eq:test error - MoE - multidimensional}) becomes 
\begin{equation}
\label{eq:test error - MoE - one dimensional}    
\mathcal{E}_{\sf test} \left({\{a_i\}_{i=0}^m, \{c_i\}_{i=1}^m}\right) = \sigma_{\epsilon}^2 + \sum_{i=1}^{m} {\int\displaylimits_{x=a_{i-1}}^{a_i}{\left(c_i - \beta(x)\right)^2 \pdfxsc(x)\dint x}}.
\end{equation}
From (\ref{eq:optimal c_i - multidimensional}) we get that the optimal expert constant for a given subinterval $[a_{i-1},a_i)$ is 
\begin{equation}
    \label{eq:optimal c_i - one-dimensional}
    c_i^{\sf opt} = \frac{\int\displaylimits_{x=a_{i-1}}^{a_i}{\beta(x) \pdfxsc(x)\dint x}}{\int\displaylimits_{x=a_{i-1}}^{a_i}{\pdfxsc(x)\dint x}}.
\end{equation}

\subsection{Small Subintervals Assumption and Locally-Linear Approximations}
We now assume that there are many experts (i.e., $m$ is high) such that the subintervals are small and, therefore, allow locally-linear approximations of $\beta$ and $\pdfxsc$. For the $\ith$ subinterval, we denote its length as $\Delta_i \triangleq a_i - a_{i-1}$ and its center as $x_i \triangleq \frac{a_{i-1}+a_i}{2}$. 
Then, the first-order Taylor approximation of $\beta$ around the $\ith$ subinterval center $x_i$ is 
\begin{equation}
    \label{eq:first-order Taylor approximation of beta - one-dimensional}
    \beta(x) = \beta(x_i) +  \beta'(x_i)\cdot\left(x-x_i\right) + R_{\beta,1}(x)
\end{equation}
where $R_{\beta,1}\in o\left(\lvert{x-x_i}\rvert\right)$ is a remainder term. 
Similarly, the approximation for the probability density function of $x$ is 
\begin{equation}
    \label{eq:first-order Taylor approximation of p - one-dimensional}
    \pdfxsc(x) = \pdfxsc(x_i) +  \pdfxsc'(x_i)\cdot\left(x-x_i\right) + R_{\pdfxsc,1}(x)
\end{equation}
where $R_{\pdfxsc,1}\in o\left(\lvert{x-x_i}\rvert\right)$ is a remainder term. Using the locally-linear approximations (\ref{eq:first-order Taylor approximation of beta - one-dimensional})-(\ref{eq:first-order Taylor approximation of p - one-dimensional}), the optimal expert constants (\ref{eq:optimal c_i - one-dimensional}) and test error (\ref{eq:test error - MoE - one dimensional}) can be formulated as follows. 
\begin{assumption}
    \label{assumption: asymptotic sequence of segementations - one-dimensional}
    Consider a sequence of segmentations $\Gamma_m=\left\{\left[a_{i-1}^{(m)},a_{i}^{(m)}\right)\right\}_{i=1}^m$ of $[0,1]$ such that 
    \begin{equation}
        \label{eq:asymptotic sequence of segementations - one-dimensional}
        \max_{i\in\{1,\dots,m\}}\Delta_{i}^{(m)}\rightarrow 0 ~~\text{as}~m\rightarrow\infty
    \end{equation}
    where $\Delta_{i}^{(m)} \triangleq a_{i}^{(m)} - a_{i-1}^{(m)}$ is the length of the $\ith$ subinterval of $\Gamma_m$. 
    To ease notations, we omit the $m$ superscripts in subsequent notation.
\end{assumption}
\begin{theorem}
\label{theorem:optimal constants and test error in sum form - one-dimensional}
    Consider Assumption \ref{assumption: asymptotic sequence of segementations - one-dimensional} and the regularity conditions from Section \ref{sec:Problem and Model Definition} for $\beta$, $\pdfxsc$ and the segmentation.  Then, for a segmentation of $[0,1]$ by $m$ subintervals $\{[a_{i-1},a_i)\}_{i=1}^m$, the optimal expert constants are 
    \begin{equation}
    \label{eq:theorem - optimal constant - one-dimensional}
        c_i^{\sf opt} = \beta(x_i) + o(\Delta_i)~~\text{as}~m\rightarrow \infty,~~ \forall i\in\{1,\dots,m\}.
    \end{equation}
    The corresponding test error can be formulated as 
    \begin{equation}
        \label{eq:theorem - test error sum formula - one-dimensional}
        \mathcal{E}_{\sf test} = \sigma_{\epsilon}^2 + \frac{1}{12}\sum_{i=1}^{m}{ \left({\beta'(x_i)}\right)^2 \pdfxsc(x_i) \Delta_i^3 + o\left({\sum_{i=1}^{m}\Delta_i^3}\right) }.
    \end{equation}
    % where $\Delta_{\sf max}= \underset{i\in\{1,\dots,m\}}{\max} \Delta_i$ is the largest subinterval length. 
\end{theorem}
The proof is in Appendix \ref{appendix:subsec:proof of the optimal constants and test error in sum form - one-dimensional}. 
The test error in (\ref{eq:theorem - test error sum formula - one-dimensional}) is the approximation error $\mathcal{E}_{\sf app}$ of the hypothesis class $\mathcal{H}_{m,1}^c\left(\{[a_{i-1},a_i)\}_{i=1}^m\right)$; recall the definitions in (\ref{eq:hypothesis class of d-dimensional MoE given segmentation- definition}), (\ref{eq:approximation error of the simpler hypothesis class given segmentation}).

\subsection{Optimizing the Routing Segmentation using Expert Segment Density}
\label{subsec:Optimizing the Routing Segmentation using Expert Segment Density}

Motivated by the density function of quantization cells in high-rate quantization theory (defined for scalar quantization by \citet{lloyd1982least} and for vector quantization by \citet{gersho1979asymptotically,na1995bennetts}), we now define a density function for the input-space segments of experts in our problem. 
\begin{assumption}
\label{assumption: expert segment density function - one-dimensional}
    There exists an expert segment density function $\lambda:[0,1]\rightarrow(0,\infty)$ which is smooth and satisfies 
    \begin{equation}
        \label{eq:expert segment density function - definition - one-dimensional}
        \sup_{x\in[0,1]}\left\lvert \lambda_m(x) - \lambda(x) \right\rvert \rightarrow 0 ~~~\text{as}~~m\rightarrow \infty
    \end{equation}
    where $\lambda_m(x) = \frac{1}{m\Delta_i}$ for $x\in[a_{i-1},a_i)$, $\forall i\in\{1,\dots,m\}$.
    Therefore, for sufficiently large $m$, at the subinterval centers, $\lambda(x_i) \approx \frac{1}{m\Delta_i}$, $\Delta_i \approx \frac{1}{m\lambda(x_i)}$, $\forall i\in\{1,\dots,m\}$.
\end{assumption}

The asymptotically small segments let the sum in (\ref{eq:theorem - test error sum formula - one-dimensional}) converge to an integral, leading to the following test error formula as a corollary of Theorem \ref{theorem:optimal constants and test error in sum form - one-dimensional}.
\begin{corollary}
\label{corollary:test error integral formula - one-dimensional}
Under the conditions of Theorem \ref{theorem:optimal constants and test error in sum form - one-dimensional} and Assumption \ref{assumption: expert segment density function - one-dimensional}, the test error for optimal expert constants asymptotically becomes continuous integration: 
       \begin{equation}
        \label{eq:corollary - test error integral formula with asymptotic remainder - one-dimensional}
        \mathcal{E}_{\sf test} = \sigma_{\epsilon}^2 + \frac{1}{12m^2}\int_{x=0}^{1}{ \frac{\left({\beta'(x)}\right)^2 \pdfxsc(x)}{\lambda^2(x)} \dint x} + o(m^{-2})~~~\text{as}~~m\rightarrow\infty.
    \end{equation}
\end{corollary}
In high-rate quantization theory, related integral formulas of the error (e.g., (\ref{appendix:scalar quantizer squared error - approximate error - integration}) in Appendix \ref{appendix:subsec:Scalar High-Rate Quantization - Approximated Error Formula}) were informally justified based on the large number of segments (quantization cells) by \citet{bennett1948spectra,panter1951quantization,lloyd1982least}, and formally justified by \citet{na1995bennetts} for vector quantizers. We prove Corollary \ref{corollary:test error integral formula - one-dimensional} in Appendix \ref{appendix:subsec:Additional Details on Corollary - test error integral formula - one-dimensional}. 

We define the asymptotic test error formula associated with (\ref{eq:corollary - test error integral formula with asymptotic remainder - one-dimensional}) as 
       \begin{equation}
        \label{eq:corollary - test error integral formula - one-dimensional}
        \widetilde{\mathcal{E}}_{\sf test} = \sigma_{\epsilon}^2 + \frac{1}{12m^2}\int_{x=0}^{1}{ \frac{\left({\beta'(x)}\right)^2 \pdfxsc(x)}{\lambda^2(x)} \dint x}.
    \end{equation}
We next study the minimization of the asymptotic test error formula from (\ref{eq:corollary - test error integral formula - one-dimensional}). In this sense, the following theorem provides the asymptotic-optimal expert segment density function $\lambda^{\sf opt}$ and its corresponding minimal value of the asymptotic test error formula.  
\begin{theorem}
    \label{theorem: optimal expert segment density function and test error - one-dimensional}
    The asymptotic test error formula $\widetilde{\mathcal{E}}_{\sf test}$ (\ref{eq:corollary - test error integral formula - one-dimensional}) is minimized by the asymptotic-optimal expert segment density function 
    \begin{equation}
        \label{eq:optimal expert segment density function - one-dimensional}
        \lambda^{\sf opt}(x)=\frac{\sqrt[3]{\pdfxsc(x)\left(\beta'(x)\right)^2 }}{\int\displaylimits_{\xi=0}^{1}{\sqrt[3]{\pdfxsc(\xi)\left(\beta'(\xi)\right)^2 }\dint\xi}}
    \end{equation}
    that attains the minimal asymptotic test error of 
    \begin{equation}
        \label{eq:optimal test error - one-dimensional}
        \widetilde{\mathcal{E}}_{\sf test}^{\sf opt} = \sigma_{\epsilon}^2 + \frac{1}{12m^2}\left(\,{ \int\displaylimits_{x=0}^{1}{\sqrt[3]{\pdfxsc(x)\left(\beta'(x)\right)^2 \dint x}} }\right)^3.
    \end{equation}    
\end{theorem}
See remark in Appendix \ref{appendix:subsec: remark for theorem on optimal expert segment density function and test error - one-dimensional}. The theorem is proved in Appendix \ref{appendix:subsec: proof of theorem on optimal expert segment density function and test error - one-dimensional}.
Recall that the approximation error $\mathcal{E}_{\sf app}$ of the hypothesis class $\mathcal{H}_{m,1}$, as defined in (\ref{eq:hypothesis class of d-dimensional MoE - definition}) and (\ref{eq:approximation error of the hypothesis class}), is the minimum test error over both the segmentation and the expert constants. 
Within the considered framework of high-rate and segment density assumptions, (\ref{eq:optimal test error - one-dimensional}) gives the asymptotic approximation error formula for the hypothesis class $\mathcal{H}_{m,1}$.

In Section \ref{appendix:subsec:Empirical Results - Optimal Segmentation for MoE with One-Dimensional Inputs} we provide and discuss the empirical evaluations that correspond to the theory in this section; showcasing the segmentations constructed from the asymptotic-optimal density (which we call asymptotic-optimal segmentations), expert constants, segment density functions, and that the asymptotic test error formula (\ref{eq:optimal test error - one-dimensional}) excellently matches the empirical evaluation on test data (unless the number of experts is too small, as reasonable due to our many-experts assumption). 

\subsection{Differences Between the 1D-MoE Learning and Scalar Quantization}
\label{subsec:Differences Between the 1D-MoE Learning and Scalar Quantization}
Let us highlight some of the prominent differences between our current analysis and the classical high-rate quantization analysis. 
Many differences between the two tasks emerge from their basic error formulations that should be minimized; i.e., see that the MoE test error (\ref{eq:test error - MoE - one dimensional}) involves both $\pdfxsc$ and $\beta$, whereas the quantization error (\ref{appendix:scalar quantizer squared error - discrete}) is based only on $\pdfxsc$ and there is no $\beta$ in the problem --- this important difference holds even before the high-rate assumptions.
Then, after applying high-rate assumptions, the optimal expert constants (\ref{eq:theorem - optimal constant - one-dimensional}) are approximately the $\beta$ evaluations at the interval midpoints, whereas the optimal quantizer representations are approximated just as the interval midpoints (\ref{appendix:eq:high-rate scalar quantization - optimal representations are interval centers}). 

Importantly, for MoE learning, the asymptotic error and its minimizing segment density and value depend on the product between the input PDF $\pdfxsc$ and the squared derivative of the input-output relation function $\left(\beta'(x)\right)^2$; whereas for scalar quantization the dependency is only on the input PDF $\pdfxsc$. To observe this, compare (\ref{eq:corollary - test error integral formula - one-dimensional}), (\ref{eq:optimal expert segment density function - one-dimensional}), (\ref{eq:optimal test error - one-dimensional}) with (\ref{appendix:scalar quantizer squared error - approximate error - integration}), (\ref{appendix:scalar quantizer squared error - optimal density}), (\ref{appendix:eq:optimal quantization error - one-dimensional}), respectively. Our minimizing segment density has a mathematical form as in scalar functional quantization \citep{misra2011distributed,sun2013distributed}; recall Section \ref{sec:Related Work - Functional Quantization}. Nevertheless, in contrast to functional quantization, our MoE learning setting additionally treats $\beta$ as unknown to the learner: the expert constants must be estimated from noisy training data. Section \ref{sec:Learning the Constant Experts} analyzes the resulting estimation error for a given segmentation and its interplay with approximation error as the number of experts varies.

\subsection{Empirical Results: Approximation Error, Optimal Segmentation and Experts}
\label{appendix:subsec:Empirical Results - Optimal Segmentation for MoE with One-Dimensional Inputs}

\begin{figure}[]
    \centering
    \subfigure[]{
    \includegraphics[width=0.36\textwidth]{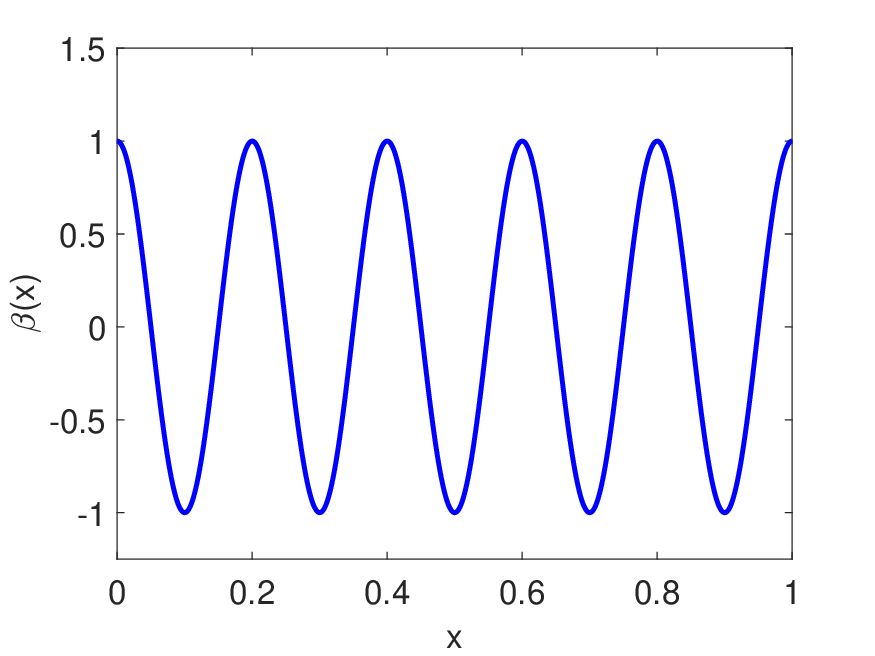}
    \label{fig:example 1 - beta}
    }
    \subfigure[]{
    \includegraphics[width=0.36\textwidth]{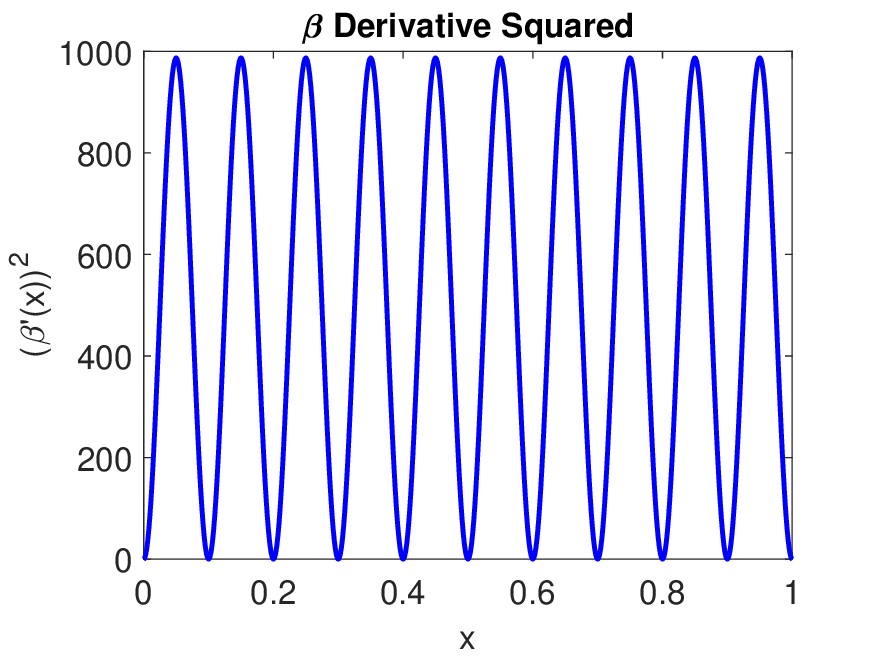}
    \label{fig:example 1 - beta derivative energy}
    }
    \\
    \subfigure[]{
    \includegraphics[width=0.36\textwidth]{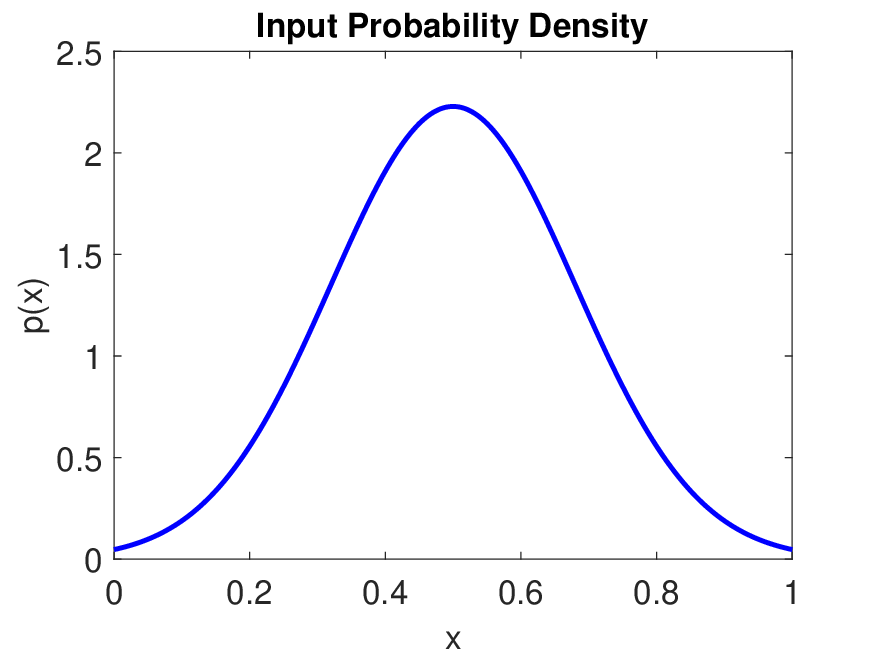}
    \label{fig:example 1 - pdf x}
    }
    \subfigure[]{
    \includegraphics[width=0.36\textwidth]{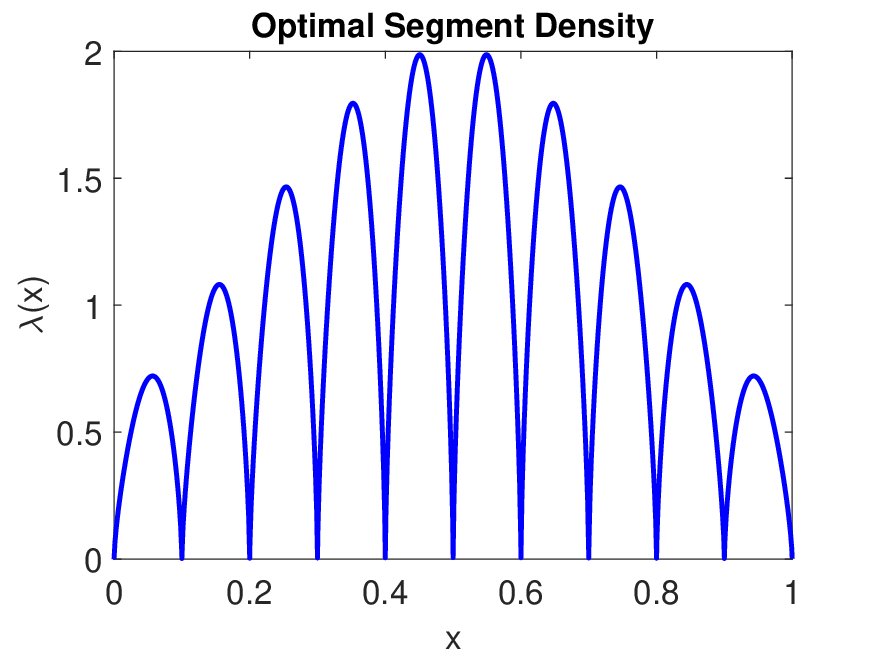}
    \label{fig:example 1 - optimal segment density}
    }
    \\
    \subfigure[]{
    \includegraphics[width=0.36\textwidth]{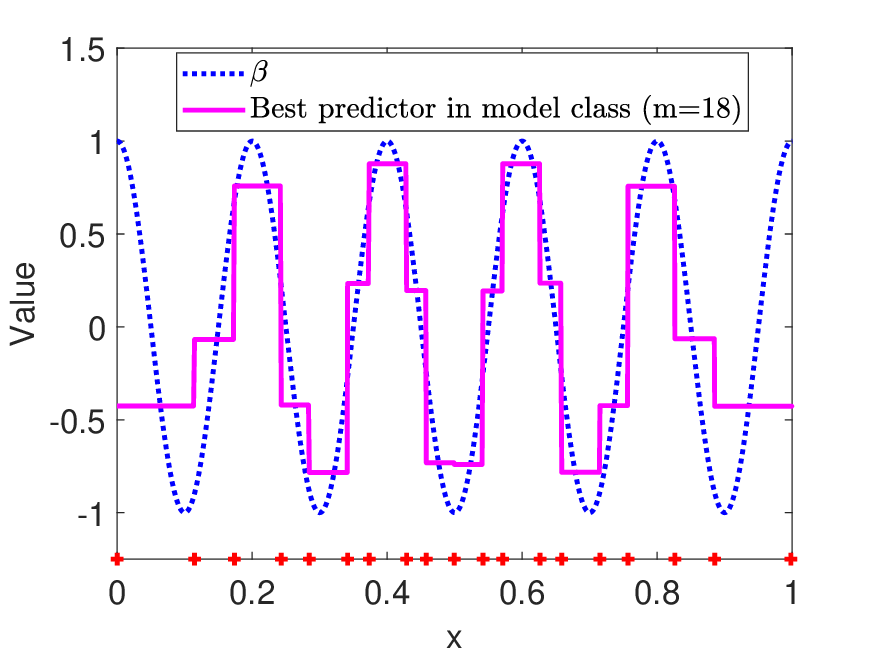}
    \label{fig:example 1 - beta reconstruction and segmentation m=18}
    }
    \subfigure[]{
    \includegraphics[width=0.36\textwidth]{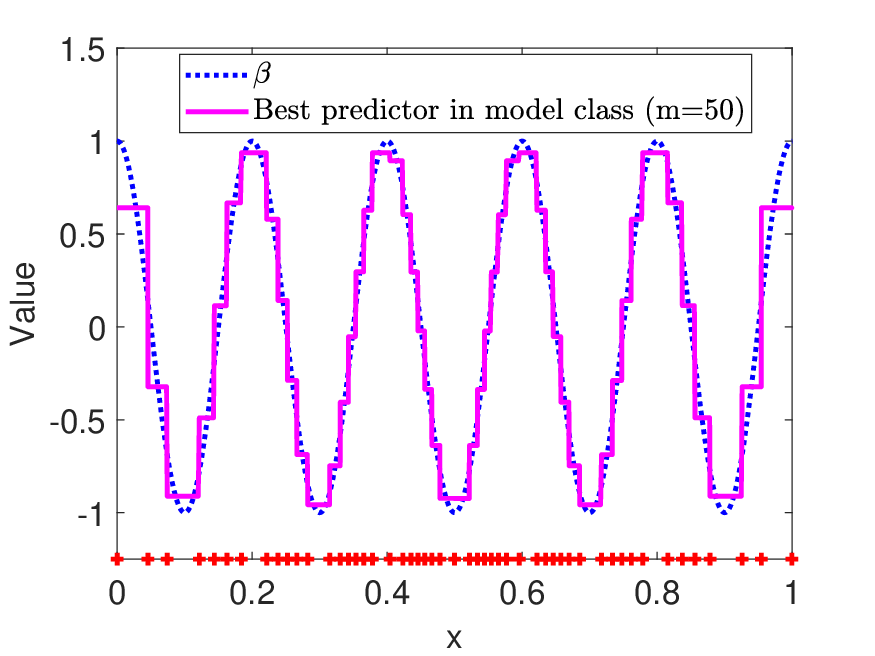}
    \label{fig:example 1 - beta reconstruction and segmentation m=50}
    }
    \\
    \subfigure[]{
    \includegraphics[width=0.36\textwidth]{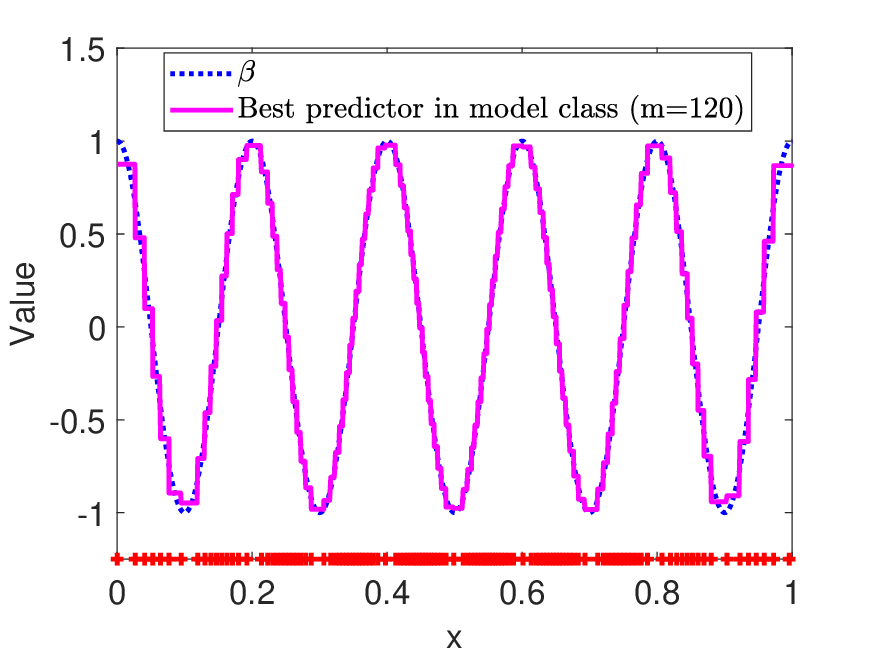}
    \label{fig:example 1 - beta reconstruction and segmentation m=120}
    }
    \subfigure[]{
    \includegraphics[width=0.36\textwidth]{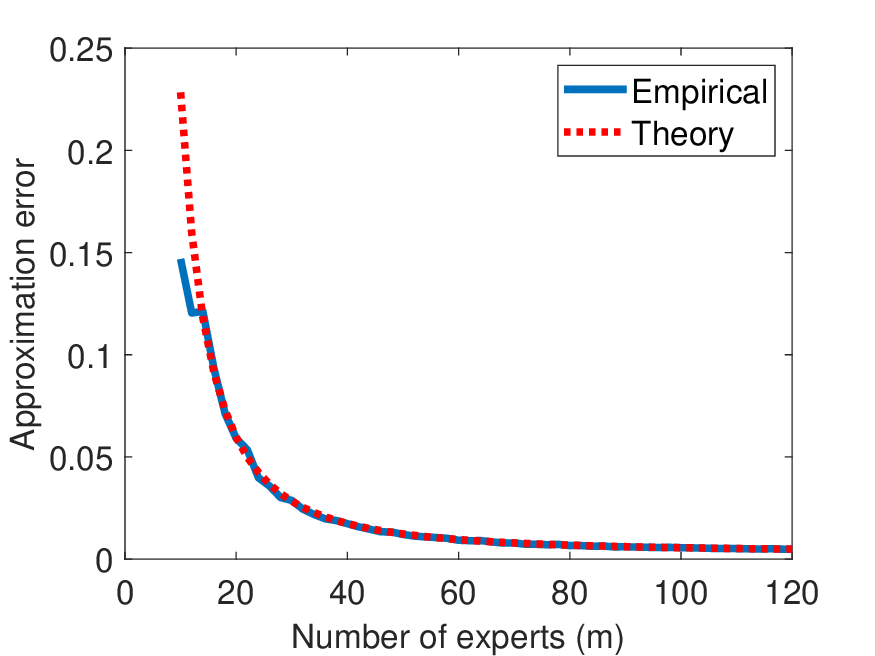}
    \label{fig:example 1 - approximation error curve vs m}
    }
    \caption{Examples for optimal segmentation, best predictor in $\mathcal{H}_{m,1}$, and approximation error curves for a \textbf{cosine} $\beta$ and truncated Gaussian $\pdfxsc$. In (e)-(g), the red markers on the $x$-axis denote the optimal segmentation points $\{a_i\}_{i=1}^m$ that were formed from the optimal segment density in (d).}
    \label{fig:empirical results - example 1 - cosine beta}
\end{figure}

\begin{figure}[]
    \centering
    \subfigure[]{
    \includegraphics[width=0.36\linewidth]{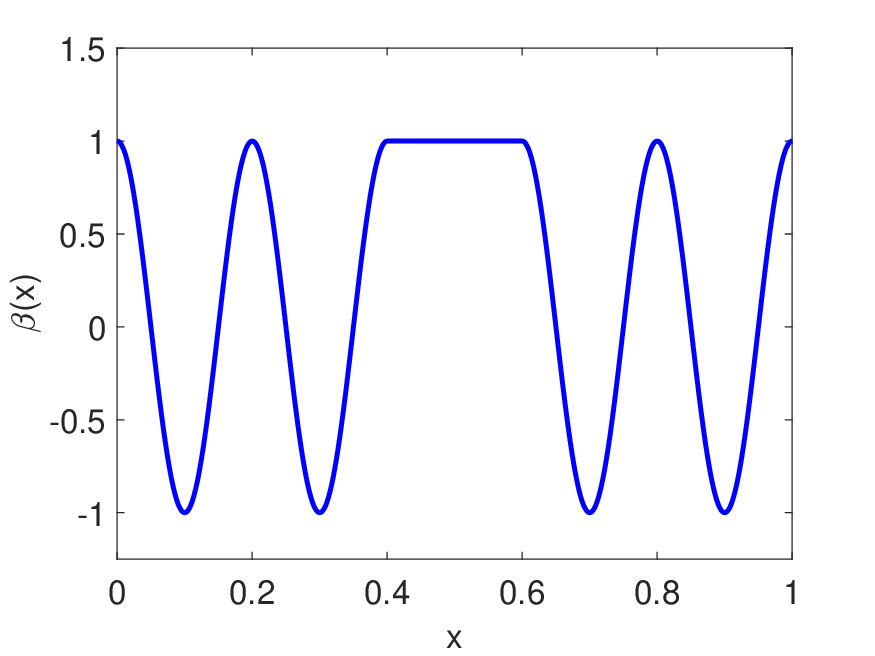}
    \label{fig:example 2 - beta}
    }
    \subfigure[]{
    \includegraphics[width=0.36\linewidth]{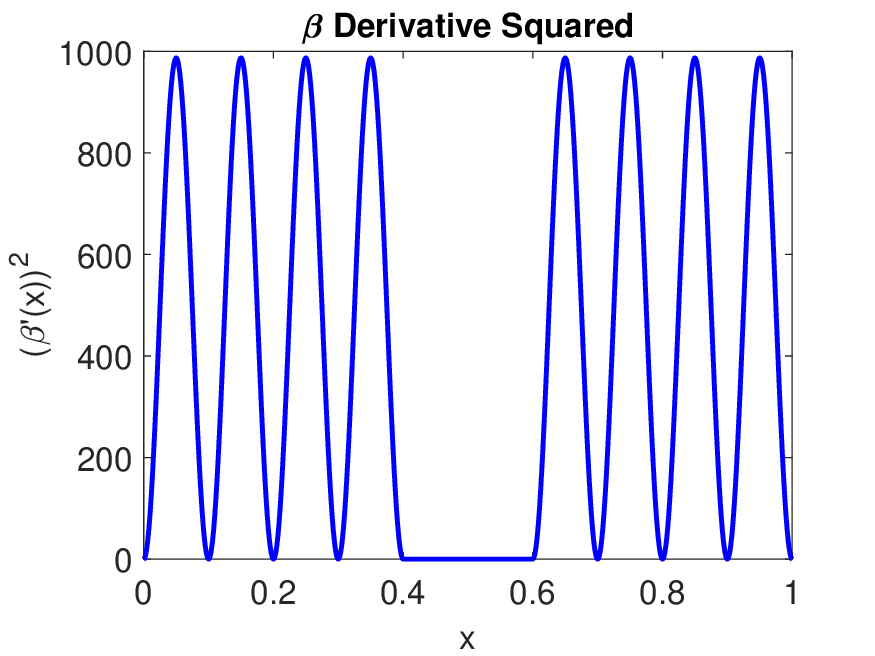}
    \label{fig:example 2 - beta derivative energy}
    }
    \\
    \subfigure[]{
    \includegraphics[width=0.36\linewidth]{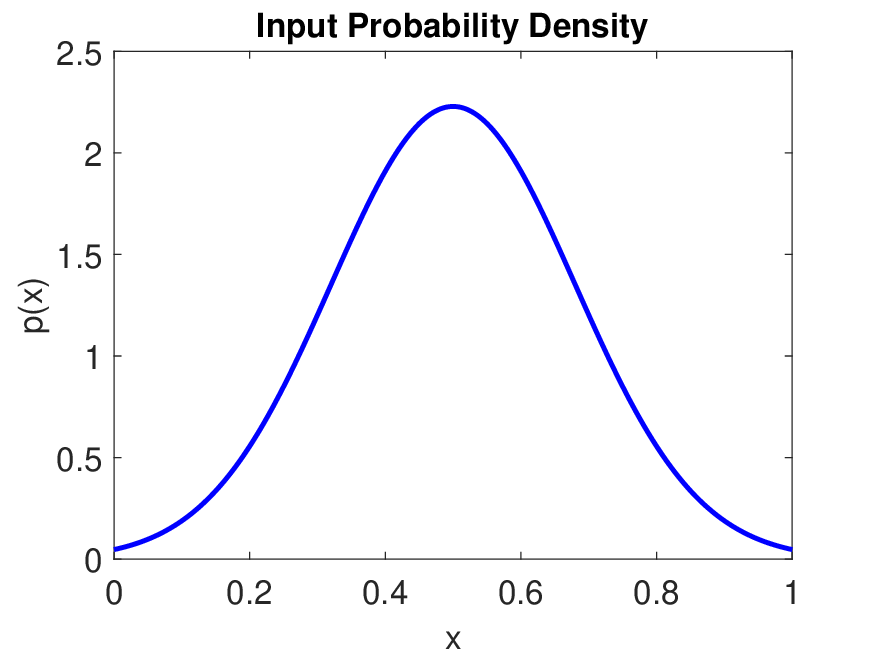}
    \label{fig:example 2 - pdf x}
    }
    \subfigure[]{
    \includegraphics[width=0.36\linewidth]{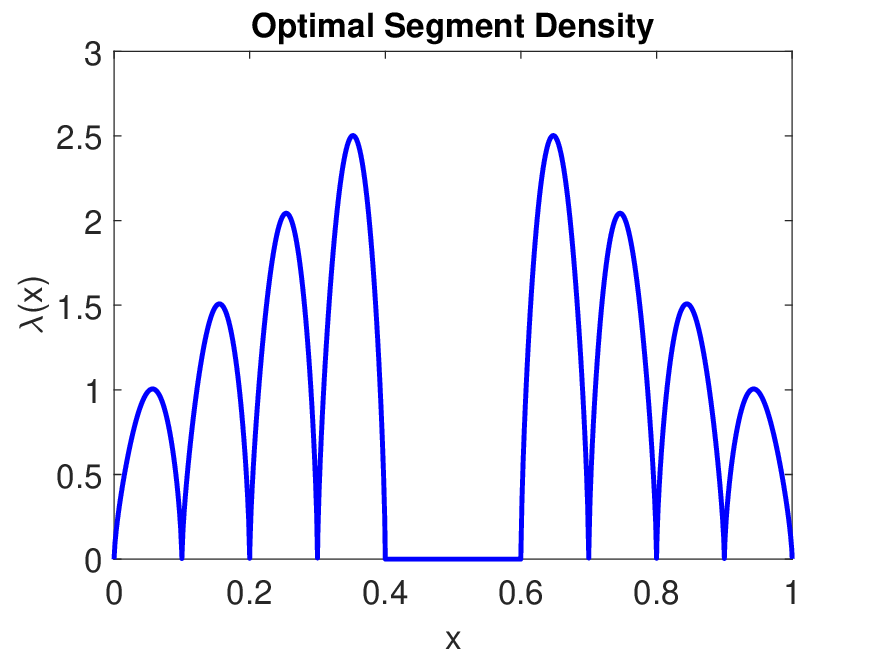}
    \label{fig:example 2 - optimal segment density}
    }
    \\
    \subfigure[]{
    \includegraphics[width=0.36\linewidth]{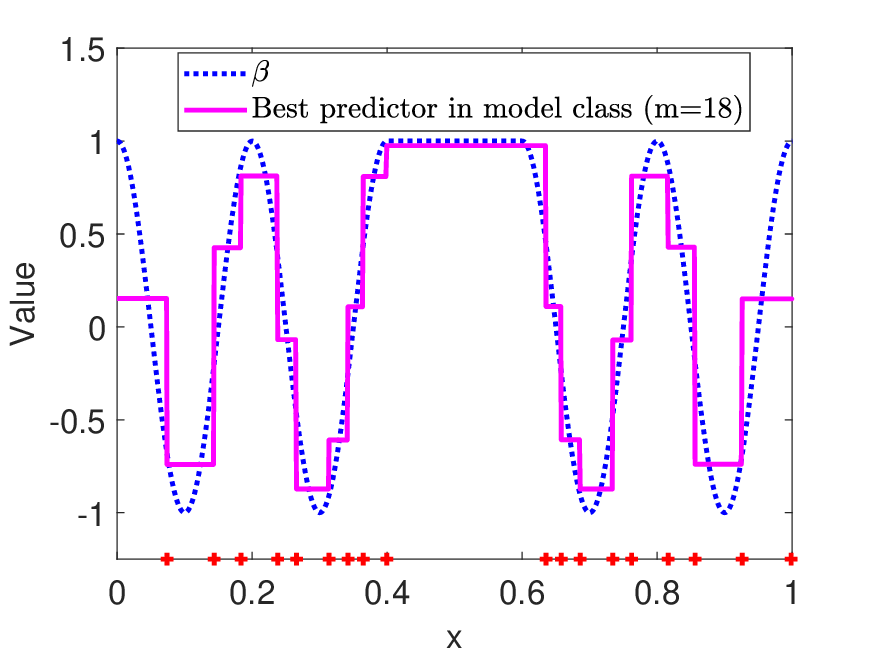}
    \label{fig:example 2 - beta reconstruction and segmentation m=18}
    }
    \subfigure[]{
    \includegraphics[width=0.36\linewidth]{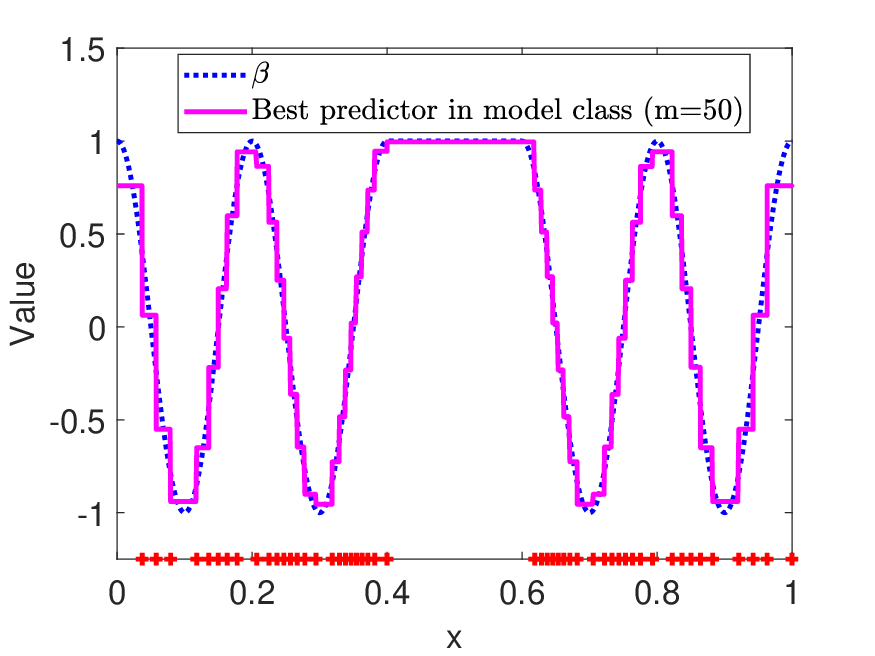}
    \label{fig:example 2 - beta reconstruction and segmentation m=50}
    }
    \\
    \subfigure[]{
    \includegraphics[width=0.36\linewidth]{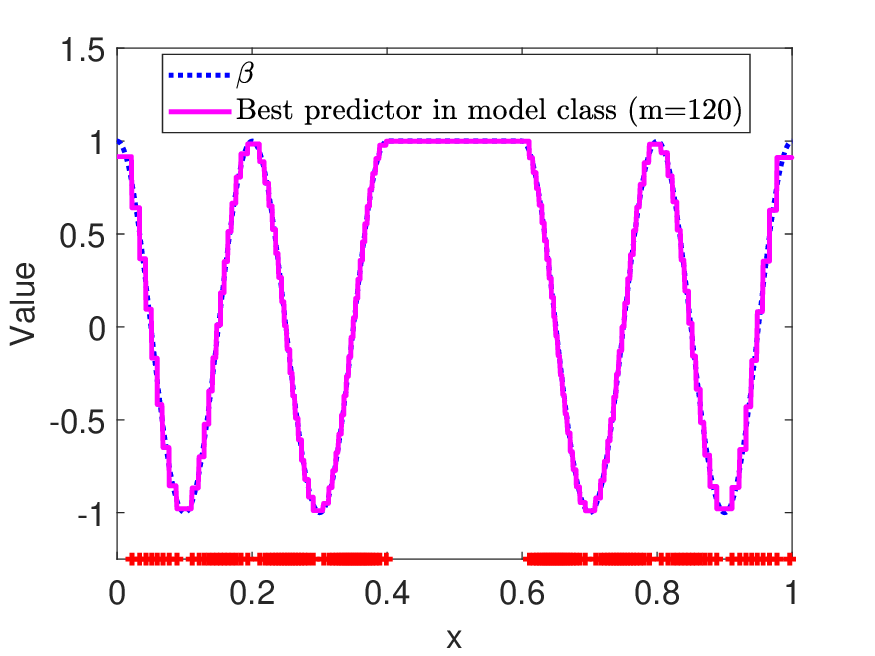}
    \label{fig:example 2 - beta reconstruction and segmentation m=120}
    }
    \subfigure[]{
    \includegraphics[width=0.36\linewidth]{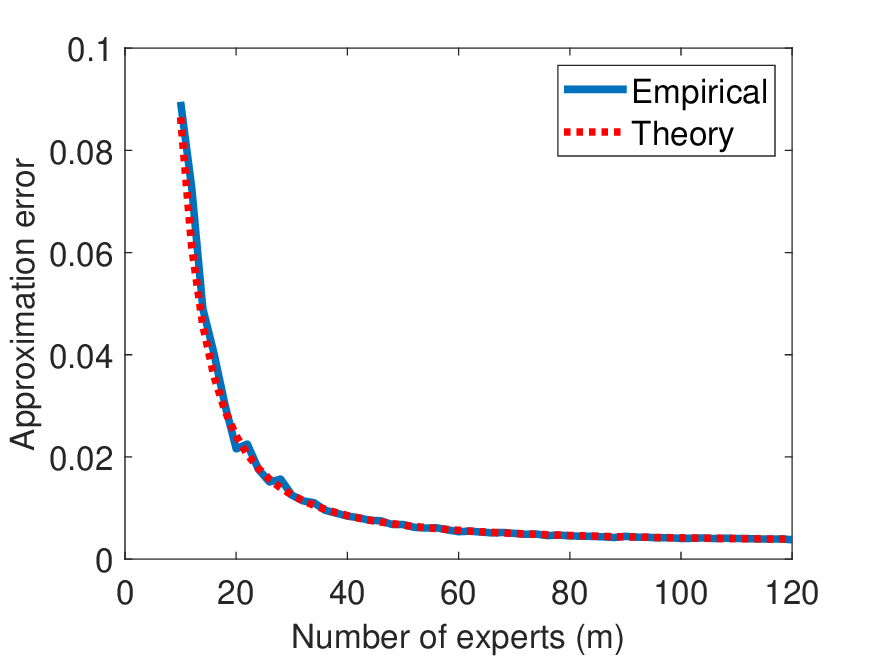}
    \label{fig:example 2 - approximation error curve vs m}
    }
    \caption{Examples for optimal segmentation, best predictor in $\mathcal{H}_{m,1}$, and approximation error curves for a \textbf{cosine with a constant segment} $\beta$ and truncated Gaussian $\pdfxsc$. In (e)-(g), the red markers on the $x$-axis denote the optimal segmentation points $\{a_i\}_{i=1}^m$ that were formed from the optimal segment density in (d).}
    \label{fig:empirical results - example 2 - cosine beta with constant segment}
\end{figure}

In Figs.~\ref{fig:empirical results - example 1 - cosine beta}-\ref{fig:empirical results - example 2 - cosine beta with constant segment} we empirically exemplify the optimal segmentation that corresponds to the optimal segment density $\lambda^{\sf opt}$ from Eq.~(\ref{eq:optimal expert segment density function - one-dimensional}) of Theorem \ref{theorem: optimal expert segment density function and test error - one-dimensional}. The discrete segmentation is formed by the formulations (\ref{appendix:eq:optimal expander function for scalar quantization})-(\ref{appendix:eq:optimal scalar quantization - optimal segmentation by expander}) of Appendix \ref{appendix:subsec:Scalar High-Rate Quantization - Forming a Quantizer from a Quantization Interval Density} but instead of the quantizer density $g^{\sf opt}$ we use our MoE density $\lambda^{\sf opt}$ from (\ref{eq:optimal expert segment density function - one-dimensional}). 

Recall that (\ref{appendix:eq:optimal expander function for scalar quantization}) has a unique solution for a density which is positive over $[0,1]$. However, even if we assume $\pdfxsc(x)>0, \forall x\in[0,1]$, our density $\lambda^{\sf opt}$ may have zero values for $x\in[0,1]$ such that $\beta'(x)=0$. As explained by \citet{dar2019high} for piecewise-constant approximation of a given function (without learning), the empirical difficulty is when $\beta$ is constant (and therefore $\beta'(x)=0$) over a continuous sub-interval of the input domain. They also showed that this can be empirically addressed by replacing the zero-derivative values with an arbitrarily-small constant. 
Accordingly, we empirically compute $\lambda^{\sf opt}$ from (\ref{eq:optimal expert segment density function - one-dimensional}) with replacing the near zero values $\pdfxsc(x)\left(\beta'(x)\right)^2 < \sf{eps}$ with $\sf{eps}$ where the numerical precision value $\sf{eps}$ is in the order of $10^{-16}$; this gives a numerically regularized version of (\ref{eq:optimal expert segment density function - one-dimensional}) and (\ref{eq:optimal test error - one-dimensional}).
In Fig.~\ref{fig:empirical results - example 2 - cosine beta with constant segment} we show results for a setting that significantly needs this mechanism. 

The experiment settings of Figs.~\ref{fig:empirical results - example 1 - cosine beta}-\ref{fig:empirical results - example 2 - cosine beta with constant segment} differ only in the true $\beta$ function. In Fig.~\ref{fig:empirical results - example 1 - cosine beta}, $\beta(x)=\cos(10\pi x)$ for $x\in[0,1]$. In Fig.~\ref{fig:empirical results - example 2 - cosine beta with constant segment}, $\beta(x)=\cos(10\pi x)$ for $x\in[0,0.4]\cup[0.6,1]$ and $\beta(x)=1$ for $x\in(0.4,0.6)$. 
The $\beta$ function is used in the one-dimensional data model $\scrand{y}=\beta(\scrand{x})+\scrand{\epsilon}$ from Section \ref{sec:One-Dimensional Inputs}; here, $\scrand{x}$ is from a truncated Gaussian distribution with mean $0.5$ and standard deviation $0.18$ (before truncation); the noise $\scrand{\epsilon}$ is a uniform random variable from the continuous range $[-0.1,0.1]$.
Both experiments demonstrate that the optimal segment density is higher for inputs whose product of the input probability density function $\pdfxsc$ and the squared derivative of $\beta$ is relatively high; see Figs.~\ref{fig:example 1 - optimal segment density} and \ref{fig:example 2 - optimal segment density}, which empirically compute $\lambda^{\sf opt}$ that was provided in Eq.~(\ref{eq:optimal expert segment density function - one-dimensional}) of Theorem \ref{theorem: optimal expert segment density function and test error - one-dimensional}. 

Figs.~\ref{fig:example 1 - beta reconstruction and segmentation m=18}-\ref{fig:example 1 - beta reconstruction and segmentation m=120}, \ref{fig:example 2 - beta reconstruction and segmentation m=18}-\ref{fig:example 2 - beta reconstruction and segmentation m=120} show the best predictors in the MoE hypothesis class $\mathcal{H}_{m,1}$, where both the segmentation and the expert constant are model parameters, as defined in (\ref{eq:hypothesis class of d-dimensional MoE - definition}); the magenta lines show the best in-class predictor, $\widehat{\beta}^{\sf opt}$ from (\ref{eq:best in-class predictor that minimizes test error}) under our high-rate assumptions. The red markers on the $x$-axis show the points $\{a_i\}_{i=1}^m$ of the optimal segmentation that was formed from the optimal density $\lambda^{\sf opt}$. It can be observed that relatively smaller segments are assigned for when $\lambda^{\sf opt}$ is higher (due to a relatively higher product of the input probability density function $\pdfxsc$ and the squared derivative of $\beta$). 

Figs.~\ref{fig:example 1 - approximation error curve vs m}, \ref{fig:example 2 - approximation error curve vs m} show the approximation error curves for a varying number of experts $m$. The empirical curve, in solid blue line, was computed as the average squared error of the best in-class predictor (for each $m$) for a dataset of 5000 test examples (note that here, in the approximation error evaluation, there is no learning from training dataset). 
The theoretical curve, in dotted red line, was numerically computed by (\ref{eq:optimal test error - one-dimensional}) of Theorem \ref{theorem: optimal expert segment density function and test error - one-dimensional}. The evaluations show excellent match between the empirical evaluation and the theoretical formula; significant differences between the two are observed in Fig.~\ref{fig:example 1 - approximation error curve vs m} for a small number of experts ($m<14$) but this is reasonable considering the theoretical assumption of having many experts (i.e., sufficiently large $m$).

\section{Approximation Error of MoE for Multidimensional Inputs}
\label{sec:Multi-Dimensional Inputs}
We now return to the general case of multidimensional inputs, as presented in Section \ref{sec:Problem and Model Definition}. 
While some of the detailed analysis naturally extends the one-dimensional case of Section \ref{sec:One-Dimensional Inputs}, the multidimensional case has important differences that affect the test error formulation and segmentation optimization. 

\subsection{Upper Bound of the Approximation Error}
\label{subsec:Multi-Dimensional Inputs - Upper Bound of the Approximation Error}
Each region has a volume $V(A_i) \triangleq \int\displaylimits_{\vec{x}\in A_i} \dint\vec{x}$, center $\vec{x}_i \triangleq \frac{1}{V(A_i)}\int\displaylimits_{\vec{x}\in A_i} \vec{x} \dint\vec{x}$, and shape that is reflected by the normalized second-moment of inertia about the center $\vec{x}_i$: 
\begin{equation}
\label{eq:normalized second-moment of inertia - definition}
    M(A_i) \triangleq \frac{\int\displaylimits_{\vec{x}\in A_i}\Ltwonormsquared{\vec{x}-\vec{x}_i}\dint\vec{x}}{d\cdot V(A_i)^{1+\frac{2}{d}}}
\end{equation}
where the normalization makes $M(A_i)$ invariant to proportional scaling of a given shape of $A_i$.
The multidimensional high-rate analysis concerns sequences of segmentations $\Gamma_{d,m}=\left\{A_{i}^{(m)}\right\}_{i=1}^m$ of $[0,1]^d$ that satisfy the next two assumptions. The superscript of $A_{i}^{(m)}$ will be often omitted to ease notation.
\begin{assumption}
\label{assumption:finite moment of inertia}
    The normalized second-moment of inertia is uniformly bounded, i.e., 
    \begin{equation}
        \sup\limits_{m,i} M\left(A_i^{(m)}\right) < \infty.
    \end{equation}
\end{assumption}
Moreover, we assume a diameter shrinkage behavior for the asymptotically small regions. The diameter of a region $A_i$ is ${\rm diam}\!\left(A_i\right)\triangleq\sup\limits_{\vec{x},\widetilde{\vec{x}}\in A_i}\Ltwonorm{\vec{x}-\widetilde{\vec{x}}}$.
\begin{assumption}
\label{assumption: vanishing region diameter}
    The region diameter uniformly shrinks over all regions: 
    \begin{equation}
        \max\limits_{i\in\{1,\dots,m\}}{\rm diam}\!\left(A_{i}^{(m)}\right)\rightarrow0~~\text{as}~m\rightarrow\infty.
    \end{equation}    
\end{assumption}

We assume that there are many experts such that their $d$-dimensional regions $\{A_i\}_{i=1}^m$ are sufficiently small to allow locally-linear approximations of $\beta$ and $\pdfxvec$. Specifically, the first-order Taylor approximations of $\beta$ and $\pdfxvec$ around the $\ith$ region center $\vec{x}_i$ are 
\begin{align}
    \label{eq:first-order Taylor approximation of beta - multidimensional}
    \beta(\vec{x}) &= \beta(\vec{x}_i) +  \nabla\beta(\vec{x}_i)^T\left(\vec{x}-\vec{x}_i\right) + R_{\beta,1}(\vec{x})
    \\
    \label{eq:first-order Taylor approximation of p - multidimensional}
    \pdfxvec(\vec{x}) &= \pdfxvec(\vec{x}_i) +  \nabla\pdfxvec(\vec{x}_i)^T \left(\vec{x}-\vec{x}_i\right) + R_{\pdfxvec,1}(\vec{x})
\end{align}
where $\nabla\beta(\vec{x}_i)$ and $\nabla\pdfxvec(\vec{x}_i)$ are the gradients of $\beta$ and $\pdfxvec$, respectively, at the region center $\vec{x}_i$;  $R_{\beta,1}\in o\left(\Ltwonorm{\vec{x}-\vec{x}_i}\right)$ and $R_{\pdfxvec,1}\in o\left(\Ltwonorm{\vec{x}-\vec{x}_i}\right)$ are remainder terms.

Using the locally-linear approximations (\ref{eq:first-order Taylor approximation of beta - multidimensional})-(\ref{eq:first-order Taylor approximation of p - multidimensional}), the optimal expert constants (\ref{eq:optimal c_i - multidimensional}) can be formulated and their test error (\ref{eq:test error - MoE - multidimensional}) can be formulated and upper bounded as follows (see proof in Appendix \ref{appendix:subsec:proof of the optimal constants and test error in sum form - multidimensional}).
\begin{theorem}
\label{theorem:optimal constants and test error in sum form - multidimensional}
    Consider the regularity conditions from Section \ref{sec:Problem and Model Definition} for $\beta$, $\pdfxvec$ and the segmentation. Under Assumptions \ref{assumption:finite moment of inertia},\ref{assumption: vanishing region diameter}, for a given segmentation of $[0,1]^d$ by $m$ regions $\{A_i\}_{i=1}^m$, the optimal expert constants are 
    \begin{equation}
    \label{eq:theorem - optimal constant - multidimensional}
        c_i^{\sf opt} = \beta(\vec{x}_i) + o\left(V(A_i)^{1/d}\right)~~\text{as}~m\rightarrow \infty, ~~ \forall i\in\{1,\dots,m\}.
    \end{equation}
    Note that $m\rightarrow \infty$ implies $V(A_i)\rightarrow 0, \forall i\in\{1,\dots,m\}$.
    The corresponding test error can be upper bounded as 
    \begin{align}
        \label{eq:theorem - test error sum formula - equality - multidimensional}
        \mathcal{E}_{\sf test} \left({\{A_i\}_{i=1}^m, \{c_i^{\sf opt}\}_{i=1}^m}\right) = \sigma_{\epsilon}^2 + \sum_{i=1}^{m}{\left(\pdfxvec(\vec{x}_i)\nabla\beta(\vec{x}_i)^T\mtx{J}_i\nabla\beta(\vec{x}_i) \right) } + o\left({\sum_{i=1}^{m}V(A_i)^{1+\frac{2}{d}}}\right) \\ 
        \label{eq:theorem - test error sum formula - multidimensional}
        \le \sigma_{\epsilon}^2 + d\sum_{i=1}^{m}{\left( \Ltwonormsquared{\nabla\beta(\vec{x}_i)} \pdfxvec(\vec{x}_i) M(A_i) V(A_i)^{1+\frac{2}{d}} \right) }+ o\left({\sum_{i=1}^{m}V(A_i)^{1+\frac{2}{d}}}\right)
    \end{align}
    where $M(A_i)$ is the normalized second-moment of inertia of $A_i$, and $\mtx{J}_i\triangleq\int\displaylimits_{\vec{x}\in A_i}{\left(\vec{x}-\vec{x}_i\right)\left(\vec{x}-\vec{x}_i\right)^T\dint\vec{x}}$ is a matrix integral. 
\end{theorem}
The test error on the left side in (\ref{eq:theorem - test error sum formula - equality - multidimensional}) is the approximation error $\mathcal{E}_{\sf app}$ of the hypothesis class $\mathcal{H}_{m,d}^{c}\left(\{A_i\}_{i=1}^m\right)$; this is due to the definitions in (\ref{eq:hypothesis class of d-dimensional MoE given segmentation- definition}), (\ref{eq:approximation error of the simpler hypothesis class given segmentation}) and the proof of (\ref{eq:theorem - test error sum formula - equality - multidimensional}) where the optimal constants are set in the test error formula \textit{before} developing the upper bound. 
Importantly, note that (\ref{eq:theorem - test error sum formula - multidimensional}) is an upper bound on the test error and not formulation with equality as in the one-dimensional input case (\ref{eq:theorem - test error sum formula - one-dimensional}); this is due to the Cauchy-Schwarz inequality applications on the inner products of gradients and centered input vectors, for more details see the proof in Appendix \ref{appendix:subsec:proof of the optimal constants and test error in sum form - multidimensional}.

Now we will define two more assumptions that refer to the limiting behavior of a sequence of segmentations $\Gamma_{d,m}=\left\{A_{i}^{(m)}\right\}_{i=1}^m$ as $m$ increases and for a fixed dimension $d$.
Motivated by high-rate vector quantization theory \citep{gersho1979asymptotically,na1995bennetts}, we define a density function for the input-space segments of experts in our multidimensional problem (extending Assumption \ref{assumption: expert segment density function - one-dimensional} for one-dimensional inputs). 
\begin{assumption}
\label{assumption: expert segment density function - multidimensional}
    There exists an expert segment density function $\lambda:[0,1]^d\rightarrow(0,\infty)$ which is smooth and satisfies 
    \begin{equation}
        \label{eq:expert segment density function - definition - multidimensional}
        \sup_{\vec{x}\in[0,1]^d}\left\lvert \lambda_m(\vec{x}) - \lambda(\vec{x}) \right\rvert \rightarrow 0 ~~~\text{as}~~m\rightarrow \infty
    \end{equation}
    where $\lambda_m(\vec{x}) =  \frac{1}{mV\left(A_{i}^{(m)}\right)}$ for $\vec{x}\in A_{i}^{(m)}$, $\forall i\in\{1,\dots,m\}$.
    Therefore, for a sufficiently large $m$, at the region centers, $\lambda(\vec{x}_i) \approx \frac{1}{mV\left(A_{i}^{(m)}\right)}$, $\forall i\in\{1,\dots,m\}$.
\end{assumption}

The multidimensional case requires to consider the segmentation region shapes, and specifically their normalized moment of inertia $\{M(A_i)\}_{i=1}^{m}$ that appear in (\ref{eq:theorem - test error sum formula - multidimensional}).
Motivated by the vector quantization analysis by \cite{na1995bennetts}, the following assumption will be analytically useful.
\begin{assumption}
\label{assumption: smooth normalized moment of inertia - multidimensional}
    There exists a smooth function $\mu:[0,1]^d\rightarrow(0,\infty)$, known as the normalized moment of inertia profile, that satisfies 
    \begin{equation}
    \label{eq:normalized moment of inertia profile for x}
        \sup_{\vec{x}\in[0,1]^d}\left\lvert \mu_m(\vec{x}) - \mu(\vec{x}) \right\rvert \rightarrow 0 ~~~\text{as}~~m\rightarrow \infty
    \end{equation}
    where $\mu_m(\vec{x}) =  M\left(A_{i}^{(m)}\right)$ for $\vec{x}\in A_{i}^{(m)}$, $\forall i\in\{1,\dots,m\}$.
    Therefore, for a sufficiently large $m$, at the region centers, $\mu(\vec{x}_i) \approx M\left(A_{i}^{(m)}\right)$, $\forall i\in\{1,\dots,m\}$.
\end{assumption}

Using Assumptions \ref{assumption: expert segment density function - multidimensional}-\ref{assumption: smooth normalized moment of inertia - multidimensional} and approximating the sum in (\ref{eq:theorem - test error sum formula - multidimensional}) as integral due to the small segment assumption, we get an approximated formula for the error upper-bound as corollary of Theorem \ref{theorem:optimal constants and test error in sum form - multidimensional}. 
\begin{corollary}
\label{corollary:test error integral formula - multidimensional}
Under Assumptions \ref{assumption:finite moment of inertia}-\ref{assumption: smooth normalized moment of inertia - multidimensional}, the upper bound on the test error for optimal constants can be approximated using continuous integration: $\mathcal{E}_{\sf test}\le \widetilde{\mathcal{E}}_{\sf test}^{\sf bound} + o\left(m^{-2/d}\right)$ as $m\rightarrow\infty$ where 
       \begin{equation}
        \label{eq:corollary - test error integral formula - multidimensional}
        \widetilde{\mathcal{E}}_{\sf test}^{\sf bound} = \sigma_{\epsilon}^2 + \frac{d}{m^{2/d}}\int_{\vec{x}\in[0,1]^d}{ \frac{\Ltwonormsquared{\nabla\beta(\vec{x})} \pdfxvec(\vec{x}) \mu(\vec{x}) }{\lambda^{2/d}(\vec{x})} \dint \vec{x}}
    \end{equation}
    where $\lambda$ and $\mu$ are as in Assumptions \ref{assumption: expert segment density function - multidimensional} and \ref{assumption: smooth normalized moment of inertia - multidimensional}, respectively. We refer to $\widetilde{\mathcal{E}}_{\sf test}^{\sf bound}$ as the asymptotic upper bound formula of the test error, which is also the approximation error associated with the given segmentation.
\end{corollary}
More details on (\ref{eq:corollary - test error integral formula - multidimensional}) are provided in Appendix \ref{appendix:subsec:Additional Details on Corollary - test error integral formula - multidimensional}.

Recall that the volume of the $d$-dimensional unit cube is 1, therefore cases where the value of  $\frac{\Ltwonormsquared{\nabla\beta(\vec{x})} \pdfxvec(\vec{x}) \mu(\vec{x}) } {\lambda^{2/d}(\vec{x})}$ for $\vec{x}\in[0,1]^d$ can be upper bounded by a constant agnostic of $m$ and $d$ motivate the following assumption. 
\begin{assumption}
    \label{assumption:upper bound of integration factor is agnostic of m and d}
    $\int_{\vec{x}\in[0,1]^d}{ \frac{\Ltwonormsquared{\nabla\beta(\vec{x})} \pdfxvec(\vec{x}) \mu(\vec{x}) }{\lambda^{2/d}(\vec{x})} \dint \vec{x}}$ can be upper bounded by a constant agnostic of $m$ and $d$.
\end{assumption}
Then, Corollary \ref{corollary:test error integral formula - multidimensional} implies the following big-O bound on the approximation error. 
\begin{corollary}
\label{corollary:test error big-O bound - multidimensional}
Under Assumptions \ref{assumption:finite moment of inertia}-\ref{assumption:upper bound of integration factor is agnostic of m and d}, the approximation error satisfies  
\begin{equation}
\label{eq:test error big-O bound - formula - multidimensional}
\mathcal{E}_{\sf app} \left(\mathcal{H}_{m,d}^{c}\left(\{A_i\}_{i=1}^m\right)\right) = \sigma_{\epsilon}^2 + \mathcal{O}\left(\frac{d}{m^{2/d}}\right)
\end{equation}
where $\{A_i\}_{i=1}^m$ is a discrete segmentation that theoretically corresponds to a given segment density $\lambda(\vec{x})$ for all $m$. 
Since $\mathcal{H}_{m,d}^{c}\left(\{A_i\}_{i=1}^m\right)\subseteq\mathcal{H}_{m,d}$, the same upper bound of (\ref{eq:test error big-O bound - formula - multidimensional}) also holds for the approximation error of $\mathcal{H}_{m,d}$.
\end{corollary}

\subsection{The Expert Segment Density that Minimizes the Asymptotic Upper Bound of the Test Error}
\label{subsec:Multi-Dimensional Inputs - The Expert Segment Density that Minimizes the Test-Error Upper Bound}
In this subsection, we continue to analyze the minimization of the upper bound by the segmentation --- although the bound of Corollary \ref{corollary:test error big-O bound - multidimensional} is sufficient for analyzing the statistical learning with $\mathcal{H}_{m,d}^{c}\left(\{A_i\}_{i=1}^m\right)$ in Section \ref{sec:Learning the Constant Experts}. 

We specialize the expression in (\ref{eq:corollary - test error integral formula - multidimensional}) to sequences of segmentations with a constant limiting normalized second-moment profile. This additional geometric assumption is motivated by a conjecture for high-rate vector quantization by \citet{gersho1979asymptotically}; see Section \ref{sec:Related Work - High-Rate Vector Quantization Theory}, Appendix \ref{appendix:sec:Related Work: Gersho Conjecture for High-Rate Vector Quantization} and \citep{gray1998quantization} for more details. 
\begin{assumption}
    \label{assumption: moment of inertia is approximately as for the optimal tessellating polytope}
    For a fixed dimension $d$, the considered sequence of segmentations $\Gamma_{d,m}=\left\{A_{i}^{(m)}\right\}_{i=1}^m$ satisfies 
    \begin{equation}
        \label{eq: moment of inertia has a constant limit}
        \max_{i\in\{1,\dots,m\}}\left\lvert M\left(A_{i}^{(m)}\right)-\mu_d\right\rvert \rightarrow0~~~\text{as}~~m\rightarrow\infty
    \end{equation}
    for some constant $\mu_d>0$. Hence, the limiting moment profile is 
    \begin{equation}
        \label{eq: moment of inertia is approximately as for the optimal tessellating polytope}
        \mu(\vec{x}) = \mu_d~~~\forall \vec{x}\in[0,1]^d.
    \end{equation}
\end{assumption}
The classical conjecture by \citet{gersho1979asymptotically} assumes an optimal convex polytope $A_{{\sf opt},d}$ that tessellates the $d$-dimensional space. Here, if the limiting moment is identified with that of $A_{{\sf opt},d}$ then $\mu_d=M(A_{{\sf opt},d})$; the following error bound for the multidimensional MoE, however, applies to any constant profile satisfying Assumption \ref{assumption: moment of inertia is approximately as for the optimal tessellating polytope}.
\begin{corollary}
\label{corollary:test error integral formula with fixed moment of inertia - multidimensional}
Under Assumption \ref{assumption: moment of inertia is approximately as for the optimal tessellating polytope}, the asymptotic upper bound formula in (\ref{eq:corollary - test error integral formula - multidimensional}) becomes 
       \begin{equation}
        \label{eq:corollary - test error integral formula with fixed moment of inertia - multidimensional}
        \widetilde{\mathcal{E}}_{\sf test}^{\sf bound} = \sigma_{\epsilon}^2 + \frac{d\cdot \mu_d}{m^{2/d}}\int_{\vec{x}\in[0,1]^d}{ \frac{\Ltwonormsquared{\nabla\beta(\vec{x})} \pdfxvec(\vec{x}) }{\lambda^{2/d}(\vec{x})} \dint \vec{x}}.
    \end{equation}
\end{corollary}
Then, the following theorem provides the upper-bound minimizing (UBM) expert segment density function $\lambda^{\sf UBM}$ and its corresponding error bound.  
Note that (\ref{eq:corollary - test error integral formula with fixed moment of inertia - multidimensional}) is an upper bound (after adding the remainder term $o(m^{-2/d})$) for the approximation error $\mathcal{E}_{\sf app}$ of the hypothesis class $\mathcal{H}_{m,d}^{c}\left(\{A_i\}_{i=1}^m\right)$.
\begin{theorem}
    \label{theorem: optimal expert segment density function and test error - multidimensional}
    The test error's asymptotic upper bound $\widetilde{\mathcal{E}}_{\sf test}^{\sf bound}$ from (\ref{eq:corollary - test error integral formula with fixed moment of inertia - multidimensional}) is minimized by the expert segment density function 
    \begin{equation}
        \label{eq:optimal expert segment density function - multidimensional}
        \lambda^{\sf UBM}(\vec{x})=\frac{\left({\pdfxvec(\vec{x})\Ltwonormsquared{\nabla\beta(\vec{x})}}\right)^{\frac{d}{d+2}}}{\int\displaylimits_{\vecgreek{\xi}\in[0,1]^d}{\left({\pdfxvec(\vecgreek{\xi})\Ltwonormsquared{\nabla\beta(\vecgreek{\xi})} }\right)}^{\frac{d}{d+2}}\dint\vecgreek{\xi}}
    \end{equation}
    that attains the minimal asymptotic upper bound of the test error:     
    \begin{equation}
        \label{eq:optimal test error - multidimensional}
        \widetilde{\mathcal{E}}_{\sf test}^{\sf bound, min} = \sigma_{\epsilon}^2 + \frac{d\cdot \mu_d}{m^{2/d}}\left({\, \int\displaylimits_{\vec{x}\in[0,1]^d}{\left({\pdfxvec(\vec{x})\Ltwonormsquared{\nabla\beta(\vec{x})} }\right)}^{\frac{d}{d+2}}\dint\vec{x} }\right)^{1+\frac{2}{d}}.
    \end{equation}    
\end{theorem}
The theorem proof is provided in Appendix \ref{appendix:subsec: proof of theorem on optimal expert segment density function and test error - multidimensional}. Note the following remarks for this theorem. The minimization in Theorem \ref{theorem: optimal expert segment density function and test error - multidimensional} is over all nonnegative probability densities on $[0,1]^d$, without imposing the smoothness and strict positivity requirements of Assumption \ref{assumption: expert segment density function - multidimensional}. In the minimization objective (\ref{eq:corollary - test error integral formula with fixed moment of inertia - multidimensional}), the integrand is defined as zero where both its numerator and denominator vanish, and as $\infty$ where only its denominator vanishes. The density formula (\ref{eq:optimal expert segment density function - multidimensional}) assumes that $\beta$ is nonconstant; if $\beta$ is constant, every density minimizes $\widetilde{\mathcal{E}}_{\sf test}^{\sf bound}$ from (\ref{eq:corollary - test error integral formula with fixed moment of inertia - multidimensional}) whose value is $\sigma_{\epsilon}^2$. Applying Corollary \ref{corollary:test error integral formula with fixed moment of inertia - multidimensional} with $\lambda=\lambda^{\sf UBM}$ requires a sequence of segmentations satisfying its assumptions with this limiting density and the constant moment profile $\mu_d$. Theorem \ref{theorem: optimal expert segment density function and test error - multidimensional} optimizes the formula of $\widetilde{\mathcal{E}}_{\sf test}^{\sf bound}$ from (\ref{eq:corollary - test error integral formula with fixed moment of inertia - multidimensional}) and does not establish the existence of such a minimizing sequence.

The segment density in (\ref{eq:optimal expert segment density function - multidimensional}) minimizes the asymptotic upper bound of the test error, and does not directly minimize the test error. Optimizing an objective's bound is a valid heuristic approach for when the objective optimization is intractable (e.g., as commonly done for variational autoencoders \citep{kingma2014auto}); the assumption is that minimization of the objective's upper bound would promote minimization of the true objective. However, usually the upper-bound minimizing solution does not achieve the minimum of the true objective. An important aspect of Theorem \ref{theorem: optimal expert segment density function and test error - multidimensional} is the connection to the corresponding results for high-rate vector quantization by \cite{gersho1979asymptotically,na1995bennetts}. 

Regardless of the suboptimality due to minimizing an upper bound, there is an important difference between the multidimensional and one-dimensional cases. In the one-dimensional case, it is possible to form an optimal segmentation from the optimal segment density function (when the density is strictly positive, see Appendix \ref{appendix:subsec:Scalar High-Rate Quantization - Forming a Quantizer from a Quantization Interval Density}); but in the multidimensional case, a segment density alone does not determine the region shapes and hence a completely optimized segmentation. More detailed discussions on this appear in the high-rate vector quantization literature \citep{gersho1979asymptotically,na1995bennetts,gray1998quantization}.

\subsection{Empirical Results: Two-Dimensional Segmentation guided by the UBM density and its Approximation Error}
\label{subsec:Multi-Dimensional Inputs - Empirical Results - Segmentation for MoE with Two-Dimensional Inputs}

\begin{figure}[]
    \centering
    \subfigure[]{
    \includegraphics[width=0.3\textwidth]{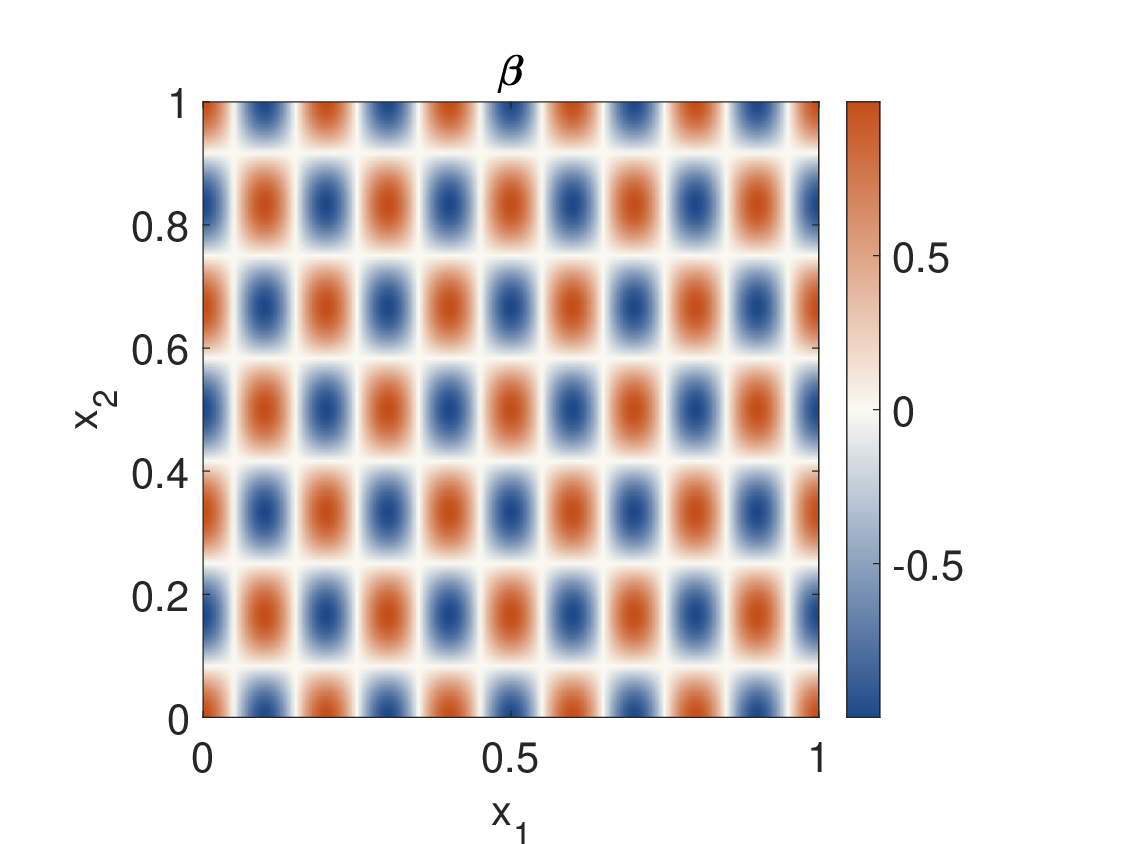}
    \label{fig:example 1 - two-dimensions - beta}
    }
    \subfigure[]{
    \includegraphics[width=0.3\textwidth]{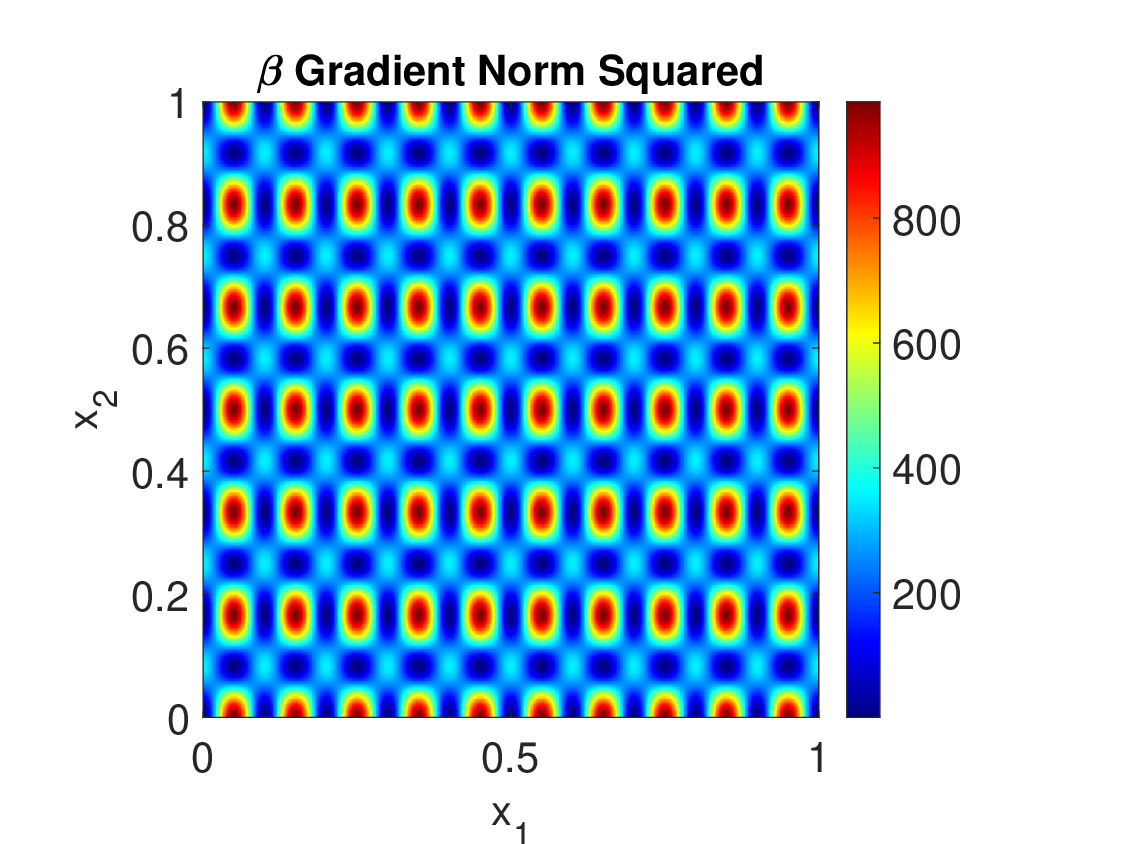}
    \label{fig:example 1 - two-dimensions - beta derivative energy}
    }
    \subfigure[]{
    \includegraphics[width=0.3\textwidth]{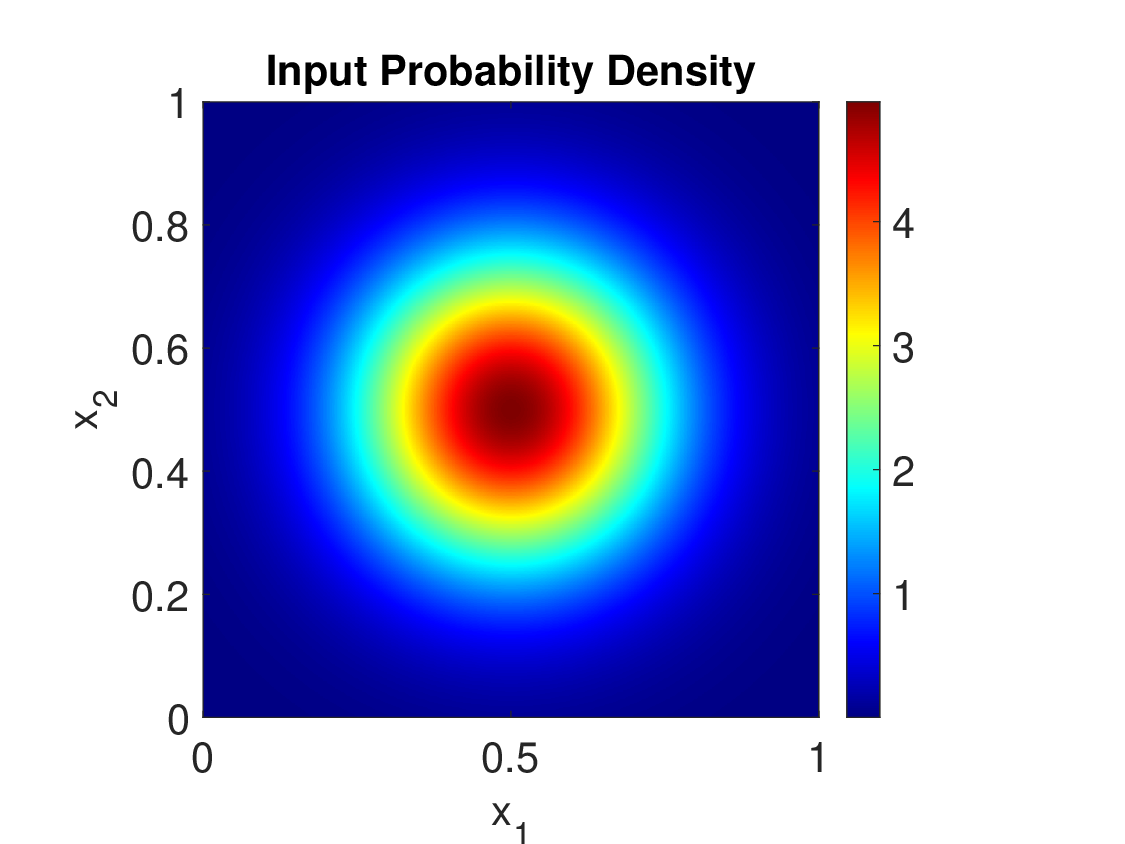}
    \label{fig:example 1 - two-dimensions - pdf x}
    }
    \\
    \subfigure[]{
    \includegraphics[width=0.3\textwidth]{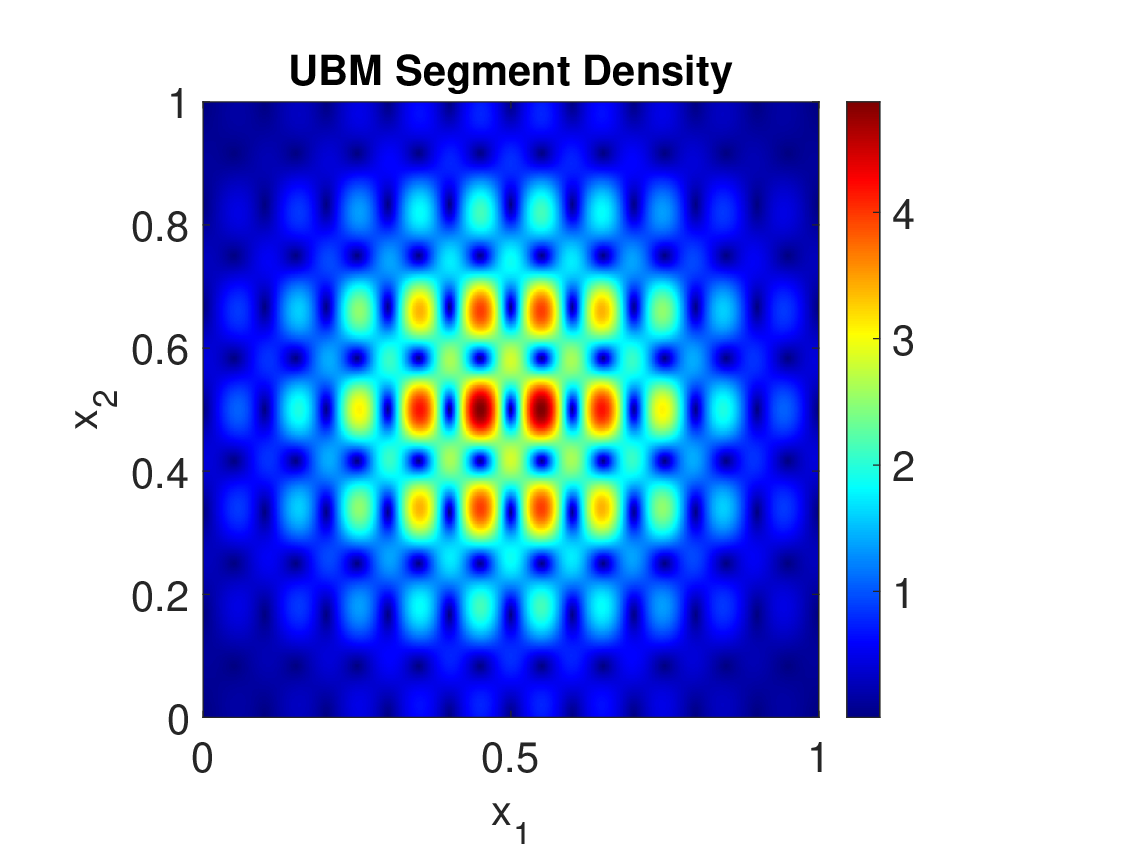}
    \label{fig:example 1 - two-dimensions - optimal segment density}
    }
    \subfigure[]{
    \includegraphics[width=0.3\textwidth]{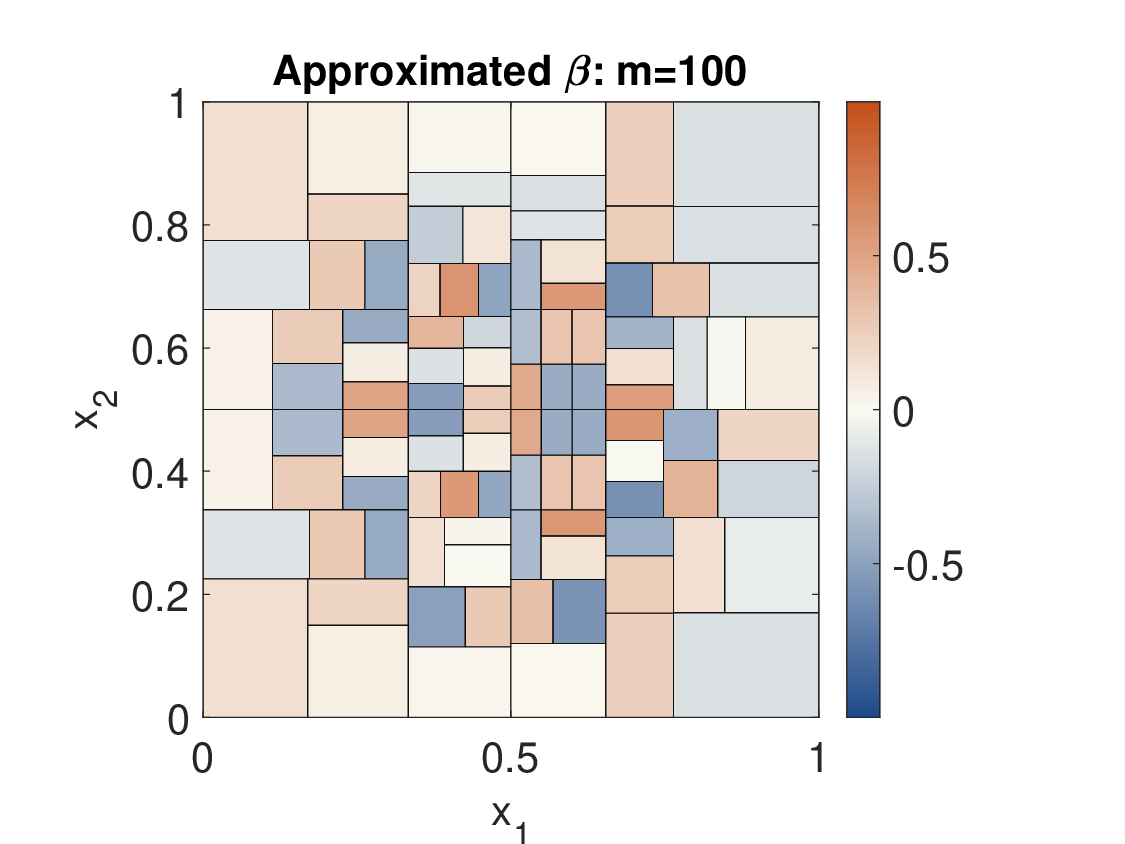}
    \label{fig:example 1 - two-dimensions - beta reconstruction and segmentation m=100}
    }
    \subfigure[]{
    \includegraphics[width=0.3\textwidth]{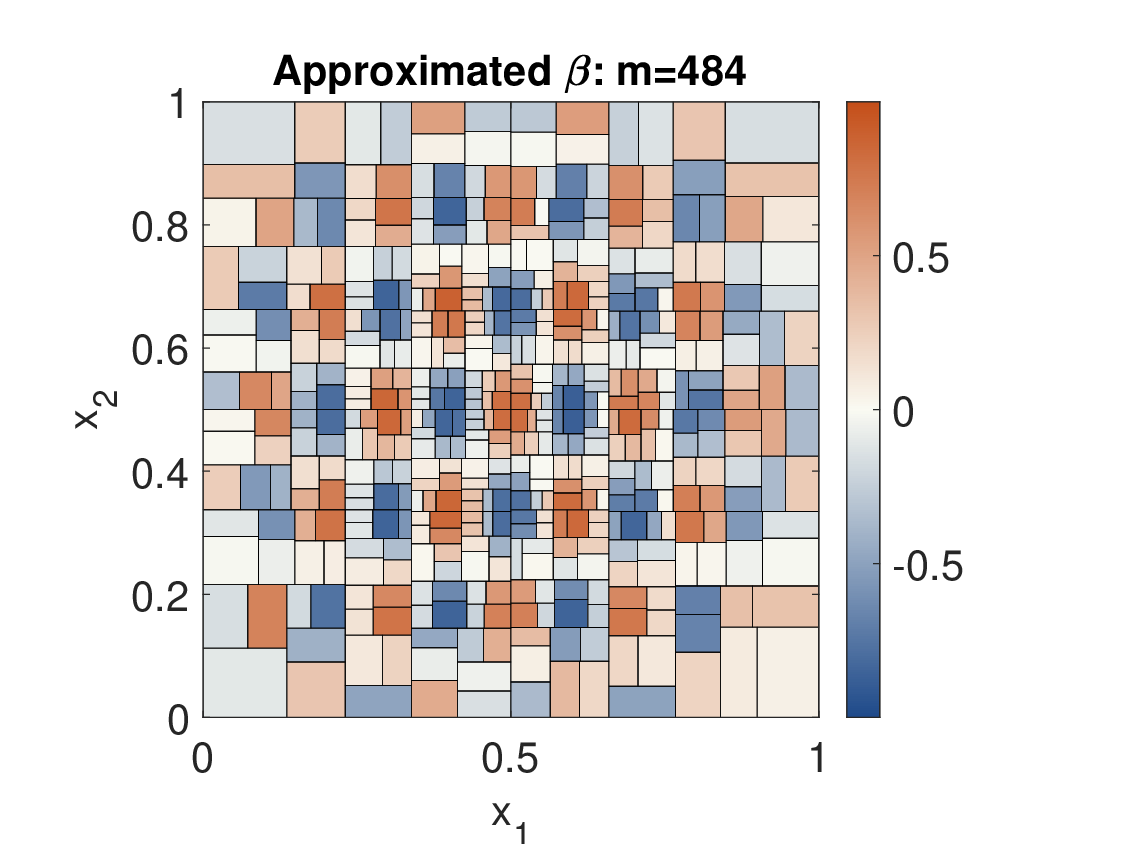}
    \label{fig:example 1 - two-dimensions - beta reconstruction and segmentation m=484}
    }
    \\
    \subfigure[]{
    \includegraphics[width=0.3\textwidth]{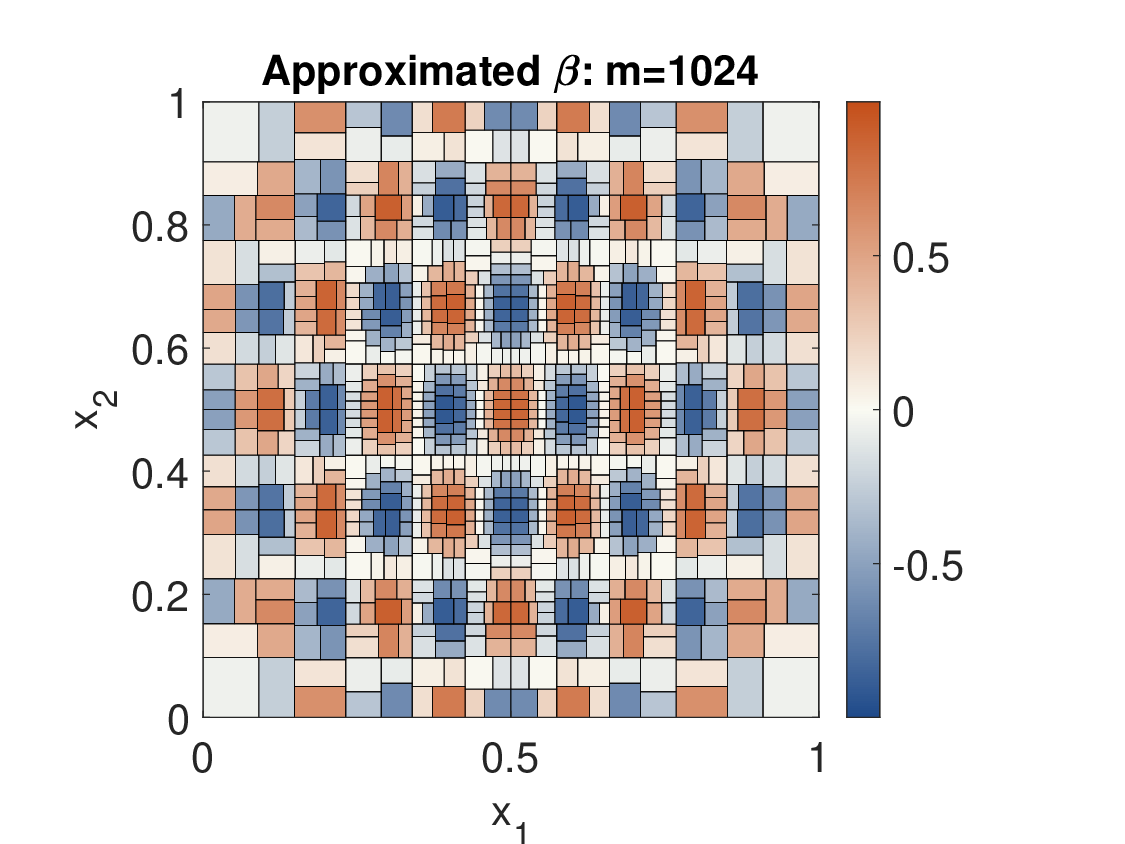}
    \label{fig:example 1 - two-dimensions - beta reconstruction and segmentation m=1024}
    }
    \subfigure[]{
    \includegraphics[width=0.3\textwidth]{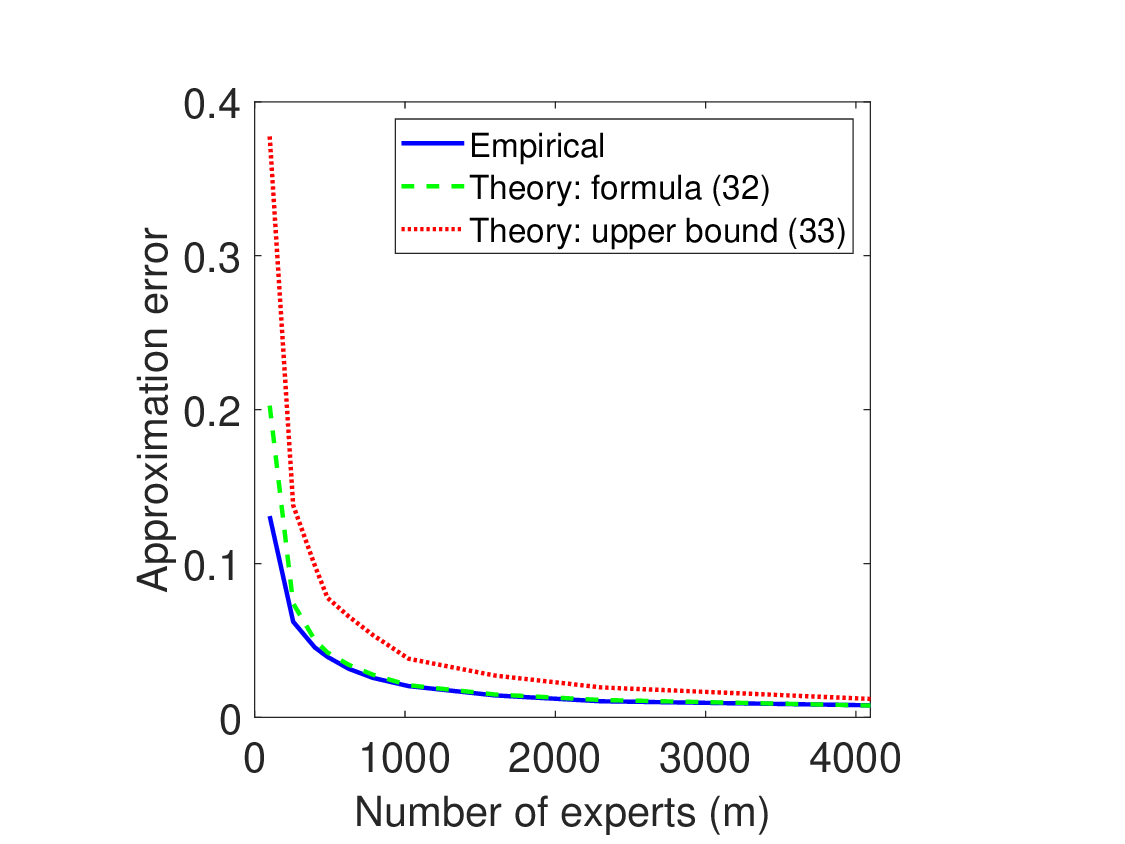}
    \label{fig:example 1 - two-dimensions - approximation error curve vs m}
    }
    \caption{Examples for UBM-density-guided segmentation for a two-dimensional input space where $\beta$ is a two-dimensional \textbf{separable cosine} and $\pdfxvec$ is a truncated isotropic Gaussian. In (\textit{e})-(\textit{g}), the black lines denote the segmentation borders of the regions $\{A_i\}_{i=1}^m$ that were formed by Algorithm \ref{algorithm:2D segmentation by bisection} using the UBM segment density in (\textit{d}).}
    \label{fig:empirical results - two-dimensions - example 1 - cosine beta}
\end{figure}

\begin{figure}[]
    \centering
    \subfigure[]{
    \includegraphics[width=0.3\textwidth]{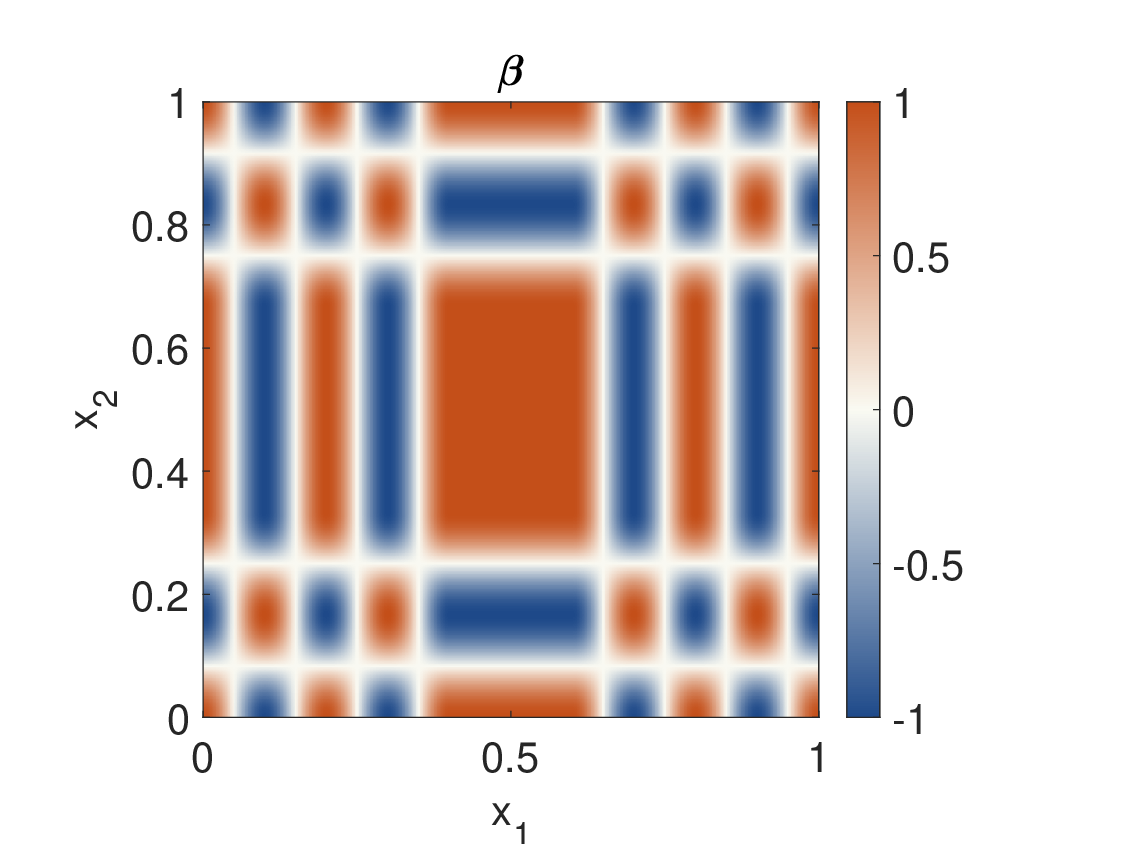}
    \label{fig:example 2 - two-dimensions - beta}
    }
    \subfigure[]{
    \includegraphics[width=0.3\textwidth]{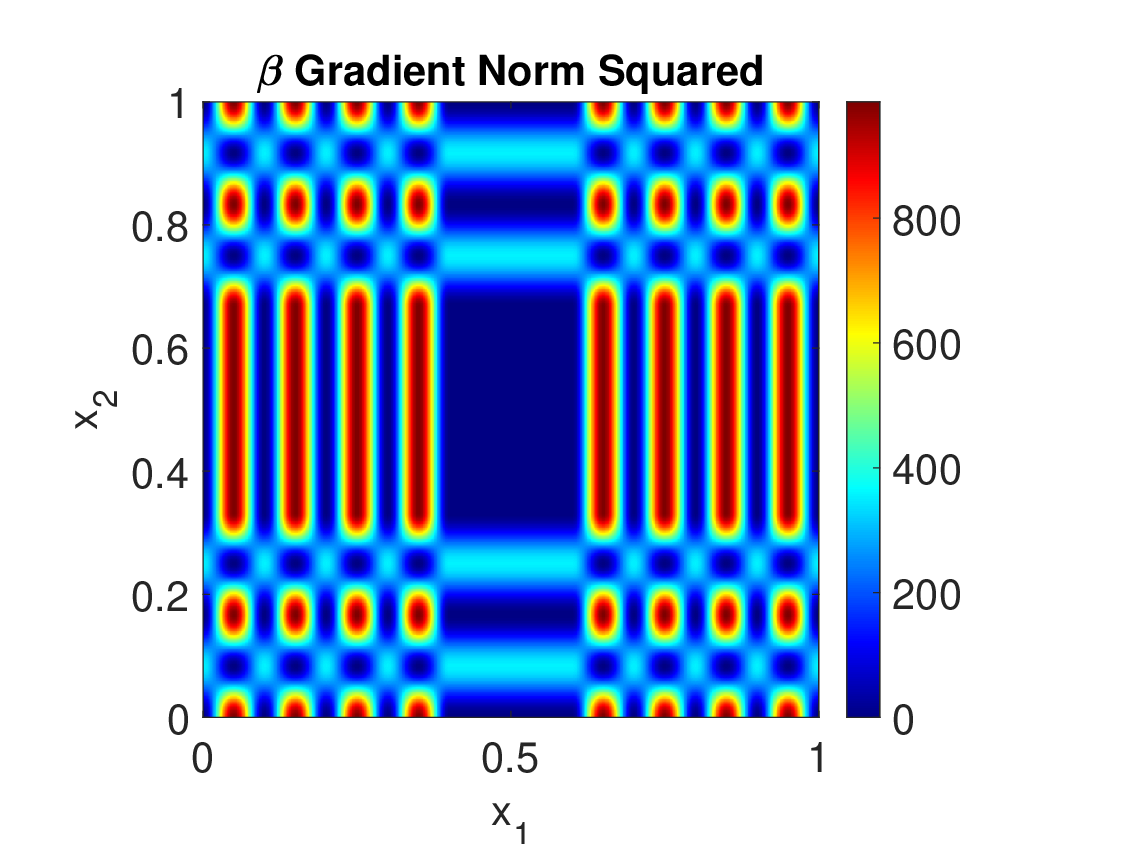}
    \label{fig:example 2 - two-dimensions - beta derivative energy}
    }
    \subfigure[]{
    \includegraphics[width=0.3\textwidth]{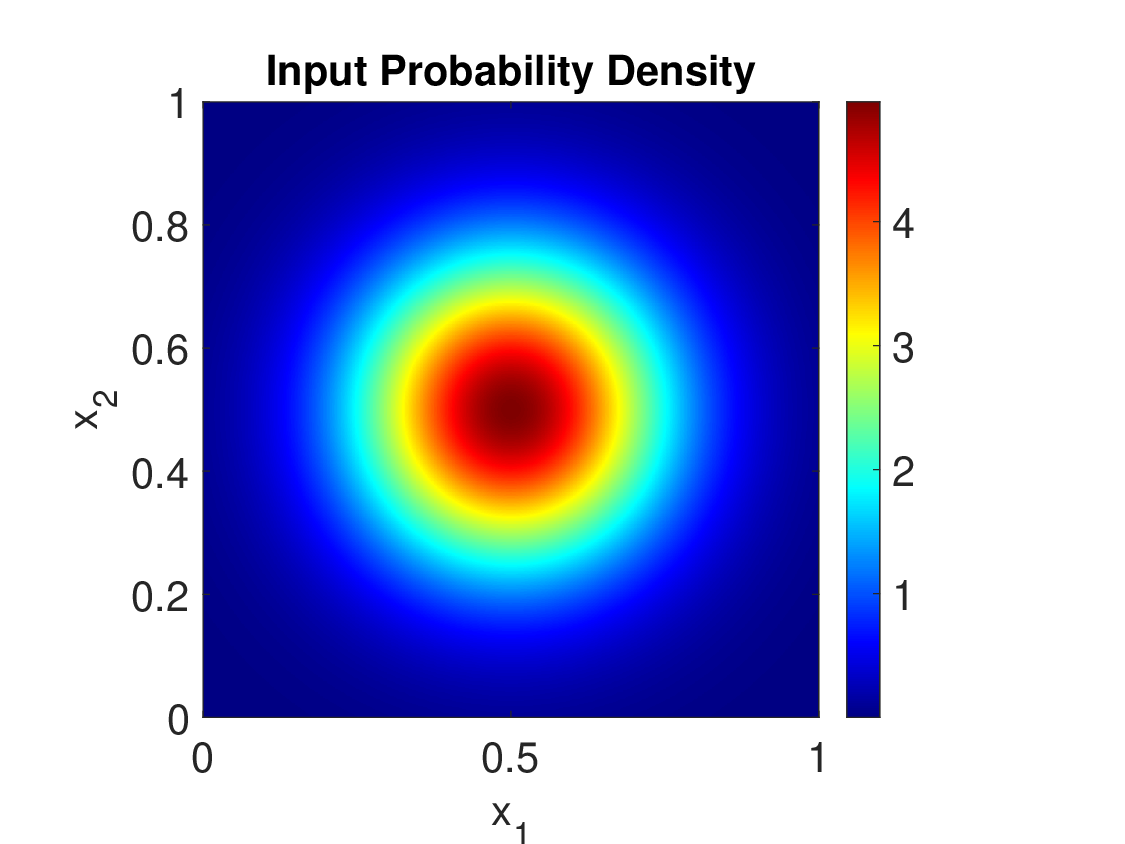}
    \label{fig:example 2 - two-dimensions - pdf x}
    }
    \\
    \subfigure[]{
    \includegraphics[width=0.3\textwidth]{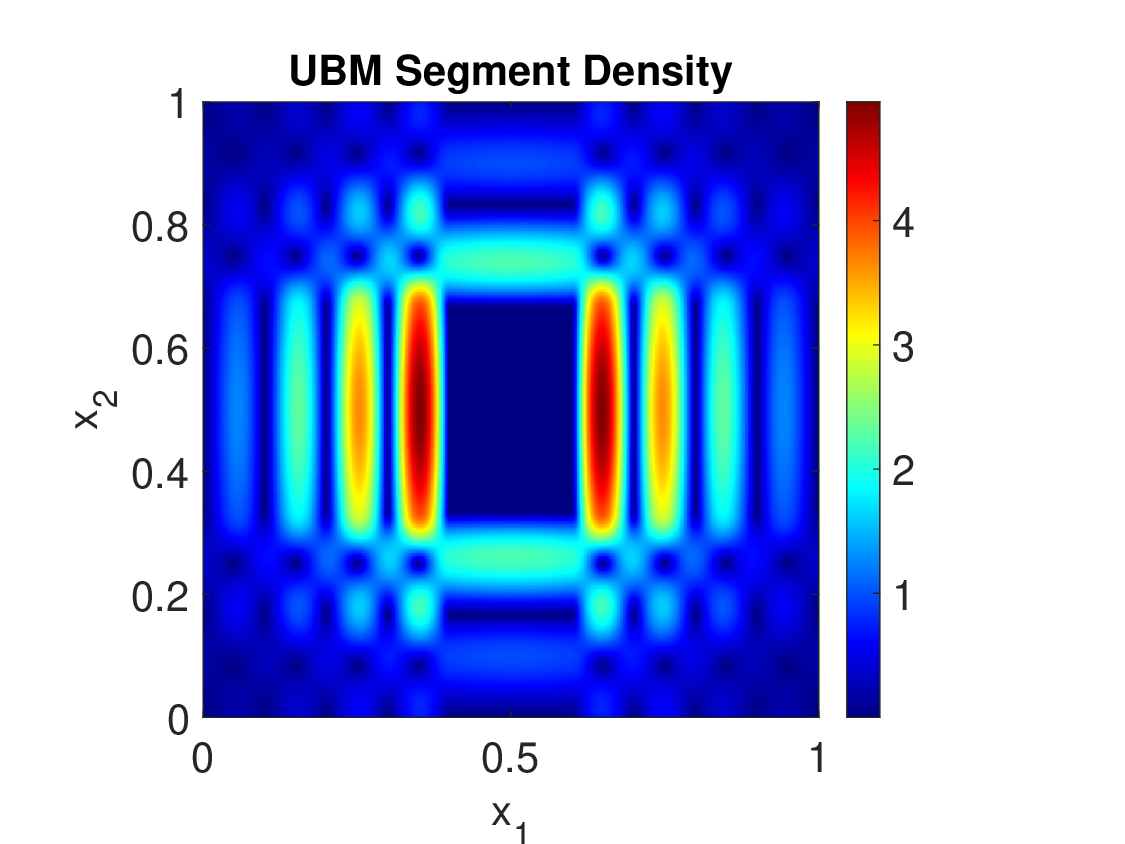}
    \label{fig:example 2 - two-dimensions - optimal segment density}
    }
    \subfigure[]{
    \includegraphics[width=0.3\textwidth]{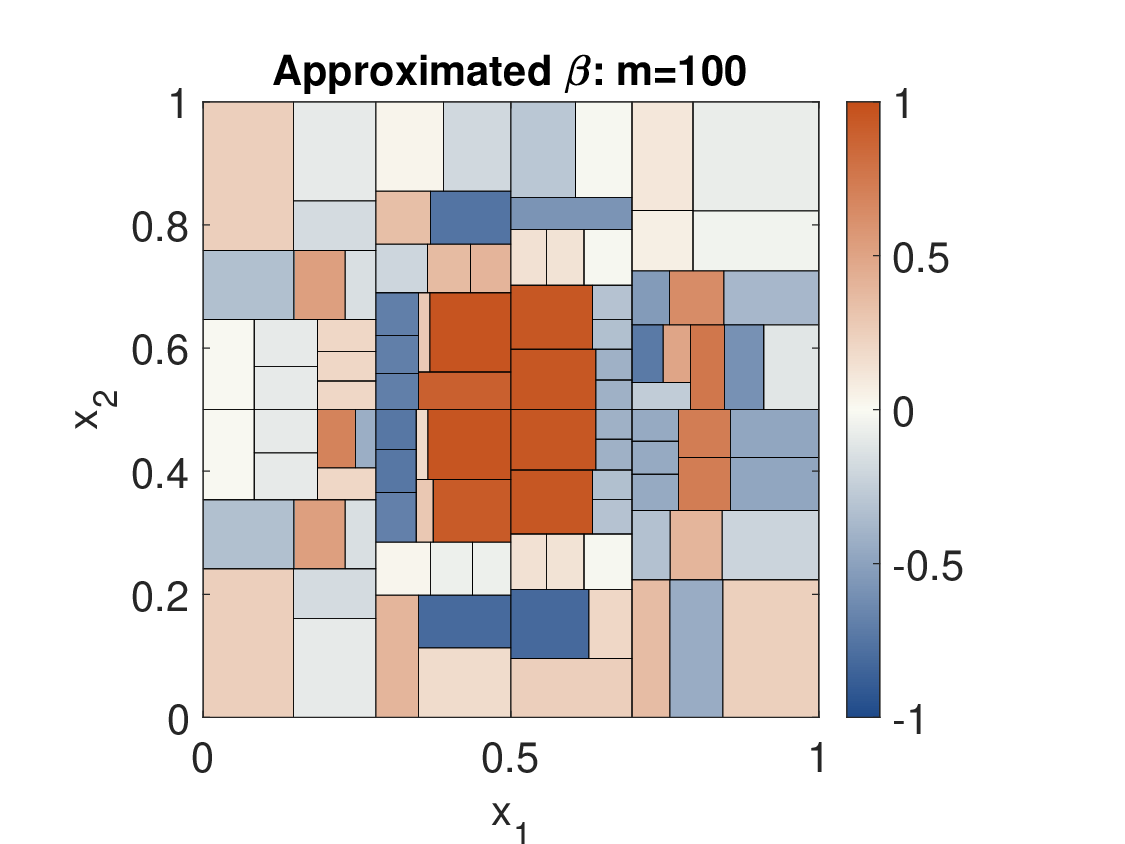}
    \label{fig:example 2 - two-dimensions - beta reconstruction and segmentation m=100}
    }
    \subfigure[]{
    \includegraphics[width=0.3\textwidth]{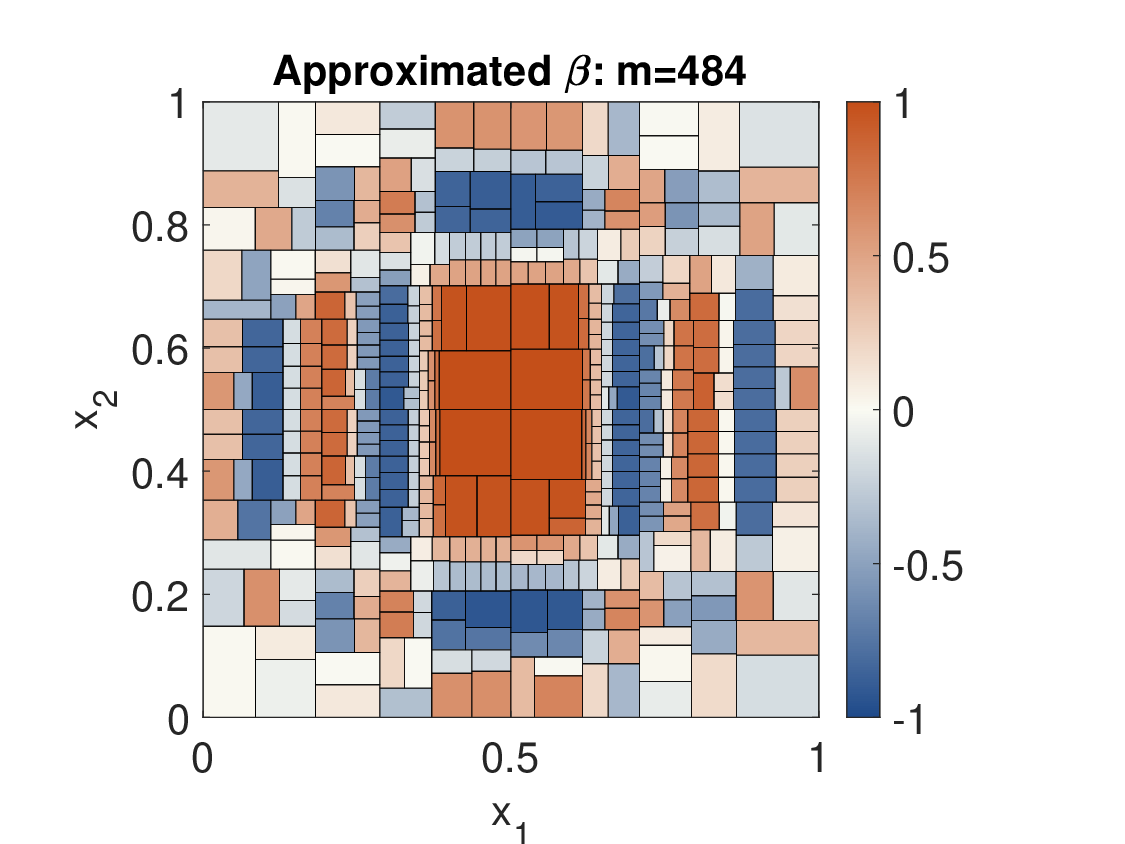}
    \label{fig:example 2 - two-dimensions - beta reconstruction and segmentation m=484}
    }
    \\
    \subfigure[]{
    \includegraphics[width=0.3\textwidth]{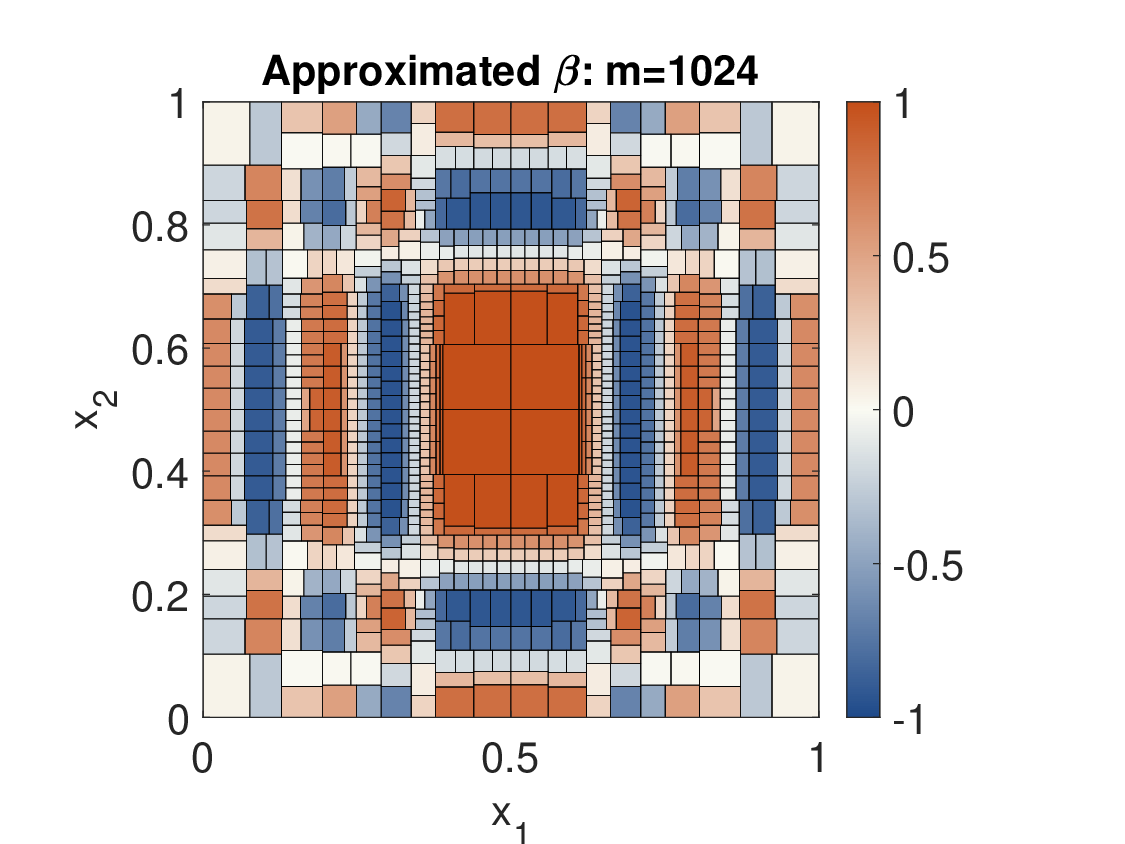}
    \label{fig:example 2 - two-dimensions - beta reconstruction and segmentation m=1024}
    }
    \subfigure[]{
    \includegraphics[width=0.3\textwidth]{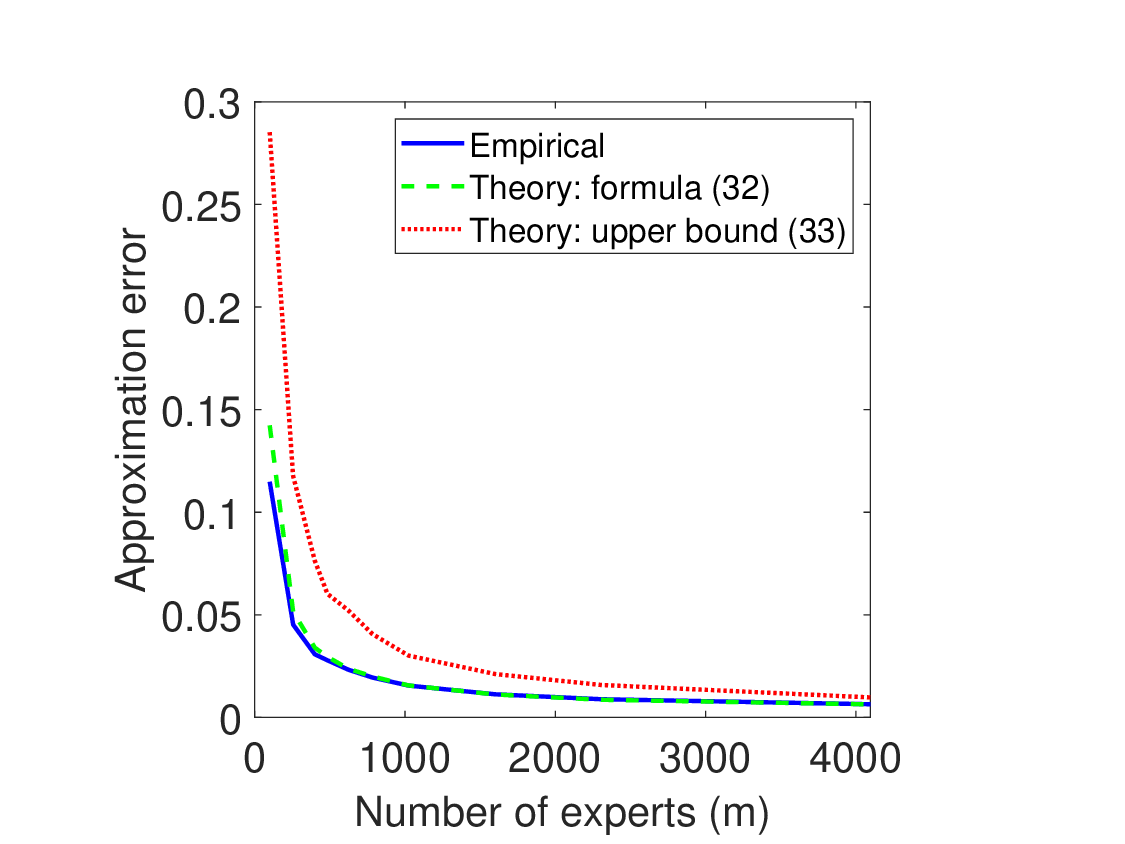}
    \label{fig:example 2 - two-dimensions - approximation error curve vs m}
    }
    \caption{Examples for UBM-density-guided segmentation for a two-dimensional input space where $\beta$ is a two-dimensional separable \textbf{cosine with a constant area} and $\pdfxvec$ is a truncated isotropic Gaussian. In (\textit{e})-(\textit{g}), the black lines denote the segmentation borders of the regions $\{A_i\}_{i=1}^m$ that were formed by Algorithm \ref{algorithm:2D segmentation by bisection} using the UBM segment density in (\textit{d}).}
    \label{fig:empirical results - two-dimensions - example 2 - cosine with constant area beta}
\end{figure}

\begin{figure}[]
    \centering
    \subfigure[]{
    \includegraphics[width=0.3\textwidth]{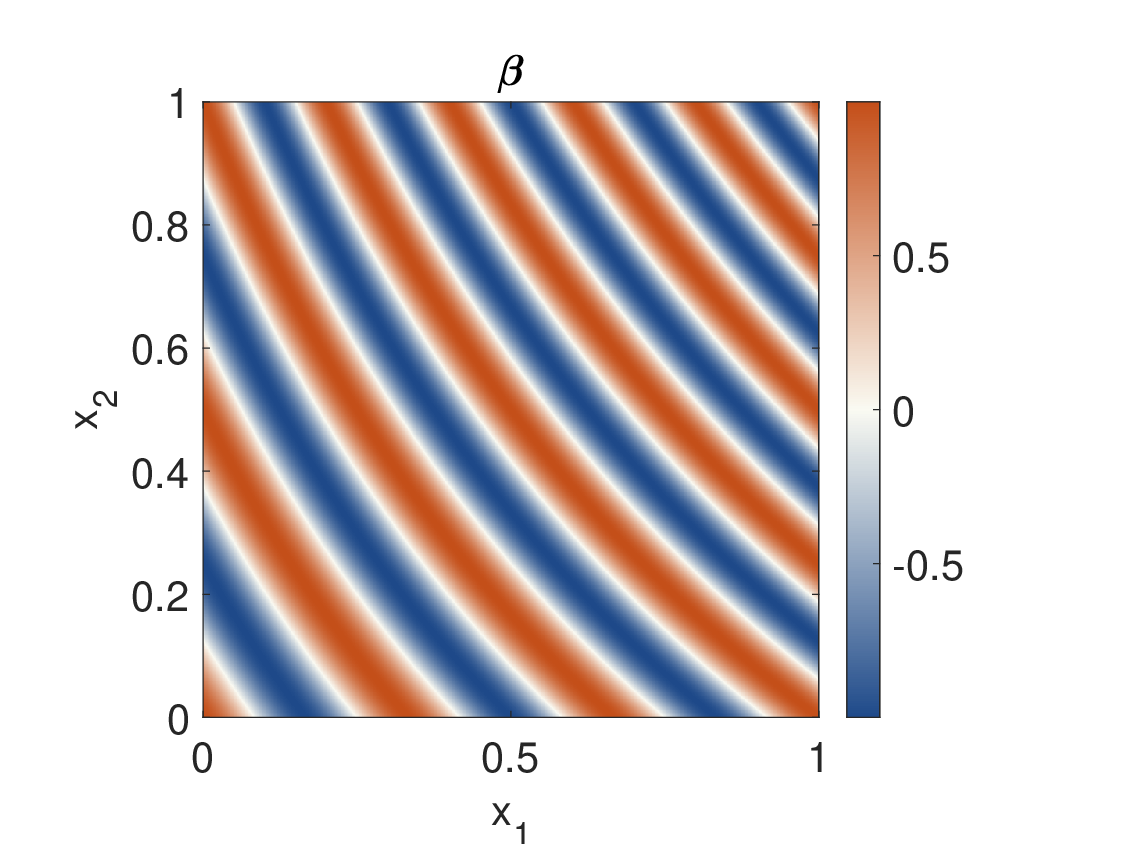}
    \label{fig:example 3 - two-dimensions - beta}
    }
    \subfigure[]{
    \includegraphics[width=0.3\textwidth]{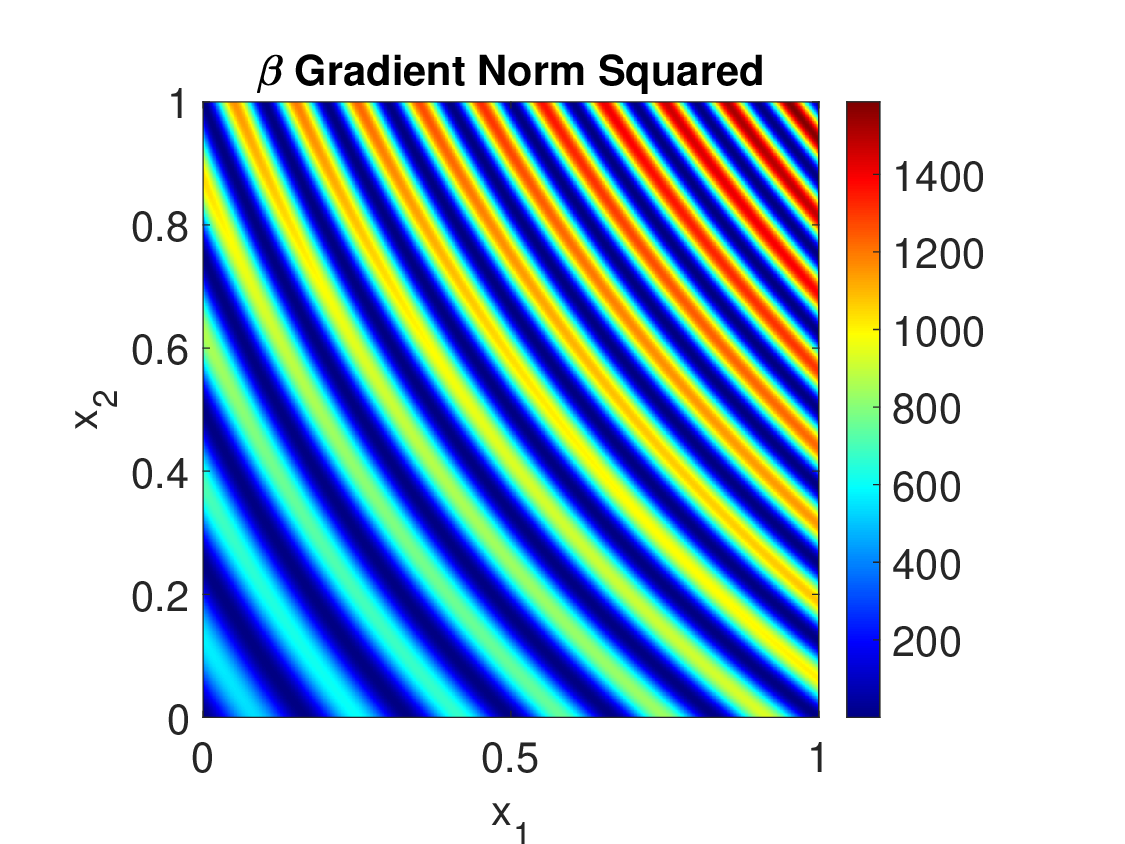}
    \label{fig:example 3 - two-dimensions - beta derivative energy}
    }
    \subfigure[]{
    \includegraphics[width=0.3\textwidth]{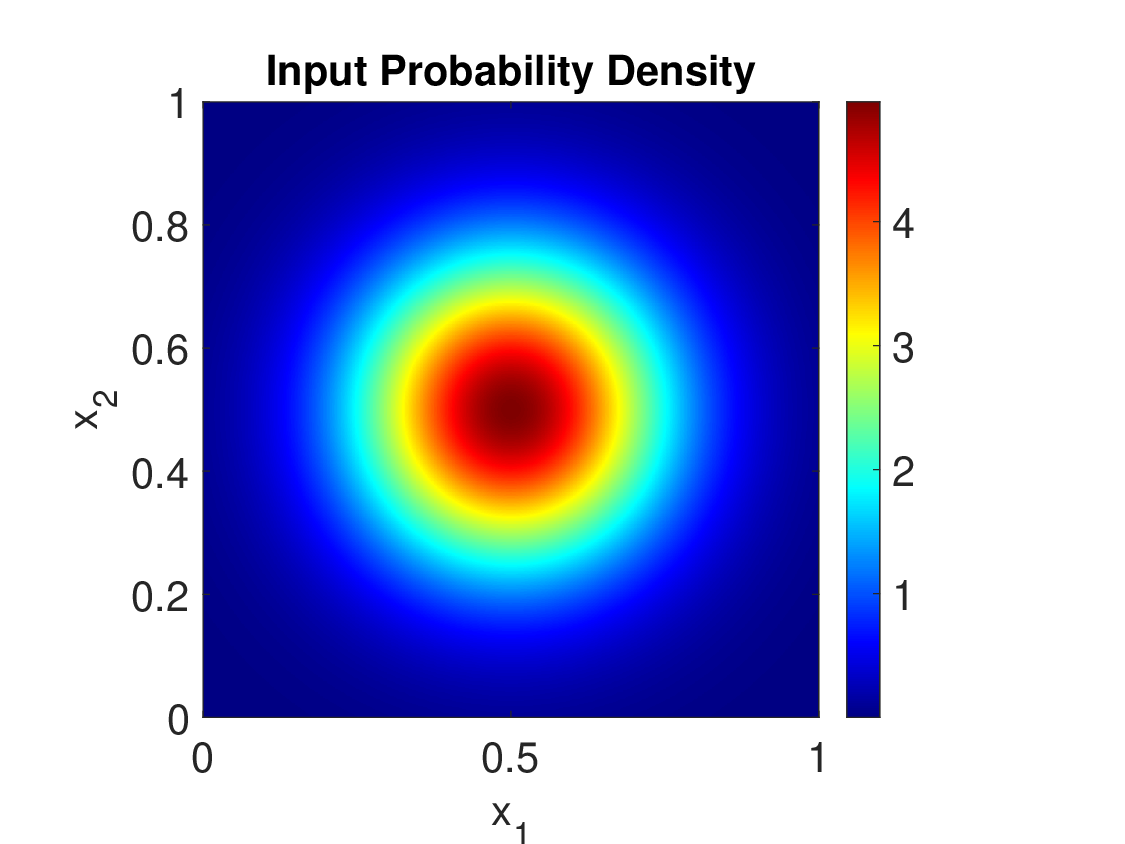}
    \label{fig:example 3 - two-dimensions - pdf x}
    }
    \\
    \subfigure[]{
    \includegraphics[width=0.3\textwidth]{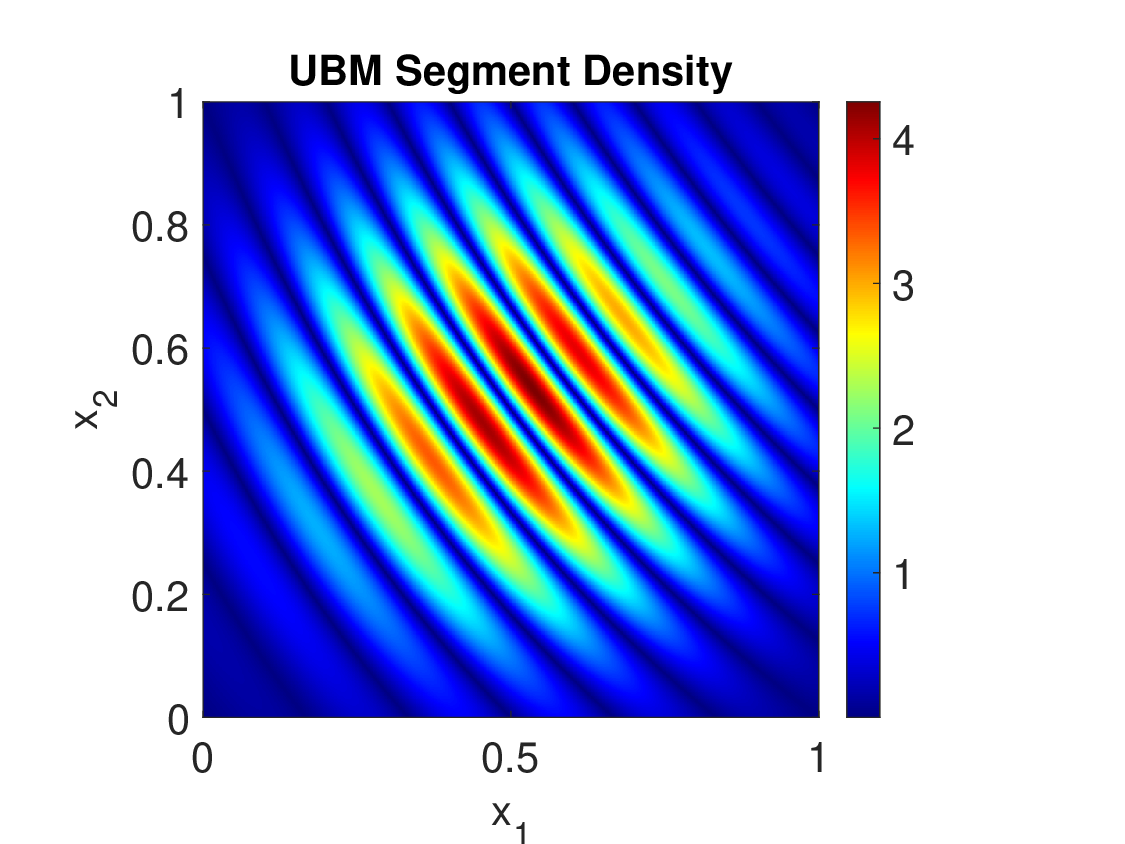}
    \label{fig:example 3 - two-dimensions - optimal segment density}
    }
    \subfigure[]{
    \includegraphics[width=0.3\textwidth]{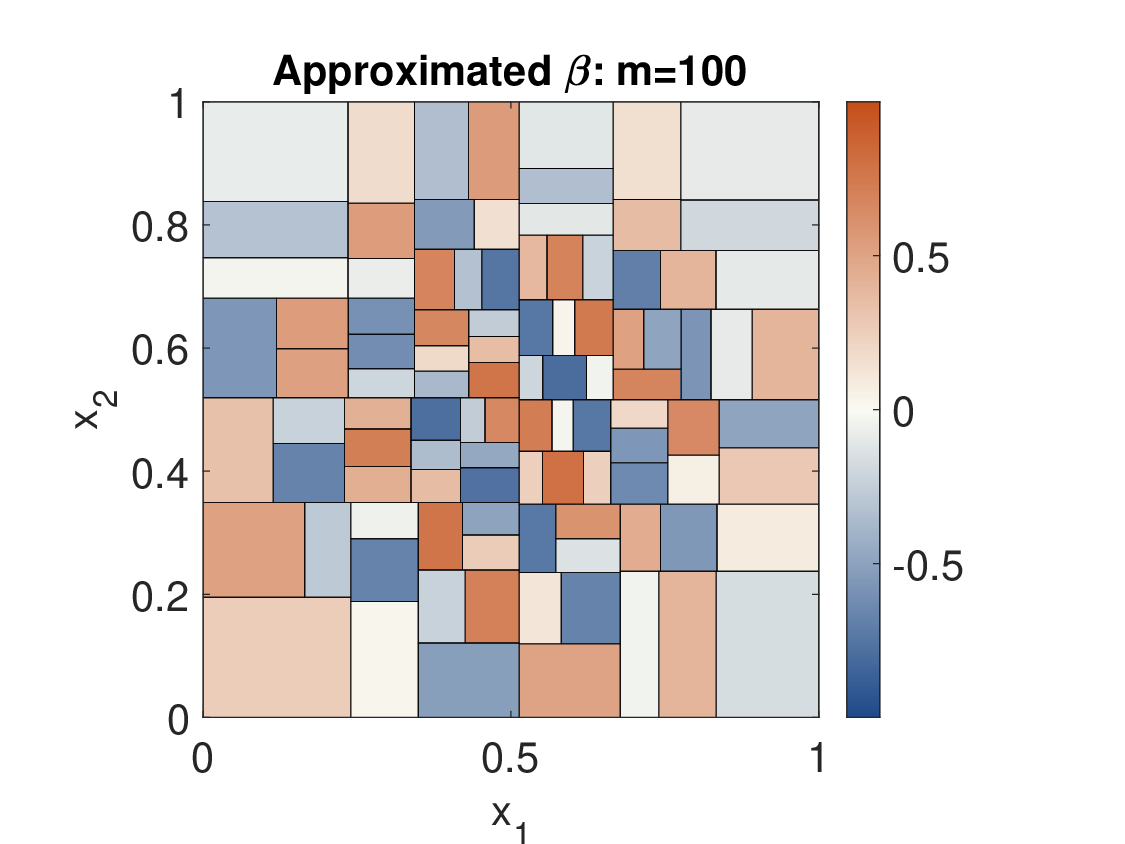}
    \label{fig:example 3 - two-dimensions - beta reconstruction and segmentation m=100}
    }
    \subfigure[]{
    \includegraphics[width=0.3\textwidth]{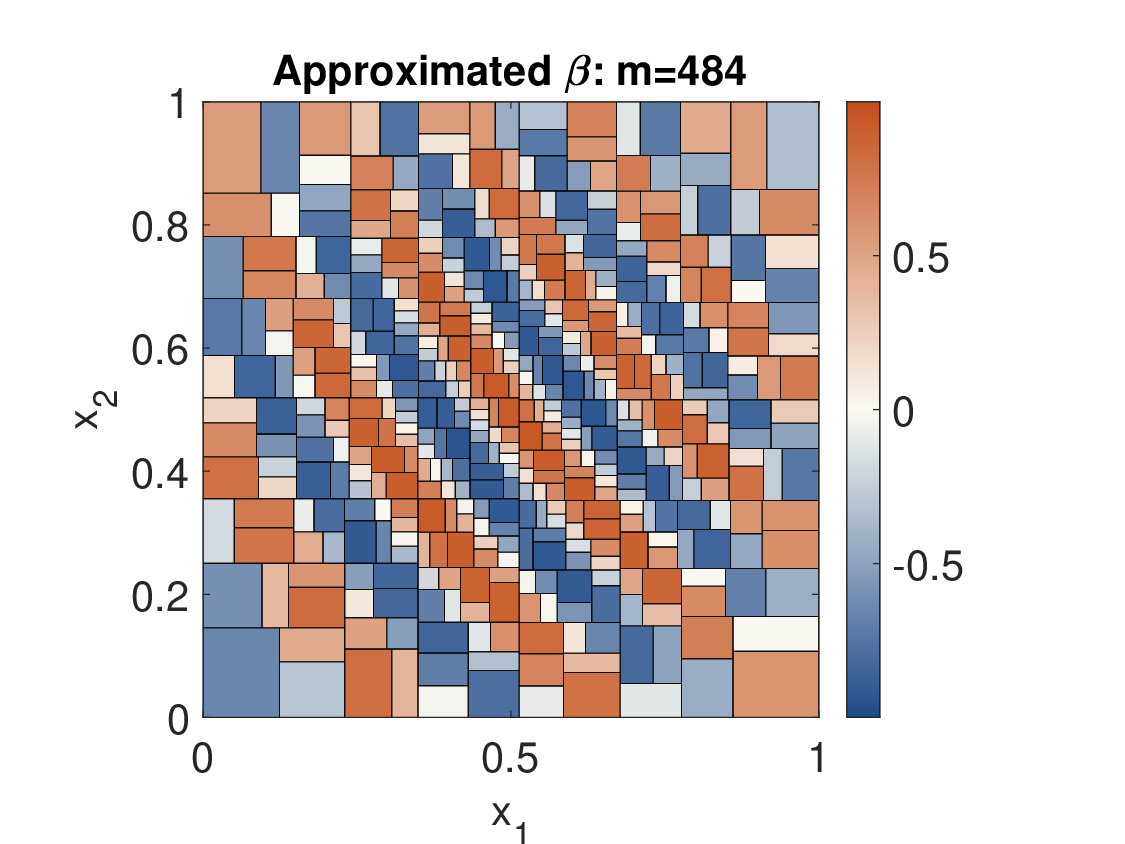}
    \label{fig:example 3 - two-dimensions - beta reconstruction and segmentation m=484}
    }
    \\
    \subfigure[]{
    \includegraphics[width=0.3\textwidth]{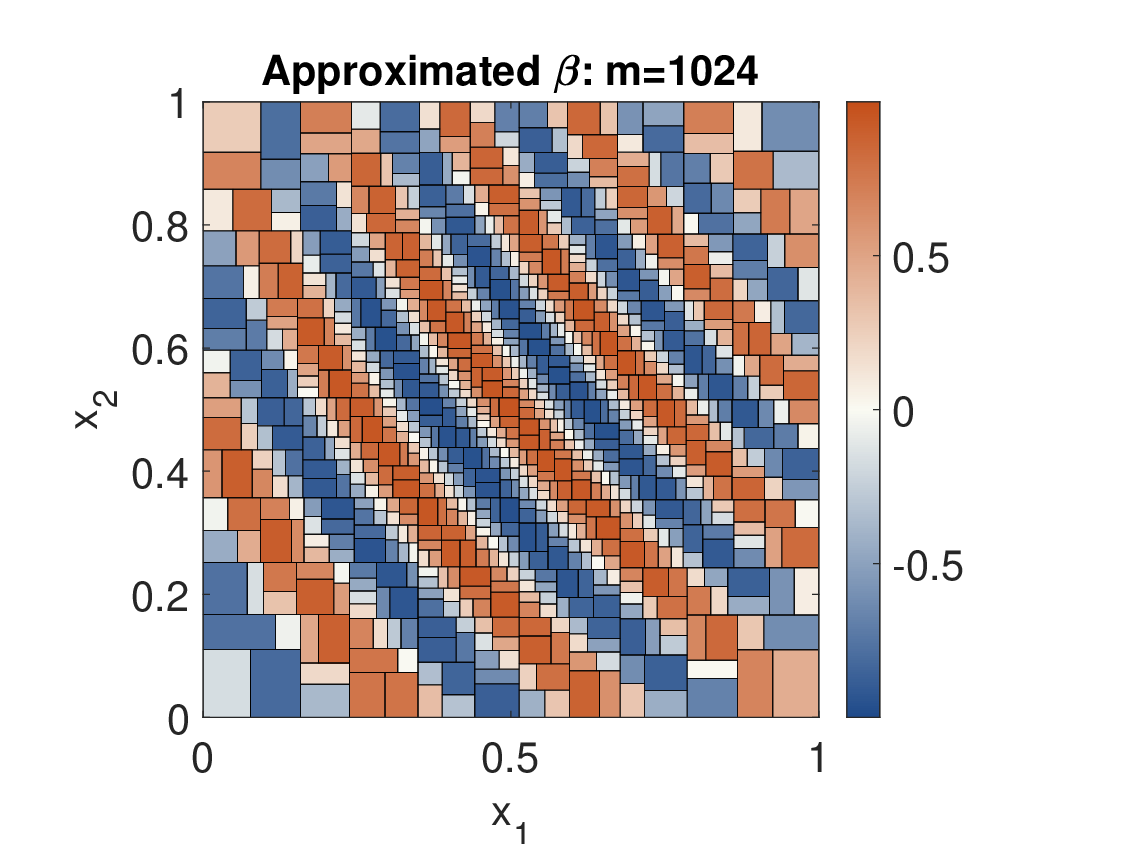}
    \label{fig:example 3 - two-dimensions - beta reconstruction and segmentation m=1024}
    }
    \subfigure[]{
    \includegraphics[width=0.3\textwidth]{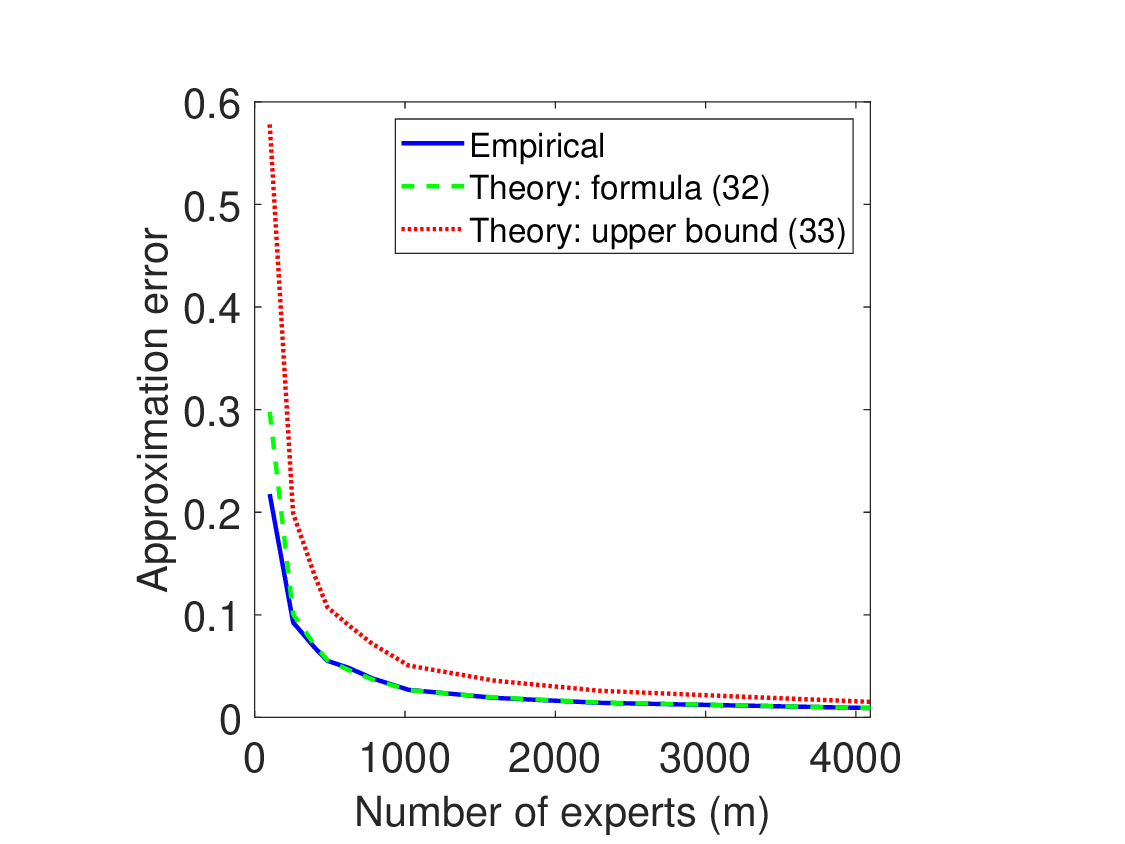}
    \label{fig:example 3 - two-dimensions - approximation error curve vs m}
    }
    \caption{Examples for UBM-density-guided segmentation for a two-dimensional input space where $\beta$ is a two-dimensional \textbf{nonseparable cosine} and $\pdfxvec$ is a truncated isotropic Gaussian. In (\textit{e})-(\textit{g}), the black lines denote the segmentation borders of the regions $\{A_i\}_{i=1}^m$ that were formed by Algorithm \ref{algorithm:2D segmentation by bisection} using the UBM segment density in (\textit{d}).}
    \label{fig:empirical results - two-dimensions - example 3 - nonseparable cosine with constant area beta}
\end{figure}

Figs.~\ref{fig:empirical results - two-dimensions - example 1 - cosine beta}, \ref{fig:empirical results - two-dimensions - example 2 - cosine with constant area beta}, \ref{fig:empirical results - two-dimensions - example 3 - nonseparable cosine with constant area beta}, empirically demonstrate segmentations that are formed for two-dimensional inputs using the UBM segment density $\lambda^{\sf UBM}$ from Eq.~(\ref{eq:optimal expert segment density function - multidimensional}) of Theorem \ref{theorem: optimal expert segment density function and test error - multidimensional}. The discrete segmentation is formed by a heuristic bisection algorithm that creates rectangular regions that contain approximately the same integrated amount of UBM segment density $\lambda^{\sf UBM}$ (see Algorithm \ref{algorithm:2D segmentation by bisection}). This equalization of segment density among the segmentation regions extends the analytical segmentation form of the one-dimensional setting; recall (\ref{appendix:eq:optimal expander function for scalar quantization})-(\ref{appendix:eq:optimal scalar quantization - optimal segmentation by expander}) of Appendix \ref{appendix:subsec:Scalar High-Rate Quantization - Forming a Quantizer from a Quantization Interval Density} that in Section \ref{appendix:subsec:Empirical Results - Optimal Segmentation for MoE with One-Dimensional Inputs} we use with our one-dimensional MoE density $\lambda^{\sf opt}$ from (\ref{eq:optimal expert segment density function - one-dimensional}). In our two-dimensional case, the one-dimensional segmentation principle is adapted for recursively dividing a rectangle into two rectangular segments.
The bisection algorithm here is an adaptation of the computational load-balancing approach of \cite{berger1987partitioning} to our current problem and segment density. 
Importantly, our specific adaptation uses the UBM segment density derived from our high-rate MoE analysis to determine each split coordinate through the one-dimensional expander construction from high-rate scalar quantization. More details are provided in Appendix \ref{appendix:subsec:MoE for Multidimensional Inputs: The Bisection Algorithm for Two-Dimensional Segmentation}.

While using an optimality criterion of equalizing density among segments, we consider here rectangular region shapes and do not search for better shapes. Omitting the shape optimization here is a reasonable choice because the UBM segment density $\lambda^{\sf UBM}$ from Eq.~(\ref{eq:optimal expert segment density function - multidimensional}) already lacks some directional information, due to the constant limiting moment profile (Assumption \ref{assumption: moment of inertia is approximately as for the optimal tessellating polytope}) and the upper bound in (\ref{eq:theorem - test error sum formula - multidimensional}) that results from bounding the directional interaction between $\beta$ gradients and the matrix integral from (\ref{eq:theorem - test error sum formula - equality - multidimensional}).

The experiment settings of Figs.~\ref{fig:empirical results - two-dimensions - example 1 - cosine beta}, \ref{fig:empirical results - two-dimensions - example 2 - cosine with constant area beta}, \ref{fig:empirical results - two-dimensions - example 3 - nonseparable cosine with constant area beta} differ only in the true $\beta$ function. In Fig.~\ref{fig:empirical results - two-dimensions - example 1 - cosine beta}, $\beta(\vec{x})=\cos(10\pi x_1)\cos(6\pi x_2)$ for $\vec{x}\in[0,1]^2$. 
Here, $x_1$ and $x_2$ are the first and second coordinates of $\vec{x}$, respectively.
In Fig.~\ref{fig:empirical results - two-dimensions - example 2 - cosine with constant area beta}, $\beta(\vec{x})=\beta_1(x_1)\beta_2(x_2)$ where $\beta_1(x_1)=1$ for $x_1\in(\frac{2}{5},\frac{3}{5})$, $\beta_1(x_1)=\cos(10\pi x_1)$  for $x_1\in[0,\frac{2}{5}]\cup[\frac{3}{5},1]$, and $\beta_2(x_2)=1$ for $x_2\in(\frac{1}{3},\frac{2}{3})$, $\beta_2(x_2)=\cos(6\pi x_2)$  for $x_2\in[0,\frac{1}{3}]\cup[\frac{2}{3},1]$. 
In Fig.~\ref{fig:empirical results - two-dimensions - example 3 - nonseparable cosine with constant area beta}, we consider a nonseparable function $\beta(\vec{x})=\cos\left(2\pi (3x_1+2x_2+2x_1x_2)\right)$ for $\vec{x}\in[0,1]^2$.
The $\beta$ function is used in the data model $\scrand{y}=\beta(\vecrand{x})+\scrand{\epsilon}$ from (\ref{eq:data model - multidimensional}); here, $\vecrand{x}$ is from a two-dimensional truncated isotropic Gaussian distribution with mean $[0.5,0.5]^T$; the noise $\scrand{\epsilon}$ is a uniform random variable from the continuous range $[-0.1,0.1]$.
All three experiments demonstrate that the optimal segment density is higher for inputs whose product of the input probability density function $\pdfxvec$ and the squared norm of $\nabla\beta$ is relatively high; see Figs.~\ref{fig:example 1 - two-dimensions - optimal segment density}, \ref{fig:example 2 - two-dimensions - optimal segment density}, \ref{fig:example 3 - two-dimensions - optimal segment density}, which empirically compute $\lambda^{\sf UBM}$ that was formulated in Eq.~(\ref{eq:optimal expert segment density function - multidimensional}).

Figs.~\ref{fig:example 1 - two-dimensions - beta reconstruction and segmentation m=100}-\ref{fig:example 1 - two-dimensions - beta reconstruction and segmentation m=1024}, \ref{fig:example 2 - two-dimensions - beta reconstruction and segmentation m=100}-\ref{fig:example 2 - two-dimensions - beta reconstruction and segmentation m=1024}, \ref{fig:example 3 - two-dimensions - beta reconstruction and segmentation m=100}-\ref{fig:example 3 - two-dimensions - beta reconstruction and segmentation m=1024} show the $\lambda^{\sf UBM}$-guided predictors whose segmentations are formed by the bisection algorithm (Algorithm \ref{algorithm:2D segmentation by bisection}) and the expert constants have the optimal values for this segmentation; the black lines show the borders that form the regions $\{A_i\}_{i=1}^{m}$. Relatively smaller regions appear where $\lambda^{\sf UBM}$ is higher (due to a relatively higher product of the input probability density function $\pdfxvec$ and the squared norm of $\nabla\beta$). 

Figs.~\ref{fig:example 1 - two-dimensions - approximation error curve vs m}, \ref{fig:example 2 - two-dimensions - approximation error curve vs m}, \ref{fig:example 3 - two-dimensions - approximation error curve vs m} show the approximation error curves of the hypothesis class $\mathcal{H}_{m,2}^{c}\left(\{A_i\}_{i=1}^m\right)$ given the segmentation $\{A_i\}_{i=1}^m$ formed by Algorithm \ref{algorithm:2D segmentation by bisection} for a varying number of experts $m$ (by the definition of the hypothesis classes in (\ref{eq:hypothesis class of d-dimensional MoE - definition}) and (\ref{eq:hypothesis class of d-dimensional MoE given segmentation- definition}), the approximation error of $\mathcal{H}_{m,2}^{c}\left(\{A_i\}_{i=1}^m\right)$ upper bounds the approximation error of its containing, larger hypothesis class $\mathcal{H}_{m,2}$). The empirical curve, in solid blue line, was computed as the average squared error of the $\lambda^{\sf UBM}$-guided predictors (for each $m$) for a dataset of 5000 test examples (recall that in the approximation error evaluation of this section, there is no learning from training dataset). 
The two theoretical curves were numerically evaluated for the  $\lambda^{\sf UBM}$-guided predictors by the theoretical error formula (\ref{eq:theorem - test error sum formula - equality - multidimensional}), which is shown as a dashed green line, and the theoretical upper bound (\ref{eq:theorem - test error sum formula - multidimensional}), which is shown as a dotted red line. 

The integral formula (\ref{eq:optimal test error - multidimensional}) is not evaluated here due to the undefined choice of a constant moment; nevertheless, this integral formula is important for formulating the UBM segment density, which does not depend on the value of the constant moment.

The shown evaluations demonstrate that the theory explains the empirical results, unless the number of experts is too small to support our asymptotic assumption by which the theoretical remainders in (\ref{eq:theorem - test error sum formula - equality - multidimensional}), (\ref{eq:theorem - test error sum formula - multidimensional}) are excluded from the numerical evaluation.

\section{Learning the Constant Experts: A Tradeoff in the Number of Experts}
\label{sec:Learning the Constant Experts}

The previous sections studied the approximation error of the MoE model classes (\ref{eq:hypothesis class of d-dimensional MoE - definition}) and (\ref{eq:hypothesis class of d-dimensional MoE given segmentation- definition}), namely, the best performance that these model classes can achieve for a given data distribution (i.e., known input and noise distributions and the true $\beta$ that defines the output distribution). 
Now, we turn to study the generalization aspect of learning the expert constants from training examples (and without knowing the true data distribution) for our MoE model with a given routing segmentation of the input space. 

For a fixed routing segmentation, least-squares learning of the constant experts corresponds to histogram regression \citep{nobel1996histogram}. Related nonasymptotic bounds for the estimation error of histogram least-squares estimators are provided by \citet{saumard2012optimal}. Here, we analyze the learning of the expert constants under our assumptions, with explicit dependence on region probabilities and the ranges of the regression function and noise. We then combine the resulting estimation-error bound with the preceding quantization-based approximation analysis to characterize a tradeoff in the test error upper bound as the number of experts varies for a given training sample size.

\subsection{Learning Expert Constants via Least Squares}
\label{subsec:Learning Expert Constants via Least Squares}

Consider a training dataset of $n$ input-output examples $\mathcal{S}_n\triangleq\left\{\left(\vec{x}\exind{j}, y\exind{j}\right)\right\}_{j=1}^n$ i.i.d.~drawn from the data model (\ref{eq:data model - multidimensional}). The $j^{\sf th}$ training example satisfies the data model 
\begin{equation}
    \label{eq: training data model}
    y\exind{j} = \beta\left(\vec{x}\exind{j}\right)+\epsilon\exind{j}
\end{equation}
where $\epsilon\exind{j}$ is the underlying noise component.

Importantly, now we do not know the true $\beta$ nor the input PDF $\pdfxvec$, therefore computation of the optimal expert constants directly via (\ref{eq:optimal c_i - multidimensional}) or (\ref{eq:theorem - optimal constant - multidimensional}) is impossible. Instead, we learn the expert constants for a given routing segmentation $\{A_i\}_{i=1}^m$. This corresponds to learning a model from the hypothesis class $\mathcal{H}_{m,d}^{c}\left(\{A_i\}_{i=1}^m\right)$, as defined in (\ref{eq:hypothesis class of d-dimensional MoE given segmentation- definition}). 

We define $q_i\exind{j}$ as an indicator for the inclusion of the $j^{\sf th}$ training input $\vec{x}\exind{j}$ in the $i^{\sf th}$ region $A_i$,
\begin{equation}
    \label{eq: q indicator function for the inclusion of the j training input in the i region Ai - definition}
    q_i\exind{j} \triangleq \mathbb{I}\left[{\vec{x}\exind{j}\in A_i}\right]
\end{equation}
where $\mathbb{I}\left[{\sf condition}\right]$ returns 1 if its ${\sf condition}$ is satisfied, and 0 otherwise. 
Accordingly, the number of training examples that are routed to the $i^{\sf th}$ expert is 
\begin{equation}
    \label{eq:number of training examples routed to expert i - definition}
    n_i \triangleq \sum_{j=1}^{n} q_i\exind{j}.
\end{equation}
Then, for a given routing segmentation $\{A_i\}_{i=1}^{m}$, the $i^{\sf th}$ expert constant is learned via least squares
\begin{equation}
    \label{eq:learned expert constants - least squares}
    \forall i\in\{1,\dots,m\} ~\text{such that}~~n_i>0,~~~~\widetilde{c}_i = \underset{v\in\mathbb{R}}{\argmin} \sum_{j: q_i\exind{j} = 1} \left({v - y\exind{j}}\right)^2 = \frac{1}{n_i} \sum_{j: q_i\exind{j} = 1} y\exind{j}.
\end{equation}
Note that the $i^{\sf th}$ expert constant is learned only from the examples that are routed to it. As we consider here a fixed, non-learned input-space segmentation, there is some probability that a region might not get training examples (i.e., $n_i=0$ for some $i$); in such regions, one may apply a global default value for the expert, for example, if $n_i=0$ set the average of all outputs in the training dataset $\widetilde{c}_i=\frac{1}{n} \sum_{j=1}^{n} y\exind{j}$. This will be useful for the empirical analysis in Section \ref{appendix:subsec:Empirical Results - Learning Experts from Training Data for MoE with Uniform Segmentation}. The theoretical analysis below will focus on the main case of (\ref{eq:learned expert constants - least squares}).

The following theorem shows that the learned expert constant $\widetilde{c}_i$ for an unknown $\beta$ (and $n_i>0$) is an unbiased estimate of the optimal expert constant $c_i^{\sf opt}$ for known $\beta$. The proof is in Appendix \ref{appendix:subsec:Proof of Theorem on the learned expert constant is unbiased}.
\begin{theorem}
    \label{theorem:the learned expert constant is unbiased}
    The expectation of the expert constant $\widetilde{c}_i$ learned via (\ref{eq:learned expert constants - least squares}) given $n_i>0$ training examples is the optimal expert constant $c_i^{\sf opt}$: ~~~$\expectation{\widetilde{c}_i | n_i>0} = c_i^{\sf opt}$.
\end{theorem}

\subsection{Test Error Analysis for MoE with Learned Expert Constants}
\label{subsec:Analysis of Test Error of MoE with Learned Expert Constants}
We start the analysis by decomposing the test error of the MoE with learned constants. The probability of the input routed to the $i^{\sf th}$ expert is denoted as 
\begin{equation}
\label{eq:definition of rho_i}
\rho_i \triangleq \eventprob{\vecrand{x}\in A_i}.
\end{equation}
\begin{lemma}
    \label{lemma:test error with learned constants formulated using test error with optimal constants}
For a given routing segmentation $\{A_i\}_{i=1}^{m}$, the test error of MoE with learned constants $\left\{\widetilde{c}_i\right\}_{i=1}^m$ can be decomposed as 
    \begin{equation}
    \label{eq:test error with learned constants formulated using test error with optimal constants}
        \mathcal{E}_{\sf test} \left({\{A_i\}_{i=1}^m, \left\{\widetilde{c}_i\right\}_{i=1}^m}\right) = \mathcal{E}_{\sf app} \left(\mathcal{H}_{m,d}^{c}\left(\{A_i\}_{i=1}^m\right)\right) + \mathcal{E}_{\sf est}(m,\mathcal{S}_n)
    \end{equation}
    where the approximation error equals the test error of MoE with optimal constants $\left\{c_i^{\sf opt}\right\}_{i=1}^m$, i.e., 
    \begin{equation}
    \mathcal{E}_{\sf app} \left(\mathcal{H}_{m,d}^{c}\left(\{A_i\}_{i=1}^m\right)\right)=\mathcal{E}_{\sf test} \left({\{A_i\}_{i=1}^m, \left\{c_i^{\sf opt}\right\}_{i=1}^m}\right),
    \end{equation}
    and the estimation error, which reflects the statistical learning of $m$ expert constants from the training dataset $\mathcal{S}_n$, is formulated as 
    \begin{equation}
        \label{eq:estimation error definition}
        \mathcal{E}_{\sf est}(m,\mathcal{S}_n) = \sum_{i=1}^{m} \left( \widetilde{c}_i - c_i^{\sf opt} \right)^2 \rho_i.
    \end{equation}
\end{lemma}
See proof in Appendix \ref{appendix:subsec:Proof of Lemma on test error with learned constants formulated using test error with optimal constants}. 
This lemma decomposes the MoE test error into approximation error and estimation error, a decomposition which is widely used in machine learning theory (see, e.g., \citep{shalev2014understanding}). 

The test error expression in Lemma \ref{lemma:test error with learned constants formulated using test error with optimal constants} motivates us to analyze the statistical behavior of the distance $\left\lvert \widetilde{c}_i - c_i^{\sf opt}\right\rvert$ between the learned constants and their optimal counterparts. First, let us formulate concentration bounds for the number $n_i$ of training examples that are routed to the $i^{\sf th}$ expert.
\begin{lemma}
    \label{lemma:concentration bounds on the number ni of training examples that are routed to expert i}
    For $\widetilde{\delta}\in(0,1)$ and given $n\ge \frac{8}{\rho_i}\ln{\left(\frac{1}{\widetilde{\delta}}\right)}$, 
    \[\eventprob{ n_i > \frac{1}{2}n\rho_i } \ge 1-\widetilde{\delta}.\]
\end{lemma}
The proof uses the Chernoff bound for a sum of independent Bernoulli variables; see Appendix \ref{appendix:subsec:Proof of Lemma concentration bounds on the number ni of training examples that are routed to expert i}.

Then, using Lemma \ref{lemma:concentration bounds on the number ni of training examples that are routed to expert i} and assuming the noise $\epsilon$ in the data model (\ref{eq:data model - multidimensional}) is from a bounded range, the distance $\left\lvert \widetilde{c}_i - c_i^{\sf opt}\right\rvert$ is statistically bounded as follows. 
\begin{theorem}
    \label{theorem:statistical bound on distance between learned constant and optimal constant}
    Consider the $i^{\sf th}$ region $A_i$, whose value range size of $\beta$ is denoted as $R_{\beta,i} \triangleq \underset{\vec{x}\in A_i}{\sup}\beta(\vec{x}) - \underset{\vec{x}\in A_i}{\inf}\beta(\vec{x})$.
    Assume a positive probability $\rho_i>0$ for the input routed to $A_i$.
    Assume the noise $\epsilon$ is from the bounded range $[\epsilon_{\sf min},\epsilon_{\sf max}]$ and denote $R_{\epsilon} \triangleq \epsilon_{\sf max} - \epsilon_{\sf min}$.
    
    \noindent Then, for $\gamma\ge 0$, $\widetilde{\delta}\in(0,1)$, and given $n\ge \frac{8}{\rho_i}\ln{\left(\frac{1}{\widetilde{\delta}}\right)}$, 
    \begin{equation}
        \label{eq:hoeffding bound on distance between learned constant and optimal constant}
        \left\lvert \widetilde{c}_i - c_i^{\sf opt}\right\rvert \le \gamma \frac{R_{\beta,i} + R_{\epsilon}}{\sqrt{n\rho_i}}
    \end{equation}
    with probability at least $1-2\exp\left(-\gamma^2\right)-\widetilde{\delta}$. 
\end{theorem}
The proof is provided in Appendix \ref{appendix:subsec:Proof of theorem on the statistical bound on distance between learned constant and optimal constant}. 

Next, using Lemma \ref{lemma:test error with learned constants formulated using test error with optimal constants} and Theorem \ref{theorem:statistical bound on distance between learned constant and optimal constant}, the estimation error of MoE with learned constants $\left\{\widetilde{c}_i\right\}_{i=1}^m$ can be upper bounded as follows (proof in Appendix \ref{appendix:subsec:Proof of theorem on test error with learned constants formulated using test error with optimal constants - upper bound}).
\begin{theorem}
    \label{theorem:test error with learned constants formulated using test error with optimal constants - upper bound}
For a given routing segmentation $\{A_i\}_{i=1}^{m}$ with $\rho_i>0~\forall i$, $\delta\in(0,1)$, and given $n\ge \frac{8}{\underset{i\in\{1,\dots,m\}}{\min}\rho_i}\ln{\left(\frac{2m}{\delta}\right)}$,
        \begin{equation}
    \label{eq:theorem estimation error upper bound}
        \mathcal{E}_{\sf est}(m,\mathcal{S}_n) \le  \frac{m}{n}\ln\left(\frac{4m}{\delta}\right) \underset{i\in\{1,\dots,m\}}{\max}\left(R_{\beta,i}+R_{\epsilon}\right)^2
    \end{equation}
    with probability at least $1-\delta$. 
\end{theorem}

For a fixed data model, the range factor is bounded independently of $m$. Consequently, for a fixed $\delta$, this gives an estimation-error upper bound of order $\frac{m\log(m)}{n}$, subject to the stated condition on the training dataset size.

By Corollary \ref{corollary:test error big-O bound - multidimensional}, we already know that the approximation error behaves as  $\mathcal{E}_{\sf app} \left(\mathcal{H}_{m,d}^{c}\left(\{A_i\}_{i=1}^m\right)\right) = \sigma_{\epsilon}^2 + \mathcal{O}\left(\frac{d}{m^{2/d}}\right)$.
Therefore, by the test error decomposition (\ref{eq:test error with learned constants formulated using test error with optimal constants}), there is a tradeoff in the upper bound of the test error as function of the number of experts $m$.
Considering a fixed input dimension $d$ and a fixed dataset size $n$, the tradeoff is based on two contradicting trends: For MoE with more experts, i.e., increasing $m$,
\begin{itemize}[leftmargin=*]
 \setlength\itemsep{-0.5em}
    \item the approximation error has a $\sigma_{\epsilon}^2 + \mathcal{O}\left(\frac{d}{m^{2/d}}\right)$ upper bound, which provides a decreasing contribution to the test error bound;   
    \item the estimation error has an $\mathcal{O}\left(\frac{m\log(m)}{n}\right)$ upper bound, under the condition on $n$ of Theorem  \ref{theorem:test error with learned constants formulated using test error with optimal constants - upper bound}, which provides an increasing contribution to the test error bound.
\end{itemize}
This theoretical tradeoff is qualitatively supported by our empirical results for one-dimensional and two-dimensional inputs in the following subsection.

\subsection{Empirical Results: Learning Experts from Training Data for MoE with Uniform Segmentation for One-Dimensional and Two-Dimensional Inputs}
\label{appendix:subsec:Empirical Results - Learning Experts from Training Data for MoE with Uniform Segmentation}

Now we proceed to experiments of learning the constant experts from a given training dataset. Similarly to our theoretical setting of this section, the considered MoE is for a given segmentation (which is not necessarily optimal) that does not need to be learned; here the segmentation will be uniform. Specifically, for the one-dimensional MoE class $\mathcal{H}_{m,1}^{c}\left(\{a_i=\frac{i}{m}\}_{i=1}^m\right)$, as was more generally defined in (\ref{eq:hypothesis class of d-dimensional MoE given segmentation- definition}), we have here the subinterval borders $a_i=\frac{i}{m}$ for $i\in\{1,\dots,m\}$. 
For the two-dimensional MoE class $\mathcal{H}_{m,2}^{c}\left(\{A_i\}_{i=1}^m\right)$, we consider $m=m_{\sf root}^2$ where $m_{\sf root}$ is a positive integer and $A_i=\left[\frac{w_i^{(1)}}{m_{\sf root}},\frac{w_i^{(1)}+1}{m_{\sf root}}\right)\times\left[\frac{w_i^{(2)}}{m_{\sf root}},\frac{w_i^{(2)}+1}{m_{\sf root}}\right)$ where $w_i^{(1)}=\left\lfloor\frac{i-1}{m_{\sf root}}\right\rfloor$ and $w_i^{(2)}=(i-1)\!\!\!\!\mod\! m_{\sf root}$; for $A_i$ whose $w_i^{(1)}=m_{\sf root}-1$, the first coordinate has a closed right side of the range; for $A_i$ whose $w_i^{(2)}=m_{\sf root}-1$, the second coordinate has a closed right side of the range.

\begin{figure}[]
    \centering
    \subfigure[]{
    \includegraphics[width=0.38\linewidth]{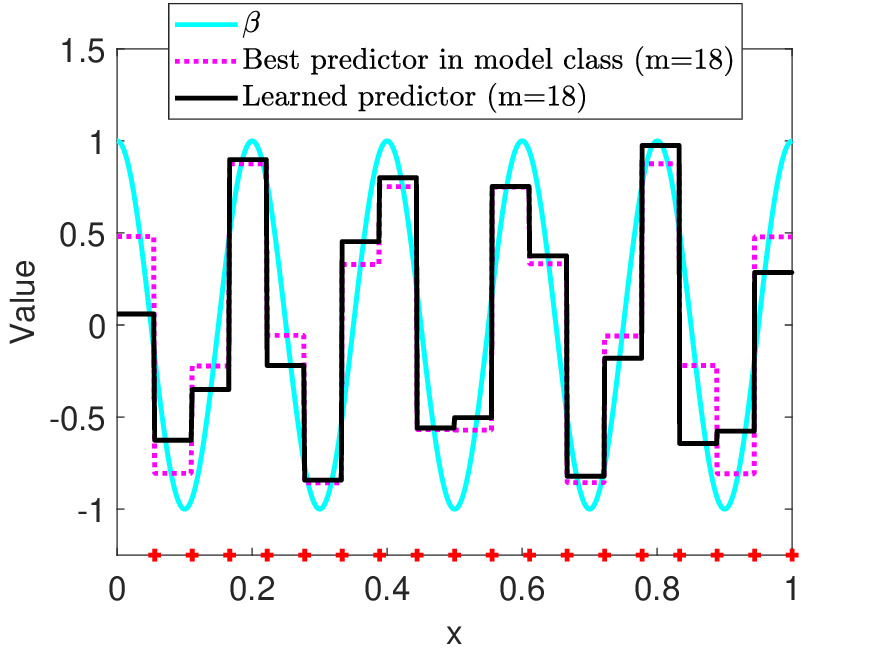}
    \label{fig:example 1 - learning experts for uniform segmentation - beta reconstruction and segmentation m=18}
    }
    \subfigure[]{
    \includegraphics[width=0.38\linewidth]{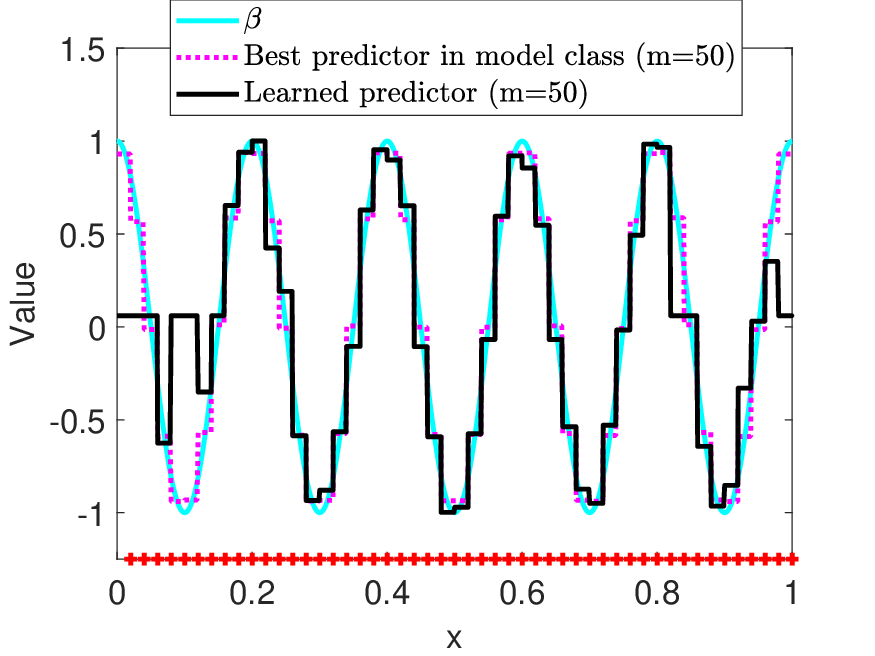}
    \label{fig:example 1 - learning experts for uniform segmentation - beta reconstruction and segmentation m=50}
    }
    \\
    \subfigure[]{
    \includegraphics[width=0.38\linewidth]{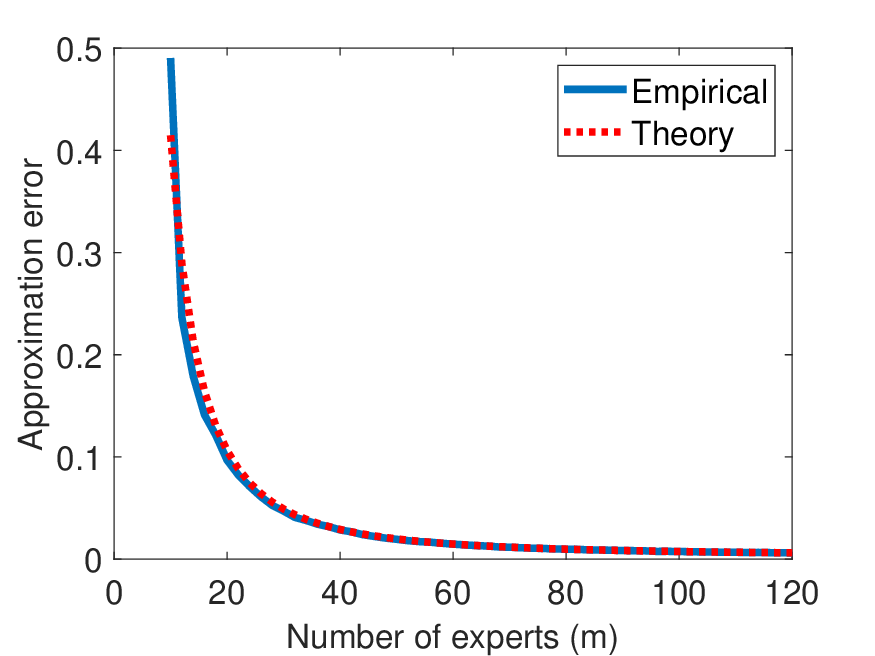}
    \label{fig:example 1 - learning experts for uniform segmentation - approximation error curve vs m}
    }
    \subfigure[]{
    \includegraphics[width=0.38\linewidth]{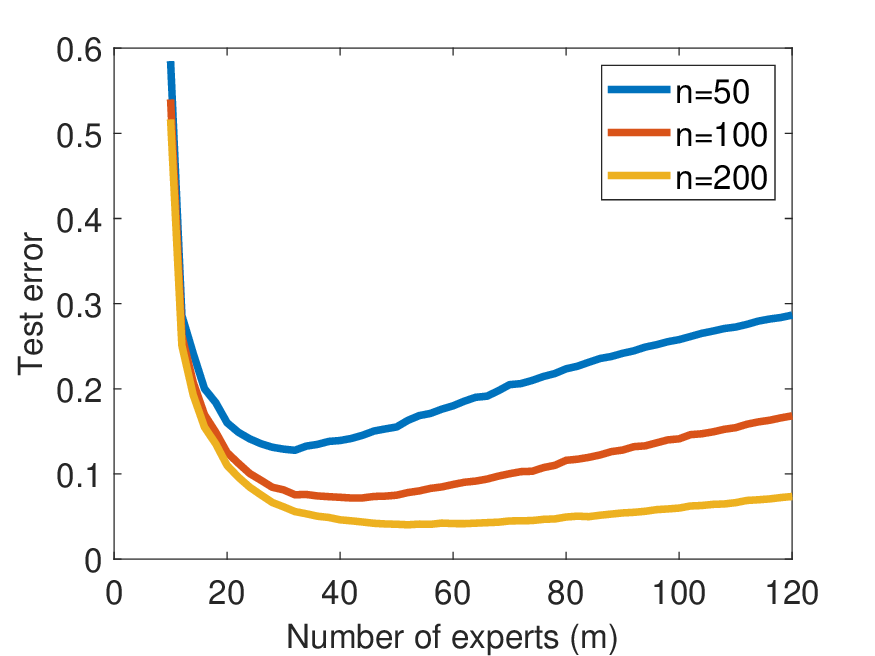}
    \label{fig:example 1 - learning experts for uniform segmentation - test error curve vs m}
    }
    \caption{\textbf{Learning experts for uniform segmentation in one dimension}. This experiment is for a \textbf{cosine} $\beta$ and truncated Gaussian $\pdfxsc$ from Figs.~\ref{fig:example 1 - beta}, \ref{fig:example 1 - pdf x}, respectively. 
    Here, in (a)-(b), the red markers on the $x$-axis denote the given uniform segmentation points $\{a_i=\frac{i}{m}\}_{i=1}^m$; the dotted magenta lines show the best predictor in $\mathcal{H}_{m,1}^{c}\left(\{a_i=\frac{i}{m}\}_{i=1}^m\right)$; the black lines show the predictor learned from a training data of 200 examples. 
    (c) shows the empirical and theoretical approximation error curves of the best predictors in $\mathcal{H}_{m,1}^{c}\left(\{a_i=\frac{i}{m}\}_{i=1}^m\right)$. 
    (d) shows the empirical test error curves for the learned predictors for three sizes of training dataset. 
    }
    \label{fig:empirical results - example 1 - learning experts for uniform segmentation - cosine beta}
\end{figure}

\begin{figure}[]
    \centering
    \subfigure[]{
    \includegraphics[width=0.38\linewidth]{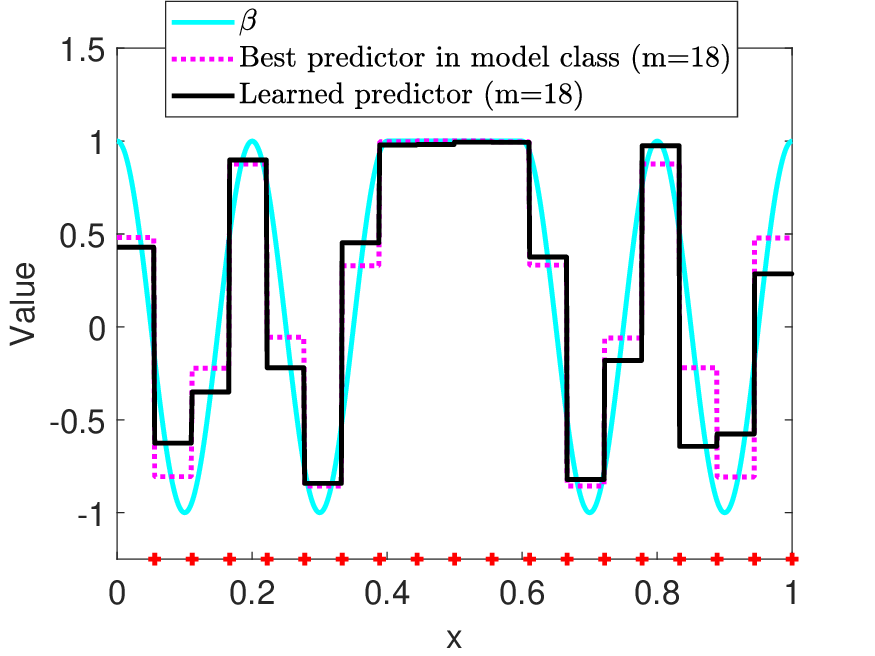}
    \label{fig:example 2 - learning experts for uniform segmentation - beta reconstruction and segmentation m=18}
    }
    \subfigure[]{
    \includegraphics[width=0.38\linewidth]{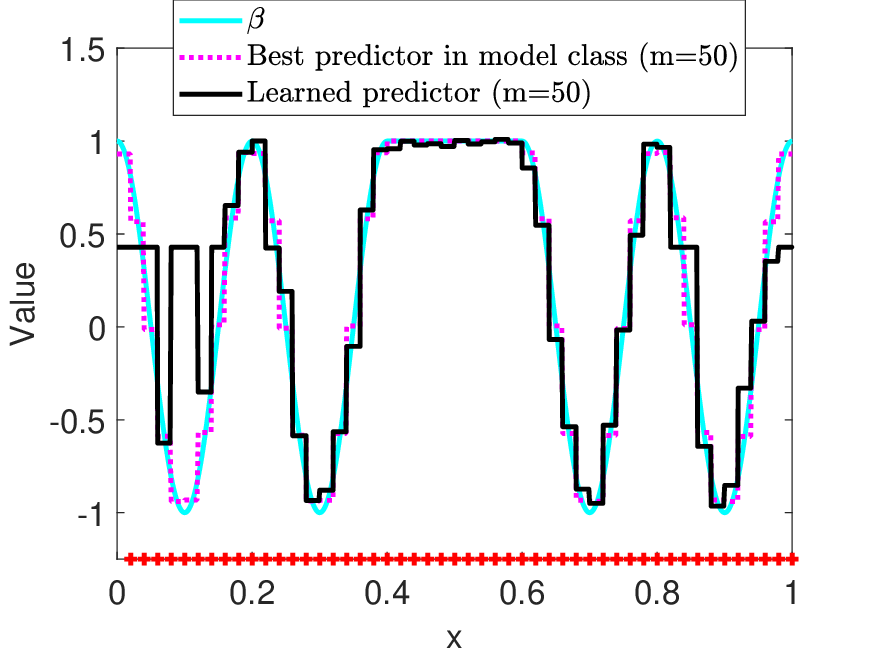}
    \label{fig:example 2 - learning experts for uniform segmentation - beta reconstruction and segmentation m=50}
    }
    \\
    \subfigure[]{
    \includegraphics[width=0.38\linewidth]{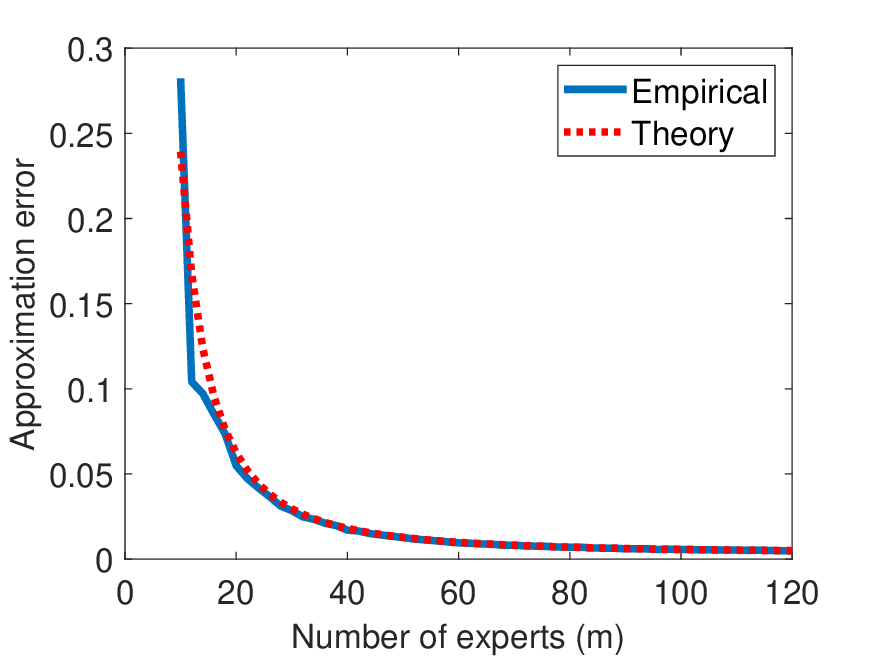}
    \label{fig:example 2 - learning experts for uniform segmentation - approximation error curve vs m}
    }
    \subfigure[]{
    \includegraphics[width=0.38\linewidth]{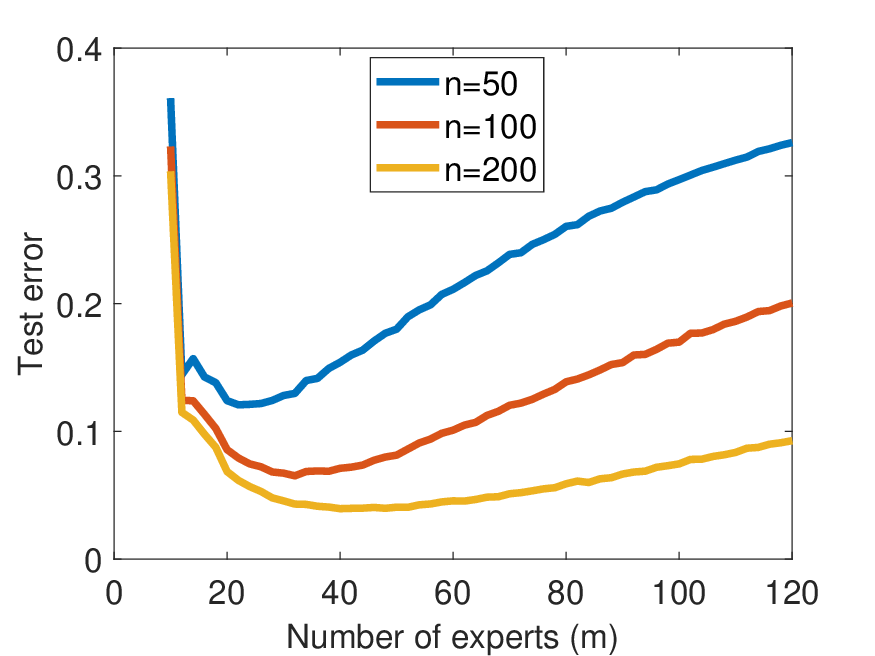}
    \label{fig:example 2 - learning experts for uniform segmentation - test error curve vs m}
    }
    \caption{\textbf{Learning experts for uniform segmentation in one dimension}. This experiment is for a \textbf{cosine with a constant segment} $\beta$ and truncated Gaussian $\pdfxsc$ from Figs.~\ref{fig:example 2 - beta}, \ref{fig:example 2 - pdf x}, respectively. 
    Here, in (a)-(b), the red markers on the $x$-axis denote the given uniform segmentation points $\{a_i=\frac{i}{m}\}_{i=1}^m$; the dotted magenta lines show the best predictor in $\mathcal{H}_{m,1}^{c}\left(\{a_i=\frac{i}{m}\}_{i=1}^m\right)$; the black lines show the predictor learned from a training data of 200 examples. 
    (c) shows the empirical and theoretical approximation error curves of the best predictors in $\mathcal{H}_{m,1}^{c}\left(\{a_i=\frac{i}{m}\}_{i=1}^m\right)$. 
    (d) shows the empirical test error curves for the learned predictors for three sizes of training dataset. 
    }
    \label{fig:empirical results - example 2 - learning experts for uniform segmentation - cosine beta with constant segment}
\end{figure}

\begin{figure}[]
    \centering
    \subfigure[]{
    \includegraphics[width=0.3\linewidth]{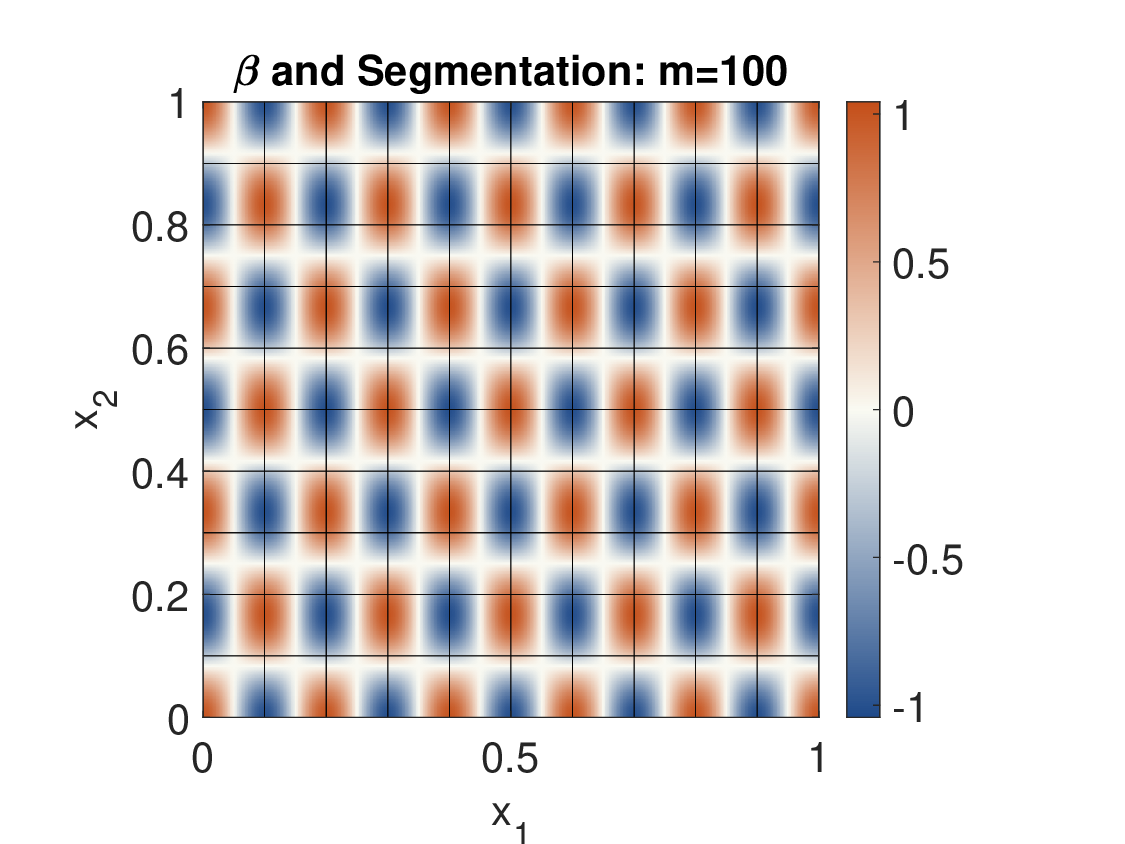}
    \label{fig:example 1 cosine - two-dimensional - learning experts for uniform segmentation - beta target and segmentation m=100}
    }
    \subfigure[]{
    \includegraphics[width=0.3\linewidth]{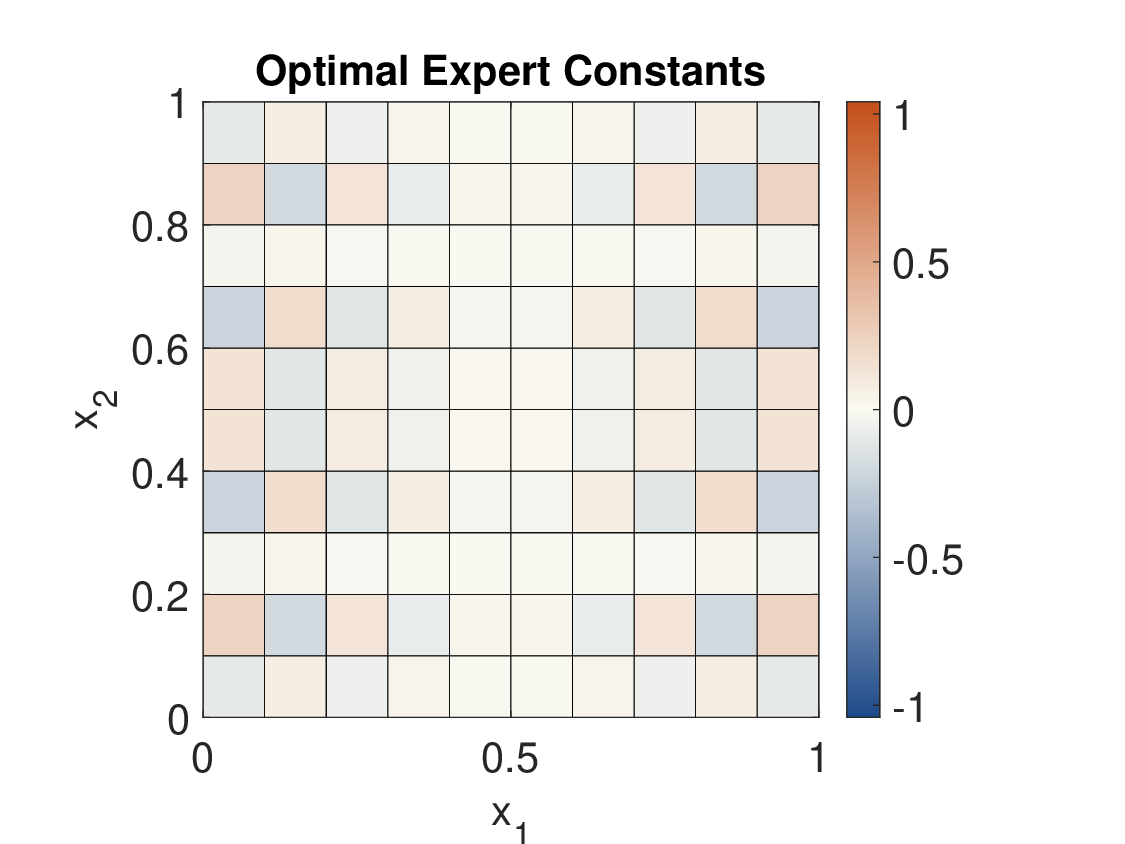}
    \label{fig:example 1 cosine - two-dimensional - learning experts for uniform segmentation - beta optimal constants m=100}
    }
    \subfigure[]{
    \includegraphics[width=0.3\linewidth]{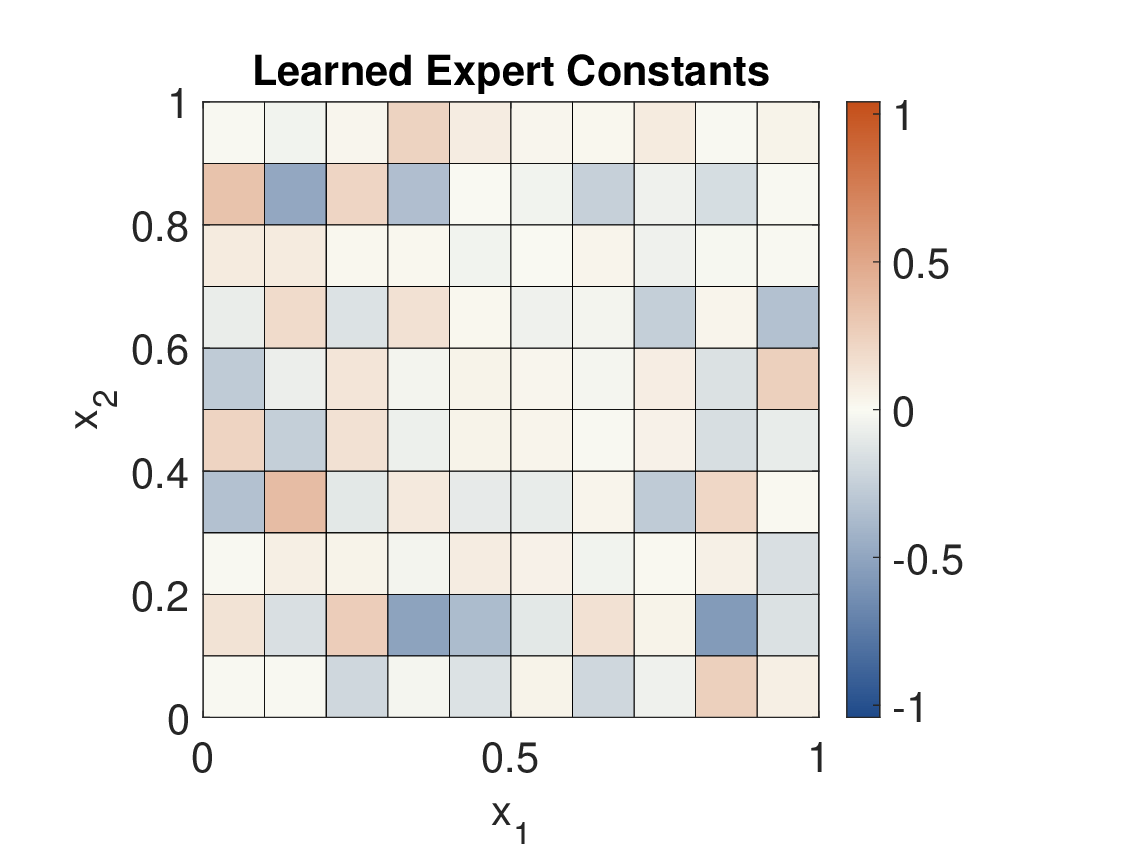}
    \label{fig:example 1 cosine - two-dimensional - learning experts for uniform segmentation - beta learned constants m=100}
    }
    \\
    \subfigure[]{
    \includegraphics[width=0.3\linewidth]{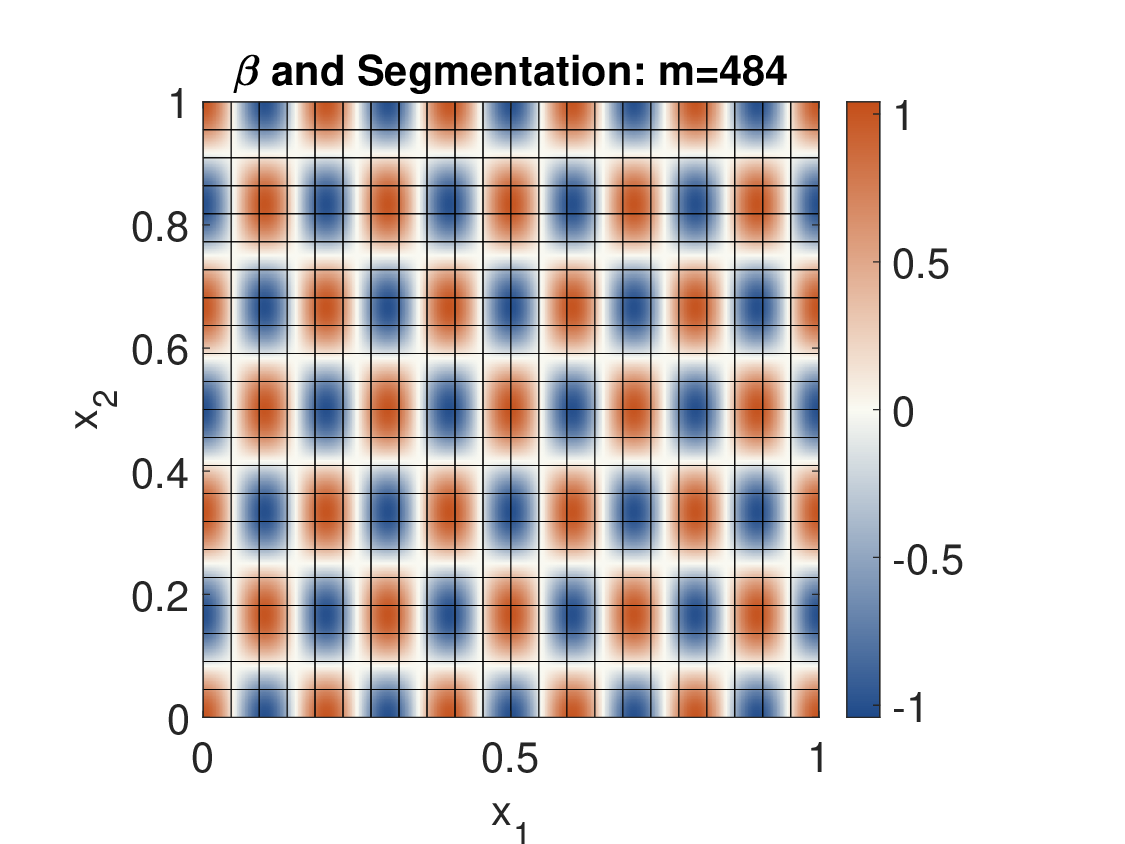}
    \label{fig:example 1 cosine - two-dimensional - learning experts for uniform segmentation - beta target and segmentation m=484}
    }
    \subfigure[]{
    \includegraphics[width=0.3\linewidth]{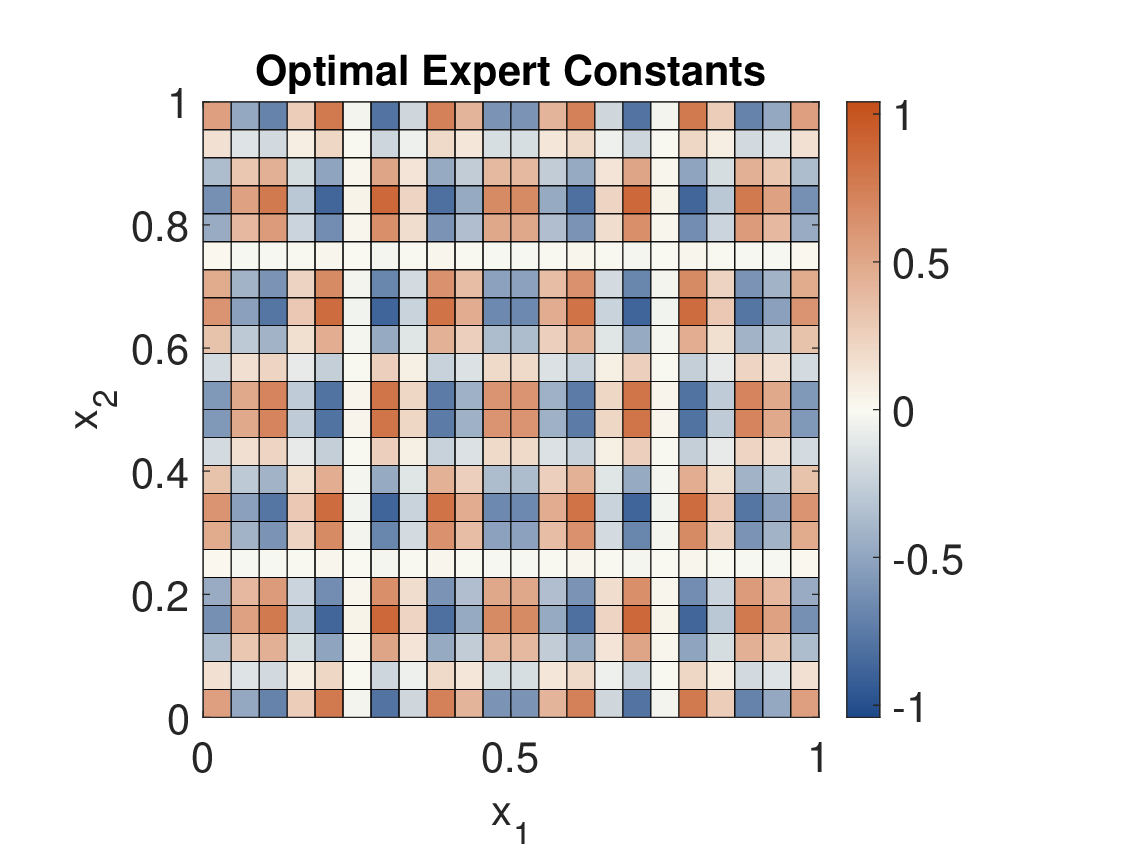}
    \label{fig:example 1 cosine - two-dimensional - learning experts for uniform segmentation - beta optimal constants m=484}
    }
    \subfigure[]{
    \includegraphics[width=0.3\linewidth]{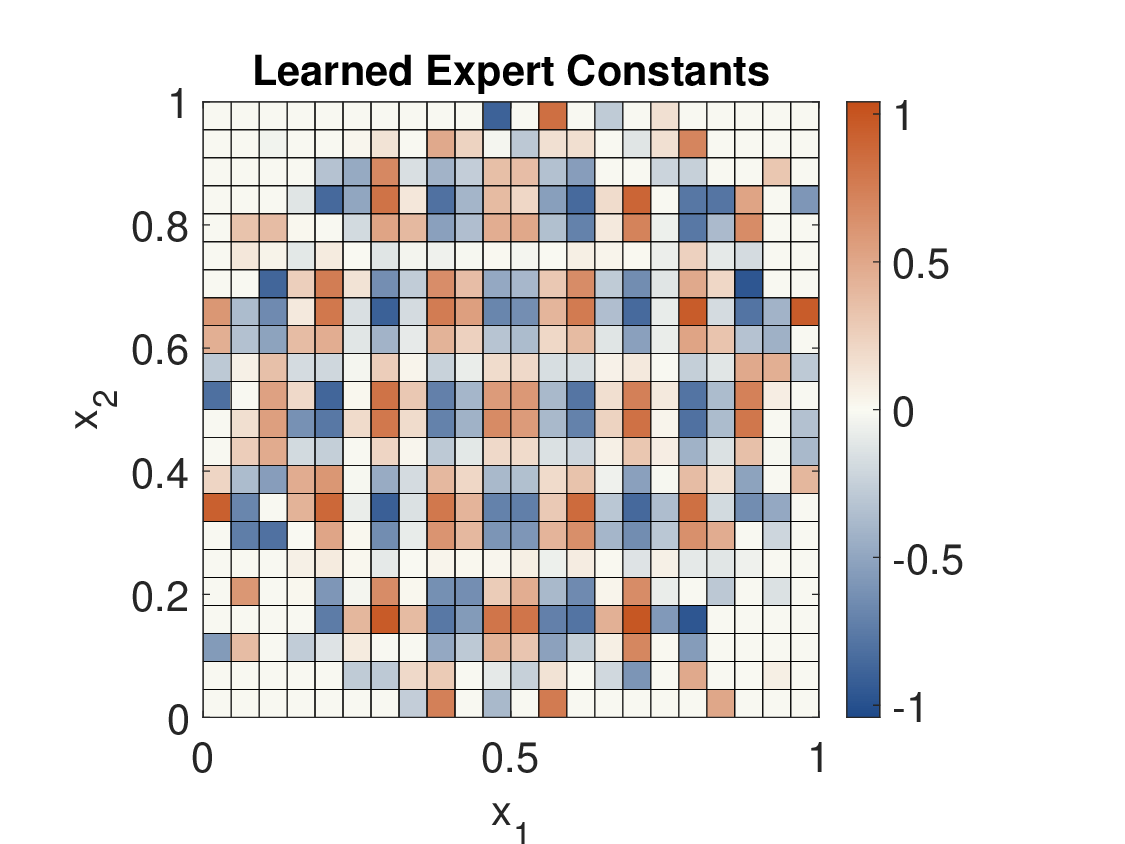}
    \label{fig:example 1 cosine - two-dimensional - learning experts for uniform segmentation - beta learned constants m=484}
    }
    \\    
    \subfigure[]{
    \includegraphics[width=0.3\linewidth]{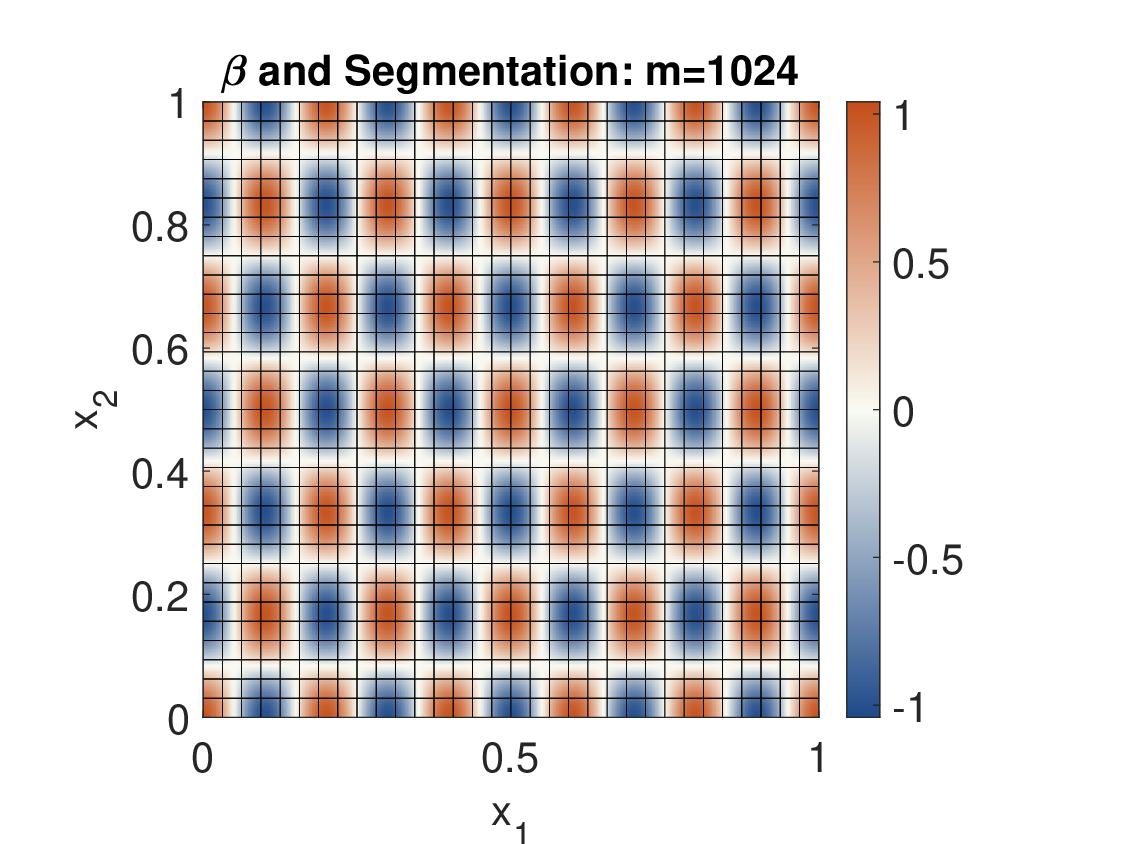}
    \label{fig:example 1 cosine - two-dimensional - learning experts for uniform segmentation - beta target and segmentation m=1024}
    }
    \subfigure[]{
    \includegraphics[width=0.3\linewidth]{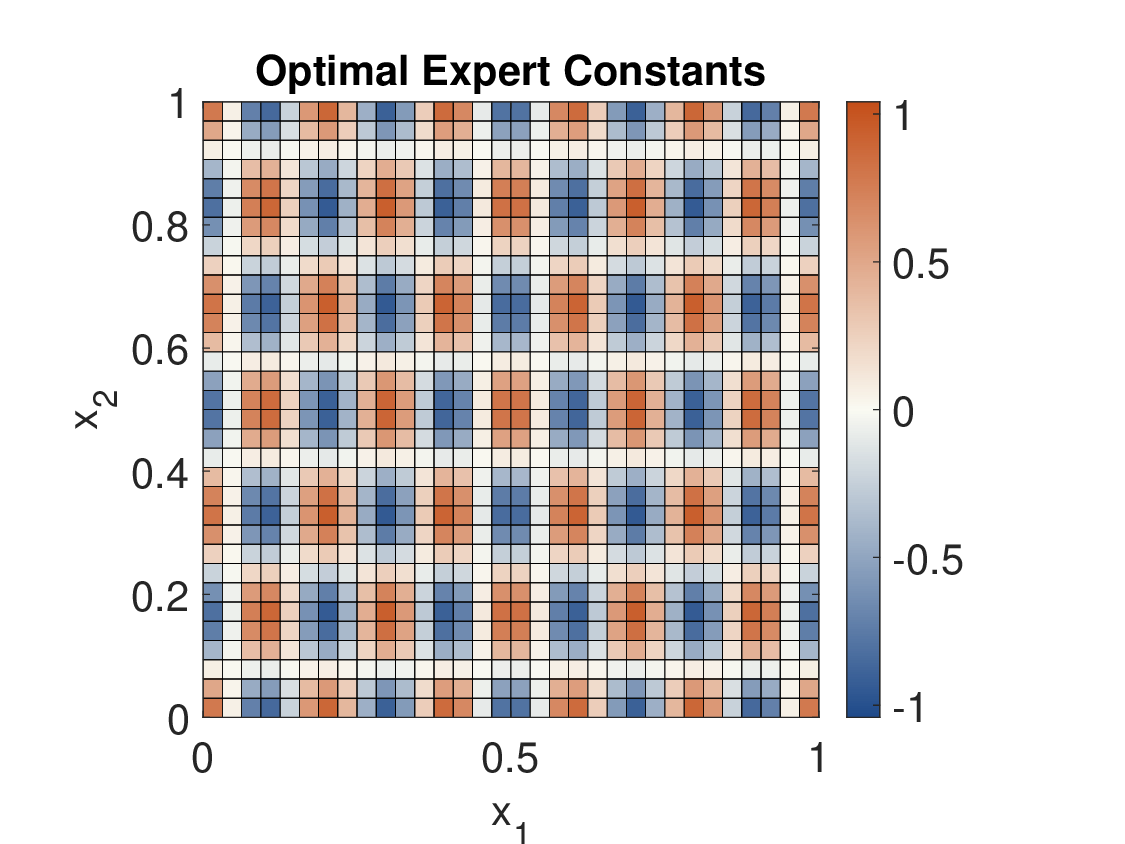}
    \label{fig:example 1 cosine - two-dimensional - learning experts for uniform segmentation - beta optimal constants m=1024}
    }
    \subfigure[]{
    \includegraphics[width=0.3\linewidth]{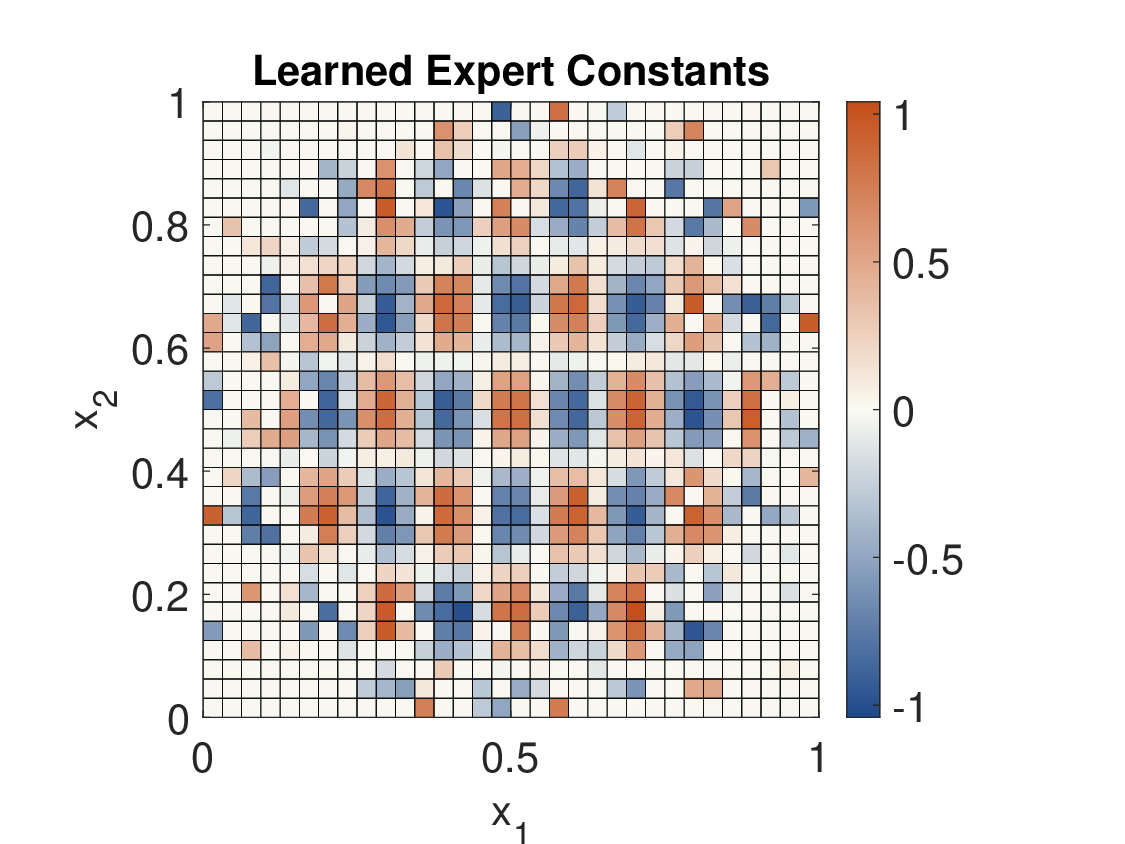}
    \label{fig:example 1 cosine - two-dimensional - learning experts for uniform segmentation - beta learned constants m=1024}
    }
    \\    
    \subfigure[]{
    \includegraphics[width=0.38\linewidth]{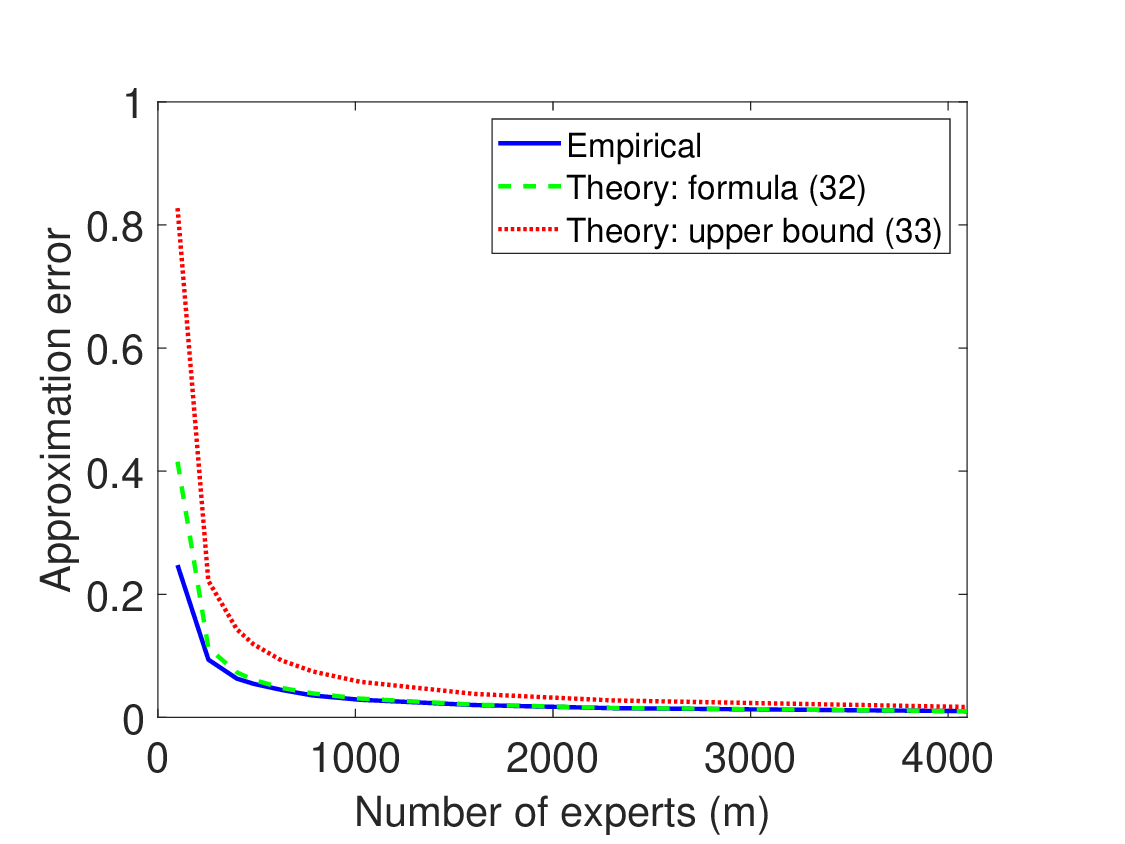}
    \label{fig:example 1 cosine - two-dimensional - learning experts for uniform segmentation - approximation error curve vs m}
    }
    \subfigure[]{
    \includegraphics[width=0.38\linewidth]{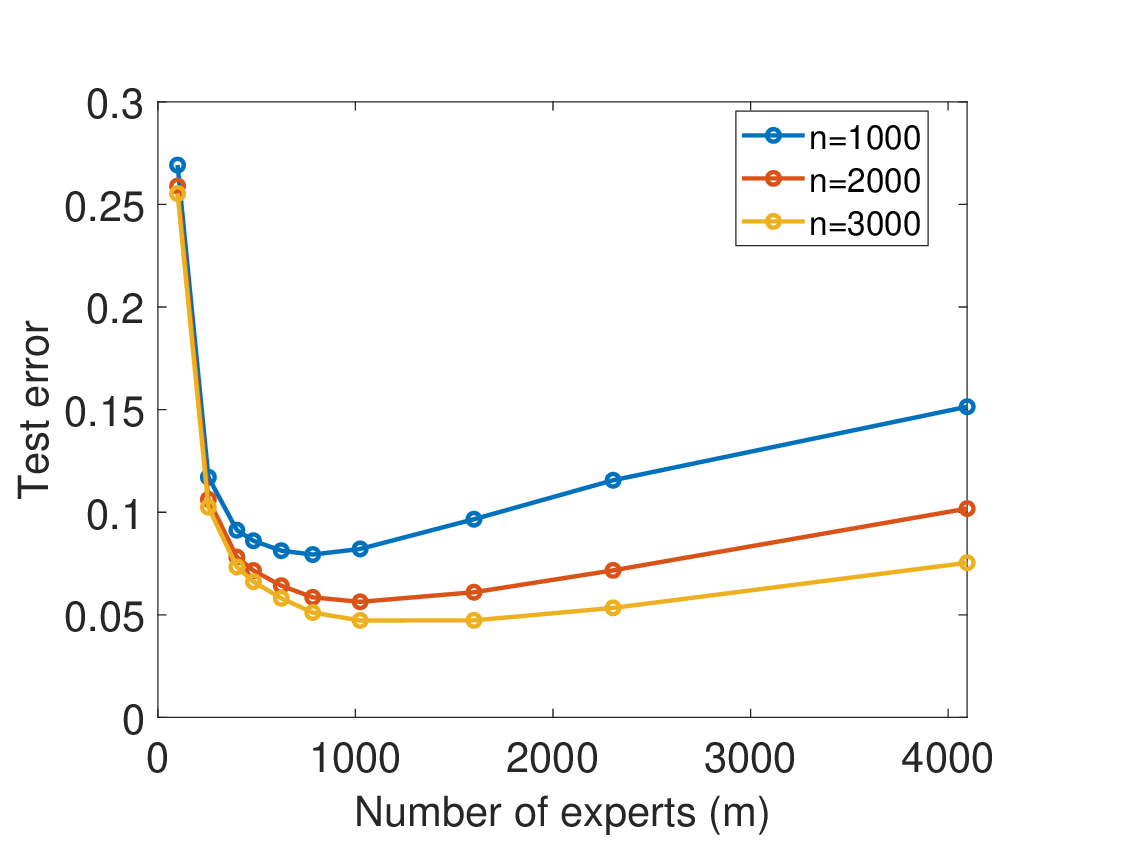}
    \label{fig:example 1 cosine - two-dimensional - learning experts for uniform segmentation - test error curve vs m}
    }
    \caption{\textbf{Learning experts for uniform segmentation in two dimensions}. This experiment is for a \textbf{separable cosine} $\beta$ and truncated Gaussian $\pdfxvec$ from Figs.~\ref{fig:example 1 - two-dimensions - beta}, \ref{fig:example 1 - two-dimensions - pdf x}, respectively. 
    In the first three rows of subfigures: Each row of subfigures shows results for uniform segmentations with a different number of experts ($m=100, 484, 1024$); the left subfigures show the true $\beta$ and the uniform segmentation (borders appear in black lines), the center subfigures show the optimal expert constants, the right subfigures show the learned expert constants from a training dataset of $n=3000$ examples. 
    Subfigure (\textit{j}) shows the empirical and theoretical approximation error curves for the uniform segmentations. Subfigure (\textit{k}) shows the empirical test error vs.~$m$ curves for three sizes of training dataset $n=1000, 2000, 3000$.
    }
    \label{fig:empirical results - two-dimensional - learning experts for uniform segmentation - cosine beta}
\end{figure}

\begin{figure}[]
    \centering
    \subfigure[]{
    \includegraphics[width=0.3\linewidth]{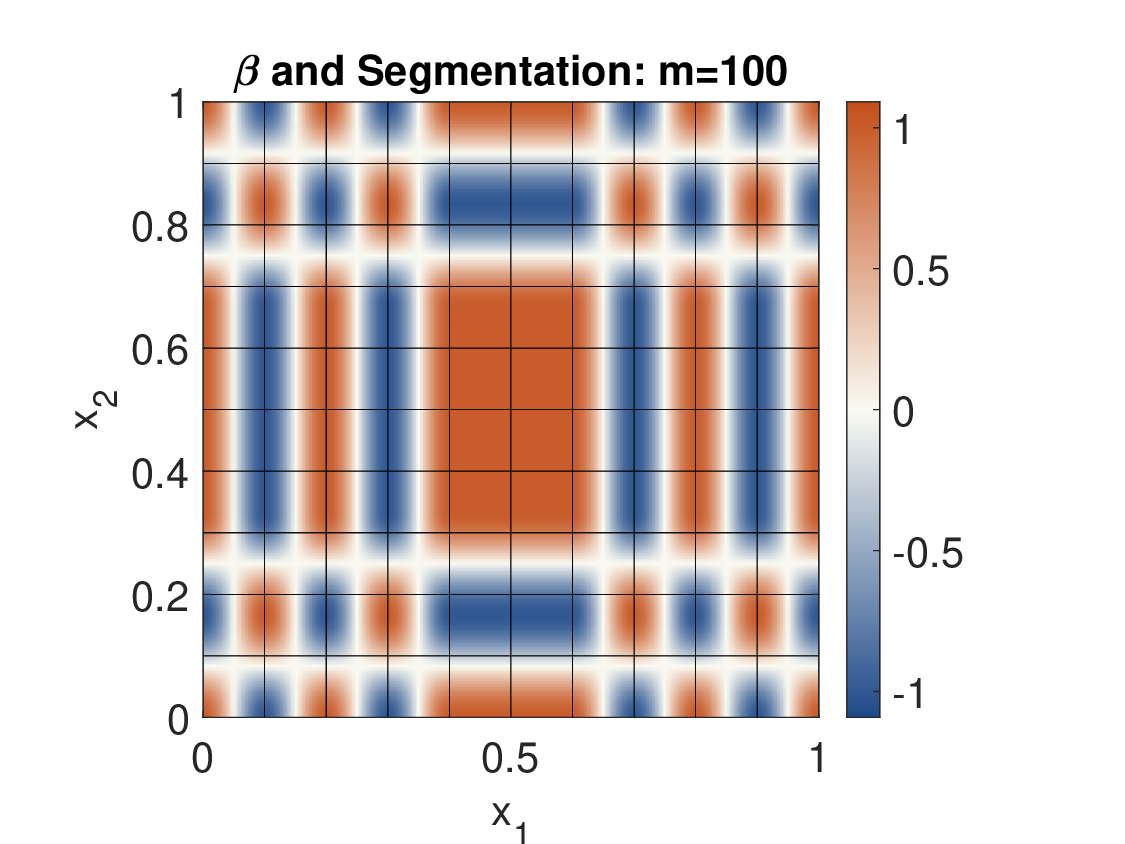}
    \label{fig:example 2 cosine with constant area - two-dimensional - learning experts for uniform segmentation - beta target and segmentation m=100}
    }
    \subfigure[]{
    \includegraphics[width=0.3\linewidth]{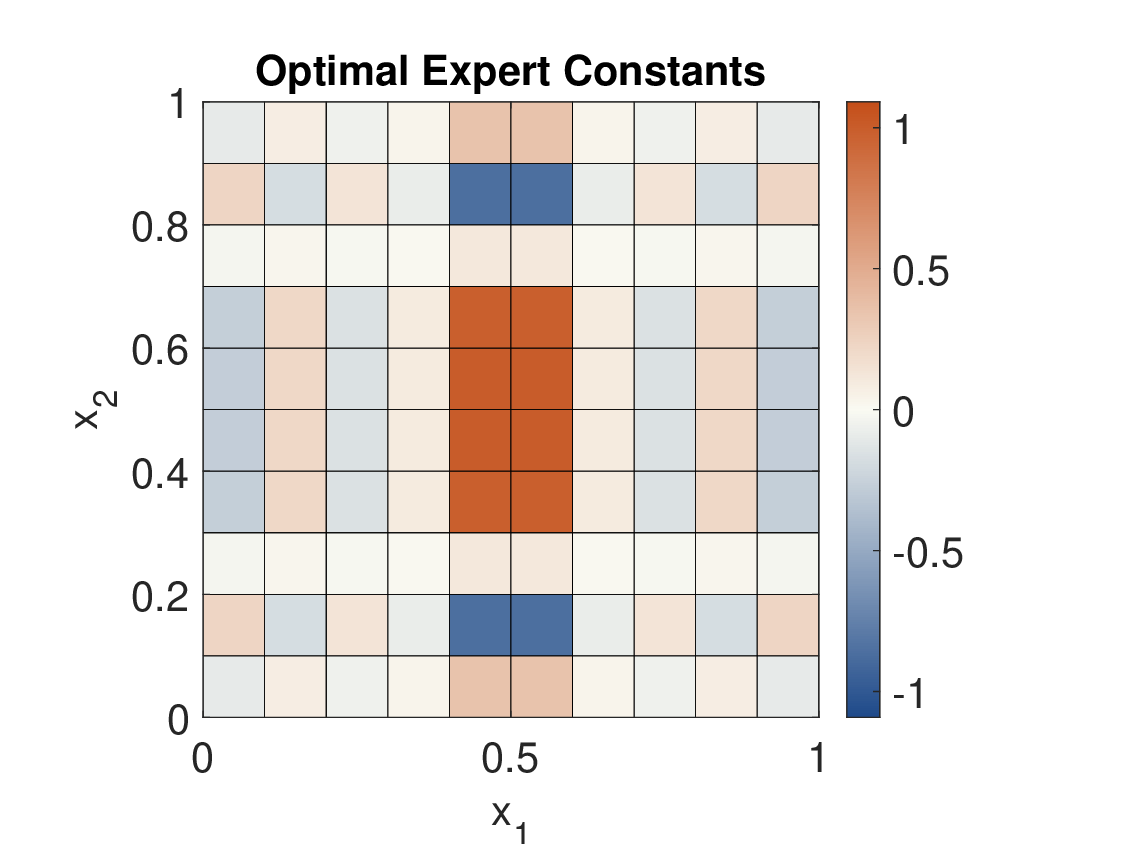}
    \label{fig:example 2 cosine with constant area - two-dimensional - learning experts for uniform segmentation - beta optimal constants m=100}
    }
    \subfigure[]{
    \includegraphics[width=0.3\linewidth]{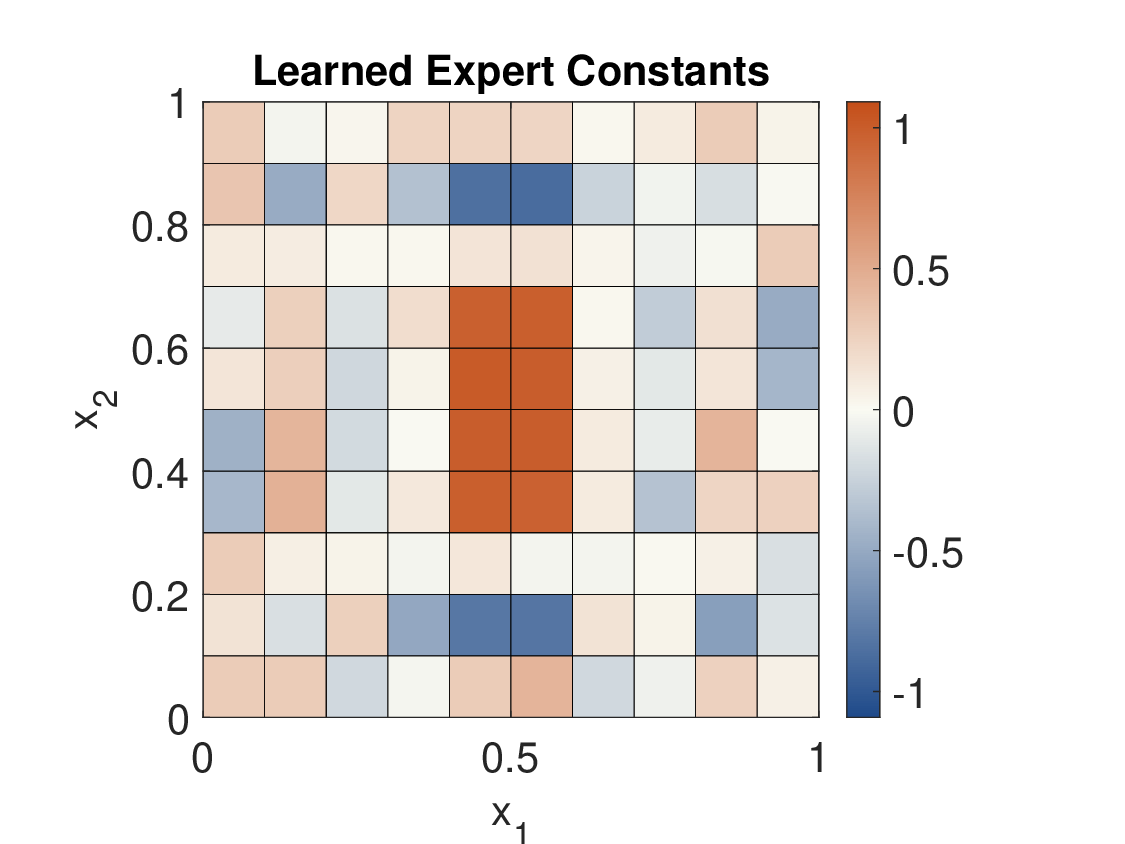}
    \label{fig:example 2 cosine with constant area - two-dimensional - learning experts for uniform segmentation - beta learned constants m=100}
    }
    \\
    \subfigure[]{
    \includegraphics[width=0.3\linewidth]{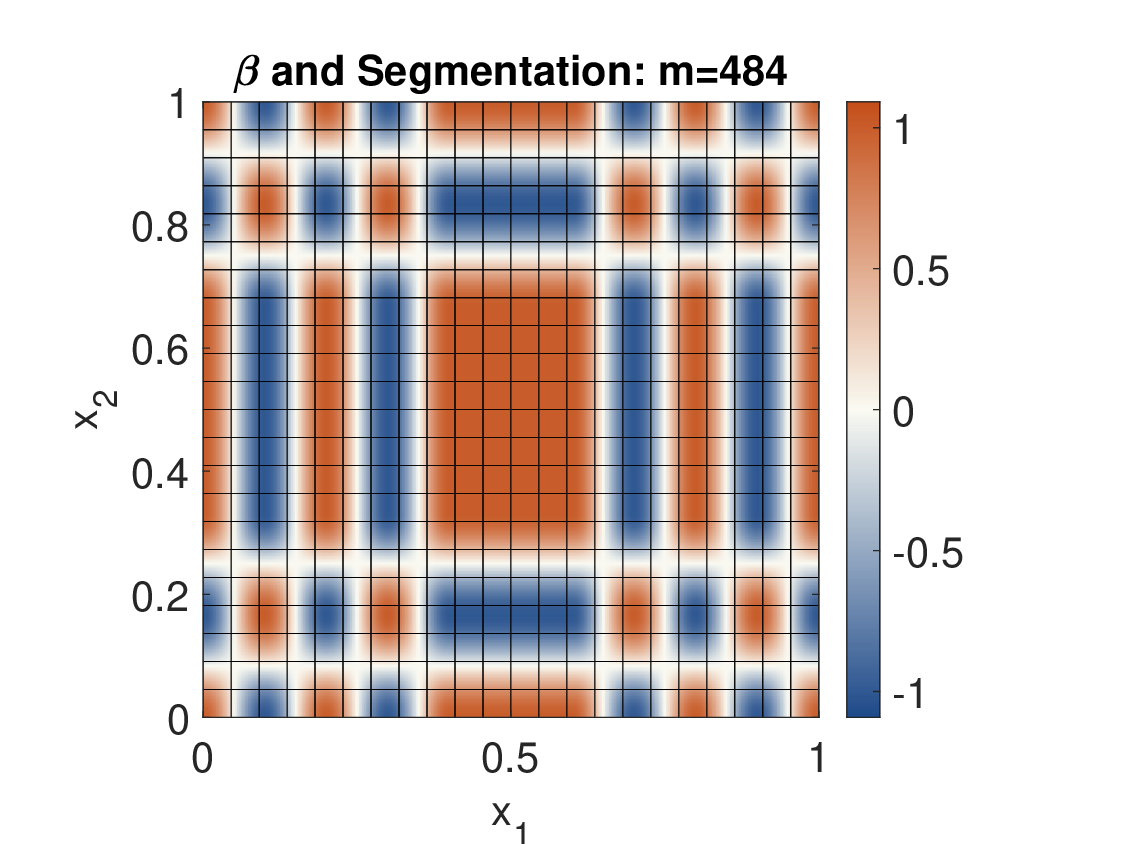}
    \label{fig:example 2 cosine with constant area - two-dimensional - learning experts for uniform segmentation - beta target and segmentation m=484}
    }
    \subfigure[]{
    \includegraphics[width=0.3\linewidth]{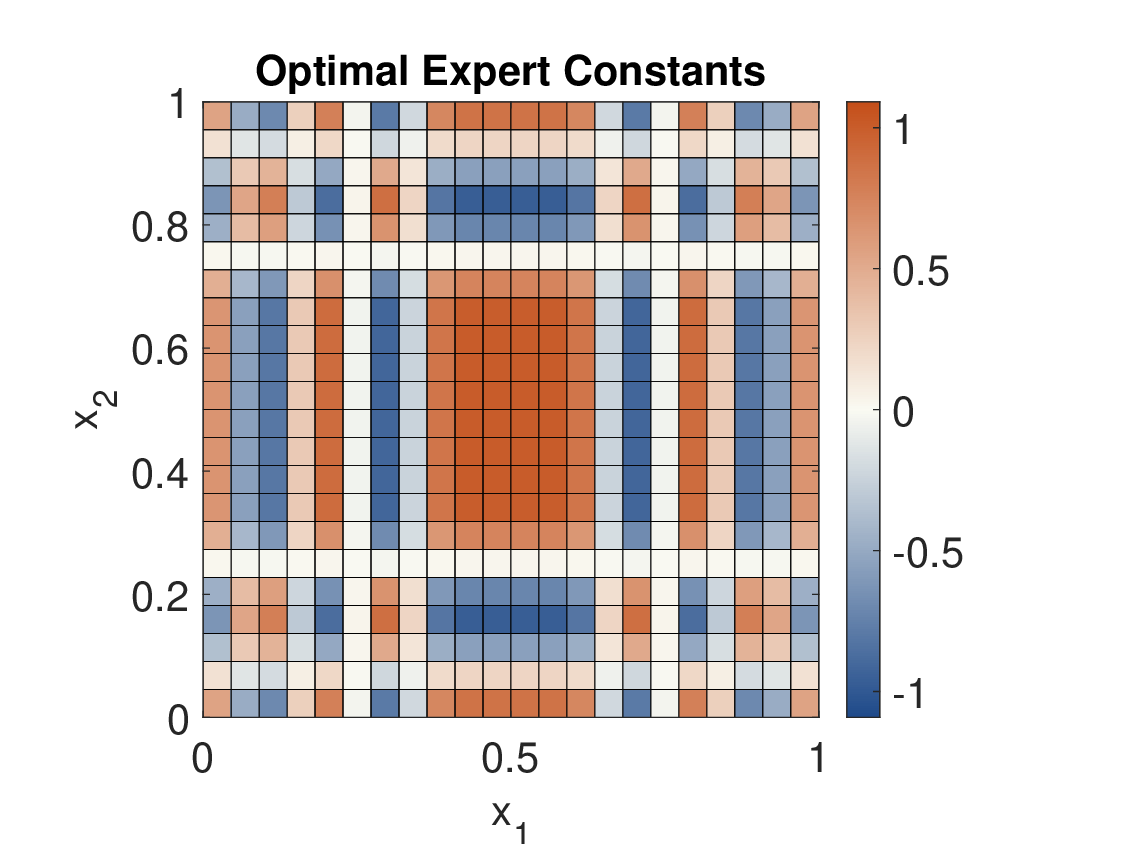}
    \label{fig:example 2 cosine with constant area - two-dimensional - learning experts for uniform segmentation - beta optimal constants m=484}
    }
    \subfigure[]{
    \includegraphics[width=0.3\linewidth]{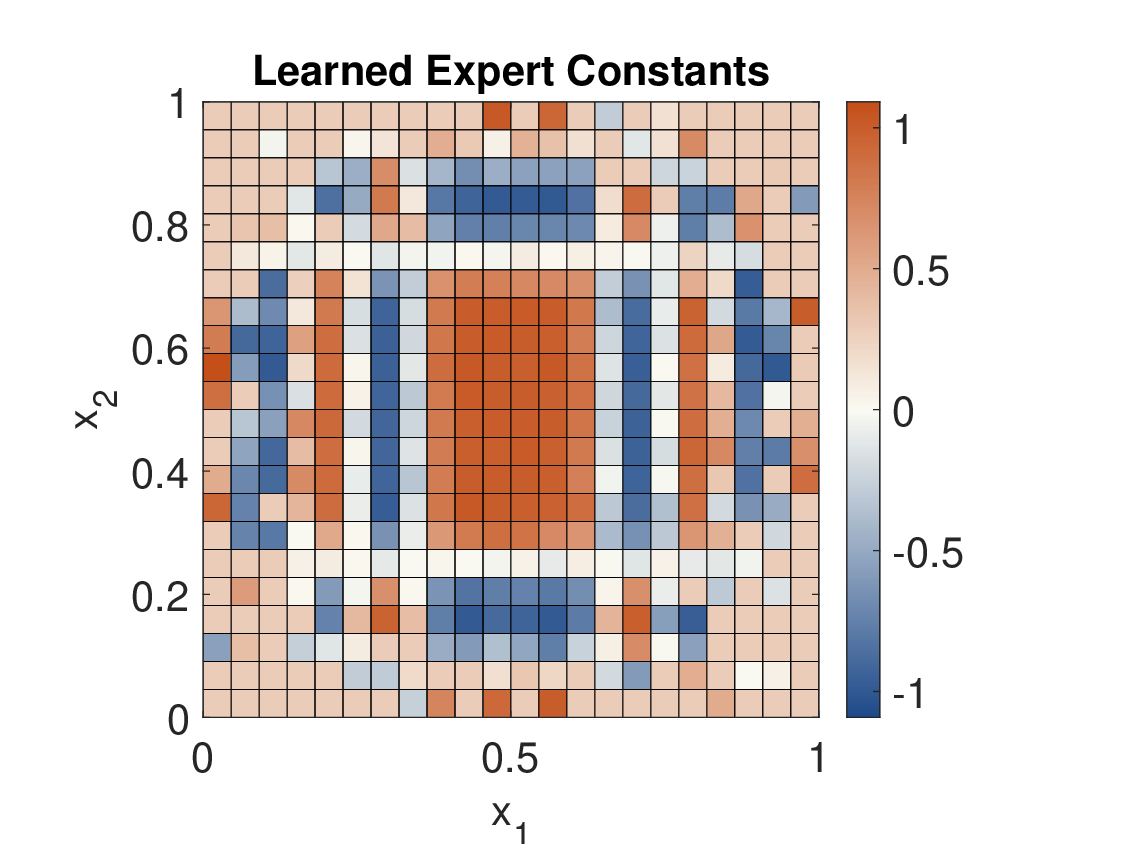}
    \label{fig:example 2 cosine with constant area - two-dimensional - learning experts for uniform segmentation - beta learned constants m=484}
    }
    \\    
    \subfigure[]{
    \includegraphics[width=0.3\linewidth]{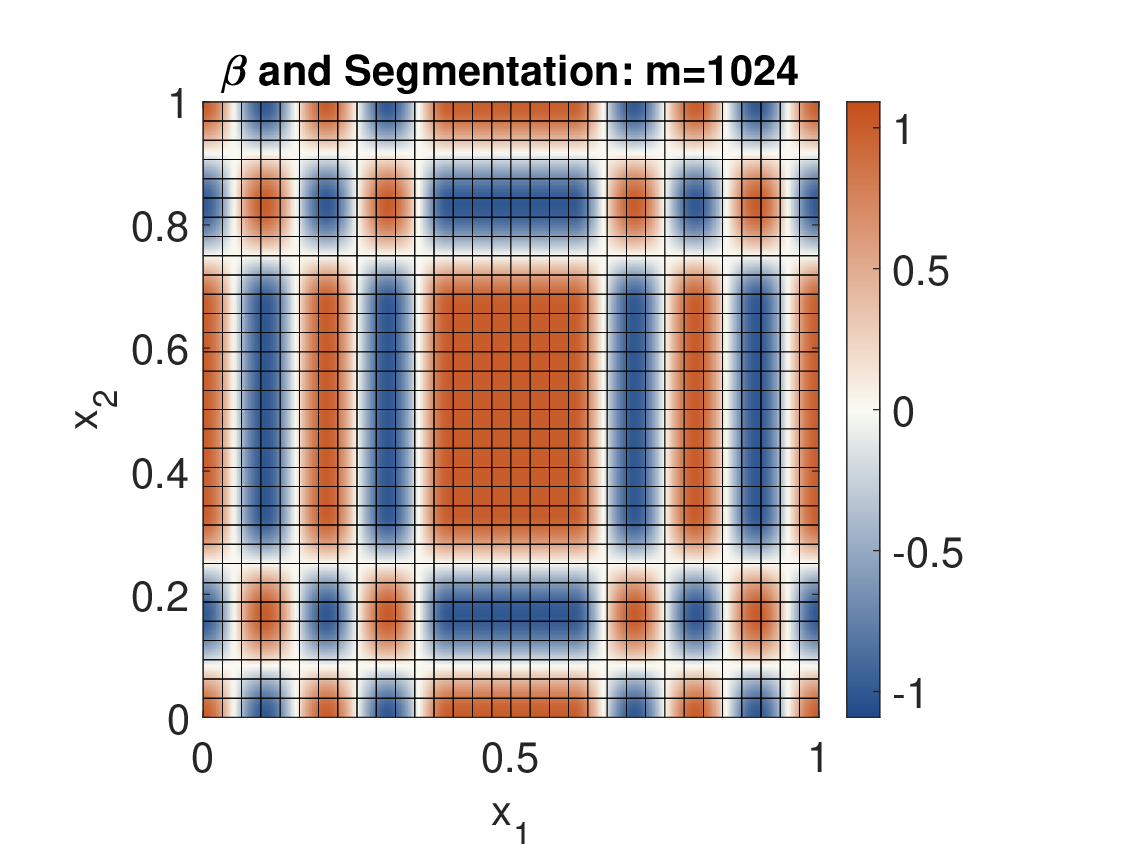}
    \label{fig:example 2 cosine with constant area - two-dimensional - learning experts for uniform segmentation - beta target and segmentation m=1024}
    }
    \subfigure[]{
    \includegraphics[width=0.3\linewidth]{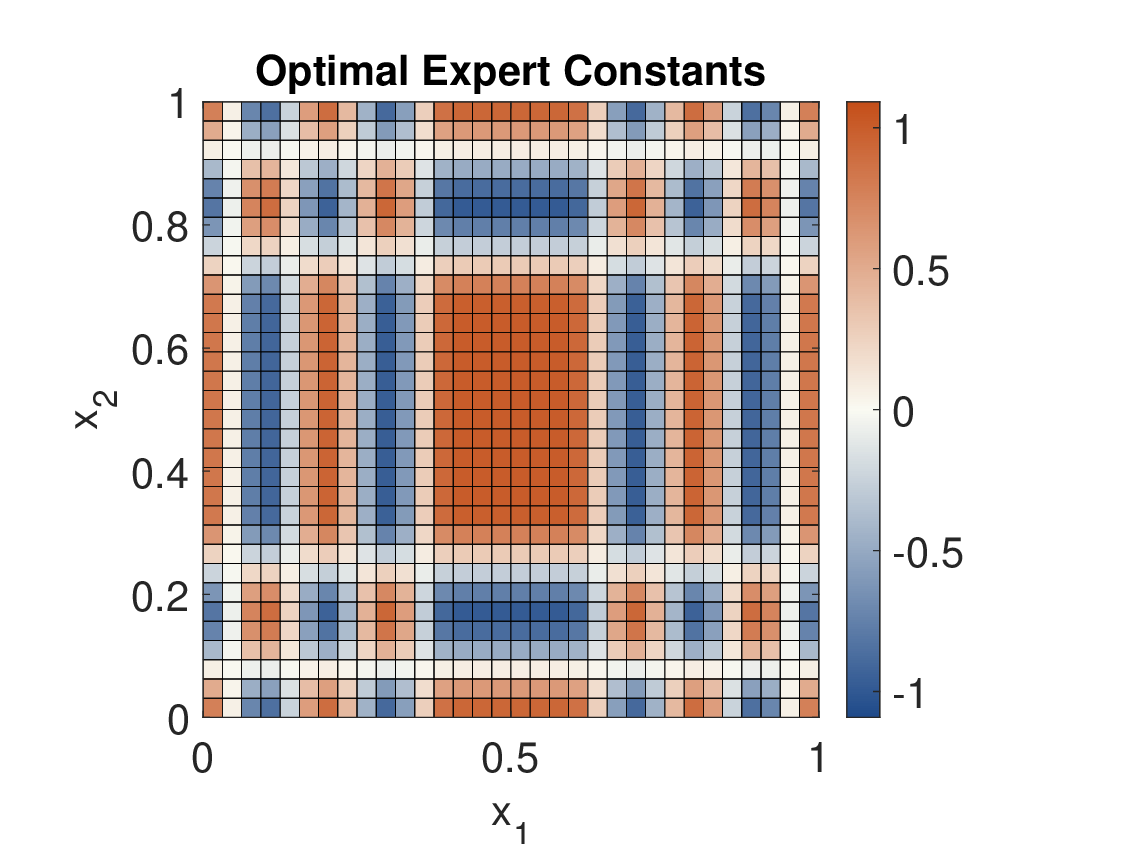}
    \label{fig:example 2 cosine with constant area - two-dimensional - learning experts for uniform segmentation - beta optimal constants m=1024}
    }
    \subfigure[]{
    \includegraphics[width=0.3\linewidth]{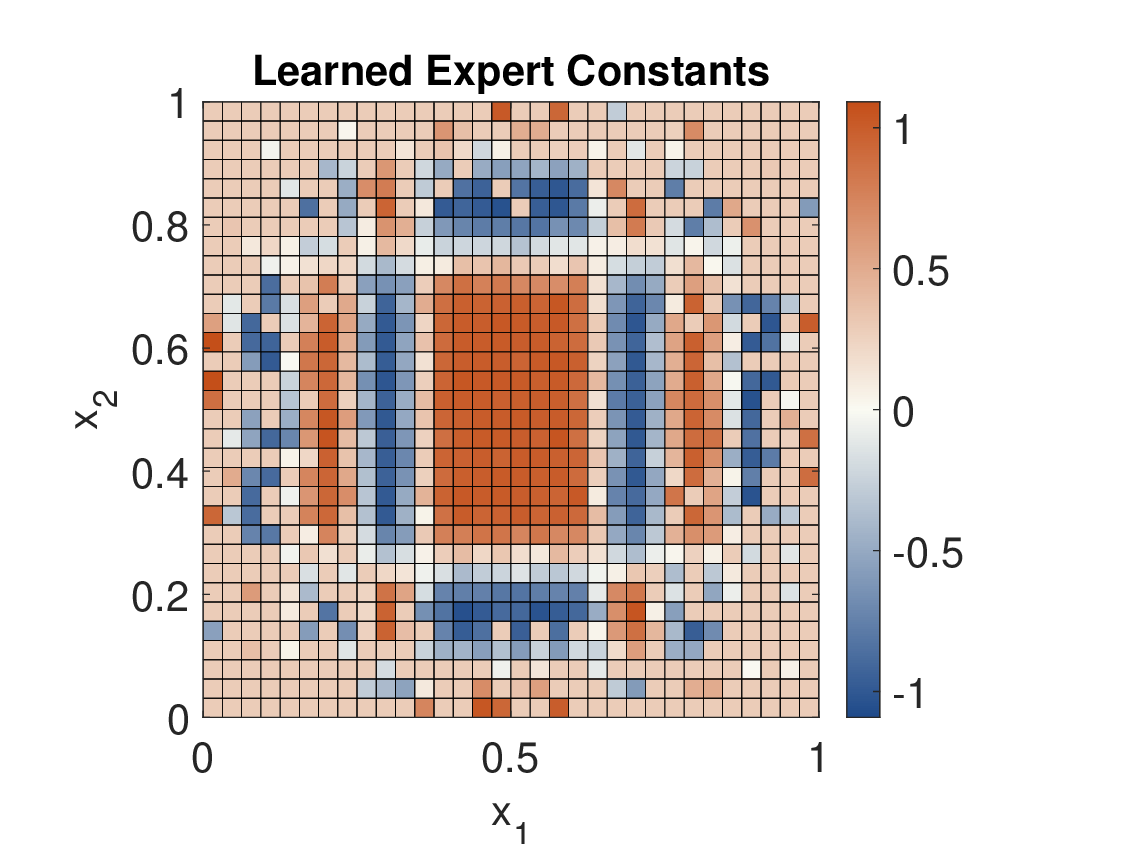}
    \label{fig:example 2 cosine with constant area - two-dimensional - learning experts for uniform segmentation - beta learned constants m=1024}
    }
    \\    
    \subfigure[]{
    \includegraphics[width=0.38\linewidth]{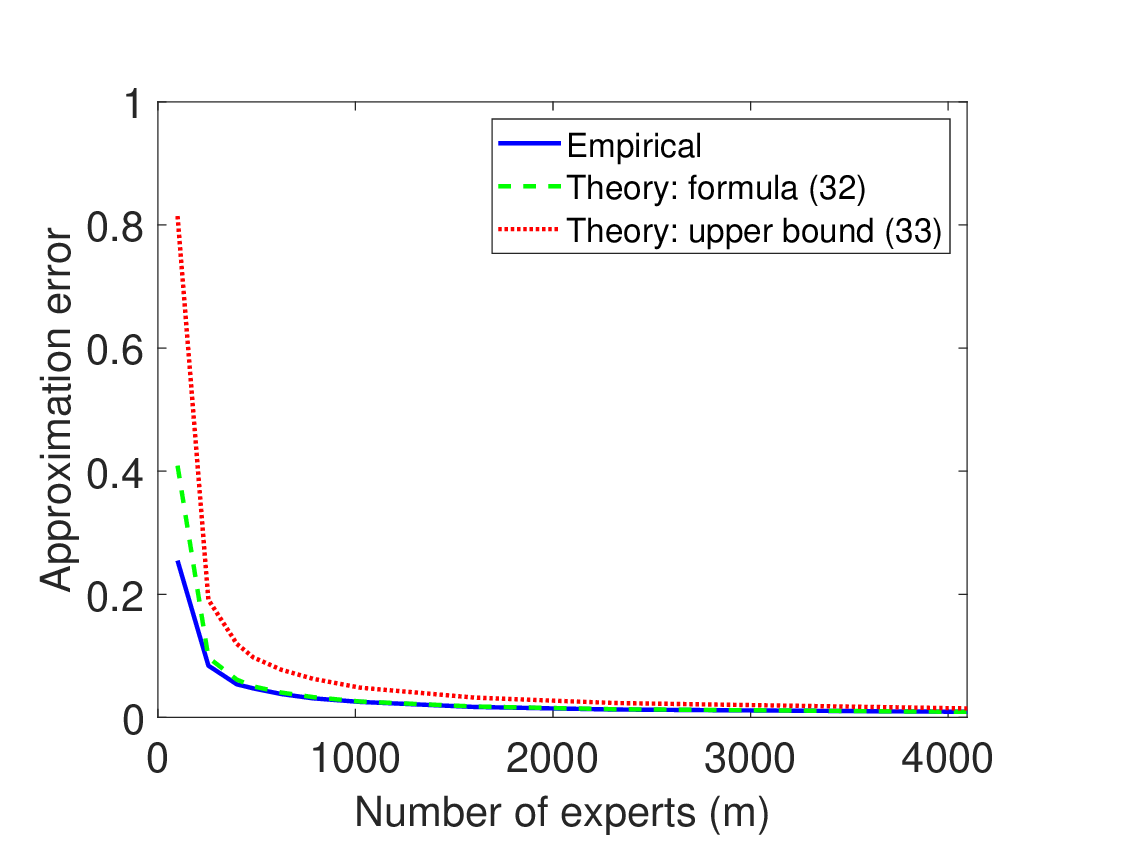}
    \label{fig:example 2 cosine with constant area - two-dimensional - learning experts for uniform segmentation - approximation error curve vs m}
    }
    \subfigure[]{
    \includegraphics[width=0.38\linewidth]{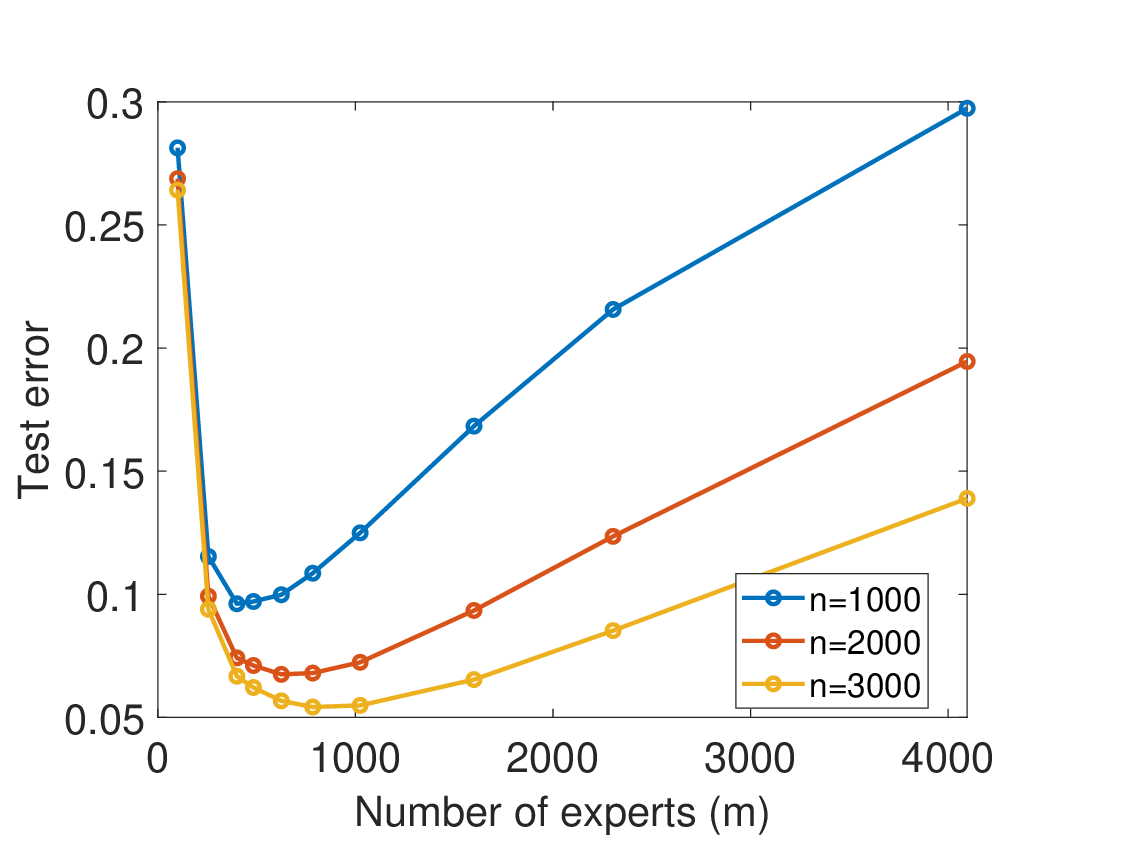}
    \label{fig:example 2 cosine with constant area - two-dimensional - learning experts for uniform segmentation - test error curve vs m}
    }
    \caption{\textbf{Learning experts for uniform segmentation in two dimensions}. This experiment is for a \textbf{cosine with a constant area} $\beta$ and truncated Gaussian $\pdfxvec$ from Figs.~\ref{fig:example 2 - two-dimensions - beta}, \ref{fig:example 2 - two-dimensions - pdf x}, respectively. 
    In the first three rows of subfigures: Each row of subfigures shows results for uniform segmentations with a different number of experts ($m=100, 484, 1024$); the left subfigures show the true $\beta$ and the uniform segmentation (borders appear in black lines), the center subfigures show the optimal expert constants, the right subfigures show the learned expert constants from a training dataset of $n=3000$ examples. 
    Subfigure (\textit{j}) shows the empirical and theoretical approximation error curves for the uniform segmentations. Subfigure (\textit{k}) shows the empirical test error vs.~$m$ curves for three sizes of training dataset $n=1000, 2000, 3000$.
    }
    \label{fig:empirical results - two-dimensional - learning experts for uniform segmentation - cosine with constant area beta}
\end{figure}

\begin{figure}[]
    \centering
    \subfigure[]{
    \includegraphics[width=0.3\linewidth]{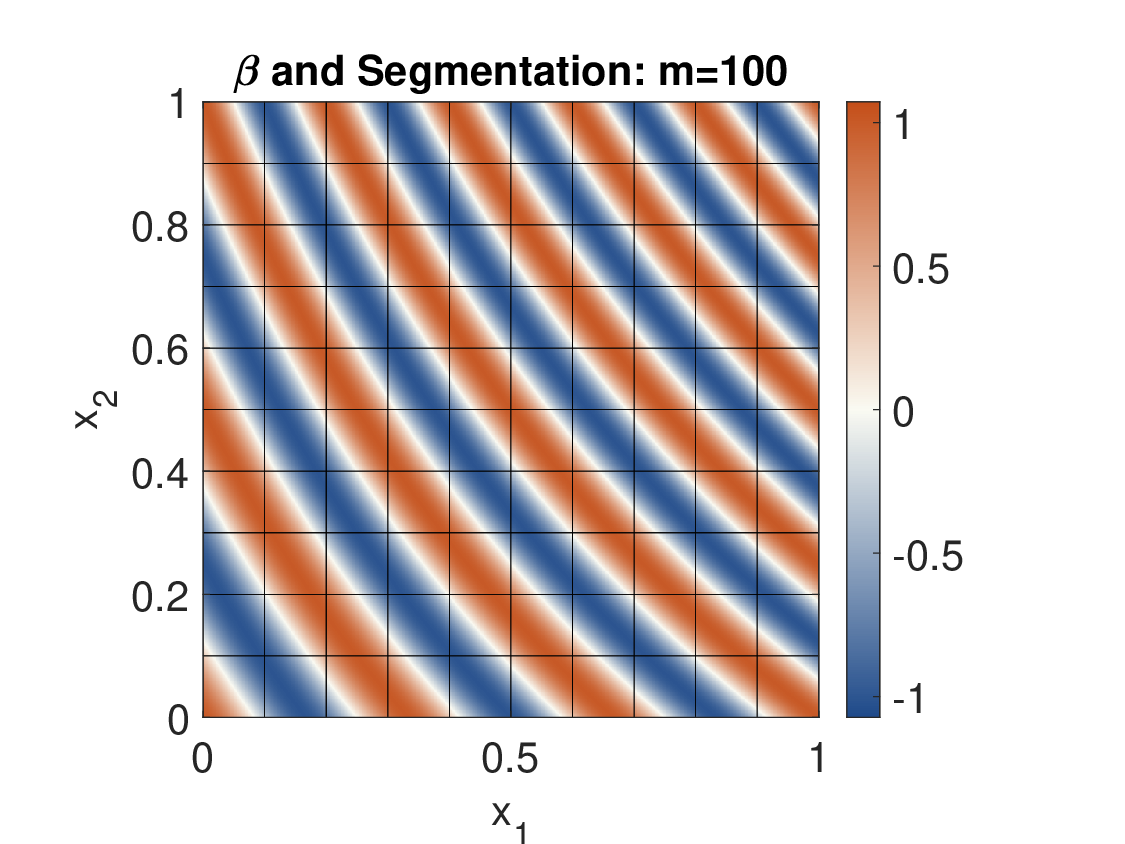}
    \label{fig:example 3 nonseparable cosine - two-dimensional - learning experts for uniform segmentation - beta target and segmentation m=100}
    }
    \subfigure[]{
    \includegraphics[width=0.3\linewidth]{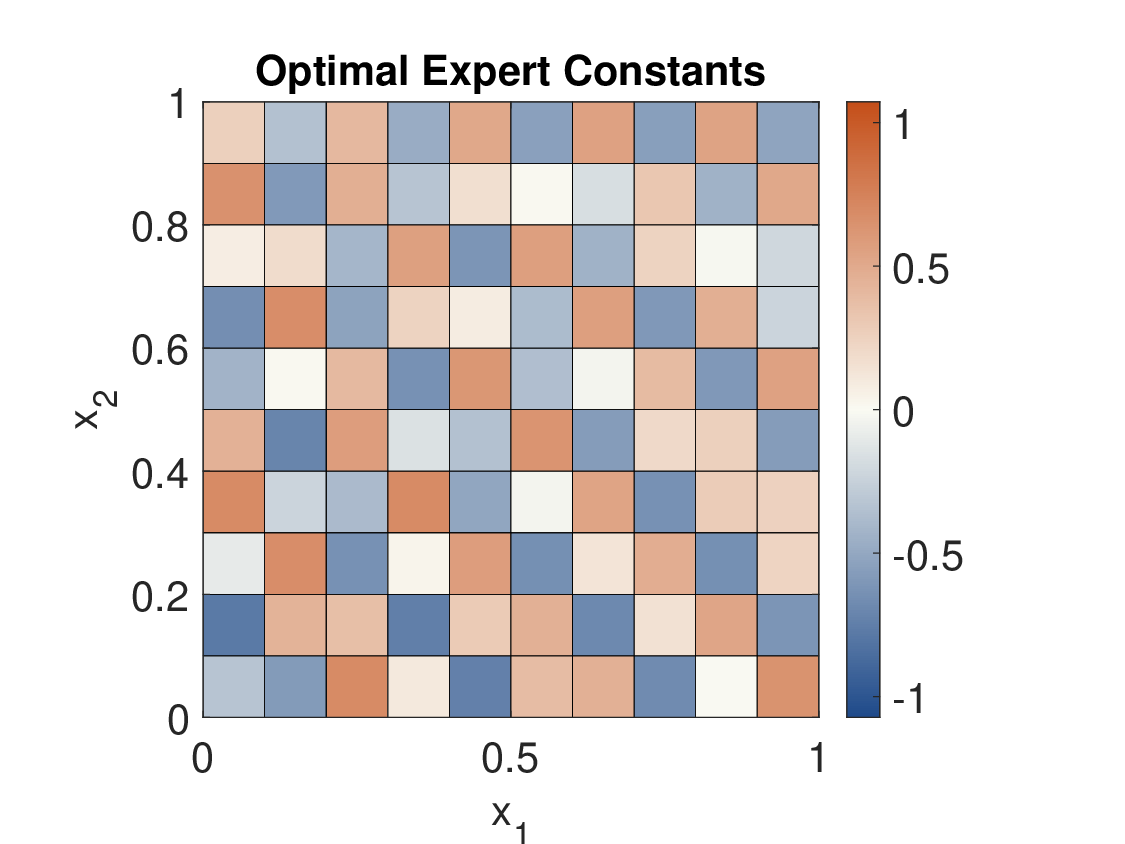}
    \label{fig:example 3 nonseparable cosine - two-dimensional - learning experts for uniform segmentation - beta optimal constants m=100}
    }
    \subfigure[]{
    \includegraphics[width=0.3\linewidth]{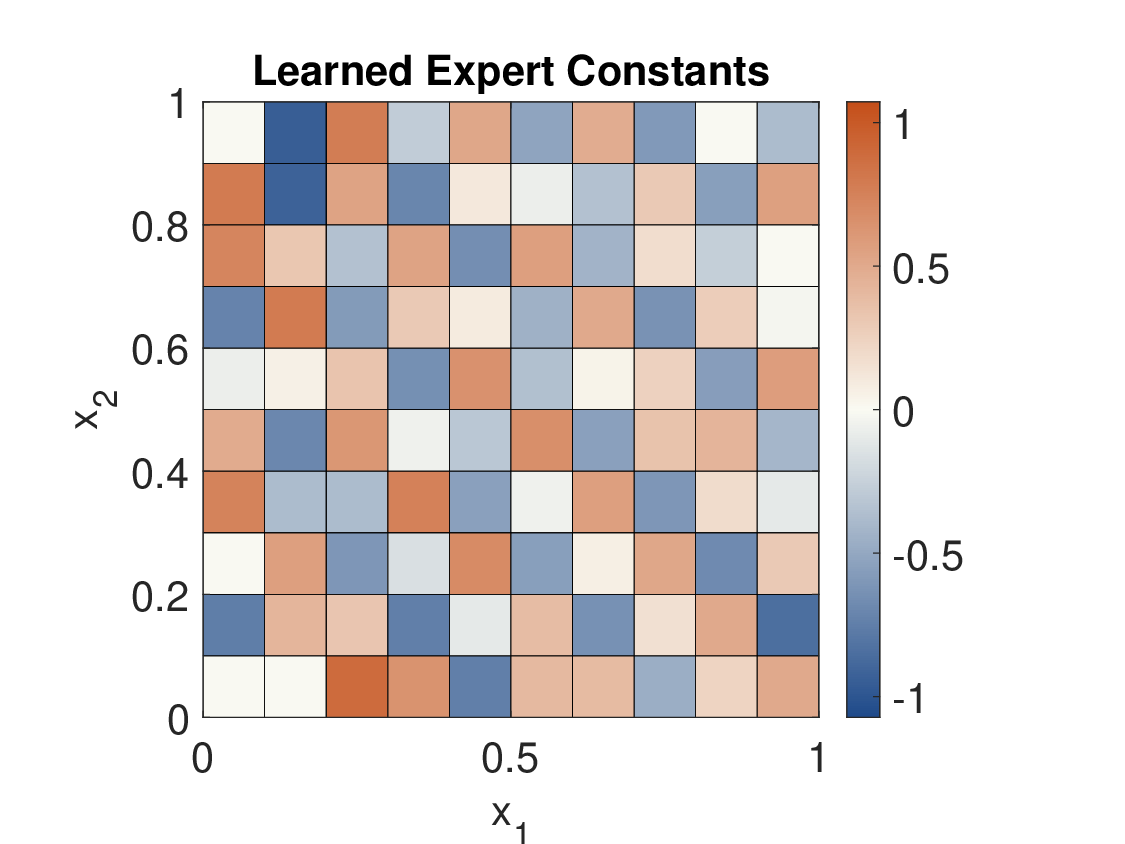}
    \label{fig:example 3 nonseparable cosine - two-dimensional - learning experts for uniform segmentation - beta learned constants m=100}
    }
    \\
    \subfigure[]{
    \includegraphics[width=0.3\linewidth]{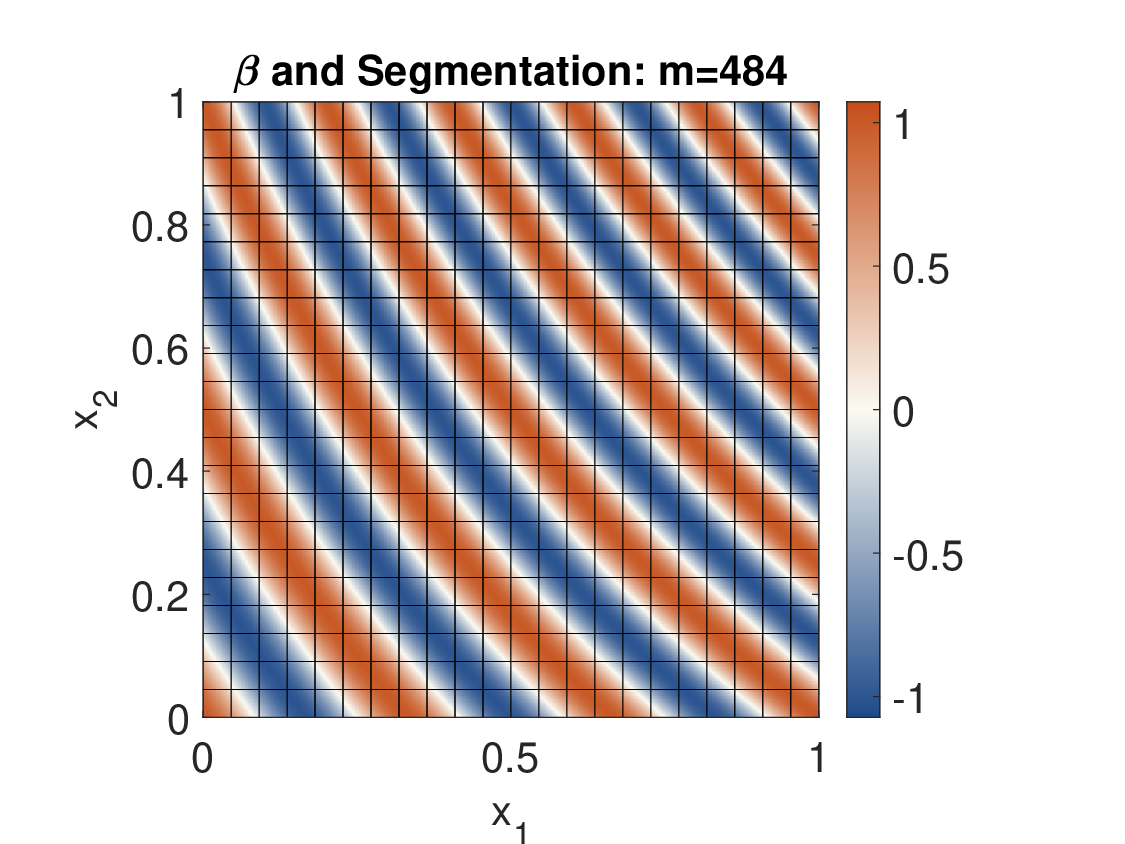}
    \label{fig:example 3 nonseparable cosine - two-dimensional - learning experts for uniform segmentation - beta target and segmentation m=484}
    }
    \subfigure[]{
    \includegraphics[width=0.3\linewidth]{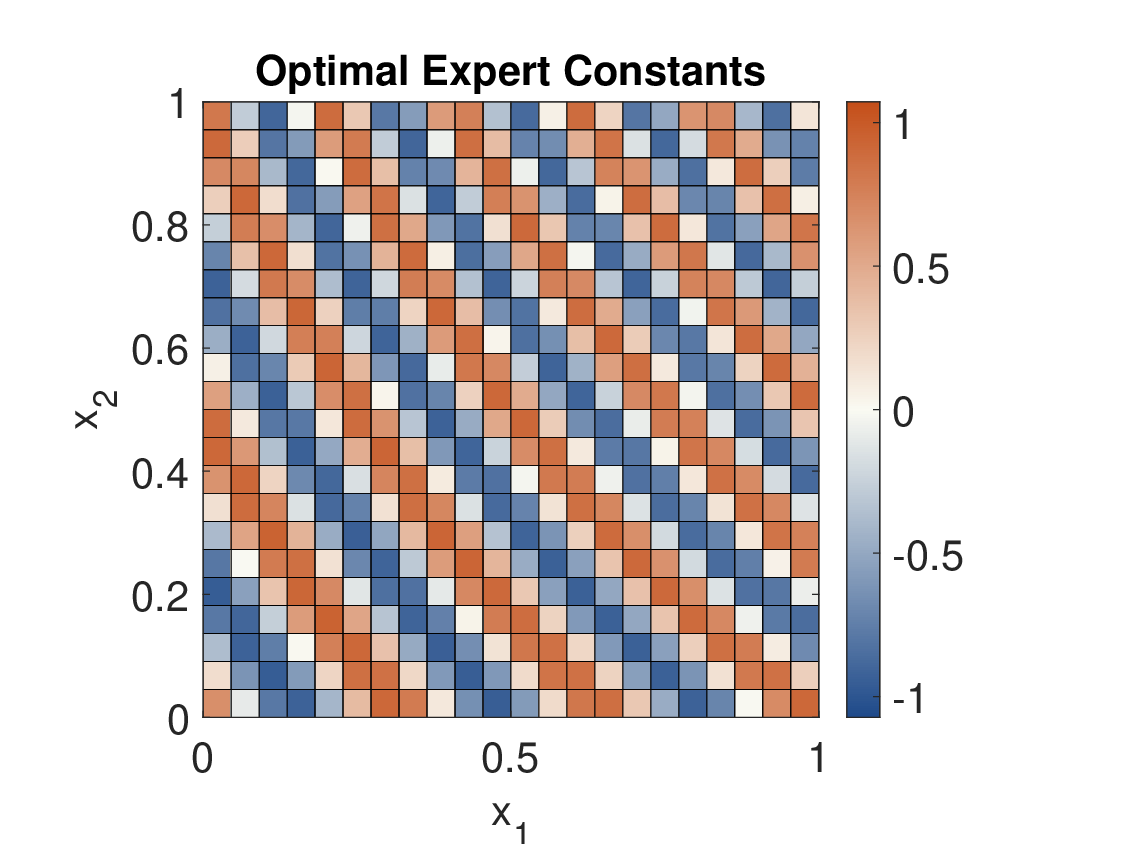}
    \label{fig:example 3 nonseparable cosine - two-dimensional - learning experts for uniform segmentation - beta optimal constants m=484}
    }
    \subfigure[]{
    \includegraphics[width=0.3\linewidth]{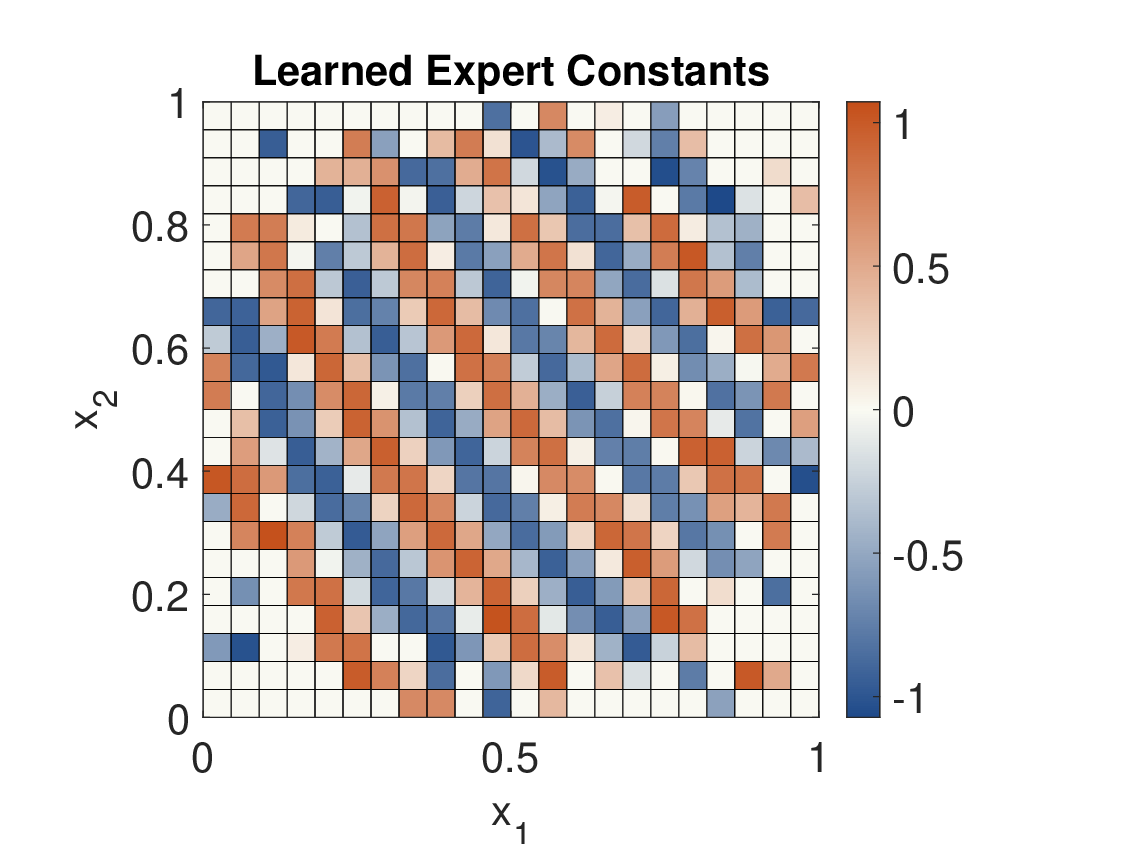}
    \label{fig:example 3 nonseparable cosine - two-dimensional - learning experts for uniform segmentation - beta learned constants m=484}
    }
    \\    
    \subfigure[]{
    \includegraphics[width=0.3\linewidth]{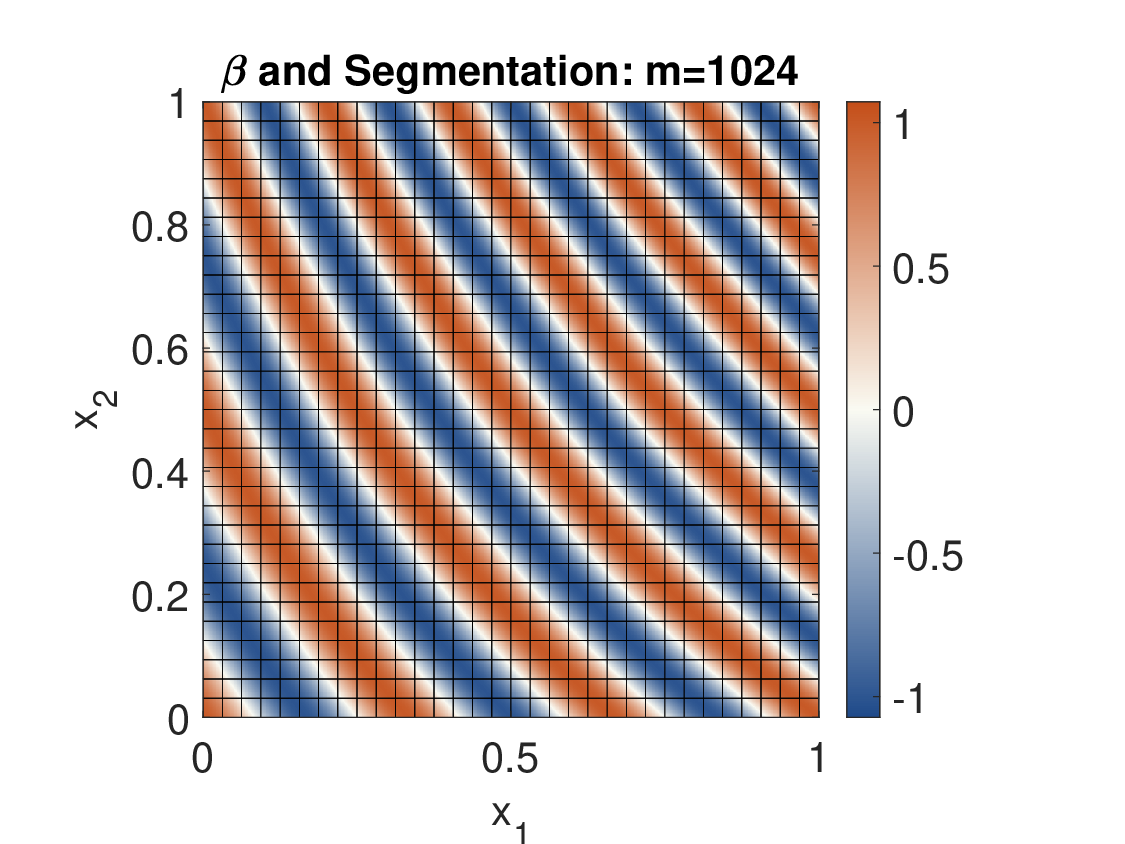}
    \label{fig:example 3 nonseparable cosine - two-dimensional - learning experts for uniform segmentation - beta target and segmentation m=1024}
    }
    \subfigure[]{
    \includegraphics[width=0.3\linewidth]{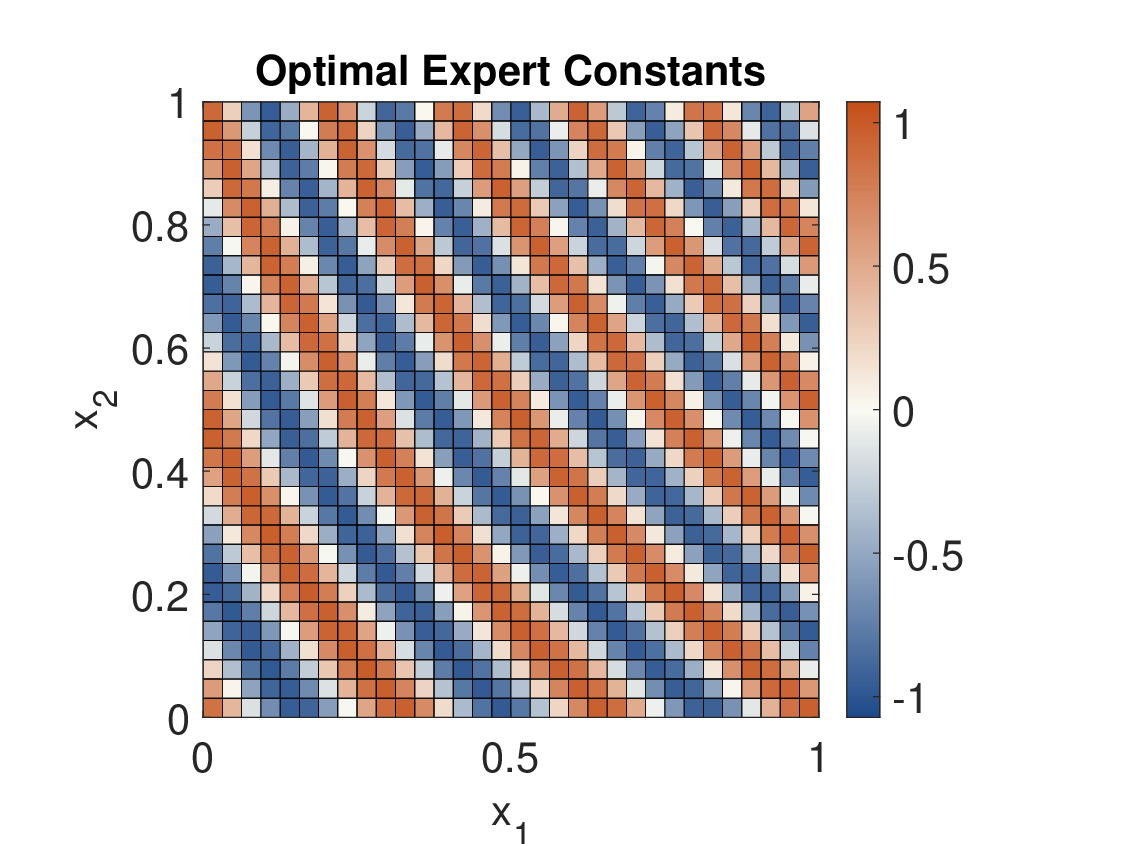}
    \label{fig:example 3 nonseparable cosine - two-dimensional - learning experts for uniform segmentation - beta optimal constants m=1024}
    }
    \subfigure[]{
    \includegraphics[width=0.3\linewidth]{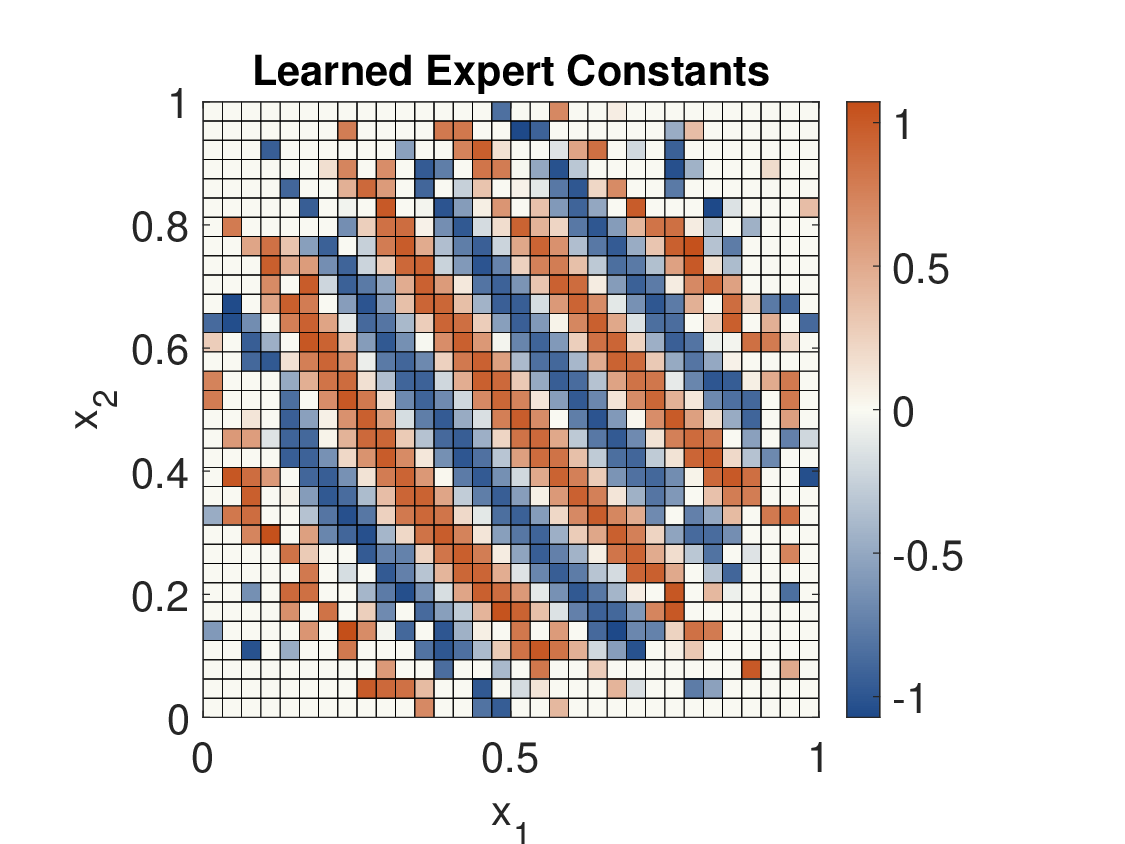}
    \label{fig:example 3 nonseparable cosine - two-dimensional - learning experts for uniform segmentation - beta learned constants m=1024}
    }
    \\    
    \subfigure[]{
    \includegraphics[width=0.38\linewidth]{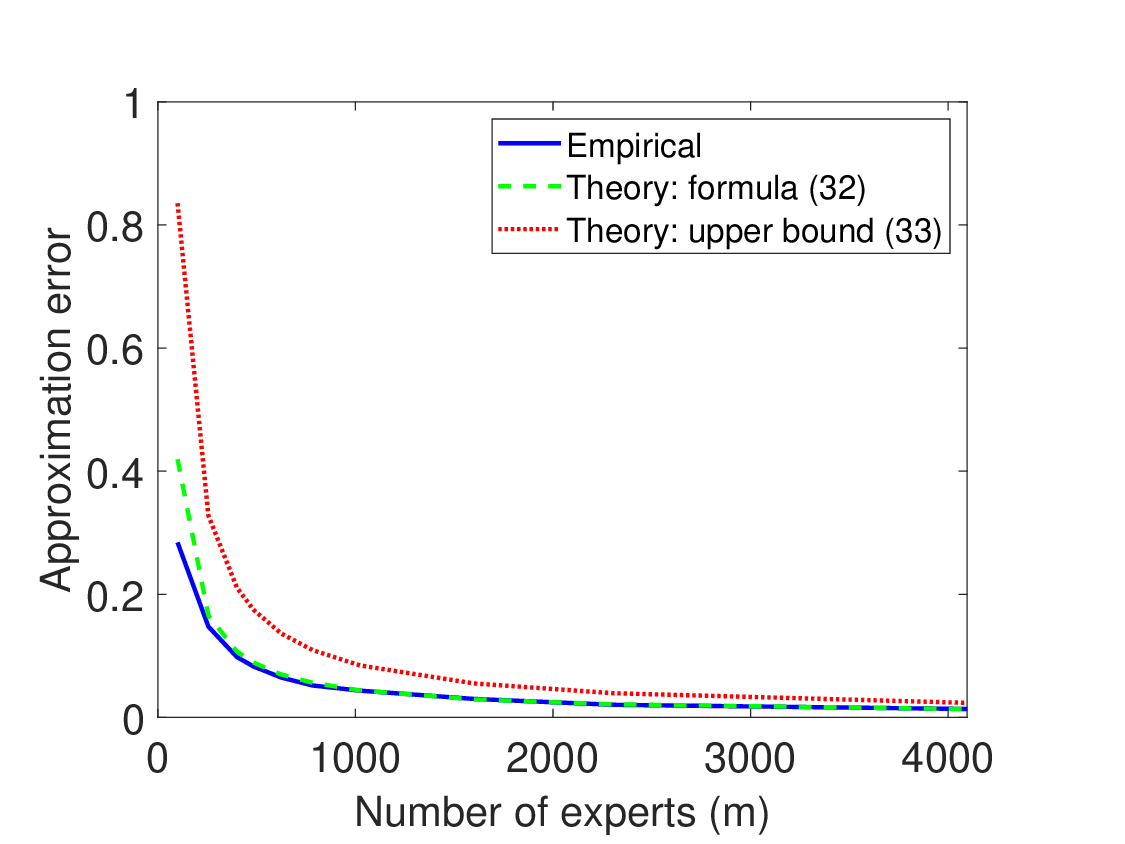}
    \label{fig:example 3 nonseparable cosine - two-dimensional - learning experts for uniform segmentation - approximation error curve vs m}
    }
    \subfigure[]{
    \includegraphics[width=0.38\linewidth]{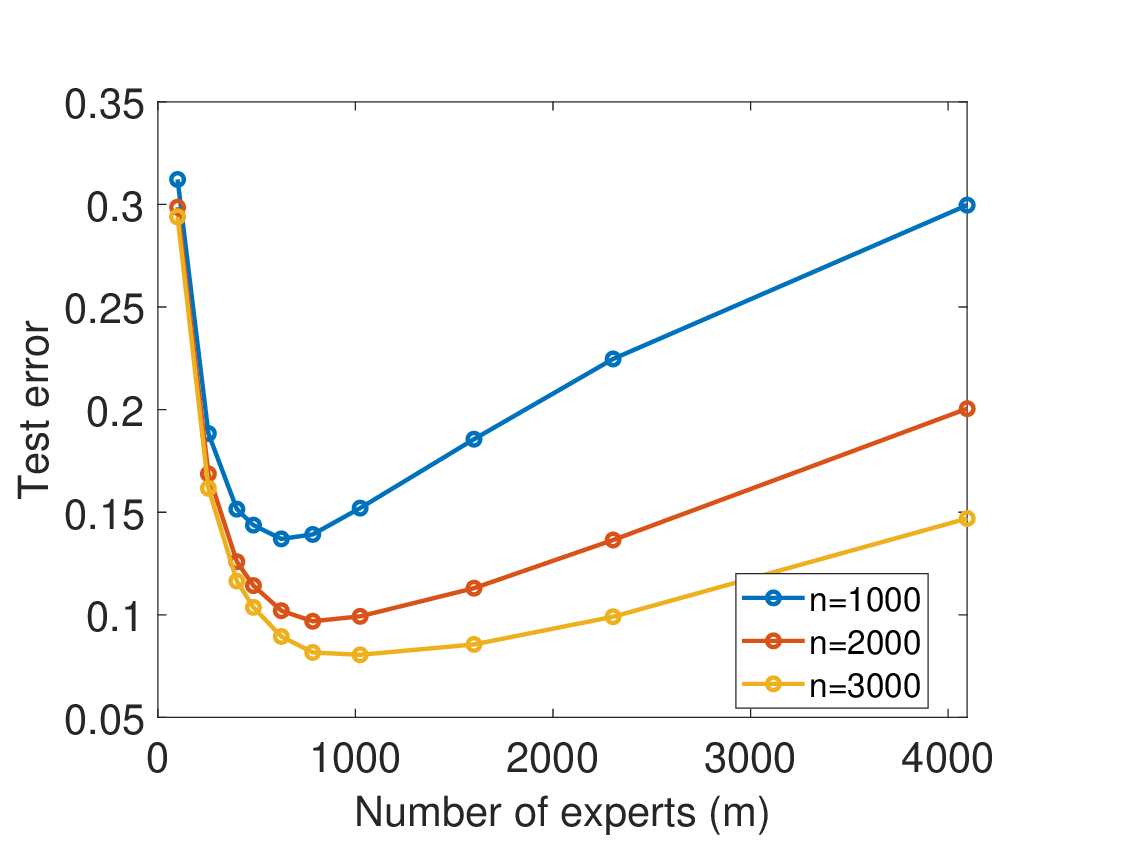}
    \label{fig:example 3 nonseparable cosine - two-dimensional - learning experts for uniform segmentation - test error curve vs m}
    }
    \caption{\textbf{Learning experts for uniform segmentation in two dimensions}. This experiment is for a \textbf{nonseparable cosine} and truncated Gaussian $\pdfxvec$ from Figs.~\ref{fig:example 3 - two-dimensions - beta}, \ref{fig:example 3 - two-dimensions - pdf x}, respectively. 
    In the first three rows of subfigures: Each row of subfigures shows results for uniform segmentations with a different number of experts ($m=100, 484, 1024$); the left subfigures show the true $\beta$ and the uniform segmentation (borders appear in black lines), the center subfigures show the optimal expert constants, the right subfigures show the learned expert constants from a training dataset of $n=3000$ examples. 
    Subfigure (\textit{j}) shows the empirical and theoretical approximation error curves for the uniform segmentations. Subfigure (\textit{k}) shows the empirical test error vs.~$m$ curves for three sizes of training dataset $n=1000, 2000, 3000$.
    }
    \label{fig:empirical results - two-dimensional - learning experts for uniform segmentation - nonseparable cosine beta}
\end{figure}

The learning of the expert constants is by the region-based least squares solution in (\ref{eq:learned expert constants - least squares}), and with the exception that expert constants of regions without training examples are set as the average of all outputs in the training dataset.
The learning experiment results are provided in Figs.~\ref{fig:empirical results - example 1 - learning experts for uniform segmentation - cosine beta}, \ref{fig:empirical results - example 2 - learning experts for uniform segmentation - cosine beta with constant segment} for the one-dimensional case, and Figs.~\ref{fig:empirical results - two-dimensional - learning experts for uniform segmentation - cosine beta}, \ref{fig:empirical results - two-dimensional - learning experts for uniform segmentation - cosine with constant area beta}, \ref{fig:empirical results - two-dimensional - learning experts for uniform segmentation - nonseparable cosine beta} for the two-dimensional case.

In the one-dimensional setting, Figs.~\ref{fig:example 1 - learning experts for uniform segmentation - beta reconstruction and segmentation m=18}-\ref{fig:example 1 - learning experts for uniform segmentation - beta reconstruction and segmentation m=50}, \ref{fig:example 2 - learning experts for uniform segmentation - beta reconstruction and segmentation m=18}-\ref{fig:example 2 - learning experts for uniform segmentation - beta reconstruction and segmentation m=50} show  the differences between the learned constants (in the learned predictors, shown as solid black lines) and the optimal constants (in the best in-class nonlearned predictors, computed by (\ref{eq:optimal c_i - one-dimensional}) for the uniform segmentation, and shown as dotted magenta lines).
Figs.~\ref{fig:example 1 - learning experts for uniform segmentation - approximation error curve vs m}, \ref{fig:example 2 - learning experts for uniform segmentation - approximation error curve vs m} show the approximation error curves for the model class $\mathcal{H}_{m,1}^{c}\left(\{a_i=\frac{i}{m}\}_{i=1}^m\right)$. There is an excellent match between the empirical evaluation (on a test dataset of 5000 examples) and the numerical evaluation of the theoretical error formula from (\ref{eq:corollary - test error integral formula - one-dimensional}) that is suitable also for suboptimal segment densities, as we have here with the uniform segmentation.

In the two-dimensional setting of Figs.~\ref{fig:empirical results - two-dimensional - learning experts for uniform segmentation - cosine beta}, \ref{fig:empirical results - two-dimensional - learning experts for uniform segmentation - cosine with constant area beta}, \ref{fig:empirical results - two-dimensional - learning experts for uniform segmentation - nonseparable cosine beta}, subfigures (c), (f), (i) show the learned constants that form the predictors using $n=3000$ training examples and three different numbers of experts $m=100, 484, 1024$. The optimal constants for these uniform segmentations are shown in subfigures (b), (e), (h); the true $\beta$ with the uniform segmentations for each of the exemplified $m$ values is shown in subfigures (a), (d), (g);  these auxiliary diagrams qualitatively demonstrate the effect of the number of experts and number of training examples on the learned constants. 

The test error curves for three different sizes of training dataset are shown in Figs.~\ref{fig:example 1 - learning experts for uniform segmentation - test error curve vs m}, \ref{fig:example 2 - learning experts for uniform segmentation - test error curve vs m} for the one-dimensional setting, and in Figs.~\ref{fig:example 1 cosine - two-dimensional - learning experts for uniform segmentation - test error curve vs m}, \ref{fig:example 2 cosine with constant area - two-dimensional - learning experts for uniform segmentation - test error curve vs m}, \ref{fig:example 3 nonseparable cosine - two-dimensional - learning experts for uniform segmentation - test error curve vs m} for the two-dimensional setting. These test error values were computed empirically as the average over 300 experiments, each with another randomly drawn training dataset. The test error curves qualitatively demonstrate the tradeoff between the approximation error and the estimation error as function of the number of experts $m$, as discussed in Section \ref{subsec:Analysis of Test Error of MoE with Learned Expert Constants}.  Specifically, the test error curve first decreases with $m$, due to the significant decrease of the approximation error that dominates the overall test error trend for sufficiently small $m$; then, the test error curve increases with $m$, due to the increase of the estimation error that dominates the overall test error trend for sufficiently large $m$. It can be observed that the minimal test error is obtained for a larger $m$ if the training dataset is larger; this is because having more training data increases the probability of having more training examples in each region, which in return allows the regions to be smaller without increasing the estimation error too much.

\section{Conclusion}
\label{sec:Conclusion}
This paper has established a new quantization-based analysis approach for MoE learning. The proposed approach uses high-rate quantization theory principles by assuming that the MoE has many constant experts, each uniquely responsible for an input-space region. These regions are assumed to be sufficiently small to allow the analysis of the discrete segmentation to regions as a continuous density, leading to a continuous analysis of the approximation error of the MoE model class for when both the experts and segmentation are learnable parameters and when only the experts are learnable parameters for a given segmentation. Moreover, our analysis eventually shows the tradeoff between the approximation error and the estimation error for MoE with learnable experts for a given segmentation. 
This paper paves the way for future research to examine the estimation error of learning the input-space segmentation, and to study more complex MoE models such as with other sparsity levels and non-constant experts.

\bibliography{mixture_of_experts_references}

\clearpage
\appendix

\section{High Rate Quantization: Formulations from Related Work}
\label{appendix:sec:High Rate Quantization - Formulations from Related Work}
This appendix provides a brief overview of some formulations and assumptions from the high-rate quantization literature. This should help to compare the formulations between the standard quantization problem and our MoE learning problem. For more details see the overview by \citet{gray1998quantization}.

\subsection{Scalar High-Rate Quantization: Approximated Error Formula}
\label{appendix:subsec:Scalar High-Rate Quantization - Approximated Error Formula}
Consider scalar quantization for a random input $\scrand{x}$ drawn from the probability density function $\pdfxsc$ over $[0,1]$. The quantizer design requires to segment the input space $[0,1]$ to $m$ intervals $\{[a_{i-1},a_i)\}_{i=1}^{m}$ where $0=a_0<a_1<\dots<a_{m-1}<a_m=1$; formally, the $m^{\sf th}$ interval is $[a_{m-1},a_m]$ but for writing simplicity it will be written as $[a_{m-1},a_m)$ as the other intervals. Each interval has a unique representation value: $v_i,~\forall i\in\{1,\dots,m\}$. The quantizer function can be written as $Q(x)=v_i$ for $x\in[a_{i-1},a_i)$. The performance criterion is the expected squared error 
\begin{equation}
    \label{appendix:scalar quantizer squared error - discrete}
    \mathcal{E}_{Q}=\expectationwrt{\left(Q(\scrand{x})-\scrand{x}\right)^2}{\scrand{x}}=\sum_{i=1}^{m} {\int\displaylimits_{x=a_{i-1}}^{a_i}{\left(v_i - x\right)^2 \pdfxsc(x)\dint x}}.
\end{equation}

Under the high-rate quantization assumption,  $m$ is large such that the intervals are sufficiently small and $\pdfxsc$ is approximately constant within each of them, i.e., 
\begin{equation}
    \label{appendix:eq:high-rate quantization assumption of locally constant pdf}
    \pdfxsc(x)\approx \pdfxsc(x_i)~~\text{for}~x\in[a_{i-1},a_i)
\end{equation}
where $x_i\triangleq \frac{a_{i-1}+a_i}{2}$ is the center of the $i^{\rm th}$ interval. This implies that, for a given segmentation, the optimal representation values are approximately the interval centers, i.e., 
\begin{equation}
    \label{appendix:eq:high-rate scalar quantization - optimal representations are interval centers}
    v_i\approx x_i,~\forall i\in\{1,\dots,m\}.
\end{equation}
Based on these, the high-rate quantization literature uses the approximated error formula: 
\begin{equation}
    \label{appendix:scalar quantizer squared error - approximate error - integration}
    \mathcal{E}_{Q}\approx\frac{1}{12m^2}\int\displaylimits_{x=0}^1{\frac{\pdfxsc(x)}{g^2(x)}\dint x} 
\end{equation}
where the smooth function $g$ has the following meanings:
\begin{itemize}
    \item In the compander design that turns a uniform quantizer (whose intervals have equal length) into a nonuniform one \citep{bennett1948spectra,panter1951quantization}, $g$ is the derivative of a nonlinear function which is applied as a preprocessing and its inverse as postprocessing to a uniform quantizer. Namely, if $Q_{\rm unif}$ is a uniform quantizer over $[0,1]$ whose quantization interval length is $\frac{1}{m}$, a nonuniform quantizer can be formed by $Q(x)=v\left(Q_{\rm unif}(u(x))\right)$ where $u$ is a preprocessing function (called compressor) and its inverse is the postprocessing function $v$ (called expander). In (\ref{appendix:scalar quantizer squared error - approximate error - integration}) we have $g(x)=u'(x)$.
    \item For nonuniform quantizers in general (i.e., also beyond the compander design), \citet{lloyd1982least} defined $g$ as the density of quantization intervals.
\end{itemize}
 It was importantly shown by \citet{panter1951quantization} that the quantization error (\ref{appendix:scalar quantizer squared error - approximate error - integration}) is minimized by 
\begin{equation}
    \label{appendix:scalar quantizer squared error - optimal density}
    g^{\rm opt} (x) = \frac{\sqrt[3]{\pdfxsc(x)}}{\int\displaylimits_{\xi=0}^{1}{\sqrt[3]{\pdfxsc(\xi)\dint\xi}}} 
\end{equation}
which attains the minimal quantization error of 
\begin{equation}
        \label{appendix:eq:optimal quantization error - one-dimensional}
        \mathcal{E}_{Q}^{\sf opt} \approx \frac{1}{12m^2}\left({ \int\displaylimits_{x=0}^{1}{\sqrt[3]{\pdfxsc(x) \dint x}} }\right)^3.
\end{equation}    

\subsection{Scalar High-Rate Quantization: Forming a Quantizer from a Quantization Segment Density}
\label{appendix:subsec:Scalar High-Rate Quantization - Forming a Quantizer from a Quantization Interval Density}
In the compander design \citep{bennett1948spectra,panter1951quantization}, the optimal compressor function is obtained by integrating $g^{\rm opt}$ from (\ref{appendix:scalar quantizer squared error - optimal density}): 
\begin{equation}
\label{appendix:eq:optimal compressor function for scalar quantization}
    u(x)=\int\displaylimits_{\xi=0}^{x}g^{\rm opt}(\xi)\dint\xi = \frac{\int\displaylimits_{\xi=0}^{x}g^{\rm opt}(\xi)\dint\xi}{\int\displaylimits_{\xi=0}^{1}g^{\rm opt}(\xi)\dint\xi}.
\end{equation}
The compander design requires also an expander function $v$, which is the inverse of the compressor $u$ from (\ref{appendix:eq:optimal compressor function for scalar quantization}): 
\begin{equation}
\label{appendix:eq:optimal expander function for scalar quantization}
    \int\displaylimits_{\xi=0}^{v(x)}g^{\rm opt}(\xi)\dint\xi = x.
\end{equation}
The equation (\ref{appendix:eq:optimal expander function for scalar quantization}) has a unique solution when $g^{\rm opt}$ is positive over $[0,1]$, here this requires that $\pdfxsc$ is positive over $[0,1]$; this will make the compressor function strictly monotonic increasing and therefore invertible. 

More generally than the compander design, the optimal quantization intervals (i.e., optimal segmentation of the input space $[0,1]$) are defined by the optimal expander from (\ref{appendix:eq:optimal expander function for scalar quantization}) as 
\begin{equation}
    \label{appendix:eq:optimal scalar quantization - optimal segmentation by expander}
    a_i^{\rm opt} = v\left(\frac{i}{m}\right),~~\forall i\in\{1,\dots,m\}.
\end{equation}

Importantly, for the multidimensional case of vector quantizers, it is generally impossible to form the optimal segmentation by extending (\ref{appendix:eq:optimal compressor function for scalar quantization})-(\ref{appendix:eq:optimal scalar quantization - optimal segmentation by expander}); this is because multidimensional quantization regions are defined by their shapes and not just by their volumes. 

More explanations about the multidimensional case of vector quantizers are provided throughout our multidimensional MoE learning analysis in Section \ref{sec:Multi-Dimensional Inputs}; for more details see Section \ref{sec:Related Work - High-Rate Vector Quantization Theory} and the detailed overview by \citet{gray1998quantization}.

\section{Related Work: Gersho's Conjecture for High-Rate Vector Quantization}
\label{appendix:sec:Related Work: Gersho Conjecture for High-Rate Vector Quantization}
Gersho's conjecture is true when the $d$-dimensional space can be optimally tessellated by a lattice; this is known to hold for $d=1$ by intervals, for $d=2$ by hexagons \citep{fejestoth1959sur,newman1982hexagon}. For $d=3$, it is informally believed that the optimal tessellation is the body-centered cubic lattice, which was proved as the optimal lattice tessellation \citep{barnes1983optimal,du2005optimal}.
For $d\ge3$, it is unknown if the optimal tessellation is by lattice and if Gersho's conjecture is accurate. Yet, as explained by \citet{gray1998quantization}, considering the relation between the conjecture to an unspecified constant from the rigorous theory by \citet{zador1982asymptotic}, the conjecture inaccuracy for $d\ge3$ is not expected to significantly affect the approximated error; moreover, a theoretical result by \citet{zamir1996lattice} implies that the relative discrepancy in the high-rate error constant resulting from the use of Gersho’s conjecture vanishes as $d\rightarrow\infty$. A detailed discussion on the conjecture is given by \citet{gray1998quantization}. 
All of these established Gersho's conjecture as a useful tool for high-rate vector quantization theory; in Section \ref{subsec:Multi-Dimensional Inputs - The Expert Segment Density that Minimizes the Test-Error Upper Bound}, Gersho’s conjecture motivates our constant moment profile assumption that we use for theoretically analyzing the segmentation that minimizes the asymptotic upper bound formula of the test error of mixture of many experts for multidimensional inputs.

\section{Proofs for Section \ref{sec:Problem and Model Definition}}

\subsection{Proof of the Test Error Formula for Multidimensional MoE in Eq.~(\ref{eq:test error - MoE - multidimensional})}
\label{appendix:subsec:Proof of the Test Error Formula for Multidimensional MoE}
The following proof is based on the data model (\ref{eq:data model - multidimensional}) and the MoE predictor form (\ref{eq:predictor - general form - multidimensional}). 
\begin{align}
    \label{appendix:eq:Proof of the Test Error Formula for Multidimensional MoE}
    \mathcal{E}_{\sf test} &= \expectationwrt{\left( \widehat{\beta}(\vecrand{x}) - \scrand{y} \right)^2}{\vecrand{x},\scrand{y}} \\ \nonumber
     &= \expectationwrt{\left( \widehat{\beta}(\vecrand{x}) - \beta(\vecrand{x}) - {\greekrand{\epsilon}} \right)^2}{\vecrand{x},\greekrand{\epsilon}} \\ \nonumber
     &= \sigma_{\epsilon}^2 + \expectationwrt{\left( \widehat{\beta}(\vecrand{x}) - \beta(\vecrand{x}) \right)^2}{\vecrand{x}} \\ \nonumber
     &= \sigma_{\epsilon}^2 + \int\displaylimits_{\vec{x}\in [0,1]^d}{ \left( \widehat{\beta}(\vecrand{x}) - \beta(\vecrand{x}) \right)^2 \pdfxvec(\vec{x})\dint\vec{x}}  \\ \nonumber
     &= \sigma_{\epsilon}^2 + \sum_{i=1}^{m} {\int\displaylimits_{\vec{x}\in A_i}{\left(c_i - \beta(\vec{x})\right)^2 \pdfxvec(\vec{x})\dint\vec{x}} }.
\end{align}

\subsection{Proof of the Optimal Constants in Eq.~(\ref{eq:optimal c_i - multidimensional})}
\label{appendix:subsec:Proof of the Optimal Constants}

For a given segmentation $\{A_i\}_{i=1}^m$, the optimal constants $\{c_i^{\sf opt}\}_{i=1}^m$ are defined by the minimization problem: 
\begin{align}
    \label{appendix:eq:optimal constants - minimization problem}
    \{c_i^{\sf opt}\}_{i=1}^m & = \underset{c_1,\dots,c_m \in\mathbb{R}}{\argmin} \mathcal{E}_{\sf test} \left({\{A_i\}_{i=1}^m, \{c_i\}_{i=1}^m}\right) \\
    & = \underset{c_1,\dots,c_m \in\mathbb{R}}{\argmin} \sum_{i=1}^{m} {\int\displaylimits_{\vec{x}\in A_i}{\left(c_i - \beta(\vec{x})\right)^2 \pdfxvec(\vec{x})\dint\vec{x}} }.
\end{align}
This is a convex minimization problem; hence, the optimal constants satisfy 
\begin{align}
    \label{appendix:eq:optimal constants - optimality condition - general}
    \frac{\partial}{\partial c_i} \mathcal{E}_{\sf test} \left({\{A_i\}_{i=1}^m, \{c_i\}_{i=1}^m}\right) = 0, ~~\forall i\in\{1,\dots,m\}
\end{align}
Our test error formula (\ref{eq:test error - MoE - multidimensional}) implies that 
\begin{align}
    \label{appendix:eq:optimal constants - optimality condition - derivative formula}
\frac{\partial}{\partial c_i} \mathcal{E}_{\sf test} \left({\{A_i\}_{i=1}^m, \{c_i\}_{i=1}^m}\right) = \int\displaylimits_{\vec{x}\in A_i}{2\left(c_i - \beta(\vec{x})\right) \pdfxvec(\vec{x})\dint\vec{x}},
\end{align}
and, therefore, (\ref{appendix:eq:optimal constants - optimality condition - general}) implies 
\begin{align}
    \label{appendix:eq:optimal constants - optimality condition - formula}
    c_i^{\sf opt} = \frac{\int\displaylimits_{\vec{x}\in A_i}{\beta(\vec{x}) \pdfxvec(\vec{x})\dint\vec{x}}}{\int\displaylimits_{\vec{x}\in A_i}{\pdfxvec(\vec{x})\dint\vec{x}}}, ~~\forall i\in\{1,\dots,m\},
\end{align}
which proves (\ref{eq:optimal c_i - multidimensional}).

\section{Proofs for the Case of One-Dimensional Inputs}
\label{appendix:sec:Proofs for the Case of One-Dimensional Inputs}
\subsection{Proof of Theorem \ref{theorem:optimal constants and test error in sum form - one-dimensional}: The Optimal Constants and Test Error in Sum Form using Locally-Linear Approximations for One-Dimensional Inputs}
\label{appendix:subsec:proof of the optimal constants and test error in sum form - one-dimensional}
Recall that $\beta$ and $\pdfxsc$ are continuously differentiable in $[0,1]$, and $\pdfxsc$ is positive valued in $[0,1]$. These, together with Assumption \ref{assumption: asymptotic sequence of segementations - one-dimensional}, enable the following developments. 

For a start, note that the first-order Taylor approximation of $\beta$ and $\pdfxsc$ from (\ref{eq:first-order Taylor approximation of beta - one-dimensional}) and (\ref{eq:first-order Taylor approximation of p - one-dimensional}) imply the approximation of their product as 
\begin{equation}
    \label{appendix:eq:first-order Taylor approximation of beta-pdfx product - one-dimensional}
    \beta(x)\pdfxsc(x) = \beta(x_i)\pdfxsc(x_i) +  \left(\beta'(x_i)\pdfxsc(x_i) + \pdfxsc'(x_i)\beta(x_i) \right)\cdot\left(x-x_i\right) + o\left(\lvert{x-x_i}\rvert\right).
\end{equation}
Then, setting the approximations (\ref{eq:first-order Taylor approximation of p - one-dimensional}), (\ref{appendix:eq:first-order Taylor approximation of beta-pdfx product - one-dimensional}) into (\ref{eq:optimal c_i - one-dimensional}) gives 
\begin{align}
    \label{appendix:eq:optimal c_i - locally linear approximation - one-dimensional}
    c_i^{\sf opt} &= \frac{\int\displaylimits_{x=a_{i-1}}^{a_i}{\left({\beta(x_i)\pdfxsc(x_i) +  \left(\beta'(x_i)\pdfxsc(x_i) + \pdfxsc'(x_i)\beta(x_i) \right)\cdot\left(x-x_i\right) + o\left(\lvert{x-x_i}\rvert\right)}\right)\dint x}}{\int\displaylimits_{x=a_{i-1}}^{a_i}{\left({\pdfxsc(x_i) +  \pdfxsc'(x_i)\cdot\left(x-x_i\right) + o\left(\lvert{x-x_i}\rvert\right)}\right)\dint x}} \nonumber \\ 
    &= \frac{\beta(x_i)\pdfxsc(x_i)\Delta_i + o\left(\Delta_i^2\right)}{\pdfxsc(x_i)\Delta_i + o\left(\Delta_i^2\right)}
\end{align}
where the following basic results were used:
\begin{align}
    \label{appendix:eq:integration of 1 is the subinterval length - one-dimensional}
    &\int\displaylimits_{x=a_{i-1}}^{a_i}\dint x = \Delta_i \\
    \label{appendix:eq:integration of x-x_i is zero - one-dimensional}
    &\int\displaylimits_{x=a_{i-1}}^{a_i} (x-x_i)\dint x = 0 \\
    \label{appendix:eq:integration of little-o of abs x-x_i is little-o of Delta_i^2 - one-dimensional}
    &\int\displaylimits_{x=a_{i-1}}^{a_i} o\left(\lvert{x-x_i}\rvert\right) \dint x \in o\left(\Delta_i^2\right) ~~\text{as}~\Delta_i\rightarrow 0.
\end{align}
The proof of (\ref{appendix:eq:integration of little-o of abs x-x_i is little-o of Delta_i^2 - one-dimensional}) is based on the little-o definition: $f(x)\in o\left(\lvert{x-x_i}\rvert\right)$ as $x\rightarrow x_i$ implies $\underset{x\rightarrow x_i}{\lim} \frac{f(x)}{\lvert{x-x_i}\rvert} = 0$ and that for every $\varepsilon>0$ there exists a $\delta>0$ such that $\lvert{x-x_i}\rvert\in(0,\delta)$ implies  $\left\lvert{ \frac{f(x)}{\lvert{x-x_i}\rvert} }\right\rvert < \varepsilon$ and hence $\left\lvert{ f(x) }\right\rvert < \varepsilon \lvert{x-x_i}\rvert$.
For an interval length $\Delta_i>0$ we have $a_i = x_i + \frac{\Delta_i}{2}$ and $a_{i-1} = x_i - \frac{\Delta_i}{2}$, then our integration over $x\in[a_{i-1},a_i)$ and the requirement $\lvert{x-x_i}\rvert\in(0,\delta)$ imply that $\frac{\Delta_i}{2}\in(0,\delta)$. Accordingly,  
\begin{align}
    \label{appendix:eq:proof of integration of little-o of abs x-x_i is little-o of Delta_i^2 - one-dimensional}
    \left\lvert{\int\displaylimits_{x=a_{i-1}}^{a_i} o\left(\lvert{x-x_i}\rvert\right)\dint x}\right\rvert &= \left\lvert{\int\displaylimits_{x=x_i - \frac{\Delta_i}{2}}^{x_i + \frac{\Delta_i}{2}} o\left(\lvert{x-x_i}\rvert\right)\dint x}\right\rvert \le \int\displaylimits_{x=x_i - \frac{\Delta_i}{2}}^{x_i + \frac{\Delta_i}{2}} \left\lvert{o\left(\lvert{x-x_i}\rvert\right)}\right\rvert \dint x  \\ \nonumber 
    &<  \int\displaylimits_{x=x_i - \frac{\Delta_i}{2}}^{x_i + \frac{\Delta_i}{2}} \varepsilon \lvert{x-x_i}\rvert \dint x = 2\varepsilon\int\displaylimits_{x=x_i}^{x_i + \frac{\Delta_i}{2}}  (x-x_i) \dint x = \varepsilon\frac{\Delta_i^2}{4} 
\end{align}
that implies 
\begin{equation}
    \frac{\left\lvert{\int\displaylimits_{x=a_{i-1}}^{a_i} o\left(\lvert{x-x_i}\rvert\right)\dint x}\right\rvert }{\Delta_i^2} < \frac{\varepsilon}{4}
\end{equation}
which holds for every $\varepsilon>0$ and its existing $\delta>0$ that were defined above. Recall that $\frac{\Delta_i}{2}\in(0,\delta)$. Consequently, we get (\ref{appendix:eq:integration of little-o of abs x-x_i is little-o of Delta_i^2 - one-dimensional}). 

We return to (\ref{appendix:eq:optimal c_i - locally linear approximation - one-dimensional}) to further develop it:
\begin{align}
    \label{appendix:eq:optimal c_i - locally linear approximation - one-dimensional - continue}
    c_i^{\sf opt} &= \frac{  \pdfxsc(x_i)\Delta_i \left({ \beta(x_i) + o\left(\Delta_i\right) }\right)}{\pdfxsc(x_i)\Delta_i \left({1 + o\left(\Delta_i\right)}\right)} = \frac{ \beta(x_i) + o\left(\Delta_i\right)}{1 + o\left(\Delta_i\right)} 
    \\
    &= \left(1 + o\left(\Delta_i\right)\right)\left({\beta(x_i) + o\left(\Delta_i\right)}\right) \label{appendix:eq: use of 1 over 1 + little o}    
    \\
    &= \beta(x_i) + o\left(\Delta_i\right)
\end{align}
where (\ref{appendix:eq: use of 1 over 1 + little o}) is due to the little-o property: if $f(x)\in o(1)$ then $\frac{1}{1+o\left(f(x)\right)} = 1+o\left(f(x)\right)$. 
By this, we complete the proof of (\ref{eq:theorem - optimal constant - one-dimensional}) in Theorem \ref{theorem:optimal constants and test error in sum form - one-dimensional}. 

The proof of the error formula (\ref{eq:theorem - test error sum formula - one-dimensional}) in Theorem \ref{theorem:optimal constants and test error in sum form - one-dimensional} is by setting the linear approximations (\ref{eq:first-order Taylor approximation of beta - one-dimensional})-(\ref{eq:first-order Taylor approximation of p - one-dimensional}) and the optimal constants for small subintervals (\ref{eq:theorem - optimal constant - one-dimensional}) in the error formula (\ref{eq:test error - MoE - one dimensional}): 
\begin{align}
\label{appendix:eq:optimal error formula development - one-dimensional}
&\mathcal{E}_{\sf test} \left({\{a_i\}_{i=0}^m, \{c_i^{\sf opt}\}_{i=1}^m}\right) = 
\\\nonumber 
&=\sigma_{\epsilon}^2 + \sum_{i=1}^{m} \int\displaylimits_{x=a_{i-1}}^{a_i}\left( \beta(x_i) + o\left(\Delta_i\right) - \left({\beta(x_i) +  \beta'(x_i)\cdot\left(x-x_i\right) + R_{\beta,1}(x)}\right)\right)^2 \\\nonumber
&\qquad\qquad\qquad\qquad\cdot\left({\pdfxsc(x_i) +  \pdfxsc'(x_i)\cdot\left(x-x_i\right) + R_{\pdfxsc,1}(x)}\right) \dint x
\\\nonumber 
&=\sigma_{\epsilon}^2 +\\\nonumber
&~~~~\sum_{i=1}^{m} {\int\displaylimits_{x=a_{i-1}}^{a_i}{\left( \beta'(x_i)\cdot\left(x-x_i\right) + R_{\beta,1}(x) + o\left(\Delta_i\right)\right)^2 \left({\pdfxsc(x_i) +  \pdfxsc'(x_i)\cdot\left(x-x_i\right) + R_{\pdfxsc,1}(x)}\right) \dint x}}
\\\nonumber 
&=\sigma_{\epsilon}^2 + \sum_{i=1}^{m} \left( {o\left(\Delta_i^3\right) + \left(\beta'(x_i)\right)^2\int\displaylimits_{x=a_{i-1}}^{a_i}{ \left(x-x_i\right)^2 \left({\pdfxsc(x_i) +  \pdfxsc'(x_i)\cdot\left(x-x_i\right)}\right) \dint x}}\right)
\\\nonumber 
&=\sigma_{\epsilon}^2 + \sum_{i=1}^{m} \left( {o\left(\Delta_i^3\right) + \left(\beta'(x_i)\right)^2 \pdfxsc(x_i)\int\displaylimits_{x=a_{i-1}}^{a_i}{ \left(x-x_i\right)^2 \dint x}}\right)
\\\nonumber 
&=\sigma_{\epsilon}^2 + \sum_{i=1}^{m} \left( {o\left(\Delta_i^3\right) + \left(\beta'(x_i)\right)^2 \pdfxsc(x_i) \cdot\frac{\Delta_i^3}{12}}\right)
\\\nonumber 
&=\sigma_{\epsilon}^2 + \sum_{i=1}^{m} \left( {\left(\beta'(x_i)\right)^2 \pdfxsc(x_i) \cdot\frac{\Delta_i^3}{12}}\right) + o\left({\sum_{i=1}^{m}\Delta_i^3}\right)
\end{align}
% where $\Delta_{\sf max}= \underset{i\in\{1,\dots,m\}}{\max} \Delta_i$ is the largest subinterval. 
The developments in (\ref{appendix:eq:optimal error formula development - one-dimensional}) use (\ref{appendix:eq:integration of x-x_i is zero - one-dimensional}) as well as 
\begin{align}
    \label{appendix:eq:integration of x-x_i cube is zero - one-dimensional}
    &\int\displaylimits_{x=a_{i-1}}^{a_i} (x-x_i)^3\dint x = 0 \\
    \label{appendix:eq:integration of little-o of abs x-x_i squared is little-o of Delta_i^3 - one-dimensional}
    &\int\displaylimits_{x=a_{i-1}}^{a_i} o\left(\lvert{x-x_i}\rvert^2\right) \dint x \in o\left(\Delta_i^3\right) ~~\text{as}~\Delta_i\rightarrow 0.
\end{align}
where (\ref{appendix:eq:integration of little-o of abs x-x_i squared is little-o of Delta_i^3 - one-dimensional}) can be proved similarly to (\ref{appendix:eq:integration of little-o of abs x-x_i is little-o of Delta_i^2 - one-dimensional}). 
By this we conclude the proof of the test error formula (\ref{eq:theorem - test error sum formula - one-dimensional}) and Theorem \ref{theorem:optimal constants and test error in sum form - one-dimensional}.

\subsection{Proof of Corollary \ref{corollary:test error integral formula - one-dimensional}}
\label{appendix:subsec:Additional Details on Corollary - test error integral formula - one-dimensional}

Following Assumption \ref{assumption: expert segment density function - one-dimensional}, define 
\begin{equation}
\label{appendix:eq:Additional Details on test error integral formula - lambda_0 - one-dimensional}
    \lambda_{\sf min} \triangleq\min_{x\in[0,1]}\lambda(x)>0.
\end{equation}
For sufficiently large $m$, $\lambda_m(x)\ge\frac{\lambda_{\sf min}}{2}$~ $\forall x\in[0,1]$; hence, 
\begin{equation}
\label{appendix:eq:Additional Details on test error integral formula - interval length bound - one-dimensional}
    \Delta_i=\frac{1}{m\lambda_m(x_i)}\le\frac{2}{m\lambda_{\sf min}}.
\end{equation}
Specifically, note that 
\begin{equation}
\label{appendix:eq:Additional Details on test error integral formula - big-O bound of Delta_max - one-dimensional}
\Delta_{\sf max}=\underset{i\in\{1,\dots,m\}}{\max} \Delta_i=\mathcal{O}(m^{-1}).    
\end{equation}
Define 
\begin{align}
    h(x)=\left({\beta'(x)}\right)^2 \pdfxsc(x) 
\end{align}
and the corresponding piecewise constant function 
\begin{align}
    h_m(x)=h(x_i)~~\forall x\in[a_{i-1},a_i).
\end{align}
Because $h$ is continuous in $[0,1]$ and $\Delta_{\sf max}\rightarrow0$, we have $h_m\rightarrow h$ uniformly. 
Moreover, Assumption \ref{assumption: expert segment density function - one-dimensional} gives $\lambda_m\rightarrow\lambda$ uniformly, with $\lambda_m(x)\ge\frac{\lambda_{\sf min}}{2}$~ $\forall x\in[0,1]$ for sufficiently large $m$. Therefore, 
\begin{equation}
    \frac{h_m}{\lambda_m^2} \rightarrow \frac{h}{\lambda^2} ~~\text{uniformly on}~[0,1]. 
\end{equation}
Using the exact definition of piecewise constant function $\lambda_m$ from Assumption \ref{assumption: expert segment density function - one-dimensional} gives 
\begin{align}
    m^2\sum_{i=1}^{m}{ \left({\beta'(x_i)}\right)^2 \pdfxsc(x_i) \Delta_i^3}&=m^2\sum_{i=1}^{m}{ h(x_i) \Delta_i^3} \nonumber\\
    &=\sum_{i=1}^{m}{\int_{x=a_{i-1}}^{a_i} \frac{h_m(x)}{\lambda_m^2(x)}\dint x } \nonumber\\
    &=\int_{x=0}^{1} \frac{h_m(x)}{\lambda_m^2(x)}\dint x \nonumber\\
    &\rightarrow \int_{x=0}^{1} \frac{h(x)}{\lambda^2(x)}\dint x=\int_{x=0}^{1}{ \frac{\left({\beta'(x)}\right)^2 \pdfxsc(x)}{\lambda^2(x)} \dint x}
    \label{appendix:eq:Additional Details on test error integral formula - limit of sum as integral - one-dimensional}
\end{align}
where the convergence of the integral is due to the uniform convergence of the integrands.

Returning to the test error formula (\ref{eq:theorem - test error sum formula - one-dimensional}), 
    \begin{align}        
    \label{appendix:eq:Additional Details on test error integral formula - test error sum formula reminder - one-dimensional}
        \mathcal{E}_{\sf test} &= \sigma_{\epsilon}^2 + \frac{1}{12}\sum_{i=1}^{m}{ \left({\beta'(x_i)}\right)^2 \pdfxsc(x_i) \Delta_i^3} + o\left({\sum_{i=1}^{m}\Delta_i^3}\right), 
    \end{align}
note that, by (\ref{appendix:eq:Additional Details on test error integral formula - big-O bound of Delta_max - one-dimensional}), 
\begin{equation}
    \sum_{i=1}^{m}\Delta_i^3\le \Delta_{\sf max}^2\sum_{i=1}^{m}\Delta_i=\mathcal{O}(m^{-2}).
\end{equation}
Hence, $o\left({\sum_{i=1}^{m}\Delta_i^3}\right)=o(m^{-2})$. Using this and (\ref{appendix:eq:Additional Details on test error integral formula - limit of sum as integral - one-dimensional}) to develop (\ref{appendix:eq:Additional Details on test error integral formula - test error sum formula reminder - one-dimensional}) gives 
    \begin{align}        
    \label{appendix:eq:Additional Details on test error integral formula - one-dimensional}
        \mathcal{E}_{\sf test} =  \sigma_{\epsilon}^2 + \frac{1}{12m^2}\int_{x=0}^{1}{ \frac{\left({\beta'(x)}\right)^2 \pdfxsc(x)}{\lambda^2(x)} \dint x} + o(m^{-2})~~~\text{as}~~m\rightarrow\infty.
    \end{align}

\subsection{Remark for Theorem \ref{theorem: optimal expert segment density function and test error - one-dimensional}}
\label{appendix:subsec: remark for theorem on optimal expert segment density function and test error - one-dimensional}
The minimization in Theorem \ref{theorem: optimal expert segment density function and test error - one-dimensional} is over all nonnegative probability densities on $[0,1]$, without imposing the smoothness and strict positivity conditions of Assumption \ref{assumption: expert segment density function - one-dimensional}. In (\ref{eq:corollary - test error integral formula - one-dimensional}), the ratio is interpreted as zero wherever both its numerator and denominator vanish, and as $\infty$ wherever only its denominator vanishes. Applying Corollary \ref{corollary:test error integral formula - one-dimensional} to the resulting optimizer requires a sequence of segmentations satisfying Assumption \ref{assumption: expert segment density function - one-dimensional} with that limiting density. Formula (\ref{eq:optimal expert segment density function - one-dimensional}) assumes that $\beta$ is nonconstant; if $\beta$ is constant, every density minimizes the asymptotic formula, which equals $\sigma_\epsilon^2$.

\subsection{Proof of Theorem \ref{theorem: optimal expert segment density function and test error - one-dimensional}: The Asymptotic-Optimal Expert Segment Density and the Corresponding Asymptotic Test Error Formula}
\label{appendix:subsec: proof of theorem on optimal expert segment density function and test error - one-dimensional}

Let us develop a lower bound for the integral $\int_{x=0}^{1}{ \frac{\left({\beta'(x)}\right)^2 \pdfxsc(x)}{\lambda^2(x)} \dint x}$ from the asymptotic test error formula (\ref{eq:corollary - test error integral formula - one-dimensional}). By H\"{o}lder's inequality, 
\begin{align}
    \label{appendix:eq:proof of theorem on optimal expert segment density function and test error - one-dimensional - 1}
    &\left({\int_{x=0}^{1}\left({ \sqrt[3]{\left({\beta'(x)}\right)^2 \pdfxsc(x)}\frac{1}{\lambda^{2/3}(x)}}\right)^3 \dint x}\right)^{1/3}  \left({\int_{x=0}^{1}\left({ \lambda^{2/3}(x)}\right)^{3/2} \dint x}\right)^{2/3} \\\nonumber 
    &\qquad\qquad\qquad\qquad\qquad\qquad\qquad\qquad\qquad\ge   \int_{x=0}^{1}\sqrt[3]{\left({\beta'(x)}\right)^2 \pdfxsc(x)} \dint x
\end{align}
where the left side can be simplified, thus, 
\begin{equation}
    \label{appendix:eq:proof of theorem on optimal expert segment density function and test error - one-dimensional - 2}
    \left({\int_{x=0}^{1} \left({\beta'(x)}\right)^2 \pdfxsc(x) \frac{1}{\lambda^2(x)} \dint x}\right)^{1/3}  \left({\int_{x=0}^{1} \lambda(x) \dint x}\right)^{2/3} \ge   \int_{x=0}^{1}\sqrt[3]{\left({\beta'(x)}\right)^2 \pdfxsc(x)} \dint x.
\end{equation}
The density $\lambda(x)$ integrates to 1, i.e., $\int_{x=0}^{1} \lambda(x) \dint x = 1$, hence, we get that 
\begin{equation}
    \label{appendix:eq:proof of theorem on optimal expert segment density function and test error - one-dimensional - 3}
    \int_{x=0}^{1} \left({\beta'(x)}\right)^2 \pdfxsc(x) \frac{1}{\lambda^2(x)} \dint x   \ge   \left({ \int_{x=0}^{1}\sqrt[3]{\left({\beta'(x)}\right)^2 \pdfxsc(x)} \dint x }\right)^3
\end{equation}
where the left side is the integral from the asymptotic test error formula (\ref{eq:corollary - test error integral formula - one-dimensional}).

H\"{o}lder's inequality in (\ref{appendix:eq:proof of theorem on optimal expert segment density function and test error - one-dimensional - 1}) is attained with equality if $\lambda(x)$ is linearly proportional to\linebreak $\left({\beta'(x)}\right)^2 \pdfxsc(x) \frac{1}{\lambda^2(x)}$. Namely, for a constant $\zeta$ such that 
\begin{align}
    \lambda(x) = \zeta \cdot \left({\beta'(x)}\right)^2 \pdfxsc(x) \frac{1}{\lambda^2(x)}
\end{align}
which, using integration over $[0,1]$ and that the density integrates to 1, implies that 
\begin{align}
    \zeta =  \frac{1}{\left({\int_{x=0}^{1}\sqrt[3]{\left({\beta'(x)}\right)^2 \pdfxsc(x) } \dint x}\right)^3}.
\end{align}
Consequently, the density that minimizes the integral in the left side of (\ref{appendix:eq:proof of theorem on optimal expert segment density function and test error - one-dimensional - 3}) is 
    \begin{equation}
        \label{appendix:eq:optimal expert segment density function - one-dimensional}
        \lambda^{\sf opt}(x)=\frac{\sqrt[3]{\pdfxsc(x)\left(\beta'(x)\right)^2 }}{\int\displaylimits_{\xi=0}^{1}{\sqrt[3]{\pdfxsc(\xi)\left(\beta'(\xi)\right)^2 \dint\xi}}}.
    \end{equation}
This $\lambda^{\sf opt}(x)$ makes  (\ref{appendix:eq:proof of theorem on optimal expert segment density function and test error - one-dimensional - 3}) to hold with equality. Then, plugging the right side of (\ref{appendix:eq:proof of theorem on optimal expert segment density function and test error - one-dimensional - 3}) into the asymptotic test error formula (\ref{eq:corollary - test error integral formula - one-dimensional}) gives the minimal asymptotic test error of 
    \begin{equation}
        \label{appendix:eq:optimal test error - one-dimensional}
        \widetilde{\mathcal{E}}_{\sf test}^{\sf opt} = \sigma_{\epsilon}^2 + \frac{1}{12m^2}\left({ \int\displaylimits_{x=0}^{1}{\sqrt[3]{\pdfxsc(x)\left(\beta'(x)\right)^2 \dint x}} }\right)^3.
    \end{equation}    
This completes the proof of Theorem \ref{theorem: optimal expert segment density function and test error - one-dimensional}.

\section{Proofs for the Case of Multidimensional Inputs}
\label{appendix:sec:Proofs for the Case of Multidimensional Inputs}

\subsection{Proof of Theorem \ref{theorem:optimal constants and test error in sum form - multidimensional}: The Optimal Constants and Test Error in Sum Form using Locally-Linear Approximations for Multidimensional Inputs}
\label{appendix:subsec:proof of the optimal constants and test error in sum form - multidimensional}
For a start, note that the first-order Taylor approximation of $\beta$ and $\pdfxsc$ from (\ref{eq:first-order Taylor approximation of beta - multidimensional}) and (\ref{eq:first-order Taylor approximation of p - multidimensional}) imply the approximation of their product as 
\begin{equation}
    \label{appendix:eq:first-order Taylor approximation of beta-pdfx product - multidimensional}
    \beta(\vec{x})\pdfxvec(\vec{x}) = \beta(\vec{x}_i)\pdfxvec(\vec{x}_i) +  \left({ \beta(\vec{x}_i)\nabla\pdfxvec(\vec{x}_i) + \pdfxvec(\vec{x}_i)\nabla\beta\left(\vec{x}_i\right) }\right)^T \left(\vec{x}-\vec{x}_i\right) + o\left(\Ltwonorm{\vec{x}-\vec{x}_i}\right).
\end{equation}
Then, setting the approximations (\ref{eq:first-order Taylor approximation of p - multidimensional}), (\ref{appendix:eq:first-order Taylor approximation of beta-pdfx product - multidimensional}) into (\ref{eq:optimal c_i - multidimensional}) gives 
\begin{align}
    \label{appendix:eq:optimal c_i - locally linear approximation - multidimensional}
    c_i^{\sf opt} &= \frac{\int\displaylimits_{\vec{x}\in A_i}{\left({\beta(\vec{x}_i)\pdfxvec(\vec{x}_i) +  \left({ \beta(\vec{x}_i)\nabla\pdfxvec(\vec{x}_i) + \pdfxvec(\vec{x}_i)\nabla\beta\left(\vec{x}_i\right) }\right)^T \left(\vec{x}-\vec{x}_i\right) + o\left(\Ltwonorm{\vec{x}-\vec{x}_i}\right)}\right)\dint\vec{x}}}{\int\displaylimits_{\vec{x}\in A_i}{\left({\pdfxvec(\vec{x}_i) +  \nabla\pdfxvec(\vec{x}_i)^T \left(\vec{x}-\vec{x}_i\right) + o\left(\Ltwonorm{\vec{x}-\vec{x}_i}\right)}\right)\dint\vec{x}}} \nonumber \\ 
    &= \frac{\beta(\vec{x}_i)\pdfxvec(\vec{x}_i)V(A_i) + o\left(V(A_i)^{1+\frac{1}{d}}\right)}{\pdfxvec(\vec{x}_i)V(A_i) + o\left(V(A_i)^{1+\frac{1}{d}}\right)}
\end{align}
where the following basic results were used:
\begin{align}
    \label{appendix:eq:integration of 1 is the region volume - multidimensional}
    &\int\displaylimits_{\vec{x}\in A_i}\dint\vec{x} = V(A_i) \\
    \label{appendix:eq:integration of x-x_i is zero - multidimensional}
    &\int\displaylimits_{\vec{x}\in A_i} (\vec{x}-\vec{x}_i)\dint\vec{x} = 0 \\
    \label{appendix:eq:integration of little-o of norm x-x_i is little-o of volume with power 1plus1overd - multidimensional}
    &\int\displaylimits_{\vec{x}\in A_i} o\left(\Ltwonorm{\vec{x}-\vec{x}_i}\right) \dint\vec{x} \in o\left(V(A_i)^{1+\frac{1}{d}}\right) ~~\text{as}~V(A_i)\rightarrow 0.
\end{align}
The proof of (\ref{appendix:eq:integration of little-o of norm x-x_i is little-o of volume with power 1plus1overd - multidimensional}) is based on the little-o definition: $f(\vec{x})\in o\left(\Ltwonorm{\vec{x}-\vec{x}_i}\right)$ as $\vec{x}\rightarrow \vec{x}_i$ implies $\underset{\vec{x}\rightarrow \vec{x}_i}{\lim} \frac{f(\vec{x})}{\Ltwonorm{\vec{x}-\vec{x}_i}} = 0$ and that for every $\varepsilon>0$ there exists a $\delta>0$ such that $\Ltwonorm{\vec{x}-\vec{x}_i}\in(0,\delta)$ implies  $\left\lvert{ \frac{f(\vec{x})}{\Ltwonorm{\vec{x}-\vec{x}_i}} }\right\rvert < \varepsilon$ and hence $\left\lvert{ f(\vec{x}) }\right\rvert < \varepsilon \Ltwonorm{\vec{x}-\vec{x}_i}$.
Recall that $\vec{x}_i$ is the center of the region $A_i$ and that we are interested in integration over $\vec{x}\in A_i$.
Accordingly,  
\begin{align}
    \label{appendix:eq:proof of integration of little-o of norm x-x_i is little-o of volume with power 1plus1overd - multidimensional}
    \left\lvert{\int\displaylimits_{\vec{x}\in A_i} o\left(\Ltwonorm{\vec{x}-\vec{x}_i}\right)\dint\vec{x}}\right\rvert & \le \int\displaylimits_{\vec{x}\in A_i} \left\lvert{o\left(\Ltwonorm{\vec{x}-\vec{x}_i}\right)}\right\rvert \dint\vec{x}  \nonumber\\  
    &<  \int\displaylimits_{\vec{x}\in A_i} \varepsilon \Ltwonorm{\vec{x}-\vec{x}_i} \dint\vec{x}.
\end{align}
Using the definition of the normalized $k^{\sf th}$ moment of inertia, 
\begin{equation}
\label{appendix:eq:normalized k-moment of inertia - definition}
    M_k(A_i) \triangleq \frac{\int\displaylimits_{\vec{x}\in A_i}\Ltwonorm{\vec{x}-\vec{x}_i}^k \dint\vec{x}}{d\cdot V(A_i)^{1+\frac{k}{d}}} 
\end{equation}
for $k=1$, we get $\int\displaylimits_{\vec{x}\in A_i}\Ltwonorm{\vec{x}-\vec{x}_i} \dint\vec{x}=M_1(A_i)\cdot d\cdot V(A_i)^{1+\frac{1}{d}}$ and set it in (\ref{appendix:eq:proof of integration of little-o of norm x-x_i is little-o of volume with power 1plus1overd - multidimensional}) to yield 
\begin{align}
    \left\lvert{\int\displaylimits_{\vec{x}\in A_i} o\left(\Ltwonorm{\vec{x}-\vec{x}_i}\right)\dint\vec{x}}\right\rvert < \varepsilon M_1(A_i)\cdot d \cdot V(A_i)^{1+\frac{1}{d}},
\end{align}
i.e., 
\begin{align}
    \label{appendix:eq:proof of integration of little-o of norm x-x_i is little-o of volume with power 1plus1overd - 2 - multidimensional}
    \frac{\left\lvert{\int\displaylimits_{\vec{x}\in A_i} o\left(\Ltwonorm{\vec{x}-\vec{x}_i}\right)\dint\vec{x}}\right\rvert}{V(A_i)^{1+\frac{1}{d}}} < \varepsilon M_1(A_i)\cdot d,
\end{align}
which holds for every $\varepsilon>0$ and its existing $\delta$ that were defined above. 

By the definition (\ref{appendix:eq:normalized k-moment of inertia - definition}) for first and second moments, and Cauchy-Schwarz inequality, we get  
\begin{equation}
    M_1(A_i)\le\sqrt{\frac{M(A_i)}{d}}\le\sqrt{\frac{\phi_M}{d}}
\end{equation}
where 
\begin{equation}
\label{appendix:eq:definition of phi_M bounding constant of second momemt}
\phi_M=\sup\limits_{m,i} M\left(A_i^{(m)}\right)<\infty    
\end{equation}
follows from Assumption \ref{assumption:finite moment of inertia}. This turns (\ref{appendix:eq:proof of integration of little-o of norm x-x_i is little-o of volume with power 1plus1overd - 2 - multidimensional}) into the following upper bound uniformly over the regions: 
\begin{align}
    \label{appendix:eq:proof of integration of little-o of norm x-x_i is little-o of volume with power 1plus1overd - 2 UNIFORM - multidimensional}
    \frac{\left\lvert{\int\displaylimits_{\vec{x}\in A_i} o\left(\Ltwonorm{\vec{x}-\vec{x}_i}\right)\dint\vec{x}}\right\rvert}{V(A_i)^{1+\frac{1}{d}}} < \varepsilon \sqrt{d\phi_M}.
\end{align}
Assumption \ref{assumption: vanishing region diameter} ensures that the regions lie within the required Taylor neighborhoods, uniformly as $m\rightarrow\infty$. The continuous differentiability assumptions from Section \ref{sec:Problem and Model Definition} make the Taylor remainder estimates uniform as well. Together, these facts justify (\ref{appendix:eq:integration of little-o of norm x-x_i is little-o of volume with power 1plus1overd - multidimensional}).

Let us return to (\ref{appendix:eq:optimal c_i - locally linear approximation - multidimensional}) to further develop it, 
\begin{align}
    \label{appendix:eq:optimal c_i - locally linear approximation - multidimensional - continue}
    c_i^{\sf opt} &= \frac{\pdfxvec(\vec{x}_i)V(A_i) \left({\beta(\vec{x}_i) + o\left(V(A_i)^{\frac{1}{d}}\right)}\right)}{\pdfxvec(\vec{x}_i)V(A_i)\left(1 + o\left(V(A_i)^{\frac{1}{d}}\right)\right)} = \frac{\beta(\vec{x}_i) + o\left(V(A_i)^{\frac{1}{d}}\right)}{1 + o\left(V(A_i)^{\frac{1}{d}}\right)}
    \\
    &= \left(1 + o\left(V(A_i)^{\frac{1}{d}}\right)\right)\left({\beta(\vec{x}_i) + o\left(V(A_i)^{\frac{1}{d}}\right)}\right) \label{appendix:eq: use of 1 over 1 + little o - multidimensional}    
    \\
    &= \beta(\vec{x}_i) + o\left(V(A_i)^{\frac{1}{d}}\right)
\end{align}
where (\ref{appendix:eq: use of 1 over 1 + little o - multidimensional}) is due to the little-o property: if $f(\vec{x})\in o(1)$ then $\frac{1}{1+o\left(f(\vec{x})\right)} = 1+o\left(f(\vec{x})\right)$. 
By this, we complete the proof of (\ref{eq:theorem - optimal constant - multidimensional}) in Theorem \ref{theorem:optimal constants and test error in sum form - multidimensional}. 

The proof of the error formula (\ref{eq:theorem - test error sum formula - equality - multidimensional}) in Theorem \ref{theorem:optimal constants and test error in sum form - multidimensional} is by setting the linear approximations (\ref{eq:first-order Taylor approximation of beta - multidimensional})-(\ref{eq:first-order Taylor approximation of p - multidimensional}) and the optimal constants for small regions (\ref{eq:theorem - optimal constant - multidimensional}) in the error formula (\ref{eq:test error - MoE - multidimensional}): 
\begin{align}
&\mathcal{E}_{\sf test} \left({\{A_i\}_{i=1}^m, \{c_i^{\sf opt}\}_{i=1}^m}\right) =  
\\\nonumber 
&=\sigma_{\epsilon}^2 + \sum_{i=1}^{m} \int\displaylimits_{\vec{x}\in A_i}\left( \beta(\vec{x}_i) + o\left(V(A_i)^{\frac{1}{d}}\right) - \left({ \beta(\vec{x}_i) +  \nabla\beta(\vec{x}_i)^T\left(\vec{x}-\vec{x}_i\right) + R_{\beta,1}(\vec{x}) }\right)\right)^2\\\nonumber
&\qquad\qquad\qquad\qquad\cdot\left({ \pdfxvec(\vec{x}_i) +  \nabla\pdfxvec(\vec{x}_i)^T \left(\vec{x}-\vec{x}_i\right) + R_{\pdfxvec,1}(\vec{x}) }\right) \dint\vec{x}
\\\nonumber 
&=\sigma_{\epsilon}^2 + \sum_{i=1}^{m} \int\displaylimits_{\vec{x}\in A_i}\left( \nabla\beta(\vec{x}_i)^T\left(\vec{x}-\vec{x}_i\right) + R_{\beta,1}(\vec{x}) + o\left(V(A_i)^{\frac{1}{d}}\right)\right)^2\\\nonumber
&\qquad\qquad\qquad\qquad\cdot \left({ \pdfxvec(\vec{x}_i) +  \nabla\pdfxvec(\vec{x}_i)^T \left(\vec{x}-\vec{x}_i\right) + R_{\pdfxvec,1}(\vec{x}) }\right) \dint\vec{x}
\\\label{appendix:eq:optimal error formula development - multidimensional - 0}
&=\sigma_{\epsilon}^2 + \sum_{i=1}^{m} \left( {o\left(V(A_i)^{1+\frac{2}{d}}\right) + \int\displaylimits_{\vec{x}\in A_i}{ \left(\nabla\beta(\vec{x}_i)^T\left(\vec{x}-\vec{x}_i\right)\right)^2 \left({ \pdfxvec(\vec{x}_i) +  \nabla\pdfxvec(\vec{x}_i)^T \left(\vec{x}-\vec{x}_i\right) }\right) \dint\vec{x}}}\right)
\\ \label{appendix:eq:optimal error formula development - multidimensional - 1}
&=\sigma_{\epsilon}^2 + \sum_{i=1}^{m} \left( o\left(V(A_i)^{1+\frac{2}{d}}\right) + \pdfxvec(\vec{x}_i)\int\displaylimits_{\vec{x}\in A_i}{ \left(\nabla\beta(\vec{x}_i)^T\left(\vec{x}-\vec{x}_i\right)\right)^2\dint\vec{x}} \right.\\\nonumber &\qquad\qquad\qquad\left.+ \int\displaylimits_{\vec{x}\in A_i}{ \left(\nabla\beta(\vec{x}_i)^T\left(\vec{x}-\vec{x}_i\right)\right)^2 \left({ \nabla\pdfxvec(\vec{x}_i)^T \left(\vec{x}-\vec{x}_i\right) }\right) \dint\vec{x}}\right).
\end{align}
The detailed proof of the equality (\ref{appendix:eq:optimal error formula development - multidimensional - 0}) is lengthy and therefore omitted here; yet, the proof is based on little-o expressions for various integrals over products of little-o terms and moment of inertia, which are technically similar to other parts of the proof that are explicitly shown in this appendix. 

To continue the proof, note that 
\begin{equation}
\label{appendix:eq:outer product equality for beta}
    \left(\nabla\beta(\vec{x}_i)^T\left(\vec{x}-\vec{x}_i\right)\right)^2= \nabla\beta(\vec{x}_i)^T\left(\vec{x}-\vec{x}_i\right)\left(\vec{x}-\vec{x}_i\right)^T\nabla\beta(\vec{x}_i)
\end{equation}
and therefore (\ref{appendix:eq:optimal error formula development - multidimensional - 1}) can be written as 
\begin{align}
\label{appendix:eq:optimal error formula development - multidimensional - equality - 1}
    &\mathcal{E}_{\sf test} \left({\{A_i\}_{i=1}^m, \{c_i^{\sf opt}\}_{i=1}^m}\right) = \sigma_{\epsilon}^2 + \sum_{i=1}^{m} \left( o\left(V(A_i)^{1+\frac{2}{d}}\right) + \pdfxvec(\vec{x}_i)\nabla\beta(\vec{x}_i)^T\mtx{J}_i\nabla\beta(\vec{x}_i) \right)+R_{\sf tot}
\end{align}
where $\mtx{J}_i\triangleq\int\displaylimits_{\vec{x}\in A_i}{\left(\vec{x}-\vec{x}_i\right)\left(\vec{x}-\vec{x}_i\right)^T\dint\vec{x}}$ is a matrix integral, and 
\begin{equation}
    R_{\sf tot}=\sum_{i=1}^{m}\int\displaylimits_{\vec{x}\in A_i}{ \left(\nabla\beta(\vec{x}_i)^T\left(\vec{x}-\vec{x}_i\right)\right)^2 \left({ \nabla\pdfxvec(\vec{x}_i)^T \left(\vec{x}-\vec{x}_i\right) }\right) \dint\vec{x}}.
\end{equation}
Then, using the Cauchy-Schwarz inequality we get 
\begin{align}
    \label{appendix:eq:Cauchy-Schwarz inequality for beta}
    &\left(\nabla\beta(\vec{x}_i)^T\left(\vec{x}-\vec{x}_i\right)\right)^2 \le \Ltwonormsquared{\nabla\beta(\vec{x}_i)}\Ltwonormsquared{\vec{x}-\vec{x}_i}
    \\
    \label{appendix:eq:Cauchy-Schwarz inequality for pdf x}
    &\nabla\pdfxvec(\vec{x}_i)^T\left(\vec{x}-\vec{x}_i\right) \le \left\lvert\nabla\pdfxvec(\vec{x}_i)^T\left(\vec{x}-\vec{x}_i\right)\right\rvert \le \Ltwonorm{\nabla\pdfxvec(\vec{x}_i)}\Ltwonorm{\vec{x}-\vec{x}_i}.
\end{align}
and can upper bound the absolute value of $R_{\sf tot}$ as 
\begin{align}
\label{appendix:eq:Rtot bound - 1}
    | R_{\sf tot} |\le \sum_{i=1}^{m}\left(\Ltwonormsquared{\nabla\beta(\vec{x}_i)} \Ltwonorm{\nabla\pdfxvec(\vec{x}_i)}\int\displaylimits_{\vec{x}\in A_i}{ \Ltwonorm{\vec{x}-\vec{x}_i}^3 \dint\vec{x}}\right)
\end{align}

The definition of the normalized second-moment of inertia (\ref{eq:normalized second-moment of inertia - definition}) implies that 
\begin{equation}
    \label{appendix:eq:second-moment integral using definition of M}
    \int\displaylimits_{\vec{x}\in A_i}{ \Ltwonormsquared{\vec{x}-\vec{x}_i} \dint\vec{x}} = M(A_i)\cdot d \cdot V(A_i)^{1+\frac{2}{d}}.
\end{equation}
Using Assumption \ref{assumption: vanishing region diameter}, we get 
\begin{equation}
    \Ltwonorm{\vec{x}-\vec{x}_i}\le {\rm diam}(A_i)\le\varphi_m
\end{equation}
where $\varphi_m\triangleq\max\limits_{i\in\{1,\dots,m\}}{\rm diam}\left(A_i\right)$.
Then, using (\ref{appendix:eq:definition of phi_M bounding constant of second momemt}), 
\begin{align}
    \int\displaylimits_{\vec{x}\in A_i}{ \Ltwonorm{\vec{x}-\vec{x}_i}^3 \dint\vec{x}} &\le \varphi_m \int\displaylimits_{\vec{x}\in A_i}{ \Ltwonorm{\vec{x}-\vec{x}_i}^2 \dint\vec{x}} \nonumber\\
    &= d \varphi_m M(A_i)V(A_i)^{1+\frac{2}{d}} \nonumber\\
    &\le d \varphi_m\phi_M V(A_i)^{1+\frac{2}{d}} \nonumber\\
    &= o \left( V(A_i)^{1+\frac{2}{d}} \right). 
\end{align}
Therefore, 
\begin{equation}
    \label{appendix:eq:third-moment integral as little o term}
    \int\displaylimits_{\vec{x}\in A_i}{ \Ltwonorm{\vec{x}-\vec{x}_i}^3 \dint\vec{x}} \in o\left(V(A_i)^{1+\frac{2}{d}}\right).
\end{equation}
Setting (\ref{appendix:eq:third-moment integral as little o term}) in (\ref{appendix:eq:Rtot bound - 1}) gives 
\begin{align}
\label{appendix:eq:Rtot bound - 2}
   | R_{\sf tot} |=o\left({\sum_{i=1}^{m}V(A_i)^{1+\frac{2}{d}}}\right)
\end{align}
because the per-region remainder estimates are uniform over the regions by the continuous differentiability assumption and Assumptions \ref{assumption:finite moment of inertia}-\ref{assumption: vanishing region diameter}, which justifies summing them into a single remainder.
Setting (\ref{appendix:eq:Rtot bound - 2}) in (\ref{appendix:eq:optimal error formula development - multidimensional - equality - 1}) and joining the remainders proves the test error formula (\ref{eq:theorem - test error sum formula - equality - multidimensional}) of Theorem \ref{theorem:optimal constants and test error in sum form - multidimensional}.

To prove the test error upper bound (\ref{eq:theorem - test error sum formula - multidimensional}) of Theorem \ref{theorem:optimal constants and test error in sum form - multidimensional}, we use the Cauchy-Schwarz inequality (\ref{appendix:eq:Cauchy-Schwarz inequality for beta}) and (\ref{appendix:eq:second-moment integral using definition of M}) to bound the $\nabla\beta(\vec{x}_i)^T\mtx{J}_i\nabla\beta(\vec{x}_i)$ terms in (\ref{appendix:eq:optimal error formula development - multidimensional - equality - 1}), using $\nabla\beta(\vec{x}_i)^T\mtx{J}_i\nabla\beta(\vec{x}_i)=\int\displaylimits_{\vec{x}\in A_i}{ \left(\nabla\beta(\vec{x}_i)^T\left(\vec{x}-\vec{x}_i\right)\right)^2\dint\vec{x}}$, getting 
\begin{align}
\label{appendix:eq:optimal error formula development - multidimensional - 3}
    &\mathcal{E}_{\sf test} \left({\{A_i\}_{i=1}^m, \{c_i^{\sf opt}\}_{i=1}^m}\right) \nonumber\\
    &\quad \le \sigma_{\epsilon}^2 + \sum_{i=1}^{m} \left( o\left(V(A_i)^{1+\frac{2}{d}}\right) + \pdfxvec(\vec{x}_i) \Ltwonormsquared{\nabla\beta(\vec{x}_i)}M(A_i)\cdot d \cdot V(A_i)^{1+\frac{2}{d}} \right)+R_{\sf tot}
    \\
    &\quad=\sigma_{\epsilon}^2 + d\sum_{i=1}^{m} \left( \pdfxvec(\vec{x}_i) \Ltwonormsquared{\nabla\beta(\vec{x}_i)}M(A_i) V(A_i)^{1+\frac{2}{d}} \right) + o\left({\sum_{i=1}^{m}V(A_i)^{1+\frac{2}{d}}}\right).
\end{align}

\subsection{Proof of Corollary \ref{corollary:test error integral formula - multidimensional}}
\label{appendix:subsec:Additional Details on Corollary - test error integral formula - multidimensional}
By the positivity of the limiting density in Assumption \ref{assumption: expert segment density function - multidimensional}, 
\begin{equation}
\label{appendix:eq:Additional Details on test error integral formula - lambda_0 - one-dimensional}
    \lambda_{\sf min} \triangleq\min_{x\in[0,1]^d}\lambda(\vec{x})>0.
\end{equation}
For sufficiently large $m$, $\lambda_m(\vec{x})\ge\frac{\lambda_{\sf min}}{2}$~ $\forall \vec{x}\in[0,1]^d$; hence, 
\begin{equation}
\label{appendix:eq:Additional Details on test error integral formula - interval length bound - multidimensional}
    V\left(A_{i}^{(m)}\right)=\frac{1}{m\lambda_m(\vec{x})}\le\frac{2}{m\lambda_{\sf min}} ~~\forall\vec{x}\in A_{i}^{(m)}.
\end{equation}
Specifically, note that 
\begin{equation}
\label{appendix:eq:Additional Details on test error integral formula - big-O bound of Delta_max - multidimensional}
V_{\sf max}\triangleq\underset{i\in\{1,\dots,m\}}{\max} V\left(A_{i}^{(m)}\right)\le\frac{2}{m\lambda_{\sf min}}.    
\end{equation}
Then, 
\begin{equation}
    \sum_{i=1}^{m}V(A_i)^{1+\frac{2}{d}}\le V_{\sf max}^\frac{2}{d}\sum_{i=1}^{m}V(A_i)=V_{\sf max}^\frac{2}{d}=\mathcal{O}\left(m^{-2/d}\right)
\end{equation}
and therefore the remainder $o\left({\sum_{i=1}^{m}V(A_i)^{1+\frac{2}{d}}}\right)$ from the test error formula (\ref{eq:theorem - test error sum formula - multidimensional}) is $o(m^{-2/d})$. 

Define 
\begin{align}
    h(\vec{x})=\Ltwonormsquared{\nabla\beta(\vec{x})} \pdfxvec(\vec{x})
\end{align}
and the corresponding piecewise constant function 
\begin{align}
    h_m(\vec{x})=h(\vec{x}_i)~~~~~\forall \vec{x}\in A_{i}^{(m)}.
\end{align}
Because $h$ is continuous in $[0,1]^d$ and the diameter shrinkage from Assumption \ref{assumption: vanishing region diameter}, we have $h_m\rightarrow h$ uniformly. 
Additionally, Assumption \ref{assumption: expert segment density function - multidimensional} gives $\lambda_m\rightarrow\lambda$ uniformly, with $\lambda_m(\vec{x})\ge\lambda_{\sf min}/2$ $\forall \vec{x}\in[0,1]^d$ for sufficiently large $m$. Assumption \ref{assumption: smooth normalized moment of inertia - multidimensional} gives $\mu_m\rightarrow\mu$ uniformly.  Therefore, 
\begin{equation}
    \frac{h_m\mu_m}{\lambda_m^{2/d}} \rightarrow \frac{h\mu}{\lambda^{2/d}} ~~\text{uniformly on}~[0,1]^d. 
\end{equation}
Using the exact definitions of piecewise constant function $\lambda_m$ and $\mu_m$ from Assumptions \ref{assumption: expert segment density function - multidimensional}, \ref{assumption: smooth normalized moment of inertia - multidimensional}, we get 
\begin{align}
    m^{2/d}\sum_{i=1}^{m}{ \Ltwonormsquared{\nabla\beta(\vec{x}_i)} \pdfxvec(\vec{x}_i) M\left(A_{i}^{(m)}\right) V\left(A_{i}^{(m)}\right)^{1+\frac{2}{d}}}&=m^{2/d}\sum_{i=1}^{m}{ h(\vec{x}_i) M\left(A_{i}^{(m)}\right) V\left(A_{i}^{(m)}\right)^{1+\frac{2}{d}}} \nonumber\\
    &=\sum_{i=1}^{m}{\int_{x\in A_{i}^{(m)}} \frac{h_m(\vec{x})\mu_m(\vec{x})}{\lambda_m^{2/d}(\vec{x})}\dint\vec{x} } \nonumber\\
    &=\int_{x\in [0,1]^d} \frac{h_m(\vec{x})\mu_m(\vec{x})}{\lambda_m^{2/d}(\vec{x})}\dint\vec{x} \nonumber\\
    &\rightarrow \int_{x\in [0,1]^d} \frac{h(\vec{x})\mu(\vec{x})}{\lambda^{2/d}(\vec{x})}\dint\vec{x}\nonumber\\
    &=\int_{x\in [0,1]^d}{ \frac{\Ltwonormsquared{\nabla\beta(\vec{x})} \pdfxvec(\vec{x})\mu(\vec{x}) }{\lambda^{2/d}(x)} \dint\vec{x}}
    \label{appendix:eq:Additional Details on test error integral formula - limit of sum as integral - multidimensional}
\end{align}
where the convergence of the integral is due to the uniform convergence of the integrands.
Setting (\ref{appendix:eq:Additional Details on test error integral formula - limit of sum as integral - multidimensional}) and the remainder form $o(m^{-2/d})$ into (\ref{eq:theorem - test error sum formula - multidimensional}) proves Corollary \ref{corollary:test error integral formula - multidimensional}.

\subsection{Proof of Theorem \ref{theorem: optimal expert segment density function and test error - multidimensional}: The Optimal Expert Segment Density and the Corresponding Test Error in the Multidimensional Case}
\label{appendix:subsec: proof of theorem on optimal expert segment density function and test error - multidimensional}

Let us develop a lower bound for the integral $\int_{\vec{x}\in[0,1]^d}{ \frac{\Ltwonormsquared{\nabla\beta(\vec{x})} \pdfxvec(\vec{x})}{\lambda^{2/d}(\vec{x})} \dint \vec{x}}$ from the test error's asymptotic upper bound formula of Corollary \ref{corollary:test error integral formula with fixed moment of inertia - multidimensional}. This will lead to the segment density that minimizes the test error upper bound. By H\"{o}lder's inequality, 
\begin{align}
    \label{appendix:eq:proof of theorem on optimal expert segment density function and test error - multidimensional - 1}
    &\left({\int_{\vec{x}\in[0,1]^d}\left({ \left({\Ltwonormsquared{\nabla\beta(\vec{x})} \pdfxvec(\vec{x})}\right)^{\frac{d}{d+2}}\frac{1}{\lambda^{\frac{2}{d+2}}(\vec{x})}}\right)^{\frac{d+2}{d}} \dint\vec{x}}\right)^{\frac{d}{d+2}}  \left({\int_{\vec{x}\in[0,1]^d}\left({ \lambda^{\frac{2}{d+2}}(\vec{x})}\right)^{\frac{d+2}{2}} \dint\vec{x}}\right)^{\frac{2}{d+2}} \nonumber\\ 
    &\qquad\qquad\qquad\qquad\qquad\qquad\qquad\qquad\qquad\qquad\qquad\ge   \int_{\vec{x}\in[0,1]^d} \left({\Ltwonormsquared{\nabla\beta(\vec{x})} \pdfxvec(\vec{x})}\right)^{\frac{d}{d+2}} \dint\vec{x}
\end{align}
where the left side can be simplified, thus, 
\begin{align}
    \label{appendix:eq:proof of theorem on optimal expert segment density function and test error - multidimensional - 2}
    &\left({\int_{\vec{x}\in[0,1]^d} \Ltwonormsquared{\nabla\beta(\vec{x})} \pdfxvec(\vec{x})\frac{1}{\lambda^{2/d}(\vec{x})} \dint\vec{x}}\right)^{\frac{d}{d+2}}  \left({\int_{\vec{x}\in[0,1]^d} \lambda(\vec{x}) \dint\vec{x}}\right)^{\frac{2}{d+2}} \nonumber\\
    &\qquad\qquad\qquad\qquad\qquad\qquad\quad \ge   \int_{\vec{x}\in[0,1]^d} \left({\Ltwonormsquared{\nabla\beta(\vec{x})} \pdfxvec(\vec{x})}\right)^{\frac{d}{d+2}} \dint\vec{x}.
\end{align}
The density $\lambda(\vec{x})$ integrates to 1, i.e., $\int_{\vec{x}\in[0,1]^d} \lambda(\vec{x}) \dint\vec{x} = 1$, hence, we get that 
\begin{equation}
    \label{appendix:eq:proof of theorem on optimal expert segment density function and test error - multidimensional - 3}
    \int_{\vec{x}\in[0,1]^d} \Ltwonormsquared{\nabla\beta(\vec{x})} \pdfxvec(\vec{x})\frac{1}{\lambda^{2/d}(\vec{x})} \dint\vec{x}  \ge   \left({ \int_{\vec{x}\in[0,1]^d} \left({\Ltwonormsquared{\nabla\beta(\vec{x})} \pdfxvec(\vec{x})}\right)^{\frac{d}{d+2}} \dint\vec{x} }\right)^{\frac{d+2}{d}}
\end{equation}
where the left side is the integral from the asymptotic upper bound formulated in Corollary \ref{corollary:test error integral formula with fixed moment of inertia - multidimensional}.

H\"{o}lder's inequality in (\ref{appendix:eq:proof of theorem on optimal expert segment density function and test error - multidimensional - 1}) is attained with equality if $\lambda(\vec{x})$ is linearly proportional to $\Ltwonormsquared{\nabla\beta(\vec{x})} \pdfxvec(\vec{x})\frac{1}{\lambda^{2/d}(\vec{x})}$. Namely, for a constant $\zeta$ such that 
\begin{align}
    \lambda(\vec{x}) = \zeta \cdot \Ltwonormsquared{\nabla\beta(\vec{x})} \pdfxvec(\vec{x})\frac{1}{\lambda^{2/d}(\vec{x})}
\end{align}
which, using integration over $[0,1]^d$ and that the density integrates to 1, implies that 
\begin{align}
    \zeta =  \frac{1}{\left({\int_{\vec{x}\in[0,1]^d}\left({\Ltwonormsquared{\nabla\beta(\vec{x})} \pdfxvec(\vec{x}) }\right)^{\frac{d}{d+2}} \dint\vec{x}}\right)^{\frac{d+2}{d}}}.
\end{align}
Consequently, the density that minimizes the integral in the left side of (\ref{appendix:eq:proof of theorem on optimal expert segment density function and test error - multidimensional - 3}) is 
    \begin{equation}
        \label{appendix:eq:optimal expert segment density function - multidimensional}
        \lambda^{\sf UBM}(\vec{x})=\frac{\left({\Ltwonormsquared{\nabla\beta(\vec{x})} \pdfxvec(\vec{x}) }\right)^{\frac{d}{d+2}}}{\int_{\vec{\xi}\in[0,1]^d}\left({\Ltwonormsquared{\nabla\beta(\vec{\xi})} \pdfxvec(\vec{\xi}) }\right)^{\frac{d}{d+2}}\dint\vec{\xi}}.
    \end{equation}
The optimal density $\lambda^{\sf UBM}(x)$ makes  (\ref{appendix:eq:proof of theorem on optimal expert segment density function and test error - multidimensional - 3}) to hold with equality. Then, plugging the right side of (\ref{appendix:eq:proof of theorem on optimal expert segment density function and test error - multidimensional - 3}) into the test error's asymptotic bound formulated in Corollary \ref{corollary:test error integral formula with fixed moment of inertia - multidimensional} gives the minimal asymptotic bound of 
    \begin{equation}
        \label{appendix:eq:optimal test error - multidimensional}
        \widetilde{\mathcal{E}}_{\sf test}^{\sf bound, min} = \sigma_{\epsilon}^2 + \frac{d\cdot \mu_d}{m^{2/d}}\left({ \int_{\vec{x}\in[0,1]^d} \left({\Ltwonormsquared{\nabla\beta(\vec{x})} \pdfxvec(\vec{x})}\right)^{\frac{d}{d+2}} \dint\vec{x} }\right)^{1+\frac{2}{d}}.
    \end{equation}    
This completes the proof of Theorem \ref{theorem: optimal expert segment density function and test error - multidimensional}.

\section{MoE for Multidimensional Inputs: The Bisection Algorithm for Two-Dimensional Segmentation using the UBM Segment Density}
\label{appendix:subsec:MoE for Multidimensional Inputs: The Bisection Algorithm for Two-Dimensional Segmentation}

Before describing the segmentation algorithm, note that the $\lambda^{\sf UBM}$ formula in (\ref{eq:optimal expert segment density function - multidimensional}) may have zero values for inputs whose $\Ltwonormsquared{\nabla\beta(\vec{x})}=0$, including where $\beta$ is constant. 
As we discussed in the one-dimensional case, we empirically compute $\lambda^{\sf UBM}$ from (\ref{eq:optimal expert segment density function - multidimensional}) with replacing the near zero values $\pdfxvec(\vec{x})\Ltwonormsquared{\nabla\beta(\vec{x})} < \widetilde{\sf{eps}}$ with $\widetilde{\sf{eps}}$ where $\widetilde{\sf{eps}}=10^{-16}$; this gives a numerically regularized version of (\ref{eq:optimal expert segment density function - multidimensional}).
Fig.~\ref{fig:empirical results - two-dimensions - example 2 - cosine with constant area beta} shows an experiment that significantly uses this mechanism due to the constant-valued $\beta(\vec{x})$ in a noticeable subarea of the input space.

The bisection algorithm is implemented by recursively applying the following procedure, which is defined as 
$\mathcal{L}=\operatorname{Bisect}\left(\widetilde{A},\lambda^{\sf UBM},\widetilde{m}\right)$. Segmenting the overall area of $[0,1]^2$ to $m$ regions is done by running $\operatorname{Bisect}\left([0,1]^2,\lambda^{\sf UBM},m\right)$.

\RestyleAlgo{ruled}
\begin{algorithm2e}[htbp]
\caption{Recursive rectangular segmentation using the UBM density}
\label{algorithm:2D segmentation by bisection}

\LinesNumbered
\DontPrintSemicolon
\SetKwInOut{Input}{Inputs}
\SetKwInOut{Output}{Outputs}
\SetKwProg{Fn}{function}{:}{end function}
\BlankLine
\Input{
    A given rectangle $\widetilde{A}=[x_{1{\sf start}},x_{1{\sf end}}]\times[x_{2{\sf start}},x_{2{\sf end}}]\subseteq[0,1]^2$, a segment density function $\lambda^{\sf UBM}$, and an integer number $\widetilde{m}\ge1$ of segments to divide the given rectangle
}
\Output{
    A list of $\widetilde{m}$ regions of $\widetilde{A}$, each region with a description of its coordinates and its integrated value of the density $\lambda^{\sf UBM}$
}
\BlankLine

\Fn{$\operatorname{Bisect}\left(\widetilde A,\lambda^{\sf UBM},\widetilde{m} \right )$}{
    \If{$\widetilde m=1$}{
        Compute the integrated density:
        $\lambda_{\mathsf{int}\widetilde A}
        = \displaystyle
        \int_{\vec{x}\in\widetilde A}
        \lambda^{\mathsf{UBM}}(\vec{x})
        \,\mathrm{d}\vec{x}$\;

        \Return
        $\mathcal{L}=\left\{\left(\widetilde{A},\lambda_{{\sf int}\widetilde{A}}\right)\right\}$\;
    }

    \eIf{$x_{1{\sf end}}-x_{1{\sf start}} \ge x_{2{\sf end}}-x_{2{\sf start}}$ (i.e., the rectangle is wide)}{
        Define the 1D density as $\widetilde{\lambda}(\widetilde{x})\triangleq \int_{x_2=x_{2{\sf start}}}^{x_{2{\sf end}}}\lambda^{\sf UBM}([\widetilde{x},x_2]^T)\dint x_2$ for the interval $\widetilde{x}\in[\widetilde{x}_{\sf start}, \widetilde{x}_{\sf end}]$ where $\widetilde{x}_{\sf start}=x_{1{\sf start}}$, $\widetilde{x}_{\sf end}=x_{1{\sf end}}$
        
         Set the split axis as $\eta=1$
    }
    {
         Define the 1D density as $\widetilde{\lambda}(\widetilde{x})\triangleq \int_{x_1=x_{1{\sf start}}}^{x_{1{\sf end}}}\lambda^{\sf UBM}([x_1,\widetilde{x}]^T)\dint x_1$  for the interval $\widetilde{x}\in[\widetilde{x}_{\sf start}, \widetilde{x}_{\sf end}]$ where $\widetilde{x}_{\sf start}=x_{2{\sf start}}$, $\widetilde{x}_{\sf end}=x_{2{\sf end}}$
         
         Set the split axis as $\eta=2$
    }

        By adapting (\ref{appendix:eq:optimal expander function for scalar quantization}) to our case, define the 1D expander function $\widetilde{v}:[0,1]\rightarrow[\widetilde{x}_{\sf start}, \widetilde{x}_{\sf end}]$ for $\widetilde{A}$ along its longer axis and the density $\widetilde{\lambda}$: 
For $\tau\in[0,1]$, $\widetilde{v}(\tau)$ is the solution to \[\int\displaylimits_{\xi=\widetilde{x}_{\sf start}}^{\widetilde{v}(\tau)}\widetilde{\lambda}(\xi)\dint\xi = \tau\cdot\int\displaylimits_{\xi=\widetilde{x}_{\sf start}}^{\widetilde{x}_{\sf end}}\widetilde{\lambda}(\xi)\dint\xi\]

        Divide the $\widetilde{m}$ to the two recursive children: $\widetilde{m}_{\sf L}=\left\lfloor\frac{\widetilde{m}}{2}\right\rfloor$ and $\widetilde{m}_{\sf R}=\widetilde{m}-\widetilde{m}_{\sf L}$

        Define the splitting coordinate of the rectangle, for an approximate equality of density according to the expander: $\widetilde{x}_{\sf s}=\widetilde{v}\left(\frac{\widetilde{m}_{\sf L}}{\widetilde{m}}\right)$

        Define the two child rectangles: $A_{\sf L}=\widetilde{A}\cap\{\vec{x}\in\widetilde{A}:x_{\eta}\le\widetilde{x}_{\sf s}\}$,~ $A_{\sf R}=\widetilde{A}\cap\{\vec{x}\in\widetilde{A}:x_{\eta}>\widetilde{x}_{\sf s}\}$

        Run the procedure recursively for the two children:\newline
 $\mathcal{L}_{\sf L}=\operatorname{Bisect}\left(A_{\sf L},\lambda^{\sf UBM},\widetilde{m}_{\sf L}\right)$\newline
 $\mathcal{L}_{\sf R}=\operatorname{Bisect}\left(A_{\sf R},\lambda^{\sf UBM},\widetilde{m}_{\sf R}\right)$

        \Return $\mathcal{L}=\operatorname{Concatenate}\left(\mathcal{L}_{\sf L}, \mathcal{L}_{\sf R}\right)$
    
}
\end{algorithm2e}

Note that the recursive splitting coordinate, which is defined by the expander in line 15 of Algorithm \ref{algorithm:2D segmentation by bisection} as $\widetilde{x}_{\sf s}=\widetilde{v}\left(\frac{\widetilde{m}_{\sf L}}{\widetilde{m}}\right)$, splits the density integral to the two children according to their eventual number of segments. This rule yields an approximately equal integrated density of $\frac{1}{m}$ for the $m$ regions of the eventual segmentation of the overall input space $[0,1]^2$.

Moreover, note that our Algorithm \ref{algorithm:2D segmentation by bisection} makes each split perpendicular to the longer side of the current rectangle, in contrast to the alternating vertical-then-horizontal partitioning approach of \cite{berger1987partitioning}.

\section{Proofs for Section \ref{sec:Learning the Constant Experts}}
\label{appendix:sec:Proofs for the Section on Learning the Constant Experts}

\subsection{Proof of Theorem \ref{theorem:the learned expert constant is unbiased}: The Learned Constant Expert is Unbiased}
\label{appendix:subsec:Proof of Theorem on the learned expert constant is unbiased}

By the learned constant formula in (\ref{eq:learned expert constants - least squares}), the formula for expectation conditioned on an event (here the event is $n_i>0$), the conditional expectation rule, the independence among $\left\{q_i\exind{j}\right\}_{j=1}^n$ for a specific region $A_i$, and the definition (\ref{eq:number of training examples routed to expert i - definition}) of $n_i$ as the sum of all the variables $\left\{q_i\exind{j}\right\}_{j=1}^n$: 
\begin{align}    
    \expectation{\widetilde{c}_i \vert n_i>0} &= \frac{\expectation{\widetilde{c}_i \cdot \mathbb{I}[n_i>0]}}{\eventprob{n_i>0}} = \frac{ \expectation{\mathbb{I}[n_i>0]\cdot\frac{1}{n_i} \sum_{j: q_i\exind{j} = 1} y\exind{j}}}{\eventprob{n_i>0}} \nonumber\\
    &= \frac{ \expectationwrt{\mathbb{I}[n_i>0]\cdot\frac{1}{n_i} \sum_{j: q_i\exind{j} = 1} \expectation{y\exind{j} | q_i\exind{j} = 1 }}{\left\{q_i\exind{j}\right\}_{j=1}^n} }{\eventprob{n_i>0}}
    \label{appendix:eq:proof of theorem on the learned expert constant is unbiased - 1}
\end{align}
where the value of $q_i\exind{j}$ is 1 in the conditional expectation of $y\exind{j}$ in the sum due to the condition $j: q_i\exind{j} = 1$ in the summation. 

By the training data model (\ref{eq: training data model}) and the zero mean of the noise, the inner conditional expectation in (\ref{appendix:eq:proof of theorem on the learned expert constant is unbiased - 1}) can be developed as follows: 
\begin{align}
    \label{appendix:eq:proof of theorem on the learned expert constant is unbiased - 2}
\expectation{y\exind{j} | q_i\exind{j}} = \expectation{ \beta\left(\vec{x}\exind{j}\right)+\epsilon\exind{j} | q_i\exind{j}} = \expectation{ \beta\left(\vec{x}\exind{j}\right) | q_i\exind{j}} + \expectation{ \epsilon\exind{j} } = \expectation{ \beta\left(\vec{x}\exind{j}\right) | q_i\exind{j}}. 
\end{align}
Note that $q_i\exind{j}$ is a Bernoulli random variable with 
\begin{equation}
    \label{appendix:eq:probability of q variable equal 1}
    \eventprob{q_i\exind{j} = 1} = \eventprob{\vec{x}\exind{j}\in A_i} = \int_{\vec{x}\in A_i} \pdfxvec(\vec{x}) \dint\vec{x}. 
\end{equation}
Then, the expectation of $y\exind{j}$ conditioned on the event $q_i\exind{j} = 1$ is 
\begin{align}
    \label{appendix:eq:proof of theorem on the learned expert constant is unbiased - 3}
\expectation{ y\exind{j} | q_i\exind{j}=1} = \expectation{ \beta\left(\vec{x}\exind{j}\right) | q_i\exind{j}=1} = \frac{\int_{\vec{x}\in A_i} \beta\left(\vec{x}\right) \pdfxvec(\vec{x}) \dint\vec{x}}{\int_{\vec{x}\in A_i} \pdfxvec(\vec{x}) \dint\vec{x}} = c_i^{\sf opt}
\end{align}
where the last equality is due to the formula of the optimal constant (without small region assumptions) from (\ref{eq:optimal c_i - multidimensional}).

Setting (\ref{appendix:eq:proof of theorem on the learned expert constant is unbiased - 3}) back in (\ref{appendix:eq:proof of theorem on the learned expert constant is unbiased - 1}) gives 
\begin{align}
    \label{appendix:eq:proof of theorem on the learned expert constant is unbiased - 4}
    \expectation{\widetilde{c}_i \vert n_i>0} &=  \frac{ \expectationwrt{\mathbb{I}[n_i>0]\cdot\frac{1}{n_i} \cdot n_i c_i^{\sf opt}}{\left\{q_i\exind{j}\right\}_{j=1}^n} }{\eventprob{n_i>0}}
    =  \frac{ \expectationwrt{\mathbb{I}[n_i>0]}{\left\{q_i\exind{j}\right\}_{j=1}^n}}{\eventprob{n_i>0}} c_i^{\sf opt} 
    \\\nonumber
    &
    =  \frac{ 1\cdot\eventprob{n_i>0}}{\eventprob{n_i>0}} c_i^{\sf opt} 
     = c_i^{\sf opt}
\end{align}
which completes the proof of Theorem \ref{theorem:the learned expert constant is unbiased}.

\subsection{Proof of Lemma \ref{lemma:test error with learned constants formulated using test error with optimal constants}}
\label{appendix:subsec:Proof of Lemma on test error with learned constants formulated using test error with optimal constants}

We develop the test error for the learned expert constants, starting from the test error formula of (\ref{eq:test error - MoE - multidimensional}): 
\begin{align}
        \mathcal{E}_{\sf test} \left({\{A_i\}_{i=1}^m, \left\{\widetilde{c}_i\right\}_{i=1}^m}\right) &= \sigma_{\epsilon}^2 + \sum_{i=1}^{m} {\int\displaylimits_{\vec{x}\in A_i}{\left(\widetilde{c}_i - \beta(\vec{x})\right)^2 \pdfxvec(\vec{x})\dint\vec{x}} } \nonumber \\ \nonumber
        &= \sigma_{\epsilon}^2 + \sum_{i=1}^{m} {\int\displaylimits_{\vec{x}\in A_i}{\left(\widetilde{c}_i - c_i^{\sf opt} + c_i^{\sf opt} - \beta(\vec{x})\right)^2 \pdfxvec(\vec{x})\dint\vec{x}} }\\ \nonumber
        &= \sigma_{\epsilon}^2 + \sum_{i=1}^{m} { \left(\widetilde{c}_i - c_i^{\sf opt}\right)^2 \int\displaylimits_{\vec{x}\in A_i}{ \pdfxvec(\vec{x})\dint\vec{x}} } \\\nonumber
        &\quad  + \sum_{i=1}^{m} {\int\displaylimits_{\vec{x}\in A_i}{\left(c_i^{\sf opt} - \beta(\vec{x})\right)^2 \pdfxvec(\vec{x})\dint\vec{x}} } \\
        &\quad + 2\sum_{i=1}^{m} {\left(\widetilde{c}_i - c_i^{\sf opt}\right)\int\displaylimits_{\vec{x}\in A_i}{\left(c_i^{\sf opt} - \beta(\vec{x})\right) \pdfxvec(\vec{x})\dint\vec{x}} }.
          \label{appendix:eq:test error with learned constants formulated using test error with optimal constants - proof - 1}
\end{align}
According to (\ref{eq:test error - MoE - multidimensional}), the second sum in (\ref{appendix:eq:test error with learned constants formulated using test error with optimal constants - proof - 1}) is related to test error of MoE with optimal constant experts (for a given routing segmentation) as 
\begin{equation}
\label{appendix:eq:test error with learned constants formulated using test error with optimal constants - proof - second sum}
\sum_{i=1}^{m} {\int\displaylimits_{\vec{x}\in A_i}{\left(c_i^{\sf opt} - \beta(\vec{x})\right)^2 \pdfxvec(\vec{x})\dint\vec{x}} } =   \mathcal{E}_{\sf test} \left({\{A_i\}_{i=1}^m, \left\{c_i^{\sf opt}\right\}_{i=1}^m}\right) -  \sigma_{\epsilon}^2. 
\end{equation}
Moreover, for the third sum in (\ref{appendix:eq:test error with learned constants formulated using test error with optimal constants - proof - 1}), note that 
\begin{align}
\label{appendix:eq:test error with learned constants formulated using test error with optimal constants - proof - for third sum}
\int\displaylimits_{\vec{x}\in A_i}{\left(c_i^{\sf opt} - \beta(\vec{x})\right) \pdfxvec(\vec{x})\dint\vec{x}} &= c_i^{\sf opt} \int\displaylimits_{\vec{x}\in A_i}{ \pdfxvec(\vec{x})\dint\vec{x}} - \int\displaylimits_{\vec{x}\in A_i}{\beta(\vec{x}) \pdfxvec(\vec{x})\dint\vec{x}}  \\ \nonumber
&= \frac{\int\displaylimits_{\vec{x}\in A_i}{\beta(\vec{x}) \pdfxvec(\vec{x})\dint\vec{x}}}{\int\displaylimits_{\vec{x}\in A_i}{\pdfxvec(\vec{x})\dint\vec{x}}} \cdot \int\displaylimits_{\vec{x}\in A_i}{ \pdfxvec(\vec{x})\dint\vec{x}} - \int\displaylimits_{\vec{x}\in A_i}{\beta(\vec{x}) \pdfxvec(\vec{x})\dint\vec{x}} 
\\ \nonumber
&= 0
\end{align}
where the optimal constant formula (\ref{eq:optimal c_i - multidimensional}) is used here. 
Setting (\ref{appendix:eq:test error with learned constants formulated using test error with optimal constants - proof - second sum}) and (\ref{appendix:eq:test error with learned constants formulated using test error with optimal constants - proof - for third sum}) back in (\ref{appendix:eq:test error with learned constants formulated using test error with optimal constants - proof - 1}) gives 
\begin{align}
\label{appendix:eq:test error with learned constants formulated using test error with optimal constants - proof - 2}
 \mathcal{E}_{\sf test} \left({\{A_i\}_{i=1}^m, \left\{\widetilde{c}_i\right\}_{i=1}^m}\right) &= 
 \mathcal{E}_{\sf test} \left({\{A_i\}_{i=1}^m, \left\{c_i^{\sf opt}\right\}_{i=1}^m}\right) + \sum_{i=1}^{m} { \left(\widetilde{c}_i - c_i^{\sf opt}\right)^2 \int\displaylimits_{\vec{x}\in A_i}{ \pdfxvec(\vec{x})\dint\vec{x}} }.
\end{align}
Using that, by (\ref{eq:approximation error of the simpler hypothesis class given segmentation}), 
\[\mathcal{E}_{\sf app} \left(\mathcal{H}_{m,d}^{c}\left(\{A_i\}_{i=1}^m\right)\right)=\mathcal{E}_{\sf test} \left({\{A_i\}_{i=1}^m, \left\{c_i^{\sf opt}\right\}_{i=1}^m}\right),\] and defining the estimation error, using $\rho_i=\eventprob{\vecrand{x}\in A_i} = \int\displaylimits_{\vec{x}\in A_i}{ \pdfxvec(\vec{x})\dint\vec{x}}$ from (\ref{eq:definition of rho_i}), as 
\[\mathcal{E}_{\sf est}(m,\mathcal{S}_n) = \sum_{i=1}^{m} \left( \widetilde{c}_i - c_i^{\sf opt} \right)^2 \rho_i\]
completes the proof of Lemma \ref{lemma:test error with learned constants formulated using test error with optimal constants}.

\subsection{Proof of Lemma \ref{lemma:concentration bounds on the number ni of training examples that are routed to expert i}}
\label{appendix:subsec:Proof of Lemma concentration bounds on the number ni of training examples that are routed to expert i}

The definition of $q_i\exind{j}$ in (\ref{eq: q indicator function for the inclusion of the j training input in the i region Ai - definition}) implies it is a Bernoulli random variable with probability $\rho_i$ (recall (\ref{eq:definition of rho_i})) of $q_i\exind{j}=1$. Then,  definition (\ref{eq:number of training examples routed to expert i - definition}) implies that $n_i$ is a sum of $n$ independent Bernoulli variables $q_i\exind{1},\dots,q_i\exind{n}$. 

A version of the Chernoff bound for a sum of independent Bernoulli variables \citep{vershynin2018high} can be written here as 
\begin{equation}
    \label{appendix:eq:chernoff bound for n_i - 1}
    \text{for}~\tau\in(0,1),~\eventprob{ n_i \le (1-\tau)\expectation{n_i} }\le \exp{\left(-\frac{\tau^2 \expectation{n_i}}{2}\right)}.
\end{equation}
According to (\ref{eq: q indicator function for the inclusion of the j training input in the i region Ai - definition}), (\ref{eq:number of training examples routed to expert i - definition}), (\ref{eq:definition of rho_i}), we have 
\begin{equation}
    \expectation{n_i} = n\rho_i
\end{equation}
and therefore (\ref{appendix:eq:chernoff bound for n_i - 1}) becomes 
\begin{equation}
    \label{appendix:eq:chernoff bound for n_i - with explicit expectation of n_i}
    \text{for}~\tau\in(0,1),~\eventprob{ n_i \le (1-\tau)n\rho_i }\le \exp{\left(-\frac{\tau^2 n\rho_i}{2}\right)}.
\end{equation}
Upper bounding the right-side probability in (\ref{appendix:eq:chernoff bound for n_i - with explicit expectation of n_i}) with $\widetilde{\delta}\in(0,1)$ yields the condition 
\begin{equation}
\label{appendix:eq:chernoff bound for n_i - condition on n}
    n\ge \frac{2}{\tau^2 \rho_i}\ln{\left(\frac{1}{\widetilde{\delta}}\right)}.
\end{equation}
Then, setting $\tau=\frac{1}{2}$ in (\ref{appendix:eq:chernoff bound for n_i - with explicit expectation of n_i}) and (\ref{appendix:eq:chernoff bound for n_i - condition on n}) gives that for $\widetilde{\delta}\in(0,1)$ and $n\ge \frac{8}{\rho_i}\ln{\left(\frac{1}{\widetilde{\delta}}\right)}$
\begin{equation}
    \label{appendix:eq:chernoff bound for n_i - end of proof}
    \eventprob{ n_i \le \frac{1}{2}n\rho_i }\le \widetilde{\delta}.
\end{equation}
The inequality in (\ref{appendix:eq:chernoff bound for n_i - end of proof}) is equivalent to $\eventprob{ n_i > \frac{1}{2}n\rho_i }\ge 1-\widetilde{\delta}$ and therefore the proof of Lemma \ref{lemma:concentration bounds on the number ni of training examples that are routed to expert i} is completed.

\subsection{Proof of Theorem \ref{theorem:statistical bound on distance between learned constant and optimal constant}}
\label{appendix:subsec:Proof of theorem on the statistical bound on distance between learned constant and optimal constant}

According to (\ref{eq:learned expert constants - least squares}), $\widetilde{c}_i$ is an average of $n_i$ independent random variables $\left\{y\exind{j}\right\}_{j: q_i\exind{j} = 1}$. 
Therefore, conditioned on $n_i=\theta_i$ where $\theta_i\in\{1,\dots,n\}$ is a non-random value, the probability distribution of the random variable $\widetilde{c}_i$ is the same as the probability distribution of the random variable 
\begin{equation}
    \label{appendix:eq:equivalent random variable to learned coefficient}
    \frac{1}{\theta_i}\sum_{\tau=1}^{\theta_i} z_i^{(\tau)} 
\end{equation}
where $z_i^{(1)},\dots,z_i^{(\theta_i)}$ are i.i.d. drawn from the distribution of $y\vert \{q_i=1\}$, such that $(\vecrand{x},y)$ are drawn from the distribution of the data model (\ref{eq:data model - multidimensional}) and $q_i=\mathbb{I}[\vecrand{x}\in A_i]$.

We use the following assumptions in addition to the data model (\ref{eq:data model - multidimensional}):
 The function $\beta$ is bounded on its entire domain. For the $i^{\sf th}$ region $A_i$, denote the value range size as $R_{\beta,i} \triangleq \underset{\vec{x}\in A_i}{\sup}\beta(\vec{x}) - \underset{\vec{x}\in A_i}{\inf}\beta(\vec{x})$.  
 The noise $\epsilon$ is from the bounded range $[\epsilon_{\sf min},\epsilon_{\sf max}]$. Denote $R_{\epsilon} \triangleq \epsilon_{\sf max} - \epsilon_{\sf min}$. Then, for $j$ such that $q_i\exind{j} = 1$, the random variable $y\exind{j}$ gets its value from the range 
 \begin{equation}
     \label{appendix:eq:value range for random variable}
     \left[\underset{\vec{x}\in A_i}{\inf}\beta(\vec{x}) + \epsilon_{\sf min},~ \underset{\vec{x}\in A_i}{\sup}\beta(\vec{x}) + \epsilon_{\sf max}\right].
 \end{equation}
 The random variables $z_i^{(1)},\dots,z_i^{(\theta_i)}$ are also from the value range (\ref{appendix:eq:value range for random variable}). 

If $R_{\beta,i}+R_{\epsilon}=0$, then every response in $A_i$ equals $c_i^{\sf opt}$. Consequently, $\widetilde{c}_i=c_i^{\sf opt}$ whenever $n_i>0$, and Lemma \ref{lemma:concentration bounds on the number ni of training examples that are routed to expert i} establishes (\ref{eq:hoeffding bound on distance between learned constant and optimal constant}) with probability at least $1-\widetilde{\delta}$. For the remainder of the proof, assume $R_{\beta,i}+R_{\epsilon}>0$.

Then, for $\gamma\ge0$ and $\theta_i\in\{1,\dots,n\}$,
\begin{align}
\label{appendix:eq:hoeffding inequality - start}
  \eventprob{ \left\lvert \widetilde{c}_i - \expectation{\widetilde{c}_i \vert n_i=\theta_i}\right\rvert \ge \gamma ~\vert~ n_i=\theta_i } &= \eventprob{ \left\lvert \frac{1}{\theta_i}\sum_{\tau=1}^{\theta_i} z_i^{(\tau)} - \frac{1}{\theta_i}\expectation{\sum_{\tau=1}^{\theta_i} z_i^{(\tau)}}\right\rvert \ge \gamma } %~\Bigg\vert~ n_i=\theta_i }
  \nonumber \\
  &\le 2\exp\left(-\frac{2\gamma^2 \theta_i}{\left(v_{{\sf max},i} - v_{{\sf min},i}\right)^2}\right)
 \end{align}
 where the last inequality is due to Hoeffding's inequality; by (\ref{appendix:eq:value range for random variable}), $v_{{\sf max},i}=\underset{\vec{x}\in A_i}{\sup}\beta(\vec{x}) + \epsilon_{\sf max}$ and $v_{{\sf min},i}=\underset{\vec{x}\in A_i}{\inf}\beta(\vec{x}) + \epsilon_{\sf min}$, therefore, 
 \begin{align}
 &v_{{\sf max},i} - v_{{\sf min},i} = R_{\beta,i} + R_{\epsilon}. 
 \end{align}

 Recall that, according to Theorem \ref{theorem:the learned expert constant is unbiased}, $\expectation{\widetilde{c}_i | n_i>0} = c_i^{\sf opt}$. It can be proved similarly to Appendix \ref{appendix:subsec:Proof of Theorem on the learned expert constant is unbiased} that 
 \begin{equation}
 \label{appendix:eq:the learned expert constant is unbiased - conditioned on n_i greater than positive eta_i}
     \expectation{\widetilde{c}_i | n_i=\theta_i} = c_i^{\sf opt},~~\text{for}~\theta_i\in\{1,\dots,n\}.
 \end{equation} 
 Then, setting (\ref{appendix:eq:the learned expert constant is unbiased - conditioned on n_i greater than positive eta_i}) in (\ref{appendix:eq:hoeffding inequality - start}) gives 
\begin{align}
\label{appendix:eq:hoeffding inequality - 2}
  \eventprob{ \left\lvert \widetilde{c}_i - c_i^{\sf opt}\right\rvert \ge \gamma ~\vert~ n_i=\theta_i } \le 2\exp\left(-\frac{2\gamma^2 \theta_i}{\left(v_{{\sf max},i} - v_{{\sf min},i}\right)^2}\right).
 \end{align}

Let us extend the conditioning such that $n_i\ge\eta_i$, i.e., $n_i\in\{\left\lceil \eta_i\right\rceil,\dots,n\}$ where $\eta_i$ is a positive non-random value from the continuous range $(0,n]$:
\begin{align}
%\label{appendix:eq:conditioning on a range of n_i}
  \eventprob{ \left\lvert \widetilde{c}_i - c_i^{\sf opt}\right\rvert \ge \gamma ~\vert~ n_i\ge\eta_i } &= \frac{\eventprob{ \left\lvert \widetilde{c}_i - c_i^{\sf opt}\right\rvert \ge \gamma,~ n_i\ge\eta_i }}{\eventprob{n_i\ge\eta_i}}
  \\
  &=\frac{\sum_{\theta_i=\left\lceil\eta_i\right\rceil}^{n} \eventprob{ \left\lvert \widetilde{c}_i - c_i^{\sf opt}\right\rvert \ge \gamma ~\vert~ n_i=\theta_i } \eventprob{n_i=\theta_i}}{\eventprob{n_i\ge\eta_i}}
  \\
  \label{appendix:eq:bound extended conditioning - 3}
  &\le\frac{\sum_{\theta_i=\left\lceil\eta_i\right\rceil}^{n} 2\exp\left(-\frac{2\gamma^2 \theta_i}{\left(v_{{\sf max},i} - v_{{\sf min},i}\right)^2}\right) \eventprob{n_i=\theta_i}}{\eventprob{n_i\ge\eta_i}}
    \\
  &\le\frac{ 2\exp\left(-\frac{2\gamma^2 \eta_i}{\left(v_{{\sf max},i} - v_{{\sf min},i}\right)^2}\right) \sum_{\theta_i=\left\lceil\eta_i\right\rceil}^{n} \eventprob{n_i=\theta_i}}{\eventprob{n_i\ge\eta_i}}
    \\
    \label{appendix:eq:bound extended conditioning - 5}
  &=\frac{ 2\exp\left(-\frac{2\gamma^2 \eta_i}{\left(v_{{\sf max},i} - v_{{\sf min},i}\right)^2}\right) \eventprob{n_i\ge\eta_i}}{\eventprob{n_i\ge\eta_i}}
    \\
  &=2\exp\left(-\frac{2\gamma^2 \eta_i}{\left(v_{{\sf max},i} - v_{{\sf min},i}\right)^2}\right).
 \end{align}
 where the bound in (\ref{appendix:eq:bound extended conditioning - 3}) is due to (\ref{appendix:eq:hoeffding inequality - 2}), and (\ref{appendix:eq:bound extended conditioning - 5}) is due to\linebreak $\eventprob{n_i\ge\eta_i}=\sum_{\theta_i=\left\lceil\eta_i\right\rceil}^{n} \eventprob{n_i=\theta_i}$; also, $\eventprob{n_i\ge\eta_i}>0$ due to the assumption that $\rho_i>0$.
Hence, we got 
 \begin{equation}
     \label{appendix:eq:Hoeffding inequality for learned constant conditioned on n_i}
     \eventprob{ \left\lvert \widetilde{c}_i - c_i^{\sf opt}\right\rvert \ge \gamma ~\vert~ n_i\ge\eta_i } \le 2\exp\left(-\frac{2\gamma^2 \eta_i}{\left(v_{{\sf max},i} - v_{{\sf min},i}\right)^2}\right)
 \end{equation}
  The concentration inequality (\ref{appendix:eq:Hoeffding inequality for learned constant conditioned on n_i}) can be equivalently written as follows. For $\eta_i\in(0,n]$, $\gamma\ge0$, 
 \begin{equation}
     \label{appendix:eq:Hoeffding inequality for learned constant conditioned on n_i - form 2}
     \eventprob{ \left\lvert \widetilde{c}_i - c_i^{\sf opt}\right\rvert \ge \gamma\frac{R_{\beta,i} + R_{\epsilon}}{\sqrt{2\eta_i}} ~\Bigg\vert~ n_i\ge\eta_i } \le 2\exp\left(-\gamma^2\right).
 \end{equation}
 Consequently, for $\gamma\ge0$, $\eta_i\in(0,n]$, and given $n_i\ge\eta_i$, 
 \begin{equation}
     \label{appendix:eq:Hoeffding inequality for learned constant conditioned on n_i - form 3}
     \left\lvert \widetilde{c}_i - c_i^{\sf opt}\right\rvert \le \gamma\frac{R_{\beta,i} + R_{\epsilon}}{\sqrt{2\eta_i}} 
 \end{equation}
 with probability at least $1-2\exp\left(-\gamma^2\right)$.
 
Recall that this theorem assumes $\rho_i>0$. Then, from Lemma \ref{lemma:concentration bounds on the number ni of training examples that are routed to expert i}, for $\widetilde{\delta}\in(0,1)$ and given $n\ge \frac{8}{\rho_i}\ln{\left(\frac{1}{\widetilde{\delta}}\right)}$, we have $n_i > \frac{1}{2}n\rho_i$ with probability at least $1-\widetilde{\delta}$. 

Hence, the probabilistic intersection of (\ref{appendix:eq:Hoeffding inequality for learned constant conditioned on n_i - form 3}) for $\eta_i=\frac{1}{2}n\rho_i$ and Lemma \ref{lemma:concentration bounds on the number ni of training examples that are routed to expert i} gives that for $\gamma\ge0$, $\widetilde{\delta}\in(0,1)$ and given $n\ge \frac{8}{\rho_i}\ln{\left(\frac{1}{\widetilde{\delta}}\right)}$, 
 \begin{equation}
     \label{appendix:eq:Hoeffding inequality for learned constant conditioned on n_i - form 4}
     \left\lvert \widetilde{c}_i - c_i^{\sf opt}\right\rvert \le \gamma\frac{R_{\beta,i} + R_{\epsilon}}{\sqrt{n\rho_i}} 
 \end{equation}
 with probability at least $1-2\exp\left(-\gamma^2\right)-\widetilde{\delta}$. 
 This completes the proof.

\subsection{Proof of Theorem \ref{theorem:test error with learned constants formulated using test error with optimal constants - upper bound}}
\label{appendix:subsec:Proof of theorem on test error with learned constants formulated using test error with optimal constants - upper bound}

The proof starts from the estimation error formula (\ref{eq:estimation error definition}) of Lemma \ref{lemma:test error with learned constants formulated using test error with optimal constants} and setting into it the upper bound of $\left\lvert \widetilde{c}_i - c_i^{\sf opt}\right\rvert$ from Theorem \ref{theorem:statistical bound on distance between learned constant and optimal constant}:
    \begin{align}
    \label{appendix:eq:proof of theorem on test error with learned constants formulated using test error with optimal constants - upper bound}
        \mathcal{E}_{\sf est}(m,\mathcal{S}_n) &= \sum_{i=1}^{m} \left( \widetilde{c}_i - c_i^{\sf opt} \right)^2 \rho_i \\
        \label{appendix:eq:proof of theorem on test error with learned constants formulated using test error with optimal constants - upper bound - 2}
        &\le \sum_{i=1}^{m} \left( \gamma\frac{R_{\beta,i} + R_{\epsilon}}{\sqrt{n\rho_i}} \right)^2 \rho_i     
        \\
        &= \frac{\gamma^2}{n}\sum_{i=1}^{m} \left(R_{\beta,i} + R_{\epsilon}\right)^2 
        \\
        \label{appendix:eq:proof of theorem on test error with learned constants formulated using test error with optimal constants - upper bound - 3}
        &\le \frac{\gamma^2}{n}m\cdot\underset{i\in\{1,\dots,m\}}{\max} \left(R_{\beta,i} + R_{\epsilon}\right)^2
    \end{align}
The inequality in (\ref{appendix:eq:proof of theorem on test error with learned constants formulated using test error with optimal constants - upper bound - 2}) is due to applying Theorem \ref{theorem:statistical bound on distance between learned constant and optimal constant} to upper bound  $\left\lvert \widetilde{c}_i - c_i^{\sf opt}\right\rvert$ for each of the $m$ experts in the sum. These $m$ applications of Theorem \ref{theorem:statistical bound on distance between learned constant and optimal constant} are for the same $\gamma>0$ and $\widetilde{\delta}\in(0,1)$. For each of the $m$ experts, the upper bound for $\left\lvert \widetilde{c}_i - c_i^{\sf opt}\right\rvert$ holds with probability at least $1-2\exp\left(-\gamma^2\right)-\widetilde{\delta}$; hence, the upper bounds hold for all the $m$ experts with probability at least $1-2m\exp\left(-\gamma^2\right)-m\widetilde{\delta}$.

For a given confidence level $1-\delta$, where $\delta\in(0,1)$, set $\widetilde\delta=\frac{\delta}{2m}$ and $\gamma^2=\ln\left(\frac{4m}{\delta}\right)$ in (\ref{appendix:eq:proof of theorem on test error with learned constants formulated using test error with optimal constants - upper bound - 3}) and in the condition on $n$ from Theorem \ref{theorem:statistical bound on distance between learned constant and optimal constant}. This yields that if
\begin{equation}
\label{eq:condition for estimation error bound by delta}
n\ge \frac{8}{\underset{i\in\{1,\dots,m\}}{\min}\rho_i}\ln{\left(\frac{2m}{\delta}\right)},    
\end{equation}
then 
\begin{equation}
\mathcal{E}_{\sf est}(m,\mathcal{S}_n) \le  \frac{m}{n}\ln\left(\frac{4m}{\delta}\right) \underset{i\in\{1,\dots,m\}}{\max}\left(R_{\beta,i}+R_{\epsilon}\right)^2  
\end{equation}
with probability at least $1-\delta$.

\end{document}